Philosophische Fakultät
Institut für Sprache und Information

Bachelorarbeit

# Integrating Supertag Features into Neural Discontinuous Constituent Parsing

Erstgutachten: Dr. Kilian Evang
Zweitgutachten: Univ.-Prof. Dr. Laura Kallmeyer

Vorgelegt von:
Lukas Michael Mielczarek, 2756129
lukas.mielczarek@uni-duesseldorf.de
Computerlinguistik, BA

Abgabe:
Düsseldorf, 14.09.2023

# Contents

# List of Figures

# List of Tables

# List of Algorithms

# 1. Introduction

## 1.1. Motivation

Syntactic parsing lies at the foundation of many natural-language processing applications. The term denotes the automatic extraction of a syntactic representation from a natural language text, usually in the form of a sequence of words and punctuation marks. The representation is designed to capture information about the elements' grammatical relations and functions. This output is traditionally used as a basis for higher-level tasks like grammar checking, semantic analysis, question answering and information extraction (Jurafsky and Martin, 2009, chapter 12).

One widely used description of syntax is so-called *constituent structure*. Rooted in X-bar theory (Chomsky, 1970)[1], the idea of constituency arises from the observation that groups of words can behave as single units (Jurafsky and Martin, 2009, chapter 12). Take for instance the noun phrases in (1.1).

(1.1) a. the memory

b. a new computer

c. the person she really likes

Despite having completely different meanings, these three terms can occur in similar syntactic environments, for example before a verb as in (1.2).

(1.2) a. the memory *fades*

b. a new computer *arrives...*

c. the person she really likes *waits...*

However, this is not true for all of the individual components of the phrases, as can be seen in (1.3). Therefore, one assumes the existence of an abstract category called a *constituent* to which one or more words can belong and for which rules like "noun phrases can be followed by verbs" can be postulated (Jurafsky and Martin, 2009, chapter 12).

(1.3) a. *the *fades*

b. *new *arrives...*

c. *likes *spends...*

Note: The asterisk marks constructions that are not grammatical.

Traditional views of constituency demand that constituents consist of adjacent words. This is rooted in a system for modelling languages called *context-free grammar* (CFG) (Chomsky, 1956), a generative device that derives sentences by using rewrite rules, starting with an initial symbol. The derivation path marks a hierarchy of constituents that can be arranged into a *derivation/parse tree*. Figure 1.1 gives an example for such a description. Note that the tree only contains non-crossing edges which is why it is called *projective* or *continuous*.

This view of constituency was adopted by many *treebanks*. Treebanks are linguistic corpora in which the sentences have been annotated with syntactic representations (e.g. constituency trees) (Jurafsky and Martin, 2009, chapter 12). However, this leads to difficulties when analysing syntactic phenomena that are believed to exhibit non-local dependencies. While the analysis of German or other languages with varying degrees of free word order evidently questions the projective approach, the English language also exhibits a fair amount of discontinuous phenomena, see (1.4) and (1.5).

[1] Please refer to Kornai and Pullum (1990) for a critical reflection on Chomsky's ideas.

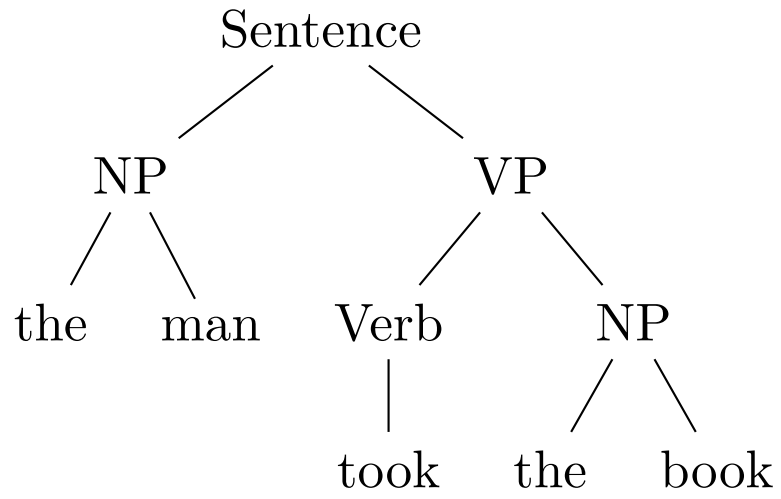

Figure 1.1: Projective derivation tree for *the man took the book* (Chomsky, 1956).

(1.4) a. sie *packte* ihre Werkzeuge *ein*
she packed her tools up

'she packed her tools'

b. *den Dank* möchte er nicht *annehmen*
the thanks want he not accept

'he does not want to accept the thanks'

c. *ohne Scham* scheinen sie *das komplette Buffet aufgegessen* zu haben
without shame seem they the complete buffet eaten to have

'they seem to have eaten the entire buffet without shame'

(1.5) a. *areas of the factory* were particularly dusty *where the crocidolite was used* (Evang, 2011)

b. *a man* entered *who was wearing a black suit* (McCawley, 1982)

Some treebanks like the English *Penn Treebank* (Marcus et al., 1993) opted to mark discontinuous phenomena through indexing and trace nodes, i.e. empty nodes in the place where a discontinuous subsection of the sentence would be expected in "standard" word order. This is motivated by the tradition of analysing discontinuities as instances of transformation or syntactic movement as introduced by Chomsky (1975). The tree in figure 1.2 features three cases of co-indexing. While such a strategy acknowledges the existence of discontinuous relationships, the long-range dependencies of the markers cannot be fully captured by traditional parsing techniques for projective grammars and often remain unused.

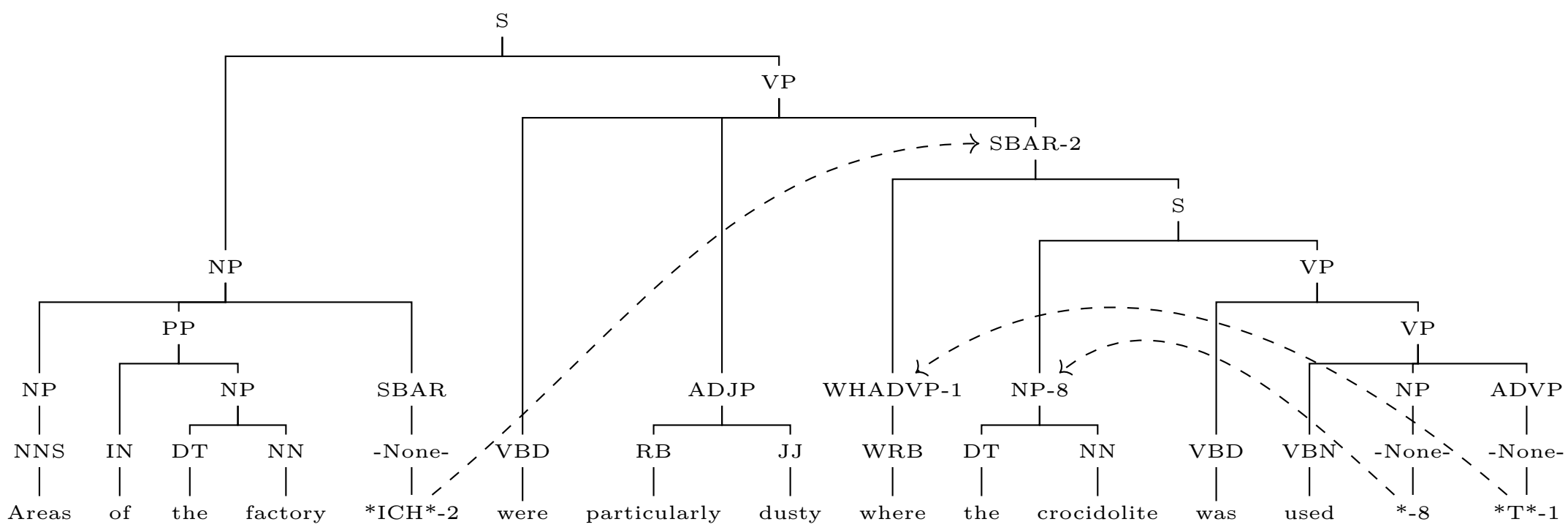

Figure 1.2: A tree from Penn Treebank, without arc-labels, adopted from Evang (2011).

In a number of treebanks like the German NeGra (Skut et al., 1998) and TIGER treebanks (Brants et al., 2004) or English *discontinuous Penn Treebank* (DPTB) (Evang and Kallmeyer, 2011) long-range dependencies are represented by crossing edges and constituents with non-adjacent elements. The resulting trees are called *discontinuous constituent trees.* Figure 1.3 shows an example from the NeGra corpus.

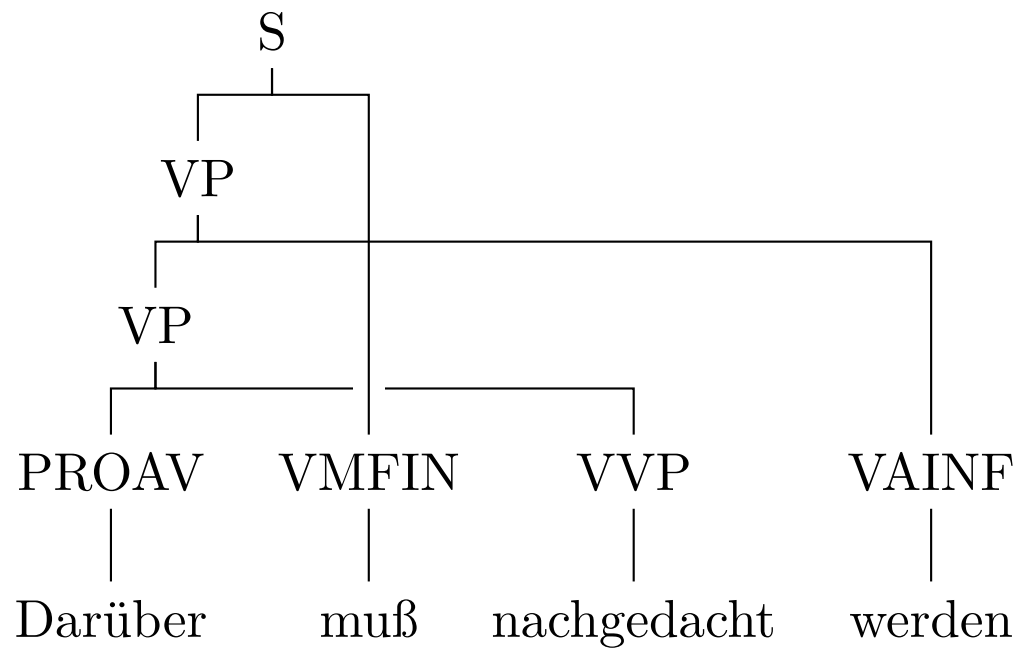

Figure 1.3: A tree from NeGra corpus, without arc-labels (Kallmeyer, 2010, chapter 6).

In order to capture syntactic descriptions with discontinuous constituents, several formalisms have been proposed. Among these *linear context-free rewriting systems* (LCFRS) (Vijay-Shanker et al., 1987) have been shown to be a natural candidate for accurately covering the notion of discontinuous constituents employed by the aforementioned treebanks.

Being a foundation for many natural language processing tasks, parsing speed is a key concern and a driver of active research. Parsing LCFRS is non-trivial in terms of efficiency. The adaptation of traditional chart parsing for LCFRS results in a polynomial time complexity of $\mathcal{O}(n^{3 \cdot dim(G)})$ where $n$ is the length of a sentence and $dim(G)$ is the maximal number of discontinuous blocks of a constituent in the grammar. Transition-based parsing approaches aim at reducing this factor by doing away with the need for an explicit grammar. Instead, an artificial neural network is trained to produce discontinuous constituent trees given only raw text input using *supervised learning* on large annotated corpora. Errors in prediction are backpropagated through the network resulting in optimisations of its parameters. In this way, it implicitly learns to identify regularities that it can effectively use to predict trees for unseen data. A recent and elegant proposal for a neural stack-free transition-based parser developed by Coavoux and Cohen (2019) successfully allows for the derivation of any discontinuous constituent tree over a sentence in worst-case quadratic time.

## 1.2. Thesis Aims

The purpose of this work is to explore the possibility of enhancing the accuracy of neural grammar-less discontinuous constituent parsing by introducing *supertag* information. In so-called *lexicalised* grammar formalisms like *lexicalised tree adjoining grammars* (LTAG) (Schabes and Joshi, 1991) and *combinatory categorial grammar* (CCG) (Steedman, 1989, 1996, 2000) informative categories are assigned to words in a sentence that act as the sole building blocks in the composition of the sentence's overarching syntactic representation. These categories are called *supertags*. The set of rules for connecting these building blocks is kept minimal. Therefore, a supertag assignment gives strong information about the structural role of the word in the sentence and about its syntactic relationship with the surrounding items.

Figure 1.4 shows a possible supertag assignment for the word *took* in LTAG. The incomplete tree serves as a primitive structure in the formalism that directly dictates the position and number of the verb's nominal argument phrases (NP). Given this assignment, a parser would not need to reconstruct the higher-level constituent structure any more. Instead, it would simply need to link a group of elementary trees.[2]

[2]This illustrative description of LTAG is of course incomplete. A second operation called *adjoining* also allows the insertion of trees into the inner structure of other trees.

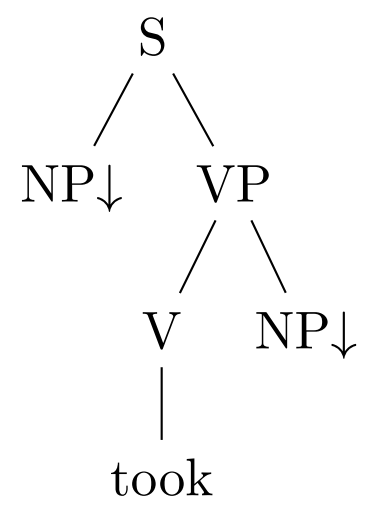

Figure 1.4: LTAG elementary tree for *likes*.

Information of this sort may aid in the analysis of discontinuous constituents. "Non-standard" word order, including discontinuity, can be expected to be directly encoded in the lexical category assignment. This work shall therefore explore its integration based on two approaches: by using the output of a dedicated supertagger as additional input for a neural parser (pipeline) and by jointly training a neural model to perform both parsing and supertagging (auxiliary/multi-task learning). Due to the public availability of a state-of-the-art supertagger for CCG (Yoshikawa et al., 2017), the focus will be on CCG supertags. Additionally, several other kinds of supertags (LTAG-spinal, LCFRS) and sequence labellings tasks (chunking, dependency parsing) will be compared in terms of their suitability as auxiliary tasks for neural discontinuous constituent parsing.

A secondary aim is to present the LCFRS formalism and to give an overview of the landscape of both grammar-based and grammar-less parsing approaches for discontinuous constituent trees. This is done to outline the advancements that led to the development of the stack-free transition system of Coavoux and Cohen (2019) used as the basis for the incorporation of supertags in this work.

## 1.3. Related Work

The potential of assigning supertags to word sequences using statistical methods as a pre-step to parsing was initially successfully explored by Bangalore and Joshi (1999) for lexicalised tree adjoining grammar (LTAG) and adapted for combinatory categorial grammar (CCG) by Clark and Curran (2010). Recently, Ruprecht and Mörbitz (2021); Ivliev (2020) have presented the first extraction algorithm for supertags based on linear context-free rewriting systems (LCFRS) and achieved state-of-the-art results in discontinuous constituent parsing.

These approaches differ from this work in that they use supertagging as an upstream task for grammar-based parsers of the respective formalism, for instance LCFRS chart-parsing (Ruprecht and Mörbitz, 2021). The extracted supertags or supertag distributions are directly utilised to reduce ambiguity and increase parsing speed by ruling out or prioritizing certain pre-terminal rule applications which effectively prunes the derivational search space. In contrast to that, I investigate the effect of supertagging in the context of a neural, grammar-less parsing approach as an input feature and as an auxiliary task. From this also follows that my choice of supertags is not structurally restricted to a certain formalism but rather to statistical correlations between supertag assignments and parsing actions.

The first thorough exploration of auxiliary/multi-task learning for natural language processing was performed by Collobert and Weston (2008) who simultaneously predict part-of-speech tags, chunk labels[3], named entity tags, semantic roles, semantic similarity and the likelihood of a sentence's well-formedness from jointly trained word embeddings. Investigations into the realm of multi-task learning that followed are mainly centred around dependency parsing[4] and in many

[3]Chunking separates a sentence into simple constituent types like NP or VP and is sometimes called *shallow parsing* (Collobert and Weston, 2008). Each word is only assigned one unique type. The sequence *a pink car* would be tagged with B-NP I-NP I-NP with B-NP marking the start of the chunk.

[4]Dependency grammars are a class of grammatical theories centred around the notion of dependency as opposed to

cases deal with incorporating higher-level semantic tasks.

Candito (2022) extracts auxiliary tasks from semantic dependency graphs like the number of dependents of a word and a concatenation of incoming arc labels to increase the accuracy of a dependency parser by training the model to predict these features from a shared bidirectional recurrent neural network. Zhou et al. (2020a) explore jointly learning dependency and span-based constituent parsing together with semantic parsing. This approach is extended by Zhou et al. (2020b) who propose improving the quality of language models like BERT (Devlin et al., 2019) with informed linguistic knowledge by jointly training on the aforementioned tasks.

These approaches differ from this work's objective through the choice of task(s) they incorporate. They are centred around bridging the gap between syntactic and semantic representations while I try to deal with the open problem of discontinuous constituency by searching for a syntactical localised lower-level task that is informative of these phenomena.

The research on multi-task approaches pertaining to supertagging is fairly limited. Yoshikawa et al. (2017) extract dependency graphs from CCG derivations and train a neural model to predict both dependency structure and CCG supertags from a shared representation to resolve derivational ambiguity in the CCG formalism. But the actual CCG parsing is performed using an A* algorithm based on the supertag probabilities predicted by the model. Zhu and Sarkar (2019) deconstruct LTAG supertags into several components like *head*, *root* or *spine* effectively bootstrapping new tasks from an existing supertag annotated corpus. They use these tasks to improve LTAG supertagging performance. Søgaard and Goldberg (2016) show that combining CCG supertagging and POS tagging using a hierarchical approach where the easier POS tagging task is predicted from a lower level yields improvements in supertagging accuracy. This work borrows from their idea of a hierarchical auxiliary task arrangement. Bingel and Søgaard (2017) explore binary multi-task relations between a range of natural language processing tasks including CCG tagging, POS tagging, chunking and semantic tasks. The effect of supertag prediction as an auxiliary task for parsing has not been examined to this date.

The work on beneficial auxiliary tasks for neural constituent parsing is also sparse. Coavoux and Crabbé (2017b) predict several word tagging tasks from a shared representation of a model for projective constituent parsing: POS tags, morphological features and functional label, i.e. the relation of a word to its head in a constituent. To the best of my knowledge, besides jointly predicted POS tags (Coavoux and Cohen, 2019), the only exploration of multi-task frameworks for grammar-less discontinuous constituent parsers to this date has been recently brought forward by Johansson and Adesam (2020) who train constituent parsing on a Swedish discontinuous treebank with a similar annotation scheme to NeGra/TIGER. They pursue parsing based on differently annotated corpora, including dependency treebanks, as an auxiliary task with predictions from a shared intermediate representation to leverage the limited availability of linguistic resources in Swedish. This differs from the work at hand in that the objectives of Johansson and Adesam (2020) are all parsing tasks predicted at the same model level while I explore the implementation of a lower-level task. Furthermore, Johansson and Adesam (2020) use a different transition system for discontinuous constituent parsing.[5]

To the best of my knowledge, the work at hand is the first exploration of supertagging in a pipeline arrangement and as an auxiliary task for grammar-less neural constituent parsers. It might therefore help to identify if a shared neural representation benefits parsing and specifically the resolvement of challenging discontinuous phenomena and if further research in this direction is a worthwile endeavour.

Note that there has been a substantial amount of research on the question of beneficial schedul-constituency. While constituent phrase structure grammars arrange words into hierarchical constituents, dependency grammars express word-to-word relations through labelled directed arcs (Jurafsky and Martin, 2009, chapter 12.7).

[5] They utilise the *Swap* action for reordering (Versley, 2014) while this work is based on the stack-free MERGE-approach proposed by Coavoux and Cohen (2019).

ing of multi-task learning (Kiperwasser and Ballesteros, 2018; Zaremoodi and Haffari, 2019) and on metrics for the exploration of well complementing auxiliary tasks (Ben-David and Schuller, 2003; Bingel and Søgaard, 2017) in the context of natural language processing. I will leave an examination of supertagging and parsing in this regard to future work.

## 1.4. Structure

The following sections are structured as follows. Section 2 gives a detailed introduction to linear context-free rewriting systems (LCFRS), a natural extension of context-free grammars that accounts for discontinuous constituents. Section 3 presents two discontinuous parsing approaches: LCFRS CYK chart-parsing and grammar-less transition-based parsers. Then, in section 4 follows a discussion of the stack-free transition system proposed by Coavoux and Cohen (2019) including its implementation. Section 5 explains the formalism of combinatory categorial grammar (CCG), a major lexicalised grammar theory that gives rise to informative supertags, and examines its account of discontinuous phenomena that can be found in the discontinuous Penn Treebank (DPTB). Finally, section 6 describes the implementations and results of several experiments for the integration of CCG supertags into the stack-free transition-based neural parser. Section 7 summarises the findings and discusses them in a broader context.

# 2. Transgressing Context-Freeness

Discontinuous constituent trees can be seen as structural descriptions produced by derivations of linear context-free rewriting systems (LCFRS). They were introduced by Vijay-Shanker et al. (1987) and are an extension of traditional context-free grammars (CFG), which do not suffice to adequately treat discontinuous structures. This section shall be concerned with defining both CFGs and LCFRSs as well as with the derivation trees that accompany them.

## 2.1. Prerequisites

The formalism of context-free grammars (CFG), developed by Chomsky (1956), has a long history as the foundation for modelling the syntactic structure of natural languages. In order to define CFGs, several prerequisites are needed. I follow the definitions in Kallmeyer (2010).

**Definition 2.1 (Alphabet, word, language).**
Let $X$ be a non-empty finite set of symbols.

1. $X$ is called an alphabet.
2. A string $x_1x_2....x_n$ with $n \in \mathbb{N}$ and $x_i \in X$ for all $i \in \{1, ..., n\}$ is called a *non-empty word over $X$*. $X^+$ is defined as the set of all non-empty words over $X$.
3. I define $X^* := X^+ \cup \{\varepsilon\}$ for a new element $\varepsilon \notin X^+$ called *empty word*. $\varepsilon$ is defined as the neutral element of concatenation on $X^*$, i.e. for $w \in X^* : w\varepsilon = \varepsilon w = w$. $w \in X^*$ is called a *word over $X$*.
4. A set $L$ is called *a language over $X$* iff $L \subseteq X^*$.

Two explanatory notes on the preceding definition: firstly, for $i, j \in \mathbb{Z}$ the expression $\{i, ..., j\}$ is used as a shorthand notation for $\{n \in \mathbb{Z} \mid i \leq n \leq j\}$. Secondly, in the context of natural language an alphabet $X$ can be a set of natural language words treated as unique symbols. For clarity, the term *word* will only refer to the formal definition given above in this work unless stated otherwise.

**Definition 2.2 (Length of a word).**
Let $X$ be an alphabet and $w \in X^*$. The *length of $w$* is denoted by $|w|$ and defined recursively:

$$|w| := \begin{cases} 1 + |w'|, & \text{if } w = xw' \text{ for some } x \in X, \\ 0, & \text{otherwise.} \end{cases}$$

**Definition 2.3 (Word repetition).**
Let $X$ be an alphabet and $w \in X^*$. Word exponentiation for $k \in \mathbb{N}_0$ is defined as word repetition:

$$w^k := \begin{cases} \varepsilon, & \text{if } k = 0, \\ ww^{k-1}, & \text{otherwise.} \end{cases}$$

**Definition 2.4 (Context-free grammar).**
A context-free grammar $G$ is defined by a 4-tuple $G = \langle N, T, P, S \rangle$ where

1. $N$ is an alphabet. The elements of $N$ are called *nonterminals*.
2. $T$ is an alphabet. The elements of $T$ are called *terminals*.
3. it holds that $N \cap T = \emptyset$.

4. $P \subset N \times (N \cup T)^*$ is a finite set of *productions/rewriting rules*. A production $\langle A, \beta \rangle \in P$ can be written as $A \to \beta$.

5. $S \in N$ is the *start symbol*.

Context-free grammars describe languages by generating the words of the language through derivation. The following definition is given by Jurafsky and Martin (2009, chapter 12):

**Definition 2.5 (Derivation).**
Let $G = \langle N, T, P, S \rangle$ be a CFG.

1. Given $w, w' \in (N \cup T)^*$, $w$ *directly derives* $w'$ iff there exists a production $A \to \beta \in P$ and there are strings $\alpha, \gamma \in (T \cup N)^*$ such that $w = \alpha A \gamma$ and $w' = \alpha \beta \gamma$. In this case we write $w \Rightarrow w'$.

2. For $w_1, w_2, ..., w_n \in (N \cup T)^*, n \in \mathbb{N}$ such that $w_1 \Rightarrow w_2, w_2 \Rightarrow w_3, ..., w_{n-1} \Rightarrow w_n$ we write $w_1 \overset{*}{\Rightarrow} w_n$ and say that $w$ *derives* $w'$. $\overset{*}{\Rightarrow}$ is the reflexive transitive closure of $\Rightarrow$.

For $w_1, w_2, w_3 \in (N \cup T)^*$ such that $w_1 \Rightarrow w_2$ and $w_2 \Rightarrow w_3$ the shorthand-notation $w_1 \Rightarrow w_2 \Rightarrow w_3$ is used.

**Definition 2.6 (Context-free language).**
For a CFG $G = \langle N, T, P, S \rangle$ the language of $G$ is denoted by $\mathscr{L}_G$ and defined as the set of all strings that consist only of terminal symbols and can be derived from the start symbol:

$$\mathscr{L}_G := \{w \mid w \in T^* \land \ S \overset{*}{\Rightarrow} w\}.$$

A language $L$ is called *context-free* iff there exists a CFG $G$ such that $L = \mathscr{L}_G$.

**Example 2.7.**
Equation 2.1 gives a set of CFG rules for the language $L$ over $\{a, b, c\}$ with $L = \{xyz \mid x, z \in \{a, b\}^+, y \in \{c\}^+\}$, numbered for easy reference. A derivation of the word $abcba \in L$ is shown in equation 2.2 with the numbers below the arrows referencing the production that was applied.

$$\begin{aligned} r_1 &: \quad S \to ACA \\ r_2 &: \quad A \to aA \\ r_3 &: \quad A \to bA \\ r_4 &: \quad A \to a \\ r_5 &: \quad A \to b \\ r_6 &: \quad C \to cC \\ r_7 &: \quad C \to c \end{aligned} \tag{2.1}$$

$$S \underset{r_1}{\Rightarrow} ACA \underset{r_2}{\Rightarrow} aACA \underset{r_5}{\Rightarrow} abCA \underset{r_7}{\Rightarrow} abcA \underset{r_3}{\Rightarrow} abcbA \underset{r_4}{\Rightarrow} abcba \tag{2.2}$$

The definition of trees is essential to our matter. I base the following definitions on Kallmeyer (2010, chapter 1).

**Definition 2.8 (Directed graph).**

1. A *directed graph* is a pair $\langle V, E \rangle$ where $V$ is a finite set of *nodes/vertices* and $E \subseteq V \times V$ a set of *edges*.

2. For $v \in V$ the *in-degree of* $v$ is defined as $|\{v' \in V \mid \langle v', v \rangle \in E\}|$.

3. For $v \in V$ the *out-degree of* $v$ is defined as $|\{v' \in V \mid \langle v, v' \rangle \in E\}|$.
4. $E^+$ is the transitive closure of $E$; $E^*$ is the reflexive transitive closure of $E$.

**Definition 2.9 (Path).**
A sequence $v_1, ..., v_n$ of vertices in $V$ with $n \in \mathbb{N}$ such that $v_k E v_{k+1}$ for all $k \in \{1, ..., n-1\}$ and all vertices in the sequence are distinct is called a *path in* $\langle V, E \rangle$.

**Definition 2.10 (Tree).**

1. A *tree* is a triple $\langle V, \triangleleft, \hat{v} \rangle$ such that[6]
   (a) $\langle V, \triangleleft \rangle$ is a directed graph and $\hat{v} \in V$ (called the *root node*),
   (b) $\triangleleft$ is acyclic, i.e., $\nexists v \in V : v \triangleleft^+ v$,
   (c) $\hat{v}$ has in-degree 0,
   (d) all nodes $v \in V \setminus \{\hat{v}\}$ have in-degree 1 and
   (e) every node is accessible from $\hat{v}$, i.e., $\forall v \in V : \hat{v} \triangleleft^* v$.
2. Given a tree $\langle V, \triangleleft, \hat{v} \rangle$,
   (a) we say that $v_1$ *directly dominates* $v_2$ iff $v_1 \triangleleft v_2$ for $v_1, v_2 \in V$. In this case, we call $v_1$ *the parent* of $v_2$ and $v_2$ *a daughter* of $v_1$.
   (b) we say that $v_1$ *dominates* $v_2$ iff $v_1 \triangleleft^* v_2$ for $v_1, v_2 \in V$.
   (c) we say that $v_2$ is a *sister* of $v_3$ iff $v_1 \triangleleft v_2$ and $v_1 \triangleleft v_3$ for $v_1, v_2, v_3 \in V$.
3. An *ordered tree* is a quadruple $\langle V, \triangleleft, \prec, \hat{v} \rangle$ such that $\langle V, \triangleleft, \hat{v} \rangle$ is a tree and $\prec$ is a *linear precedence* relation with the following properties:
   (a) $\prec$ is irreflexive, antisymmetric and transitive,
   (b) $\forall v_1, v_2 \in V : (v_1 \not\triangleleft^* v_2 \wedge v_2 \not\triangleleft^* v_1) \Rightarrow v_1 \prec v_2 \vee v_2 \prec v_1$,
   (c) $\forall v_1, v_2 \in V : (v_1 \not\triangleleft^* v_2 \wedge v_2 \not\triangleleft^* v_1) \wedge ((\exists v_3 \in V : v_3 \triangleleft v_1 \wedge v_3 \prec v_2) \vee (\exists v_4 \in V : v_4 \triangleleft v_2 \wedge v_1 \prec v_4)) \Rightarrow v_1 \prec v_2$,
   (d) nothing else is in $\prec$.

A node with out-degree 0 is called a *leaf*. All other nodes are called *internal*.

**Definition 2.11 (Labelling).**
A *labelled directed graph* is a triple $\langle V, E, \lambda \rangle$ such that $\langle V, E \rangle$ is a directed graph and $\lambda$ is a function $\lambda : V \to A$ called the *node labelling of the graph over signature* $A$. This definition naturally extends to *labelled trees* $\langle V, \triangleleft, \hat{v}, \lambda \rangle$ and *labelled ordered trees* $\langle V, \triangleleft, \prec, \hat{v}, \lambda \rangle$.

For ordered and for labelled trees I explicitly include the precedence relation and the labelling function in the tuple defining the respective tree type since it will facilitate working with these structures in the following sections.

**Definition 2.12 (Syntactic tree).**
Let $N$ and $T$ be disjoint alphabets (of nonterminal and terminal symbols).
A *syntactic tree over* $N$ *and* $T$ is an ordered finite labelled tree $\langle V, \triangleleft, \prec, \hat{v}, \lambda \rangle$ such that $\lambda(v) \in N$ for each internal node $v$ and $\lambda(u) \in (N \cup T \cup \{\varepsilon\})$ for each leaf $u$.

**Definition 2.13 (Parse tree, derivation tree, tree language of a CFG).**
Let $G = \langle N, T, P, S \rangle$ be a CFG.

[6] The use of the symbol $\triangleleft$ was adopted from Versley (2014). I naturally extend the notation to $v_1 \not\triangleleft v_2 :\Leftrightarrow \langle v_1, v_2 \rangle \notin \triangleleft$.

1. A syntactic tree $\langle V, \lhd, \prec, \hat{v}, \lambda \rangle$ over $N$ and $T$ is a *parse tree in* $G$ iff
   (a) $\lambda(v) \in (T \cup \{\varepsilon\})$ for every $v \in V$ if $v$ is a leaf,
   (b) for every $v_0, v_1, ..., v_n \in V, n \in \mathbb{N}$ such that $v_0 \lhd v_i$ for all $i \in \{1, ..., n\}$ where there is no $u \in V : v_0 \lhd u$ and $u \notin \{v_1, ..., v_n\}$ and where $v_j \prec v_{j+1}$ for all $j \in \{1, ..., n-1\}$, it holds that $\lambda(v_0) \to \lambda(v_1), ..., \lambda(v_n) \in P$.
2. A parse tree $\langle V, \lhd, \prec, \hat{v}, \lambda \rangle$ is a *derivation tree* in $G$ iff $\lambda(\hat{v}) = S$.
3. The *tree language* of $G$ is defined as $\mathscr{L}_G^{\mathcal{D}} := \{D \mid D \text{ is a derivation tree in } G\}$.

Following Chomsky (1963) the notion of generative capacity can be introduced:

**Definition 2.14 (Generative capacity).**

- The *weak generative capacity of a grammar* $G$ is defined by all strings it can generate, i.e. by $\mathscr{L}_G$.
- The *strong generative capacity of a grammar* $G$ is defined by all structural descriptions it produces. For a grammar where the notion of a tree language is defined this amounts to $\mathscr{L}_G^{\mathcal{D}}$.

**Example 2.15.**
Figure 2.1 depicts a derivation tree for the word *abcba* in the grammar given in example 2.7.

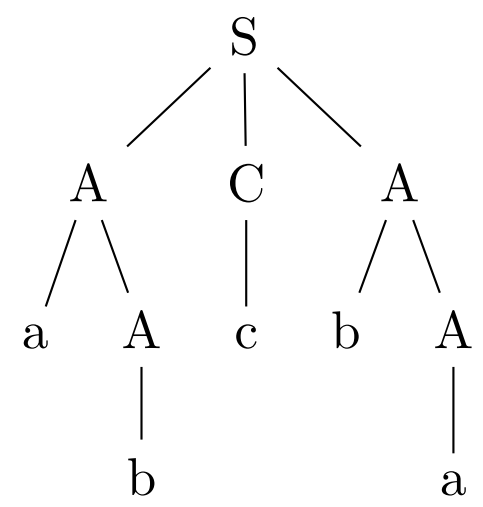

Figure 2.1: Derivation tree for the word *abcba* using the rules from example 2.7.

## 2.2. Insufficiency of Context-Free Grammars

Since the early 1980s, many researchers argued that CFGs are inadequate for describing natural language. Many of these arguments, however, were based on the strong generative capacity of grammars, e.g. Bresnan et al. (1982), who analysed non-local dependencies in Dutch. They relied on assumptions about the underlying structure and were therefore susceptible to critique. Primarily, this style of argument does not rule out the possibility that a different context-free grammar formalism (with different structural descriptions) might be able to generate the language in question (Kallmeyer, 2010, chapter 2).

The matter was finally resolved when Shieber (1985) showed that nesting relative clauses with crossing dependencies in Swiss German also surpass the weak generative capacity of CFGs. These findings led to the search of appropriate grammatical formalisms strong enough to generate long-distance relationships and at the same time restricted enough to allow for efficient parsing. In this context, Joshi (1985) coined the term *mild context-sensitivity* for a category of formalisms that possess certain desirable properties. Kallmeyer (2010, chapter 2) give the following definition for mild context-sensitivity:

**Definition 2.16 (Mild context-sensitivity).**

1. A set of languages $\mathcal{L}$ over an alphabet $X$ is *mildly context-sensitive* iff

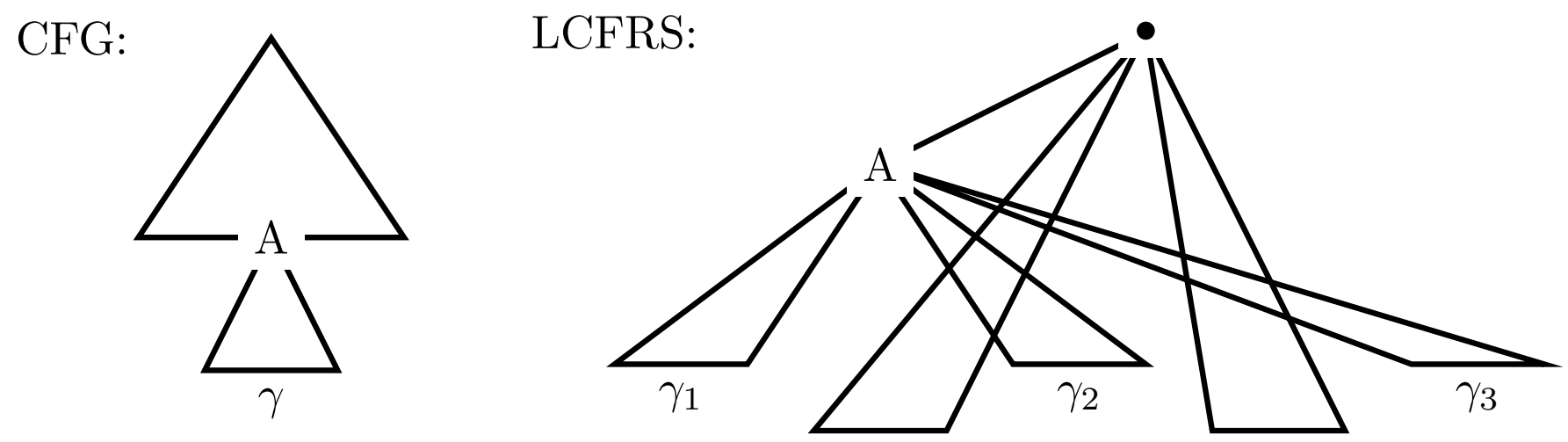

Figure 2.2: Illustration of the domain of locality of CFG and LCFRS, inspired by Kallmeyer (2010, chapter 1).

(a) $\mathcal{L}$ contains all context-free languages,

(b) $\mathcal{L}$ can describe a limited amount of cross-serial dependencies, i.e. there is an $n \geq 2$ such that $\{w^k \mid w \in T^*\} \in \mathcal{L}$ for all $k \leq n$,

(c) the languages in $\mathcal{L}$ can be parsed in polynomial time and

(d) the languages in $\mathcal{L}$ have the *constant growth property.*

2. a formalism $F$ is mildly context-sensitive iff the set of languages it can describe $\{\mathscr{L}_G \mid G \in F\}$ is mildly context-sensitive.

The constant-growth property means, that if one orders the lengths of the words in a language, the difference in length between two consecutive elements is restricted by a constant (Kallmeyer, 2010, chapter 2). This is defined formally by Weir (1988) in the following way:

**Definition 2.17 (Constant growth property).**
Let $L$ be a language over an alphabet $X$. $L$ has the *constant growth property* iff there is a constant $c_0 > 0$ and a finite set of constants $C \subset \mathbb{N}$ such that for all $w \in L$ with $|w| > c_0$, there is a $w' \in L$ with $|w| = |w'| + c$ for some $c \in C$.

Several formalisms in the scope of mild context-sensitivity have been proposed. Among these are tree adjoining grammar (TAG) (Joshi et al., 1975), *head grammar* (HG) (Pollard, 1984), *linear indexed grammars* (LIG) (Gazdar, 1988) and linear context-free rewriting systems (LCFRS) (Vijay-Shanker et al., 1987) as well as certain variants of combinatory categorial grammar (CCG) (Steedman, 1989, 1996, 2000).

LCFRSs were shown to naturally arise from (discontinuous) treebank representations (Kallmeyer, 2010, chapter 1). They are an adequate formalism for describing discontinuous constituents utilised in treebanks and will therefore be a subject for further inspection in the following sections.

## 2.3. Defining LCFRSs

Linear context-free rewriting systems (LCFRS) are a powerful type of mildly context-sensitive grammar formalism that extends the rewriting-nature of CFGs. Like context-free grammars, the formalism possesses rewriting rules with *left-hand sides* (LHS) and *right-hand sides* (RHS). Words are derived through recursive application of the rules which in turn constitute the constituent structure desirable from a linguist's point of view.

The key difference is the fact that nonterminals do not yield a string of terminals, but a tuple of strings. The strings in one tuple may be intertwined with elements of other tuples. This allows an element to have a yield of non-adjacent strings. The size of the tuples is fixed for each nonterminal $A$ and is called its *fan-out.* Thus, the fan-out of $A$ gives the maximum number of (discontinuous) components $A$ dominates. The fan-out of the start symbol is set to 1. Figure 2.2 shows a tree where the nonterminal $A$ yields three non-adjacent components $\gamma_1, \gamma_2, \gamma_3$.

LCFRSs are equivalent to simple *range concatenation grammars* (SRCG) (Boullier, 1998b) and to *multiple context-free grammars* (MCFG) (Seki et al., 1991). A formal definition for LCFRS was first outlined by Vijay-Shanker et al. (1987). In the following I will reference the characterisation given by Evang (2011) who adapts the SRCG notation from Boullier (1998b) for LCFRS.

**Definition 2.18 (Linear context-free rewriting system).**
A *linear context-free rewriting system* (LCFRS) is a 5-tuple $G = \langle N, T, V, P, S\rangle$ where

1. $N, T$ and $V$ are alphabets. The elements of $N$ are called *nonterminals*, those of $T$ *terminals* and those of $V$ *variables*. $T$ and $V$ are disjoint.
2. each nonterminal $A \in N$ has an associated *fan-out* denoted by a function $dim : N \to \mathbb{N}$.
3. the maximum fan-out is denoted by $dim(G) := \max(\{dim(A) \mid A \in N\})$. For $k = dim(G)$, $G$ is said to have fan-out $k$ and is called a $k$-LCFRS.
4. $P \subset (N \times ((T \cup V)^*)^{\leq dim(G)}) \times (N \times V^{\leq dim(G)})^{<\mathbb{N}_0}$ is a finite set of *productions/rewriting rules.*[7] A rule

   $$r = \langle\langle A_0, \langle \chi_{0,1}, ..., \chi_{0,dim(A_0)}\rangle\rangle, \langle\langle A_1, \langle \chi_{1,1}, ..., \chi_{1,dim(A_1)}\rangle\rangle, ..., \langle A_m, \langle \chi_{m,1}, ..., \chi_{m,dim(A_m)}\rangle\rangle\rangle$$

   can be written as

   $$A_0(\langle \chi_{0,1}, ..., \chi_{0,dim(A_0)}\rangle) \to A_1(\langle \chi_{1,1}, ..., \chi_{1,dim(A_1)}\rangle)...A_m(\langle \chi_{m,1}, ..., \chi_{m,dim(A_m)}\rangle)$$

   where the part left of the arrow is called left-hand side (LHS) and denoted by $LHS(r)$ and the part right of the arrow is called right-hand side (RHS) and denoted by $RHS(r)$. Furthermore:

   (a) Every variable occurring in a rule occurs exactly once on the LHS and exactly once on the RHS.

   (b) There is a function

   $$rank : P \to \mathbb{N}, \qquad r \mapsto |RHS(r)|,$$

   associated with the grammar which assigns the number of its RHS's components to a rule. Furthermore, *rank* is also defined for the grammar as the maximum rank occurring for any rule:

   $$rank(G) := \max(\{rank(r) \mid r \in P\}).$$

   (c) The strings $\chi_{0,1}, ..., \chi_{0,dim(A_0)} \in (T \cup V)^*$ are called the LHS *arguments*. $\langle \chi_{0,1}, ..., \chi_{0,dim(A_0)}\rangle$ is called the LHS *argument tuple*.

   (d) The variables $\chi_{1,1}, ..., \chi_{1,dim(A_1)}, ..., \chi_{m,1}, ..., \chi_{m,dim(A_m)} \in V$ are called the RHS *arguments*. The tuples $\langle \chi_{1,1}, ..., \chi_{1,dim(A_1)}\rangle, ..., \langle \chi_{m,1}, ..., \chi_{m,dim(A_m)}\rangle$ are called the RHS *argument tuples*.

   (e) if $rank(r) = 0$, the RHS is written as $\varepsilon$.

By convention $A_k(\langle \chi_{n,1}, ..., \chi_{n,dim(A_k)}\rangle)$ is equivalent to $A_k(\chi_{n,1}, ..., \chi_{n,dim(A_k)})$ and a tuple can be denoted by a variable in bold, e.g. $\boldsymbol{\alpha}$. $\boldsymbol{\alpha}^{[i]}$ refers to the $i$-th element of the tuple. The fan-out of a grammar gives the maximum number of non-adjecent strings in the yield of a nonterminal.

In this formalism, rewriting rules describe how the yield of the nonterminal on the LHS can be computed using the yields of the RHS nonterminals. The notion of a yield can be defined formally (Evang and Kallmeyer, 2011):

[7] $X^{\leq n}$ denotes the set of all tuples with maximum number of components $n$ over a set $X$. $X^{<\mathbb{N}_0}$ is the set of all finite tuples over $X$.

**Definition 2.19 (Yield and string language of an LCFRS).**
Let $G = \langle N, T, V, P, S \rangle$ be an LCFRS.

1. The function $yield : N \to \mathcal{P}((T^*)^{\leq dim(G)})$ is defined as follows: for every $A \in N$:
   (a) for every $A(\boldsymbol{\alpha}) \to \varepsilon \in P$, it holds that $\boldsymbol{\alpha} \in yield(A)$.
   (b) for every rule $A_0(\chi_{0,1}, ..., \chi_{0,dim(A_0)}) \to A_1(\chi_{1,1}, ..., \chi_{1,dim(A_1)})...A_m(\chi_{m,1}, ..., \chi_{m,dim(A_m)}) \in P$ and for all $\boldsymbol{\alpha}_i \in yield(A_i)$ with $i \in \{1, ..., m\}$ it holds that $\langle f(\chi_{0,1}), ..., f(\chi_{0,dim(A)}) \rangle \in yield(A)$ where the *composition function* $f$ is defined as follows:
      i. $f(t) = t$ for all $t \in T$,
      ii. $f(\chi_{i,j}) = \boldsymbol{\alpha}_i^{[j]}$ if $\chi_{i,j} \in V$ for all $i \in \{1, ..., m\}$ and all $j \in \{1, ..., dim(A_i)\}$.
      iii. $f(\beta\gamma) = f(\beta)f(\gamma)$ for all $\beta, \gamma \in (T \cup V)^+$.
   (c) nothing else is in $yield(A)$.
2. The string language of $G$ is $\mathscr{L}_G = \{w \mid \langle w \rangle \in yield(S)\}$.

The *yield*-function maps each nonterminal $A$ to its set of possible yields. Every element in the yield is the result of a distinct combination of available yields of the RHS nonterminals of a rule where $A$ stands on the LHS. Each of these elements is a tuple of sequences of terminals.

The function $f$ replaces each variable with the corresponding component of the yield of the RHS argument it also occurs in. Since for every $\chi_{0,k} \in V$ with $k \in \{1, ..., dim(A_0)\}$ there exists exactly one $\chi_{i,j}$ among the RHS arguments with $\chi_{i,j} = \chi_{0,k}$ and unique indices $i, j$ with $i \in \{1, ..., m\}$ and $j \in \{1, ..., dim(A_i)\}$ (definition 2.18 (3.f)), it follows that $f(\chi_{0,k}) = \boldsymbol{\alpha}_i^{[j]}$. Thus, the function is well-defined.

**Example 2.20.**
Equation 2.3 gives a set of LCFRS production for the language $L$ over $\{a, b, c\}$ with $L = \{wcccw \mid w \in \{a, b\}^+\}$, numbered for easy reference.

| number | LHS | | RHS |
|---|---|---|---|
| $r_1$ | $S(UVWX)$ | $\to$ | $A(U, X)B(V, W)$ |
| $r_2$ | $A(aU, aX)$ | $\to$ | $A(U, X)$ |
| $r_3$ | $A(bU, bX)$ | $\to$ | $A(U, X)$ |
| $r_4$ | $A(a, a)$ | $\to$ | $\varepsilon$ |
| $r_5$ | $A(b, b)$ | $\to$ | $\varepsilon$ |
| $r_6$ | $B(c, cc)$ | $\to$ | $\varepsilon$ |
| $r_7$ | $B(cc, c)$ | $\to$ | $\varepsilon$ |

(2.3)

The production $A(b, b) \to \varepsilon$ specifies that the tuple $\langle b, b \rangle$ is in the yield of $A$ and $A(aU, aX) \to A(U, X)$ means that it is possible to generate a new tuple in the yield of $A$ from an already existing one by prepending $a$ to both components.

It holds that $\langle abcccab \rangle \in yield(S)$ since:

$$\begin{array}{llll} \langle b, b \rangle & \in yield(A), & \text{using } r_5 & \\ \langle ab, ab \rangle & \in yield(A), & \text{using } r_2, & \text{identifying } \langle U, X \rangle \text{ with } \langle b, b \rangle \in yield(A), \\ \langle c, cc \rangle & \in yield(B), & \text{using } r_6 & \\ \langle abcccab \rangle & \in yield(S), & \text{using } r_1, & \text{identifying } \langle V, W \rangle \text{ with } \langle c, cc \rangle \in yield(B) \text{ and} \\ & & & \langle U, X \rangle \text{ with } \langle ab, ab \rangle \in yield(A). \end{array} \tag{2.4}$$

Similar to context-free grammars it is possible to define derivation trees. Derivation trees model the construction of a word beginning with the start symbol. Unlike phrase-structure trees of context free-grammars, LCFRS derivation trees are unordered. For two nodes it is not possible to say that one precedes the other since their children might be intertwined, leading to crossing branches. Nevertheless, LCFRS derivation trees are usually drawn so that the positions of the leaves correspond to their positions in the word. A *range* is assigned to each leaf that uniquely identifies its position in the word. For two internal nodes $v_1, v_2$, usually $v_1$ is drawn left of $v_2$ if it dominates some leaf that precedes all leaves dominated by $v_2$ in the word.

An internal node with a nonterminal as label together with its children represents a rule application. The parent corresponds to the LHS nonterminal while the children consist of other internal nodes labeled with the RHS nonterminals and/or leaves that correspond to terminals on the LHS.

Evang (2011) gives a formal definition for derivation trees that I outline in the following. It requires defining the concepts of *range*, *range concatenation* and *rule instance*, which are taken from range concatenation grammars (Boullier, 1998b). I modified the range notation slightly to account for its use in other contexts in this work.

**Definition 2.21 (Indices).**
Let $X$ be an alphabet. The set of all indices for a word $w$ over $X$ is denoted as $Ind(w) := \{1, ..., |w|\}$.

**Definition 2.22 (Positions).**
Let $X$ be an alphabet. The set of all positions for a word $w$ over $X$ is denoted as $Pos(w) := Ind(w) \cup \{0\}$.

**Definition 2.23 (Tuple projection).**

1. Let $X$ be an alphabet. For some tuple of positive integers $\langle i_1, ..., i_n \rangle \in \mathbb{N}^{<\mathbb{N}_0}$ the projection function $\pi_{\langle i_1,...,i_n \rangle}$ is defined on all words $w \in X^*$ where $i_1, ..., i_n \in Ind(w)$ as

$$\pi_{\langle i_1,...,i_n \rangle}(w) = w_{i_1}, ..., w_{i_n}.$$

$\pi_{\langle i_1,...,i_n \rangle}(w)$ is usually written as $w_{\langle i_1,...,i_n \rangle}$. For the 0-tuple $\langle\ \rangle$ it naturally holds that $w_{\langle\ \rangle} = \varepsilon$.

2. Let $X$ be an alphabet. For some tuple of tuples of positive integers

$$\langle \langle i_{1,1}, ..., i_{1,n_1} \rangle, ..., \langle i_{m,1}, ..., i_{m,n_m} \rangle \rangle \in (\mathbb{N}^{<\mathbb{N}_0})^{<\mathbb{N}_0}$$

the second-degree tuple projection function $\pi^2_{\langle \langle i_{1,1},...,i_{1,n_1} \rangle,...,\langle i_{m,1},...,i_{m,n_m} \rangle \rangle}$ is defined on all words $w \in X^*$ where $i_{1,1}, ..., i_{1,n_1}, ..., i_{m,1}, ..., i_{m,n_m} \in Ind(w)$ as

$$\pi^2_{\langle \langle i_{1,1},...,i_{1,n_1} \rangle,...,\langle i_{m,1},...,i_{m,n_m} \rangle \rangle}(w) = \langle \pi_{\langle 1,1,...,i_{1,n_1} \rangle}(w), ..., \pi_{\langle i_{m,1},...,i_{m,n_m} \rangle}(w) \rangle.$$

$\pi^2_{\langle \langle i_{1,1},...,i_{1,n_1} \rangle,...,\langle i_{m,1},...,i_{m,n_m} \rangle \rangle}(w)$ is usually written as $w_{\langle \langle i_{1,1},...,i_{1,n_1} \rangle,...,\langle i_{m,1},...,i_{m,n_m} \rangle \rangle}$. For the 0-tuple $\langle\ \rangle$ it naturally holds that $\pi^2_{\langle\ \rangle}(w) = \langle\ \rangle$. It is not abbreviated to $w_{\langle\ \rangle}$ to avoid confusion with the first-order 0-tuple projection.

**Definition 2.24 (Range).**

1. A tuple $\langle a, b \rangle \in \mathbb{N}_0 \times \mathbb{N}_0$ with $a \leq b$ is called a *range*. A range $\langle a, b \rangle$ with $a = b$ is called an *empty range*.

2. The set of all ranges is denoted by *Ranges*.

3. The function $\langle : \rangle : Ranges \to \mathbb{N}^{<\mathbb{N}}$ maps a range $\langle a, b \rangle$ to a tuple where all natural numbers from $a + 1$ to $b$ are arranged in order, i.e.:

$$\langle a, b \rangle \mapsto \langle a + 1, a + 2, ..., b - 1, b \rangle$$

$\langle:\rangle(\langle a,b\rangle)$ can be written as $\langle a:b\rangle$.

4. $\langle:\rangle$ is extended to tuples of ranges, e.g. $\langle:\rangle(\langle\langle a_1,b_1\rangle,\langle a_2,b_2\rangle\rangle) = \langle\langle a_1:b_1\rangle,\langle a_2:b_2\rangle\rangle$.

5. The length of a range $\langle a,b\rangle$ is defined as $|\langle a,b\rangle| := b-a$.

6. Two ranges $\langle a_1,b_1\rangle$ and $\langle a_2,b_2\rangle$ are called *non-overlapping* iff $b_1 \leq a_2 \vee b_2 \leq a_1$.

7. The precedence relation $<_{\text{range}} \;\subset \mathit{Ranges} \times \mathit{Ranges}$ is defined as follows:

$$\langle a_1,b_1\rangle <_{\text{range}} \langle a_2,b_2\rangle :\Leftrightarrow b_1 \leq a_2$$

**Definition 2.25 (Ranges in a word).**
Let $X$ be an alphabet and $w$ a word over $X$.

1. A range $\langle a,b\rangle$ is called *a range in* $w$ iff $a,b \in \mathit{Pos}(w)$.

2. The set of all ranges in $w$ is denoted by $\mathit{Ranges}(w)$.

3. For some range $\langle a,b\rangle$ in $w$, $w_{\langle a:b\rangle}$ is called a *substring of* $w$ and conventionally written as $w_{a:b}$.

**Definition 2.26 (Range concatenation).**

1. The concatenation operation on two ranges $\langle a_1,b_1\rangle$ and $\langle a_2,b_2\rangle$ in a word $w$ over an alphabet $X$ where $b_1 = a_2$ is naturally defined as $\langle a_1,b_1\rangle \circ \langle a_2,b_2\rangle = \langle a_1,b_2\rangle$. It is undefined if $b_1 \neq a_2$.

2. A string of ranges $\rho_1...\rho_n \in \mathit{Ranges}(w)^*$ where $\rho_i \circ \rho_{i+1}$ is defined for all $i \in \{1,...,n-1\}$, is called *contiguous*. For such a string, range concatenation is defined as follows:

$$\bigcirc(\rho_1...\rho_n) := \begin{cases} \varepsilon & \text{if } n = 0, \\ \rho_1 & \text{if } n = 1, \\ \bigcirc((\rho_1 \circ \rho_2)\rho_3...\rho_n) & \text{otherwise.} \end{cases}$$

It follows directly from this definition that for two ranges $\rho_1,\rho_2$ in a word $w$ over some alphabet $X$ where the concatenation of $\rho_1$ and $\rho_2$ is defined, the following holds: $w_{\langle:\rangle(\rho_1)} w_{\langle:\rangle(\rho_2)} = w_{\langle:\rangle(\rho_1\circ\rho_2)}$.

**Definition 2.27 (Rule instance).**
Let $G = \langle N,T,V,P,S\rangle$ be an LCFRS, $r = A_0(\chi_{0,1},...,\chi_{0,dim(A)}) \to A_1(\chi_{1,1},...,\chi_{1,dim(A_1)})...$ $A_m(\chi_{m,1},...,\chi_{m,dim(A_m)}) \in P$ a rule and $w \in T^*$ a word.

1. $r$ can be transformed into a *rule instance of* $r$ *with respect to* $w$ by replacing the characters of the argument strings $\chi_{0,1},...,\chi_{0,dim(A_0)},\chi_{1,1},...,\chi_{1,dim(A_1)},...,\chi_{m,1},...,\chi_{m,dim(A_m)}$ of $p$ in the following fashion:

   (a) All terminals $t$ get replaced with some range $\langle a-1,a\rangle$ in $w$ such that $w_{a-1:a} = t$.

   (b) All variables get replaced with some range $\langle a,b\rangle$ in $w$.

   (c) All empty arguments, i.e. if $\chi_{i,j} = \varepsilon$ for some $i,j$ with $i \in \{0,...,m\}$ and $j \in \{1,...,dim(A_i)\}$, get replaced with some range $\langle a,a\rangle$ in $w$.

   The resulting arguments are strings of ranges. Each of these strings $\rho_{i,j}$ must be contiguous and gets replaced by $\bigcirc(\rho_{i,j})$.

2. The resulting rule instance has the form $A_0(\boldsymbol{\rho}_0) \to A_1(\boldsymbol{\rho}_1),...,A_m(\boldsymbol{\rho}_m)$ where $\boldsymbol{\rho}_k \in \mathit{Ranges}(w)^{dim(A_k)}$ is a tuple of ranges for all $k \in \{1,...,m\}$.

3. The set of all rule instances of $r$ wrt. $w$ is called $Instances(r, w)$.

The process described in definition 2.27 converts each of the arguments to a range in the given word where all the terminal symbol match an occurrence in the word. The range for variables can be freely chosen since there are no indications as to the substring they span. Each of the resulting sequences of ranges needs to be contiguous, i.e. the chosen ranges need to be adjacent, since a tuple component always has to match a continuous substring in the word.

**Example 2.28.**
Given the LCFRS from example 2.20 and the word $w = abcccab$, the following are some examples for rule instances for rule $r_2 = A(aU, aX) \to A(U, X)$ along with the two substrings of $w$ their LHS spans:

$$\begin{aligned}
&A(\langle 0:2\rangle, \langle 5:7\rangle) \to A(\langle 1:2\rangle, \langle 6:7\rangle), && \underline{ab}ccc\underset{\cdot\cdot}{ab} \\
&A(\langle 0:5\rangle, \langle 5:7\rangle) \to A(\langle 1:5\rangle, \langle 6:7\rangle), && \underline{abccc}\underset{\cdot\cdot}{ab} \\
&A(\langle 0:2\rangle, \langle 5:6\rangle) \to A(\langle 1:2\rangle, \langle 6:6\rangle), && \underline{ab}ccc\underset{\cdot}{a}b \\
&A(\langle 5:7\rangle, \langle 0:2\rangle) \to A(\langle 6:7\rangle, \langle 1:2\rangle), && \underset{\cdot\cdot}{ab}ccc\underline{ab} \\
&A(\langle 5:7\rangle, \langle 0:7\rangle) \to A(\langle 6:7\rangle, \langle 1:7\rangle) \in Instances(w, r_2) && \underset{\cdot\cdot\cdot\cdot\cdot\cdot\cdot}{abccc\underline{ab}}
\end{aligned} \tag{2.5}$$

Note that a rule instance can map a terminal to any matching occurrence in the word regardless of order. Two ranges in a range tuple can even overlap. However, as established, the start symbol must span one continuous range. Thus, all spans in a derivation must be reducible through concatenation to a single range. This is checked by determining if all of the RHS elements of a rule instance, i.e. assumptions about the span of its variables, can be found as LHS of some instantiated rule. Such a derivation naturally forms a tree.

**Definition 2.29 (LCFRS derivation tree).**
Let $G = \langle N, T, V_G, P, S\rangle$ be an LCFRS and $w = w_1...w_n \in T^*$ a word.

1. Let $D = \langle V, \lhd, \hat{v}, \lambda\rangle$ be a labelled tree such that there are pairwise different leaves $v_1, ..., v_n$ in $D$ with $\lambda(v_i) = \langle i-1, i\rangle$ for $i \in \{1, ..., n\}$ and all other leaves have a label $\langle i, i\rangle$ for some $i \in \{0, ..., n\}$. The function $n\text{-}yield : V \to \mathcal{P}(Ranges(w)^{\leq dim(G)})$ assigns a set of tuples of ranges to each node in the tree and is defined as follows:
   (a) For every leaf $u \in V : n\text{-}yield(u) = \{\langle\lambda(u)\rangle\}$.
   (b) For every internal node $v_0 \in V$, for every order $v_1, ..., v_m$ of all pairwise different internal nodes $v_1, ..., v_m \in V$, where $v_0 \lhd v_i$ and $\lambda(v_i) = A_i \in N$ for all $i \in \{1, ..., m\}$, for every rule $r = A_0(\boldsymbol{\alpha}_0) \to A_1(\boldsymbol{\alpha}_1), ..., A_m(\boldsymbol{\alpha}_m) \in P$, for every instance $A_0(\boldsymbol{\rho}_0) \to A_1(\boldsymbol{\rho}_1), ..., A_m(\boldsymbol{\rho}_m) \in Instances(r, w)$, it holds that $\boldsymbol{\rho}_0 \in n\text{-}yield(v_0)$ if:
      i. $\boldsymbol{\rho}_i \in n\text{-}yield(v_i)$ for all $i \in \{1, ..., m\}$,
      ii. for every terminal $t$ occurring in $\boldsymbol{\alpha}_0$, there exists a leaf $u \in V$ such that $v_0 \lhd u$ and $\lambda(u)$ is the range with which $t$ was replaced to obtain $\boldsymbol{\rho}_0$.
      iii. for all $i \in \{1, ..., dim(A_0)\}$ such that $\boldsymbol{\alpha}_0^{[i]}$ is empty, there is a $u \in V$ such that $v_0 \lhd u$, $u$ is a leaf and $\lambda(u) = \boldsymbol{\rho}_0^{[i]}$.

      Nothing else is in $n\text{-}yield(v_0)$.
2. $D$ is a *derivation tree of* $w$ *in* $G$ iff $\lambda(r) = S$ and $\langle\langle 0:n\rangle\rangle \in n\text{-}yield(r)$. We write $\mathcal{D}_G^w$ for the set of all derivation trees of $w$ in $G$.

3. The tree language of $G$ is defined as $\mathscr{L}_G^{\mathcal{D}} := \{D \mid \exists w \in T^* : D \in \mathcal{D}_G^w\}$.

It will prove useful to define the class of syntactic representations arising from LCFRS derivation trees independently of the formalism. It can be seen that every LCFRS derivation tree is a *range-labelled tree* and that for any *range-labelled tree*, an LCFRS and a word exist for which it is a derivation tree.

**Definition 2.30 (Partial range-labelled tree).**
Let $n \in \mathbb{N}_0$ be an integer and $N$ an alphabet (of nonterminal symbols).

1. A *partial range-labelled tree over $N$ in length $n$* is an unordered finite labelled tree $\langle V, \lhd, \hat{v}, \lambda \rangle$ such that
   (a) $\lambda(v) \in N$ for each internal node $v \in V$,
   (b) $\lambda(u) = \langle i-1, i \rangle$ with $i \in \{1, ..., n\}$ or $\lambda(u) = \langle i, i \rangle$ with $i \in \{0, ..., n\}$ for every leaf node $u \in V$,
   (c) for all leaf nodes $u_1, u_2 \in V$ if $|\lambda(u_1)| = |\lambda(u_2)| = 1$ and $u_1 \neq u_2$ it must hold that $\lambda(u_1) \neq \lambda(u_2)$.
2. A *partial $\varepsilon$-free range-labelled tree over $N$ in length $n$* is a partial range-labelled tree over $N$ in length $n$ such that either
   (a) $n > 0$ and no leaf is labelled with an empty range or
   (b) $n = 0$ and there is only one leaf.

**Definition 2.31 (Range-labelled tree).**
Let $n \in \mathbb{N}_0$ be an integer and $N$ an alphabet (of nonterminal symbols).

1. A *range-labelled tree over $N$ spanning length $n$* is a partial range-labelled tree $\langle V, \lhd, \hat{v}, \lambda \rangle$ such that for every $i \in \{1, ..., n\}$ there exists a leaf $u \in V$ such that $\lambda(u) = \langle i-1, i \rangle$. This definition naturally extends to $\varepsilon$-free partial range-labelled trees.
2. Given an alphabet $T$ and a word $w \in T^*$ such that $|w| = n$, a range-labelled tree $D$ over $N$ spanning length $n$ is called *a range-labelled tree for $w$*.

In essence, a range labelled tree is an unordered tree with an ordering defined on its leaves. In the following, the term *discontinuous tree* will also be used to refer to this class of representations.

**Example 2.32.**
Given the LCFRS from Example 2.20 and the word *abcccab*, a possible derivation tree is shown in figure 2.3.

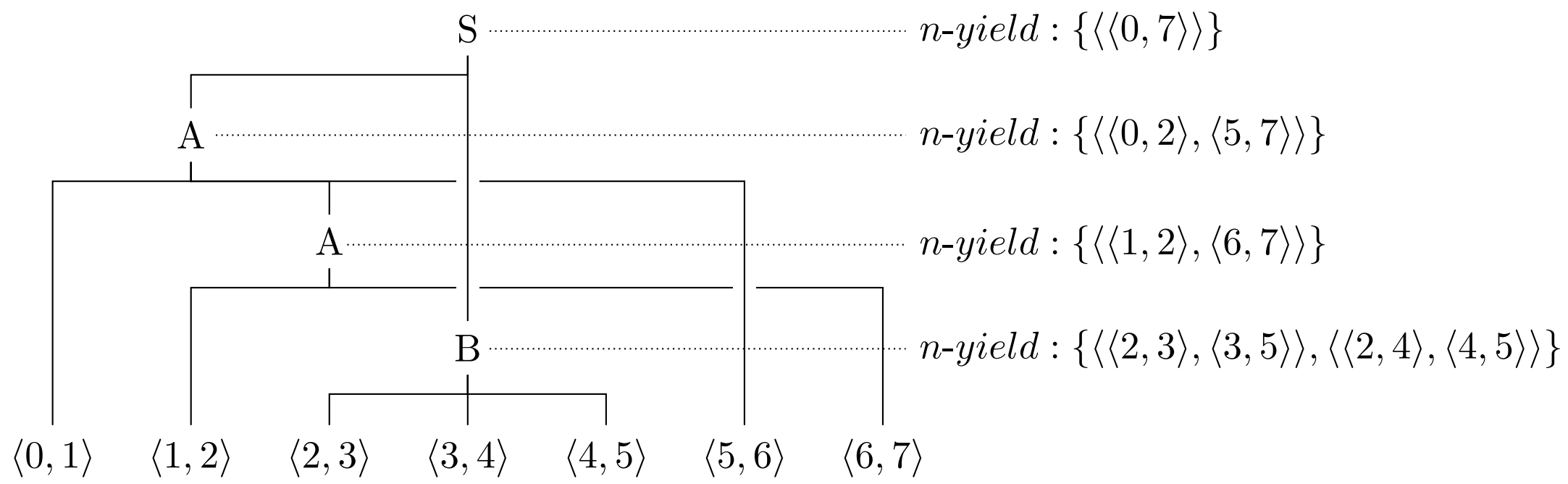

Figure 2.3: Derivation tree for the word *abcccab* in the grammar from example 2.20, including *n-yields*.

This tree illustrates that it would not suffice to map $n\text{-}yield(v)$ for $v \in V$ to a single tuple of ranges since adjacent terminals like in the case of B can allow for different combinations of ranges in the components.

## 2.4. From Treebanks to LCFRSs

As shown in the last section, LCFRS derivation trees result in the desired tree representations of non-overlapping discontinuous constituents found in constituency treebanks with discontinuities. These trees can be directly interpreted as LCFRS derivation trees. Søgaard and Maier (2008) describe an algorithm for extracting probabilistic LCFRSs (PLCFRS) from discontinuous treebanks. Kallmeyer and Maier (2010) give the following definition for PLCFRSs.

**Definition 2.33 (Probablistic LCFRS).**
A probabilistic LCFRS (PLCFRS) is a 6-tuple $\langle N, T, V, P, S, q\rangle$ such that $\langle N, T, V, P, S\rangle$ is an LCFRS and $q : P \to [0, 1]$[8] is a function such that for all $A \in N$:

$$\sum_{A(\boldsymbol{\alpha})\to\Phi\in P} q(A(\boldsymbol{\alpha}) \to \Phi) = 1$$

The function $q$ assigns a conditional probability to each rule which is based on the frequency with which instances of this rule have been found in the treebank. For more information on the extraction algorithm, please refer to Søgaard and Maier (2008).

Related research includes the work of Kuhlmann and Satta (2009) who have shown that it is possible to extract LCFRSs from dependency treebanks. Evang and Kallmeyer (2011) describe a technique to convert context-free phrase structure trees in the English language Penn Treebank into a discontinuous format using inherent additional information about non-local dependencies. This research established the discontinuous Penn Treebank (DPTB).

[8] For $i, j \in \mathbb{R}$ the expression $[i, j]$ shall equal $\{n \in \mathbb{R} \mid i \leq n \leq j\}$.

# 3. Parsing LCFRS

Parsers for LCFRSs can be utilised for producing discontinuous constituent trees from text. In the following, I will outline two main parsing approaches: traditional chart parsing and transition-based parsing. Both approaches are usually *data-driven* in that they extract their syntactic knowledge from treebank descriptions. In the case of chart-parsers, this extraction is explicit. To build the parser, LCFRS rules are automatically extracted from an annotated corpus. Transition-based parsers allow the implementation of grammar-less parsing. They are usually built on a neural-network model that is directly trained on a treebank.

## 3.1. CYK Parser

Traditional *Cocke-Younger-Kasami* (CYK) parsers were first introduced for CFG by Sakai (1961). Their underlying idea is to assert the applicability of rules in a bottom-up fashion, starting with the terminal symbols and progressively validating derivation steps on the basis of the set of rule applications that has been recovered already. Seki et al. (1991) extended the algorithm to MCFGs. It was adapted for probabilistic LCFRS by Kallmeyer and Maier (2010). I will discuss the latter approach in more detail.

### 3.1.1 Prerequisites

The input of the CYK PLCFRS parser consists of a grammar and an input word. The parser requires a PLCFRS $\langle N, T, V, P, S, p\rangle$ to have the following properties:[9]

1. *binary*: For every rule $r \in P : rank(r) \leq 2$ (Gómez-Rodríguez et al., 2009).
2. *terminal-restricted*: Terminals occur only in terminating rules of the form $A(a) \to \varepsilon$ where $a$ is a terminal. The LHS of all other rules contains only variables (Evang, 2011).
3. *gap-explicit*: Two different variables in a RHS element of a rule cannot appear next to each other in the same component on the LHS side. This means that a separation of two variables always entails that there is a gap between them (Kallmeyer and Maier, 2010).
4. *ordered*: The order of the variables within each RHS element is the same as their order of occurrence in the LHS (Villemonte de la Clergerie, 2002).
5. *$\varepsilon$-free*: There exists no rule with some empty LHS argument or there exists exactly one $\varepsilon$-rule: $S(\varepsilon) \to \varepsilon$ and $S$ does not occur in the RHS of any rule (Boullier, 1998a).

The algorithm in Søgaard and Maier (2008) produces PLCFRSs that fulfil properties 2, 3, 4 and 5. Note that for every LCFRS there is an equivalent LCFRS (i.e. generating the same language) that is $\varepsilon$-free (Boullier, 1998a), ordered (Kallmeyer, 2010, chapter 7), binary (Gómez-Rodríguez et al., 2009) and terminal-restricted (trivial).

Furthermore, for every LCFRS there exists an equivalent LCFRS that is *gap-explicit*. An algorithm for converting an ordered LCFRS into this form is given below. Thus, the properties above can all be assumed without loss of generality in the following sections.

The algorithm uses the fact that the LCFRS is ordered: variables $X_1, ..., X_n$ that appear next to each other in the same component in the LHS argument tuple and as components in the same argument tuple on the RHS cannot have another component in between on the RHS since this would conflict with the order. Therefore only the case $\langle ..., X_1, ..., X_n, ...\rangle$ for the RHS is relevant.

The algorithm traverses all rules and all elements $A_k(...)$ on the rule's RHS. For each element, the occurrence of useless variables is checked. For overlapping sequences of variables adjacent

[9] For lack of established nomenclature, I introduce the terms *terminal-restricted* and *gap-explicit*.

**Algorithm 3.1** Removing useless variable separation

**Input:** an ordered LCFRS $\langle N, T, V, P, S \rangle$

**Output:** an equivalent LCFRS that is *gap-explicit*

**for all** rules $r = A_0(\boldsymbol{\alpha}_0) \rightarrow A_1(\boldsymbol{\alpha}_1)...A_m(\boldsymbol{\alpha}_m) \in P$ **do**
    **for all** $k \in \{1, ..., m\}$ **do**
        $u \leftarrow \{\langle l, r \rangle \mid r{-}l \geq 1, \boldsymbol{\alpha}_k^{[l]}\boldsymbol{\alpha}_k^{[l+1]}...\boldsymbol{\alpha}_k^{[r]}$ is a max. adjacent substr. of some LHS argument$\}$
        **if** not $A_k^u \in N$ **then**
            introduce new nonterminal $A_k^u$ to $N$ with $dim(A_k^u) = dim(A_k) - \sum_{\langle l,r \rangle \in u} r - l$
            **for all** rules of form $A_k(\boldsymbol{\gamma}) \rightarrow \Phi \in P$ **do**
                $\boldsymbol{\gamma}' \leftarrow \boldsymbol{\gamma}$
                **for all** $\langle l, r \rangle \in u$ **do**
                    merge $\boldsymbol{\gamma}'^{[l]}, ..., \boldsymbol{\gamma}'^{[r]}$ into a single component $\boldsymbol{\gamma}'^{[l]}...\boldsymbol{\gamma}'^{[r]}$
                **end for**
                add new rule $A_k^u(\boldsymbol{\gamma}')$ to $P$
            **end for**
        **end if**
        **for all** $\langle l, r \rangle \in u$ **do**
            pick a new variable $X$
            replace components $\boldsymbol{\alpha}_k^{[l]}, ..., \boldsymbol{\alpha}_k^{[r]}$ with a single component $X$
            replace corresponding substring on the LHS with $X$
        **end for**
        replace the nonterminal symbol $A_k$ on the RHS with $A_k^u$
    **end for**
**end for**
Remove useless productions

on the LHS, the sequence with the largest number of elements is considered (*maximal adjacent substring*). A set $u$ is introduced to mark the start and end index in the RHS tuple for each sequence of useless components.

Now, we can replace all of these occurrences with new single variables both on the LHS and on the RHS and postpone the variable split to the productions governed by $A_k$.

To achieve this, a new nonterminal $A_k^u$ marked with the information which components were merged when deriving it from $A_k$, is introduced. All of the rules in $P$ where $A_k$ stands on the LHS are duplicated for $A_k^u$ with all of the components described by $u$ concatenated into single components since they were already asserted to be adjacent in the rule deriving $A_k$. In order to avoid generating new nonterminals for the same $A_k$ with the same merged components multiple times, no new nonterminal is introduced if $A_k^u$ has been created already.

Through the introduction of $A_k^u$, the original nonterminal $A_k$ might have become unreachable if it cannot be found in the RHS of some other rule. Useless productions can be eliminated in a final step with the algorithm given in Kallmeyer (2010, chapter 7).

Finally, the parser demands that the input already be assigned with part-of-speech (POS) tags, given as unary productions for each token. Therefore, when respecting all of the assumptions given above, every rule has to be of one of the following forms (Kallmeyer and Maier, 2013):

$$\begin{aligned}
&A(a) &&\rightarrow \varepsilon, &&\text{with } A \text{ being a POS tag}, a \in T \\
&A_0(\boldsymbol{\alpha}) &&\rightarrow A_1(\boldsymbol{\alpha}), &&\text{with } \boldsymbol{\alpha} \in V^{dim(A_0)},\ dim(A_0) = dim(A_1) \\
&A_0(\boldsymbol{\alpha}_0) &&\rightarrow A_1(\boldsymbol{\alpha}_1)A_2(\boldsymbol{\alpha}_2), &&\text{with } \boldsymbol{\alpha}_0 \in (V^+)^{dim(A_0)}, \boldsymbol{\alpha}_1 \in V^{dim(A_1)}, \boldsymbol{\alpha}_2 \in V^{dim(A_2)}
\end{aligned} \tag{3.1}$$

### 3.1.2 Deduction Rules

Parsing is formulated as a set of deduction rules which facilitates proofs of correctness and the examination of complexity. Deduction rules describe how to infer new elements from existing ones.

Figure 3.1 shows the general form of a deduction rule. It dictates that the *consequent item* is derivable from the *antecedent items* if the *side condition* is fulfilled. The side condition can be empty. Then no condition needs to be fulfilled.

$$\textsc{Rule} \quad \frac{antecedent_1, ..., antecedent_n}{consequent} \; side\ condition$$

Figure 3.1: Deduction principle.

An *axiom* can be expressed as a deduction rule with no antecedent items. A *parsing schema* is comprised of deduction rules with one or more axiom(s) and one or several *goal item(s)*. A successful parse is characterised by the deducibility of a goal item for a given input (Kallmeyer, 2010, chapter 3).

Following Sikkel and Nijholt (1997), I define a parsing system and the inference relation:

**Definition 3.1 (Parsing system).**
A parsing system $\mathfrak{P}$ for some grammar $G$ and a word $w$ is a triple $\langle \mathcal{I}, H, R \rangle$ such that:

1. $\mathcal{I}$ is a set of items,
2. $H$ is a set of axioms/*hypotheses*,
3. $R \subseteq \mathcal{P}_{fin}(\mathcal{P}_{fin}(\mathcal{I} \cup H) \times \mathcal{I})$ is a set of deduction rules.[10,11]

**Definition 3.2 (Inference relation).**
Let $\mathfrak{P} = \langle \mathcal{I}, H, R \rangle$ be a parsing system for some grammar $G$ and a word $w$. The following relations are defined on $\mathcal{P}_{fin}(\mathcal{I} \cup H) \times \mathcal{I}$.

1. We write $Y \vdash_r c$ iff there is $Y' \subseteq Y$ such that $c$ can be deduced from the elements in $Y'$ using rule $r \in R$, i.e. such that $\langle Y', c \rangle \in r$.
2. We write $Y \vdash c$ iff $\exists r \in R : Y \vdash_r c$.
3. We call $Y, c_1, c_2, ..., c_n$ a deduction sequence iff $Y \cup \{c_1, ..., c_{i-1}\} \vdash c_i$ for all $i \in \{1, ..., n\}$ with $n \in \mathbb{N}$. We conventionally write $Y \vdash c_1 \vdash c_2 \vdash ... \vdash c_n$.
4. We write $Y \vdash^* c$ iff $c \in Y$ or $Y \vdash ... \vdash c$.
5. An item $c \in \mathcal{I}$ is called *valid* iff $H \vdash^* c$.

In the context of LCFRS CYK parsing, the items of the parsing system are *intermediate parsing results*:

**Definition 3.3 (PLCFRS CYK intermediate parsing result).**
Let $G = \langle N, T, V, P, S, q \rangle$ be a PLCFRS, $A \in N$ a nonterminal, $\boldsymbol{\rho} \in Ranges^{dim(A)}$ and $m \in \{n \in \mathbb{R} \mid n \geq 0\}$. $m : [A, \boldsymbol{\rho}]$ is called an intermediate parsing result or an item. $m$ is called the *weight* of the item. The set of all items for $G$ is called $\mathcal{I}_{\text{CYK}}$.

The tuple of ranges $\boldsymbol{\rho}$ characterises all components in the span of $A$. $m$ represents a logarithmically encoded probability that is used to determine the relative probability of different parses.

A set of deduction rules constrains the set of valid intermediate parsing results for an input word $w$. The idea here is that a valid item should represent a parse for a part of $w$ that is licensed by the grammar. This property is proven in section 3.1.5. The full set of deduction rules for PLCFRS CYK parsing is given in figure 3.2.

[10] $\mathcal{P}_{fin}$ denotes the powerset containing only finite sets.
[11] Note that Sikkel and Nijholt (1997) do not differentiate between different deduction rules but subsume all deductions in a single set in the definition of $\mathfrak{P}$.

$$
\begin{array}{ll}
\textsc{Scan} & \dfrac{}{0 : [A,\ \langle\langle i, i+1\rangle\rangle]}\ A \text{ is a POS tag of } w_{i+1} \\
\textsc{Unary} & \dfrac{in : [B, \boldsymbol{\rho}]}{in + |\log(q(r))| : [A, \boldsymbol{\rho}]}\ r = A(\boldsymbol{\alpha}) \to B(\boldsymbol{\alpha}) \in P \\
\textsc{Binary} & \dfrac{in_B : [B, \boldsymbol{\rho}_B],\ in_C : [C, \boldsymbol{\rho}_C]}{in_B + in_C + |\log(q(r))| : [A, \boldsymbol{\rho}_A]}\ \exists r \in P : (A(\boldsymbol{\rho}_A) \to B(\boldsymbol{\rho}_B)C(\boldsymbol{\rho}_C)) \in Instances(r, w) \\
\textit{goal} & m : [S, \langle\langle 0 : |w|\rangle\rangle]
\end{array}
$$

Figure 3.2: Weighted CYK deduction system for parsing $w \in T^*$ with a PLCFRS $\langle N, T, V, P, S, q\rangle$. $R_{\text{CYK}} = \{\textsc{Unary}, \textsc{Binary}\}, H_{\text{CYK}} = \{0 : [A, \langle\langle i, i+1\rangle\rangle] \mid A \text{ is a POS tag of some } w_{i+1}\}$.

The deduction system works bottom-up. First Scan provides the axioms. These are the terminating rules that have RHS $\varepsilon$. Then the derivation tree is traversed bottom-up. Unary and Binary find ranges for rules of rank 1 or 2 respectively. Any rule instance where the pairs of nonterminals and range tuples in the RHS have been found can be applied. Therefore, its LHS is deduced. The contiguity constraint of the rule instances ensures that only elements can be chosen that can be joined in a way that is compatible with the rule to deduce, i.e. adjacent elements on the LHS are only deduced from adjacent and non-overlapping rule instances already found.

**Definition 3.4 (PLCFRS-CYK parsing system).**
The parsing schema **PLCFRS-CYK** is defined by the parsing system $\mathfrak{P}_{\text{CYK}} = \langle \mathcal{I}_{\text{CYK}}, H_{\text{CYK}}, R_{\text{CYK}}\rangle$ for any PLCFRS $G = \langle N, T, V, P, S, p\rangle$ and any word $w \in T^*$.

### 3.1.3 Chart Parsing

Natural language grammars are highly ambiguous. For a given input word $w$, a parsing schema may give rise to a variety of items spanning different, possibly overlapping, portions of $w$. There can be several unique sequences of deductions leading to a goal item for $w$. On the other hand, different analyses can be based on a common sub-analysis for a part of $w$ (Kallmeyer, 2010, chapter 3).

*Chart parsing* is a framework rooted in dynamic programming that allows to easily store intermediate parsing results and to reuse them at any time for the deduction of new elements (Grune and Jacobs, 2010, chapter 7). For this purpose, a table - called a *chart* $\mathcal{C}$ - is used. For CYK parsing, where an intermediate parsing result $[A, i, j]$ spans a projective substring of $w$, the table can have three dimensions with the first dimension corresponding to a unique index for each nonterminal. The second and third dimension have length $|w| + 1$ and correspond to the start and end index of the item's range. $[A, i, j]$ would be represented by a mark in cell $(index(A), i, j)$ of $\mathcal{C}$ (Kallmeyer, 2010, chapter 3). The cell $(index(S), 0, |w|)$ corresponds to the goal item.

Furthermore, a list called an *agenda* $\mathcal{A}$ is used. It consists of the items that are yet to be checked for possible deductions in combination with the asserted items in $\mathcal{C}$. Axioms and new items found by deduction are added to $\mathcal{A}$ (Kallmeyer, 2010, chapter 3).

A *recogniser* can be converted into a parser by saving references with each entry in $\mathcal{C}$ that refer to the items that were used to deduce it. In this way, one can retrieve the parse tree from the goal item. The references are called *backpointers* (Kallmeyer, 2010, chapter 3).

In our context, intermediate parsing results can span more than one consecutive substring and therefore a chart with three dimensions does not suffice. The number of dimensions needed is determined by the fan-out of the grammar $G$. The maximum number of range borders is given by $2 \cdot dim(G)$. When representing nonterminals in a separate dimension, the chart dimensionality is $1 + 2 \cdot dim(G)$. The goal item then is $(index(S), 0, |w|, n, n, ..., n, n)$ where $n = |w| + 1$ is a special index used for those nonterminals $A$ where $dim(A) < dim(G)$. Since this does not allow for an easy visualisation, I will represent $\mathcal{C}$ as a list containing intermediate parsing results.

### 3.1.4 Algorithm

A deductive engine can be used to deduce every valid item from the set of axioms, eventually finding the goal item with the highest weight. This naive approach would be computationally inefficient due to the large number of useless intermediate parsing results generated. Kallmeyer and Maier (2010) give a strategy for fast exploration of the best goal item based on weighted deductive parsing (Nederhof, 2003). It is reproduced in an adapted form in algorithm 3.2.

**Algorithm 3.2** Weighted deductive CYK parsing for PLCFRS.

**Input:** a parsing system $\langle I_{\text{CYK}}, H_{\text{CYK}}, R_{\text{CYK}} \rangle$ for a grammar $G$ and a word $w$
**Output:** a filled chart containing the best goal item

$\mathcal{A} \leftarrow H_{\text{CYK}}$
**while** $\mathcal{A} \neq \emptyset$ **do**
  $x : I \leftarrow$ best item in $\mathcal{A}$  ▷ select the item with the lowest weight
  $\mathcal{A}$.remove($x : I$)
  $\mathcal{C}$.add($x : I$)
  **if** $I = [S, \langle\langle 0 : |w| \rangle\rangle]$ **then**
    stop
  **else**
    **for all** $y : I'$ in (all items deducible from $x : I$ and from items in $\mathcal{C}$) **do**
      **if** $\nexists z$ with $z : I' \in \mathcal{C} \cup \mathcal{A}$ **then**
        $\mathcal{A}$.add($y : I'$)
      **else if** $\exists z$ with $z : I' \in \mathcal{A}$ **then**
        $\mathcal{A}$.remove($z : I'$)
        $\mathcal{A}$.add(min($y, z$) $: I'$)
      **end if**
    **end for**
  **end if**
**end while**

In each step, the best item in the agenda $x : I$ is written to the chart and checked for deducibility of new items by combining with the items present on the chart. This amounts to retrieving all $y : I'$ such that $\mathcal{C} \cup \{x : I\} \vdash y : I'$ since no new elements can be generated purely from items on $\mathcal{C}$ according to the logic of the algorithm. All newly generated items $y : I'$ that are not yet present in the agenda or on the chart are put on the agenda. If an item already exists on the agenda but the newly generated one has a lower (i.e. better) weight, the weight gets updated with the lower one.

By always retrieving the agenda item with the best score, the algorithm guarantees that the first goal item derived is the best one (Knuth, 1977).

**Example 3.5.**
Let $G$ be a PLCFRS characterised by the productions and probabilities in equation 3.2 with $\mathscr{L}_G = \{aaa\}$.

| number | LHS | | RHS | probability |
|---|---|---|---|---|
| $r_1$ | $S(XYZ)$ | $\rightarrow$ | $A(X,Y)B(Z)$ | 0.1 |
| $r_2$ | $S(XYZ)$ | $\rightarrow$ | $A(X,Z)B(Y)$ | 0.9 |
| $r_3$ | $A(X,Y)$ | $\rightarrow$ | $B(X)B(Y)$ | 1 |
| $r_4$ | $B(a)$ | $\rightarrow$ | $\varepsilon$ | 1 |

(3.2)

For an ordered grammar, the algorithm can be optimised to avoid deriving elements with ranges in their arguments that do not respect precedence. Furthermore, for trivial reasons, the algorithm does not need to compute tuples of ranges that contain overlaps. Figure 3.3 shows the sequence of parsing steps for the word *aaa* and PLCFRS $G$.

| $\mathcal{A}$ | $\mathcal{C}$ | new items |
|---|---|---|
| $B_0^{0:1}, B_0^{1:2}, B_0^{2:3}$ | $\emptyset$ | |
| $B_0^{1:2}, B_0^{2:3}$ | $\underline{B_0^{0:1}}$ | |
| $B_0^{2:3}$ | $\underline{B_0^{1:2}}, \underset{\cdots}{B_0^{0:1}}$ | $A_0^{0:1,1:2}$ |
| $A_0^{0:1,1:2}$ | $\underline{B_0^{2:3}}, \underset{\cdots}{B_0^{1:2}}, \underset{\cdots}{B_0^{0:1}}$ | $A_0^{1:2,2:3}, A_0^{0:1,2:3}$ |
| $A_0^{1:2,2:3}, A_0^{0:1,2:3}$ | $\underline{A_0^{0:1,1:2}}, \underset{\cdots}{B_0^{2:3}}, B_0^{1:2}, B_0^{0:1}$ | $S_{\lvert log(0.1)\rvert}^{0:3}$ |
| $A_0^{0:1,2:3}, S_{\lvert log(0.1)\rvert}^{0:3}$ | $\underline{A_0^{1:2,2:3}}, A_0^{0:1,1:2}, B_0^{2:3}, B_0^{1:2}, B_0^{0:1}$ | |
| $S_{\lvert log(0.1)\rvert}^{0:3}$ | $\underline{A_0^{0:1,2:3}}, A_0^{1:2,2:3}, A_0^{0:1,1:2}, B_0^{2:3}, \underset{\cdots}{B_0^{1:2}}, B_0^{0:1}$ | $S_{\lvert log(0.9)\rvert}^{0:3}$ |
| $S_{\lvert log(0.1)\rvert}^{0:3}$ | $\underline{S_{\lvert log(0.9)\rvert}^{0:3}}, A_0^{0:1,2:3}, A_0^{1:2,2:3}, A_0^{0:1,1:2}, B_0^{2:3}, B_0^{1:2}, B_0^{0:1}$ | |

Figure 3.3: CYK parse for word *aaa* and LCFRS $G$ from equation 3.2 according to algorithm 3.2; representing intermediate parsing results $m : [A, \boldsymbol{\rho}]$ as $A_m^{\boldsymbol{\rho}}$; the best result taken from the agenda is underlined, the chart item(s) it combines with are underlined with dots; visualising backpointers at the last step.

### 3.1.5 Soundness and Completeness

Up to this point, I assumed that the parser indeed does what it claims to do. But generally, parser *correctness* has to be proven by showing *soundness* and *completeness* for the parsing algorithm (Kallmeyer, 2010, chapter 3).

In the following, a recogniser algorithm $A$ is generalised as a function on a grammar $G$ and a word $w$ returning *true* or *false*. A class of grammars could be, for instance, the set of all CFGs or in this case the set of all ordered, binary, gap-explicit, $\varepsilon$-free, terminal-restricted LCFRSs.

**Definition 3.6 (Recogniser soundness, completeness, correctness).**
Let $A$ be a recogniser algorithm.

1. $A$ is *sound* for a class of grammars $\mathbb{G}$, iff for every grammar $G \in \mathbb{G}$ and every input word $w$, it holds that $A(G, w) = \mathit{true} \Rightarrow w \in \mathscr{L}_G$.

2. $A$ is *complete* for a class of grammars $\mathbb{G}$, iff for every grammar $G \in \mathbb{G}$ and every input word $w$, it holds that $w \in \mathscr{L}_G \Rightarrow A(G, w) = \mathit{true}$.

3. $A$ is *correct* for a class of grammars $\mathbb{G}$, iff $A$ is sound and complete for $\mathbb{G}$.

For a recogniser, *soundness* means that only the words of the language of the given grammar are recognised as correct while *completeness* is defined by the fact that all of the words of the language are recognised by the algorithm. If both properties hold, the system is called *correct*.

This notion can be extended to parsers. Here, $A$ is understood as a function on a grammar $G$ and a word $w \in \mathscr{L}_G$ returning a set of final syntactic representations (usually trees spanning the input and rooted at S).

**Definition 3.7 (Parser soundness, completeness, correctness).**
Let $A$ be a parsing algorithm.

1. $A$ is *sound* for a class of grammars $\mathbb{G}$, iff for every $G \in \mathbb{G}$, every input word $w$, $\forall D \in A(G, w)$ : $D \in \mathcal{D}_G^w$.
2. $A$ is *complete* for a class of grammars $\mathbb{G}$, iff for every grammar $G \in \mathbb{G}$ and every input word $w$, $\forall D \in \mathcal{D}_G^w : D \in A(G, w)$.
3. $A$ is *correct* for a class of grammars $\mathbb{G}$ iff it is sound and complete for $\mathbb{G}$.

For the deductive chart-parsing approach, the proof of correctness is two-fold:

1. proving soundness and completeness for $\mathbb{G}$ where $A(G, w)$ tells the presence of a goal item in the set of valid items (recogniser) or gives the set of valid goal items including backpointers (parser) (Gómez-Rodríguez et al., 2008) and
2. showing that the deductive engine actually explores all possible deductions - or the best one in case of weighted deductive parsing.

Usually, 1. is shown by proving a stronger result: soundness and completeness for the intermediate parsing results as is shown in the following.

**Corollary 3.8.**

1. Given a fixed LCFRS $G$ with the properties from section 3.1.1 and an input sentence $w = w_1...w_n$: $[A, \boldsymbol{\rho}]$ iff $w_{\langle:\rangle(\boldsymbol{\rho})} = \langle w_{\langle:\rangle\,\boldsymbol{\rho}^{[1]}}, ..., w_{\langle:\rangle\,\boldsymbol{\rho}^{[|\boldsymbol{\rho}|]}} \rangle \in \mathit{yield}(A)$.
2. The **CYK-LCFRS** parsing schema is correct for the class of all LCFRSs with the properties given in Section 3.1.1.

*Proof.* Part 1 is proven via induction over the sum of lengths for the range tuple $\boldsymbol{\rho}$ (i.e. the number of terminals the item spans) defined by $s = \sum_{i=1}^{|\boldsymbol{\rho}|} |\boldsymbol{\rho}^{[i]}|$. This is inspired by the proof for CFG CYK parsing given by Kallmeyer (2010, chapter 3).

1. Case $s = 1$:
   (⇒) Given: $[A, \boldsymbol{\rho}]$.
   The item must have form $[A, \langle\langle i-1, i\rangle\rangle]$ since empty ranges do not occur.
      i. If the item is an axiom, then by axiom definition, it holds that $w_{i-1:i} = w_i \in \mathit{yield}(A)$.
      ii. If the item was derived using UNARY, the antecedent must have form $[B, \langle\langle i-1, i\rangle\rangle]$. If the induction claim holds for the antecedent, then the UNARY side-condition claims the existence of rule $A(X) \to B(X)$ for some $X \in V$. Thus, by definition of the yield, it follows that $\langle w_i \rangle \in \mathit{yield}(A)$. The induction claim must hold for the antecedent $B$ (by induction over the deduction sequence) since it must have been derived either by another UNARY or be an axiom because all valid items $c$ must satisfy $H_{\mathrm{CYK}} \vdash^* c$ and $s$ cannot decrease by application of deductions rules.

   (⇐) Given: There is some range $\langle i-1, i\rangle$ such that $\langle w_i \rangle \in \mathit{yield}(A)$. The definition of *yield* allows two options:
      i. Either there exists a rule $A(w_i) \to \varepsilon \in P$. Then by the definition of $H_{\mathrm{CYK}}$ it holds that $[A, \langle\langle i-1, i\rangle\rangle]$.
      ii. Or there is some $A(X) \to B(X) \in P$ with $X \in V$ such that $w_i \in \mathit{yield}(B)$. In the same fashion as above, let us assume that $[B, \langle\langle i-1, i\rangle\rangle]$. Then it follows by the UNARY side-condition that $[A, \langle\langle i-1, i\rangle\rangle]$.
2. Case $s > 1$:

(⇒) Given: $[A, \boldsymbol{\rho}]$.
The item is not in $H_{\text{CYK}}$ since it spans more than one symbol.

i. If the item was derived via Binary from two items $[B, \boldsymbol{\rho}_B], [C, \boldsymbol{\rho}_C]$, the induction claim must hold for both antecedents since their individual range length sum must be smaller (productions with empty ranges are not allowed). Thus, $w_{\langle:\rangle(\boldsymbol{\rho}_B)} \in \mathit{yield}(B)$ and $w_{\langle:\rangle(\boldsymbol{\rho}_C)} \in \mathit{yield}(C)$. Now, since by the side condition there exists an instance $A(\boldsymbol{\rho}) \to B(\boldsymbol{\rho}_B)C(\boldsymbol{\rho}_C)$, it follows that there is a composition function such that $w_{\langle:\rangle(\boldsymbol{\rho})} \in \mathit{yield}(A)$.

ii. If the item was derived via Unary from an item $[B, \boldsymbol{\rho}]$: assume that the claim holds for $[B, \boldsymbol{\rho}]$. Since the side condition demanded the existence of a unary production $A(\boldsymbol{\alpha}) \to B(\boldsymbol{\alpha})$, we can assert that also $w_{\langle:\rangle(\boldsymbol{\rho})} \in \mathit{yield}(A)$. Now, by induction over the deduction sequence, the claim must hold for $[B, \boldsymbol{\rho}]$ since there must be valid items $c_x, c_y, c_z$ such that $\{c_x, c_y\} \vdash_{\text{Binary}} c_z$ and $\{c_z\} \vdash^* [B, \boldsymbol{\rho}]$ because only Binary can compose items with larger $s$ than their antecedents.

(⇐) Given: There is a range tuple $\boldsymbol{\rho}$ with $w_{\langle:\rangle(\boldsymbol{\rho})} \in \mathit{yield}(A)$. The definition of *yield* allows two options:

i. There is some binary production
$A(\chi_{A,1}, ..., \chi_{A,dim(A)}) \to B(\chi_{B,1}, ..., \chi_{B,dim(B)})C(\chi_{C,1}, ..., \chi_{C,dim(C)})$ such that $w_{\langle:\rangle(\boldsymbol{\rho})}$ was composed from the elements of a yield $w_{\langle:\rangle(\boldsymbol{\rho}_B)} \in \mathit{yield}(B)$ and $w_{\langle:\rangle(\boldsymbol{\rho}_C)} \in \mathit{yield}(C)$ according to the production. From this composition follows that there is a rule instance $A(\boldsymbol{\rho}_A) \to B(\boldsymbol{\rho}_B)C(\boldsymbol{\rho}_C)$. Furthermore, since there are no empty non-terminal arguments allowed, both of these yields span a number of tokens smaller than $s$. Thus, by induction claim there must be $[B, \boldsymbol{\rho}_B]$ and $[C, \boldsymbol{\rho}_C]$. Therefore, Binary gives $[A, \boldsymbol{\rho}]$.

ii. Or there is a unary production $A(\boldsymbol{\alpha}) \to B(\boldsymbol{\alpha})$ and also $w_{\langle:\rangle(\boldsymbol{\rho})} \in \mathit{yield}(B)$. Like above, we can assume that the induction claim holds for this antecedent. Therefore, $[B, \boldsymbol{\rho}]$ and by use of the side-condition also $[A, \boldsymbol{\rho}]$.

Part 2 follows from this result. The deduction schema is correct since for every LCFRS $G$ and for every $w$,

1. if $[S, \langle\langle 0, |w|\rangle\rangle]$ it follows that $\langle w_{0:|w|}\rangle = \langle w\rangle \in \mathit{yield}(S)$ i.e. $w \in \mathscr{L}_G$ (sound) and
2. if $w \in \mathscr{L}_G$, i.e. $\langle w\rangle \in \mathit{yield}(S)$, then $[S, \langle\langle 0 : |w|\rangle\rangle]$ (complete).

□

Correctness of the parser built by saving backpointers for each item follows directly from this proof in combination with definition 2.29. The items represent nodes of the tree, an item's non-terminal is its node label and the result of the *n-yield* function is given by the range tuple of the item.

A proof of correctness for the weighted deductive parsing algorithm for finding the best scoring derivation (algorithm 3.2) is given by Knuth (1977) for systems where the weight is non-decreasing. This is the case for the deduction rules given in figure 3.2.

### 3.1.6 Complexity

The separation of parsing algorithm and schema allows for a concise examination of complexity. Given a grammar $G$ and an input word $w$ one has to consider the maximum number of possible deduction rule applications (Kallmeyer, 2010, chapter 3). This depends on the most complex rule, in this case: Binary with $A(\boldsymbol{\rho}_A) \to B(\boldsymbol{\rho}_B)C(\boldsymbol{\rho}_C)$.

There is a maximum of $dim(G)$ variables and therefore $2 \cdot dim(G)$ range boundaries for each of the antecedents. For two variables adjacent in a LHS argument, the number of independent range boundaries reduces by one since for ranges $\rho_1, \rho_2$ the concatenation $\rho_1 \circ \rho_2$ must be defined, i.e. the right boundary of $\rho_1$ is the left boundary of $\rho_2$. In the worst case, $dim(A) = dim(B) = dim(C) = dim(G)$ and each component of the LHS contains exactly two variables. More adjacent variables would lead to a greater boundary reduction. Each LHS component now gives rise to 3 independent boundaries, which means that the total maximum number of boundaries is $3 \cdot dim(G)$. For a word of length $n$ this leads to $n^{3 \cdot dim(G)}$ possibilities to choose these boundaries and thus a time complexity of $\mathcal{O}(n^{3 \cdot dim(G)})$ for the algorithm (Evang, 2011).

Note that the time complexity of parsing LCFRSs is fundamentally dependent on the maximum number of discontinuous blocks for a nonterminal. Søgaard and Maier (2008) show that for a typical treebank LCFRS fan-out is around 3, which makes CYK parsing very computationally expensive.

### 3.1.7 Related Work

Several strategies were explored to speed up LCFRS CYK chart parsing. Kallmeyer and Maier (2013) use $A^*$ search. A coarse-to-fine approach was presented by Van Cranenburgh (2012) while Angelov and Ljunglöf (2014) propose a new cost estimation for ranking parser items. Kallmeyer (2010, chapter 7) presents several approaches for optimisations on LCFRS like optimal binarisation and elimination of useless rules. Furthermore she explains the use of filters during parsing that reject items that cannot lead to a goal item. Kallmeyer and Maier (2009) present an incremental Earley parser for SRCG.

Maier et al. (2012) focus on optimizing the extraction from treebanks by limiting the fan-out of the resulting grammar to 2. They note that in the DPTB, which features constituents with at most three discontinuous blocks, the overwhelming majority of the cases with three blocks is caused by punctuation. They show that by changing punctuation annotation the extraction of a 2-LCFRS with minimal informational loss is possible.

## 3.2. Transition-Based Parsing

Parsing with transitions is based on a pseudo-deterministic approach and widely used for dependency parsing. The parser traverses the input from left to right and performs one local action from a set of possible transitions at each step to incrementally build a graph. In the case of constituent parsing, this graph is a tree. While chart parsers handle ambiguities by maintaining multiple analyses in parallel through the use of dynamic programming and narrowing down the search space with a statistical model, transition-based parsers maintain only one (incomplete) analysis of the input sequence at a time. This is called a *greedy* strategy. They usually utilise a classifier trained on treebank-data to predict the next action when faced with a choice (Nivre, 2008). While this narrows down the search space to the one best scored action at each step, it has been shown that by using neural context-aware classifiers, this approach can yield state-of-the-art results.[12]

In the following subsections, I will first provide some necessary prerequisites and define transition-based parsing formally. Then, a short outline of the traditional projective constituent *shift-reduce* parser is given, followed by an exploration of two extensions of this approach for discontinuous constituent parsing: Swap and Gap.

### 3.2.1 Transition Systems

The following definitions for *transition systems* and *transition sequences* are based on Nivre (2008). They were introduced for dependency parsing but are, with small modifications, applicable to constituent parsing.

[12] See Section 6.3 for a comparison of results for discontinuous constituent parsing.

**Definition 3.9 (Transition system).**
Let $T$ be an alphabet of terminals and $N$ an alphabet of nonterminals. A *transition system wrt. $T$ and $N$* is a 5-tuple $\mathfrak{S} = \langle \mathcal{C}, \mathcal{T}, \sigma, F, d \rangle$, where

1. $\mathcal{C}$ is a set of *configurations*,
2. $\mathcal{T}$ is a set of transitions, which are (partial) functions $\tau : \mathcal{C} \to \mathcal{C}$,
3. $\sigma$ is an *initialisation function* that maps an input word to a configuration, i.e. $\sigma : T^* \to \mathcal{C}$,
4. $F \subseteq \mathcal{C}$ is a set of *terminal configurations*.
5. $d$ is an *output function* that maps a terminal configuration to a syntactic representation labelled over the signature $N$.

The use of the terms *terminal* and *nonterminal* is disconnected from any formal grammar here and only occurs to allude to their function as the input sequence alphabet and as the signature of the resulting tree.

Usually, a configuration possesses a *stack* $S$ that contains the constructed constituents and a *buffer* $B$ of remaining words. $S$ and $B$ can be thought of as ordered lists of elements where access is only allowed on one end. Only the top element can be retrieved. For $S$, a new element can be pushed onto the structure to constitute the new top while $B$ is read-only. Usually, $S$ is written with its top facing to the right while $B$ has its top on the left side. The symbol $|$ aids in marking the top element. $S|s$ denotes a stack where $s$ is the top element and $S$ is the (possibly empty) rest of the stack. $b|B$ denotes a buffer with $b$ as its top element. The initialisation function $\sigma$ returns an initial configuration where the input sentence words are contained in the buffer and where the stack is empty. The stack and the buffer can be written down in list notation $[x_1, ..., x_n]$ which is done to specify the initial buffer configuration. $[\,]$ denotes an empty structure.

The use of a stack and a buffer originates in *push-down automata* (PDA), a design of automata that is equivalent to CFGs (Hopcroft et al., 2007, chapter 6). The transitions can also be expressed as a formal deduction system. Then, $\sigma(w)$ for an input sequence $w$ serves as an axiom. The goal of the transition system is to construct a terminal configuration by means of transitions. Since the deductions have only one antecedent item, deduction sequences are purely linear. Therefore, the following notation differs from the inference relation given in definition 3.2 in that it is not closed under addition of antecedents.

**Definition 3.10 (Transition derivability relation).**
Let $T$ be an alphabet of terminals and $N$ an alphabet of nonterminals. Let $\mathfrak{S} = \langle \mathcal{C}, \mathcal{T}, \sigma, F, d \rangle$ be a transition system wrt. $T$ and $N$. Let $c, c' \in \mathcal{C}$ be configurations.

- The relation $c \underset{\tau}{\Rightarrow} c'$ holds iff $c' = \tau(c)$ for some $\tau \in \mathcal{T}$
  $c \underset{\tau}{\Rightarrow} c'$ reads: $c'$ is *directly derivable from $c$ using $\tau$.*
- The relation $c \Rightarrow c'$ holds iff there exists any transition $\tau \in \mathcal{T}$ such that $c \underset{\tau}{\Rightarrow} c'$.
  $c \Rightarrow c'$ reads: $c'$ is *directly derivable from $c$.*
- $\overset{*}{\Rightarrow}$ is the reflexive and transitive closure of $\Rightarrow$.
  $c \overset{*}{\Rightarrow} c'$ reads: $c'$ is *derivable from $c$.*

For $c_1 \Rightarrow c_2, c_2 \Rightarrow c_3$ we conventionally write $c_1 \Rightarrow c_2 \Rightarrow c_3$. This also holds for $\underset{\tau}{\Rightarrow}$ and $\overset{*}{\Rightarrow}$.

**Definition 3.11 (Transition sequence).**
Let $T$ be an alphabet of terminals and $N$ an alphabet of nonterminals. Let $\mathfrak{S} = \langle \mathcal{C}, \mathcal{T}, \sigma, F, d \rangle$ be a transition system and $w \in T^*$ a word. A *transition sequence for $w$ in $\mathfrak{S}$* is a sequence $c_{0:n} = c_1, c_2, ..., c_n$ of configurations with $n \in \mathbb{N}$, such that

1. $c_1 = \sigma(w)$,
2. $c_n \in F$,
3. $c_1 \Rightarrow c_2 \Rightarrow ... \Rightarrow c_n$

**Definition 3.12 (Valid configuration).**
Let $T$ be an alphabet of terminals and $N$ an alphabet of nonterminals. Let $\mathfrak{S} = \langle \mathcal{C}, \mathcal{T}, \sigma, F, d \rangle$ be a transition system wrt. $T$ and $N$ and $w \in T^*$ a word. A configuration $c \in \mathcal{C}$ is called *valid for w* iff there exists a transition sequence $c_{0:n}$ for $w$ in $\mathfrak{S}$ such that $c = c_i$ for some $i \in \mathbb{N}, 1 \leq i \leq n$.

**Definition 3.13 (Parse).**
Let $T$ be an alphabet of terminals and $N$ an alphabet of nonterminals. Let $\mathfrak{S} = \langle \mathcal{C}, \mathcal{T}, \sigma, F, d \rangle$ be a transition system wrt. $T$ and $N$ and $c_{0:n}$ a transition sequence for $w \in T^*$ in $\mathfrak{S}$. $d(c_n)$ is called the *parse* assigned to $w$ by $c_{0:n}$

Oracles act as a system for searching and predicting the next transition given a configuration. Nivre (2008) defines oracles in the following way:

**Definition 3.14 (Oracle).**
Let $T$ be an alphabet of terminals and $N$ an alphabet of nonterminals. Let $\mathfrak{S} = \langle \mathcal{C}, \mathcal{T}, \sigma, F, d \rangle$ be a transition system wrt. $T$ and $N$.

1. An *oracle o for* $\mathfrak{S}$ is a function $o : \mathcal{C} \to \mathcal{T}$.
2. A transition sequence $c_{0:n}$ for some word $w \in T^*$ in $\mathfrak{S}$ is *licensed by o* iff for every $i \in \{1, ..., n-1\}$ it holds that $c_{i+1} = (o(c_i))(c_i)$.

In data-driven parsing, the oracle has access to the gold trees of a given corpus and can base its predictions on them. A *scorer* is trained with the goal of approximating the oracle. It can take the form of a neural classifier and receives input features at each step that are associated with certain elements in the configuration.[13]

Given a transition system $\mathfrak{S} = \langle \mathcal{C}, \mathcal{T}, \sigma, F, d \rangle$ wrt. an alphabet of terminals $T$ and an alphabet of nonterminals $N$ and an oracle $o$ for $\mathfrak{S}$, the following simple algorithm for deterministic parsing is defined (Nivre, 2008):

**Algorithm 3.3** Parse($w$)

**Input:** a sentence $w \in T^*$
**Output:** a parse assigned to $w$

```
c ← σ(w)
while c ∉ F do
    c ← (o(c))(c)
end while
return d(c)
```

The benefit of this formal characterisation is the separation of oracles, transitions and configurations which facilitates the analysis of the complexity and other formal properties of the system. According to Nivre (2008) one can usually assume that $o(c)$ and $\tau(c)$ have constant time complexity for all transitions and configurations. Therefore, the complexity of the transition system is given by the worst-case maximum length of transition sequences for an input sentence. The space complexity is characterised by the maximum size of a configuration since the system only needs to store one configuration at a time.

The examination of correctness differs from that of grammar-based parsers. Correctness for a transition system is characterised not by a set of syntactic representations for a specific grammar

[13] A more detailed account of scorers is given in section 3.2.7.

(i.e. its tree language) but by a class of syntactic representations (e.g. the set of all possible labelled trees). Nivre (2008) gives the following definition of soundness and completeness which I slightly modified to account for the tree labelling.

**Definition 3.15 (Soundness, completeness, correctness).**
Let $T$ be an alphabet of terminals and $N$ an alphabet of nonterminals. Let $\mathfrak{S} = \langle \mathcal{C}, \mathcal{T}, \sigma, F, d \rangle$ be a transition system wrt. $T$ and $N$.

1. $\mathfrak{S}$ is *sound* for a class $\mathbb{D}$ of syntactic representations, iff for every word $w \in T^*$ and every transition sequence $c_{0:n}$ for $w$ in $\mathfrak{S}$, it holds that $d(c_n) \in \mathbb{D}$.
2. $\mathfrak{S}$ is *complete* for a class $\mathbb{D}$ of syntactic representations iff for every syntactic representation $D_w \in \mathbb{D}$ for some word $w$, there is a transition sequence $c_{0:n}$ for $w$ in $\mathfrak{S}$ such that $d(c_n) = D_w$.
3. $\mathfrak{S}$ is *correct* for a class $\mathbb{D}$ of syntactic representations iff it is sound and complete for $\mathbb{D}$.

For the class of range-labelled trees correctness of the transition system ensures that every tree can be built using the system and that every syntactic representation built is indeed formally a range-labelled tree. Additionally, to match the notion of parser correctness given in definition 3.7, it needs to be proven that the optimal steps predicted by the oracle lead indeed to the desired tree for an input.

Following Goldberg and Nivre (2012a) one can define oracle correctness as follows:

**Definition 3.16 (Oracle correctness).**
Let $T$ be an alphabet of terminals, $N$ an alphabet of nonterminals and $\mathfrak{S} = \langle \mathcal{C}, \mathcal{T}, \sigma, F, d \rangle$ be a transition system wrt. $T$ and $N$ that is correct for a class $\mathbb{D}$ of syntactic representations and $o : \mathcal{C} \to \mathcal{T}$ an oracle for $\mathfrak{S}$.
$o$ is *correct* iff for every word $w \in T^*$ and a syntactic representation $D_w \in \mathbb{D}$ for $w$ the transition sequence $c_{0:n}$ for $w$ licensed by $o$ assigns exactly the parse $d(c_n) = D_w$.

Note that a proof of correctness can only be done for the oracle and not for the final model trained to approximate the oracle. Its properties rely crucially on the quality of the neural model and its learning algorithm.

### 3.2.2 Comparing Chart Parsing and Transition-Based Parsing

In this section I will point out several differences between chart parsers and transition-based parsers. In chart parsers the items are intermediate parsing results which represent a partial analysis. One or more items can be used to deduce a new analysis for some part of the input. In a transition-based system the items are configurations that in turn contain all available partial analyses for the input. They can be seen as a data structure that represents the state of analysis at a certain time-step, comparable to a view of chart and agenda. But unlike in chart-parsing, analyses cannot conflict with each other.

A deduction can be used to alter the inner structure of the item that restricts which elements are available for derivation (e.g. shift an element from the buffer to the stack) or to derive a new partial analysis from certain elements in the item. When deriving a new partial analysis from elements using a transition, they are “used up” and no longer available for future derivation steps. Thus, by design there is no derivational ambiguity in the retrieved analysis. This reduces the formalism’s complexity. The parsing process works bottom-up and the transition system can be seen as a framework to keep track of active partial analyses which have not been used in a derivation yet.

While both approaches are usually data-driven, the weighted CYK chart parser works with an explicitly formalised grammar that was extracted from a treebank in a preceding step. Weights are assigned to productions based on their observed frequency of occurrence. Transition-based parsers are generally built on a neural network framework. They are trained on a treebank and learn to replicate the patterns found in it. The transition system specifies the order and manner in which this replication takes place (e.g. bottom-up retrieval of derivation steps, left-to-right processing). Thus, one can view the model as implicitly learning a weighted grammar that licenses the treebank. Adopting this view, the model can be said to assign the highest score to the chain of actions that leads to the analysis with the best overall weight.

Note however that while a chart-parser based on an extracted grammar can only apply derivations that were observed during the extraction process on a corpus, a neural approach yields probabilities for all derivations allowed by the transition system including those that it did not observe during training. This has two consequences: on the one hand, it greatly expands the field of potential parsing errors far beyond the scope of derivational ambiguity inherent to grammar-based approaches. On the other hand, it enables the parser to generalise over constituents that exhibit a similar behaviour and to predict the correct action if faced with a context that is only attested for a portion of these constituents in the training data.

In contrast to the production weights used in weighted chart parsing, the likelihood of a production in a transition-based neural model is not independent of surrounding items. A prediction is computed using a complex neural function based on features like items present on the stack and the buffer or previously derived items. This mitigates the disadvantage that arises from pursuing a greedy 1-best strategy. The parameters of this function are adjusted during training to produce higher scores for derivations found in the gold derivation trees. Improbable derivations are penalised by the cost-function of the learning algorithm.

### 3.2.3 Standard Shift-Reduce Projective Parsing

Shift-reduce parsing is a bottom-up left-to-right parsing method for CFGs. It serves as the basis for most transition-based parsers. Here, a configuration is a tuple $\langle S, B \rangle$ with a stack $S$ and a buffer $B$. The configuration is visualised in figure 3.5.

The stack consists of derived constituents usually characterised by a nonterminal and a tuple of pointers to its daughters. The goal is to construct a single item via transitions from which an ordered derivation tree can be recovered. This mirrors the use of backpointers in charts.

**Definition 3.17 (Recursive labelled ordered tree).**
Let $X$ be an alphabet of labels.

1. A symbol $A \in X$ is a *recursive ordered tree labelled over signature* $X$, more specifically a leaf.
2. A tuple $\langle A_1, \mathbf{D}_1 \rangle$ is a recursive ordered tree labelled over signature $X$ if
   (a) $A_1 \in X$,
   (b) $\mathbf{D}_1$ is a tuple of recursive ordered trees labelled over signature $X$ and
   (c) there is no cyclic sequence of recursive ordered trees $\langle A_1, \mathbf{D}_1 \rangle, \langle A_2, \mathbf{D}_2 \rangle, ..., \langle A_n, \mathbf{D}_n \rangle$ where $\mathbf{D}_i^{[j]} = \langle A_{i+1}, \mathbf{D}_{i+1} \rangle$ for some $j \in \{1, ..., |\mathbf{D}_i|\}$ for all $i \in \{1, ..., n-1\}$ and $\langle A_n, \mathbf{D}_n \rangle = \langle A_1, \mathbf{D}_1 \rangle$.
3. Nothing else is a recursive ordered tree labelled over signature $X$.

The notation in definition 3.17 is an alternative to the traditional graph notation $\langle V, \triangleleft, \prec, \hat{v}, \lambda \rangle$ of an ordered tree with a node labelling over signature $X$ introduced in section 2.1. It is the common way to implement trees as a data structure. A recursive ordered tree $\langle A, \mathbf{D} \rangle$ can be likewise treated as an ordered tree with a root node that is labelled with $A$ and has daughters defined by the subtrees in $D$ for which precedence is determined by their position in $\mathbf{D}$. In the following, I will use the two notations interchangeably.

The transition system of Sagae and Lavie (2005) is used as the foundation for many transition-based constituent parsers. They work with binarised trees with headedness information.[14] Coavoux and Crabbé (2017a) give the following reduced set of transitions for unlexicalised parsing:

1. Shift moves the first element from the buffer onto the stack.
2. ReduceUnary-X creates a new constituent, removes the top element from the stack, assigns it as the single daughter of the new constituent and puts the new constituent onto the stack.
3. Reduce-X creates a new constituent, removes the two topmost elements from the stack, assigns them as the daughters of the new constituent and puts the new constituent onto the stack.

The formalisations as deductions are presented in figure 3.4. Cross and Huang (2016b) use a similar set of transitions that allows non-binary trees.

$$\begin{array}{rl}
\textit{axiom} & \dfrac{}{\langle\ [\ ],\ [w_1, ..., w_n]\ \rangle} \\
\textsc{Shift} & \dfrac{\langle\ S,\ w_i|B\ \rangle}{\langle\ S|w_i,\ B\ \rangle} \\
\textsc{Reduce-X} & \dfrac{\langle\ S|s_1|s_0,\ B\ \rangle}{\langle\ S|\langle X, \langle s_1, s_0\rangle\rangle,\ B\ \rangle} \\
\textsc{ReduceUnary-X} & \dfrac{\langle\ S|s_0,\ B\ \rangle}{\langle\ S|\langle X, \langle s_0\rangle\rangle,\ B\ \rangle} \\
\textit{goal} & \langle\ [\textit{ROOT}],\ [\ ]\ \rangle
\end{array}$$

Figure 3.4: Shift-reduce as deductions, $\mathcal{T}^{\mathrm{SR}} = \{\textsc{Shift}, \textsc{Reduce-X}, \textsc{ReduceUnary-X}\}$.

**Definition 3.18 (Shift-reduce transition system).**
A Shift-Reduce transition system for an alphabet of terminals $T$ and an alphabet of nonterminals $N$ is a 5-tuple $\mathfrak{S}^{\mathrm{SR}} = \langle \mathcal{C}^{\mathrm{SR}}, \mathcal{T}^{\mathrm{SR}}, \sigma, F, d\rangle$ where

1. $\mathcal{C}^{\mathrm{SR}} = \{\langle S, B\rangle \mid \forall c \in S \cup B : c$ is a binary ordered tree with leaf labels from $T$ and internal labels from $N\}$,[15]
2. $\mathcal{T}^{\mathrm{SR}}$ is the set of transitions defined for $\mathcal{C}^{\mathrm{SR}}$ as given in figure 3.4,
3. $\sigma(w_{0:n}) = \langle [\ ], [w_1, ..., w_n]\rangle$ for every $w_{0:n} \in T^+$,
4. $F = \{\langle [\mathrm{ROOT}], [\ ]\rangle \mid \mathrm{ROOT}$ is a binary orderd tree with leaf labels from $T$ and internal labels from $N\}$ and
5. $d$ gives the sole element in $S$ for every $\langle S, B\rangle \in F$.

**Example 3.19.**
Figure 3.6 shows an example transition sequence for the sentence *the man took the book* with the derivation tree depicted in figure 1.1.

[14] A discussion on the matter of head-information and lexicalisation is given in section 3.2.6.
[15] In a slight abuse of notation, set operators are used here to refer to all elements in the two sequences.

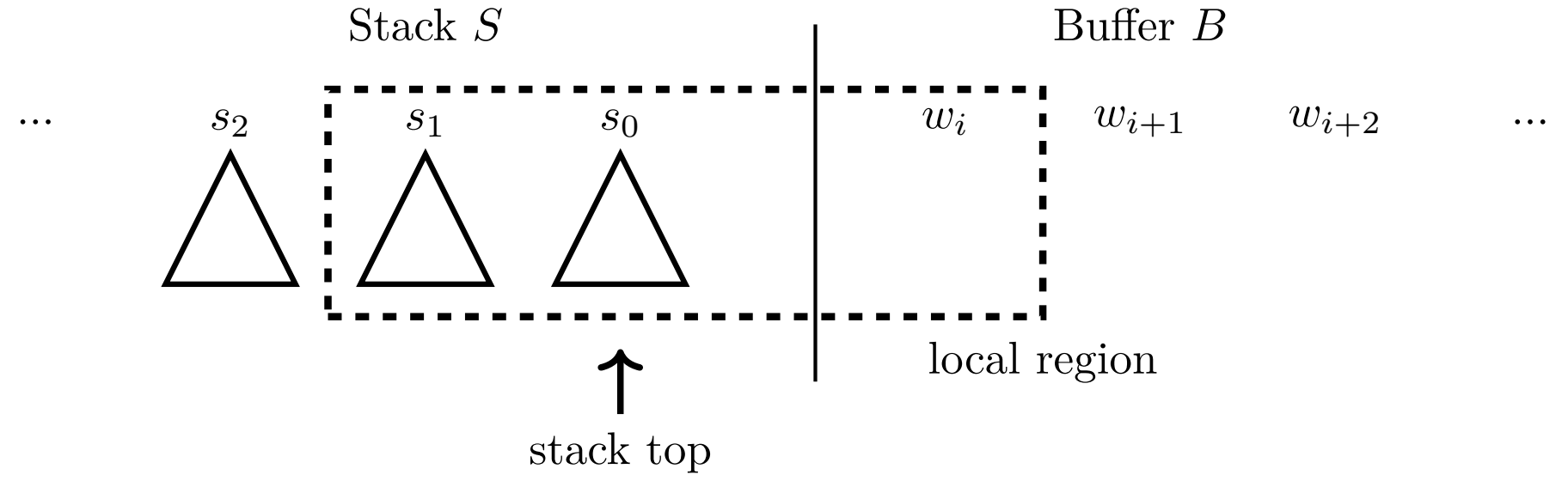

Figure 3.5: Illustration of a configuration in $\mathcal{C}^{\mathrm{SR}}$. Transitions involve changes to the local part of the configuration (dashed line). Illustration reproduced from Coavoux and Cohen (2019).

| Step | Stack | Buffer | Action |
|---|---:|---|---|
| 0 | | the, man, took, the, book | SHIFT |
| 1 | the | man, took, the, book | SHIFT |
| 2 | the, man | took, the, book | REDUCE-NP |
| 3 | $\langle$NP, $\langle\bullet,\bullet\rangle\rangle$ | took, the, book | SHIFT |
| 4 | $\langle$NP, $\langle\bullet,\bullet\rangle\rangle$, took | the, book | REDUCEUNARY-Verb |
| 5 | $\langle$NP, $\langle\bullet,\bullet\rangle\rangle$, $\langle$Verb, $\langle\bullet\rangle\rangle$ | the, book | SHIFT |
| 6 | $\langle$NP, $\langle\bullet,\bullet\rangle\rangle$, $\langle$Verb, $\langle\bullet\rangle\rangle$, the | book | SHIFT |
| 7 | $\langle$NP, $\langle\bullet,\bullet\rangle\rangle$, $\langle$Verb, $\langle\bullet\rangle\rangle$, the, book | | REDUCE-NP |
| 8 | $\langle$NP, $\langle\bullet,\bullet\rangle\rangle$, $\langle$Verb, $\langle\bullet\rangle\rangle$, $\langle$NP, $\langle\bullet,\bullet\rangle\rangle$ | | REDUCE-VP |
| 9 | $\langle$NP, $\langle\bullet,\bullet\rangle\rangle$, $\langle$VP, $\langle\bullet,\bullet\rangle\rangle$ | | REDUCE-Sentence |
| 10 | $\langle$Sentence, $\langle\bullet,\bullet\rangle\rangle$ | | |

Figure 3.6: Shift-Reduce parse for *the man took the book* producing the tree from Figure 1.1.

For an input and a parse tree to derive, there exists only one transition sequence. If the topmost element's root $v_0$ is the unary daughter of another constituent or if $v_0$ and the second topmost element's root $v_1$ represent a gold constituent, a reduction must take place. If not, SHIFT must be performed. Shifting when elements are reducible would move a new item onto the stack making $v_0$ inaccessible. From this step on, the top of the stack can only be occupied by elements succeeding $v_0$ in the precedence relation. This minimal ambiguity makes deriving an oracle a trivial task.

The elements in a valid configuration for a word $w$ represent a forest of derivation trees that are subtrees of some possible tree $D_w$ for $w$. The roots of the forest elements are the surface elements on the stack and in the buffer. They represent a *tree-cut* of $D_w$. A REDUCE transition is a transition from a tree-cut $v_1, ..., v_i, v_{i+1}, ..., v_m$ to another tree-cut $v_1, ..., v_p, ..., v_m$ where $v_p$ is the parent of $v_i, v_{i+1}$. This illustrates why no two conflicting analyses of constituents can co-occur in this framework. Items dominated by an element in the current tree-cut are not present at surface-level in the configuration and cannot be used to compose alternate analyses.

The following definition of a tree-cut is based on Versley (2014).

**Definition 3.20 (Tree-cut).**
Let $D = \langle V, \triangleleft, \hat{v} \rangle$ be a tree. A *tree-cut in D* is a sequence $v_1, ..., v_n$ with $v_k \in V$ for all $k \in \{1, ..., n\}$ for some $n \in \mathbb{N}$ such that for all paths from $\hat{v}$ to a leaf, exactly one node from the path is contained in the sequence.

**Definition 3.21 (Partial parse).**
Let $T$ be an alphabet of terminals and $w \in T^*$ a word. A *partial parse* for $w$ is an ordered forest $D_{0:n}$ where there exists some tree $D_w$ for $w$ such that the roots of $D_1, ..., D_n$ are a tree-cut in $D_w$ and such that $D_1, ..., D_n$ are subtrees of $D_w$.

**Example 3.22.**
Figure 3.7 shows two possible tree-cuts in a tree. These demonstrate the reduction of two nodes in a tree-cut: Verb and NP in the left tree and their parent VP in the right tree.

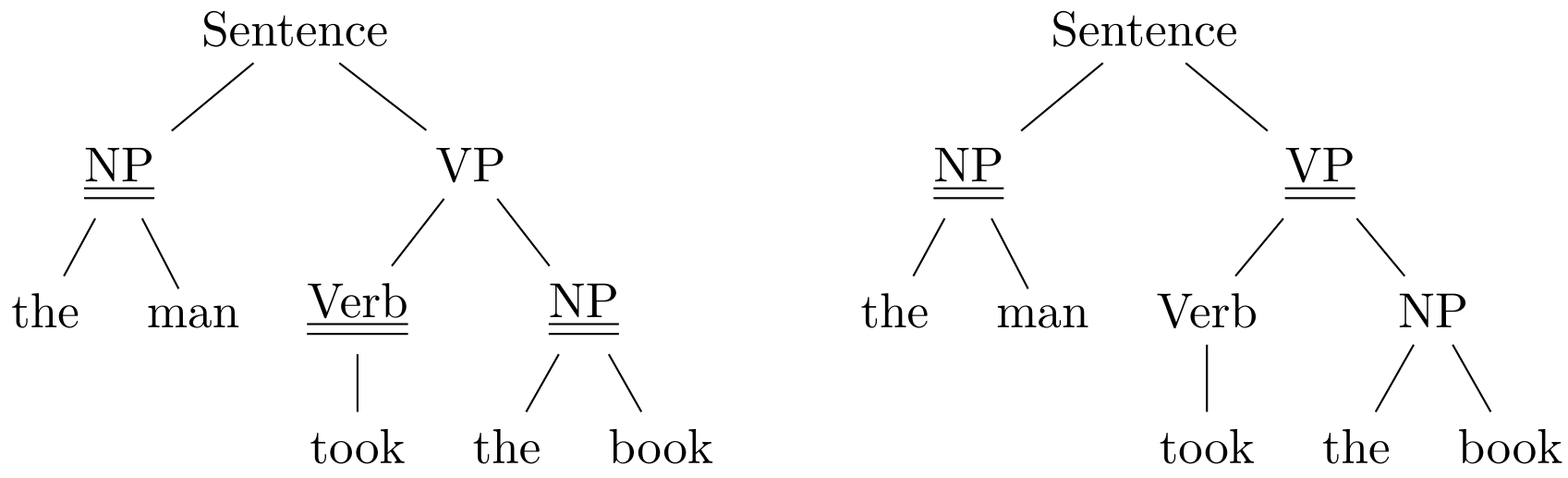

Figure 3.7: Two possible tree-cuts (underlined) in a projective tree. The tree is taken from Chomsky (1956).

Note that this set of transitions cannot predict $\varepsilon$-productions. There is no deduction rule to predict empty leaves between elements or to reduce them. Therefore, the initialisation function also does not accept the empty word. One could add a configuration with empty stack and buffer to the set of terminal configurations to enable the latter but this shall not be of any concern here as the corpora at hand do not feature empty leaves.

What is significant, however, is this system's inability to deduce crossing edges. It can only reduce adjacent elements into constituents. Several strategies were developed to solve this problem: new transitions like Swap and Gap reorder items and make discontinuous parts of a constituent adjacent in order to reduce them. In the following, I will discuss these approaches in more detail.

### 3.2.4 Reordering with the Swap Action

Versley (2014) was the first to adapt the Swap reordering action from dependency parsing to constituent parsing. The underlying idea is that for every non-projective tree, there is a projective tree that can be constructed by reordering the input words. This projective tree can be parsed with traditional methods. Swap can be used to change the order of two elements and thus allows for out-of-order processing. In the following, this approach will be referred to as Shift-Reduce-Swap or SR-Swap.

Versley (2014) uses an *easy-first* approach inspired by Goldberg and Elhadad (2010) where a parsing action or a Swap action can occur anywhere in an input string based on the prediction of a scorer. Maier (2015) adapted the Swap action to the shift-reduce transition system. I will outline this approach in more detail.

Discontinuous trees are not ordered (cf. definition 2.29). Only the leaves have the range precedence relation $<_{\text{range}}$ defined on them that describes the surface order of the input tokens. Therefore, we need to modify the definition of a transition constituent for discontinuous constituent parsing. The list of daughters for a constituent no longer needs to be ordered and can therefore be defined as a set.

**Definition 3.23 (Recursive unordered tree).**
Let $X$ be a set of labels.

1. A symbol $A \in X$ is a *recursive unordered tree labelled over signature* $X$, more specifically a leaf,

2. A tuple $\langle A_1, \mathcal{D}_1 \rangle$ is a recursive unordered tree labelled over signature $X$ if
   (a) $A_1 \in N$,
   (b) $\mathcal{D}_1$ is a set of recursive unordered trees labelled over signature $X$ and
   (c) there is no cyclic sequence of recursive unordered trees $\langle A_1, \mathcal{D}_1 \rangle, \langle A_2, \mathcal{D}_2 \rangle, ..., \langle A_n, \mathcal{D}_n \rangle$ such that $\langle A_{i+1}, \mathcal{D}_{i+1} \rangle \in \mathcal{D}_i$ for $i \in \{1, ..., n-1\}$ and $\langle A_n, \mathcal{D}_n \rangle = \langle A_1, \mathcal{D}_1 \rangle$.
3. Nothing else is a recursive unordered tree labelled over signature $X$.

This notation is equivalent to the graph-based definition $\langle V, \triangleleft, \hat{v}, \lambda \rangle$ of an unordered labelled tree and the two notations will be used interchangeably within the following paragraphs. The Reduce-X and ReduceUnary-X transitions are modified accordingly to reflect the new notation. Furthermore, the parser introduced in Maier (2015) also marks the lexical head of a constituent. I omit this aspect for reasons of brevity.

$$\textsc{Swap} \quad \frac{\langle\ S|s_1|s_0,\ \ B\ \rangle}{\langle\ S|s_0,\ \ s_1|B\ \rangle}$$

Figure 3.8: Swap transition.

The central innovation is the introduction of the Swap transition shown in figure 3.8. Using this transition, it is possible to derive discontinuous constituents by reordering items such that for two elements that have to be reduced, every element in between that blocks this reduction by not being a sister node is moved to a different position. The SR-Swap transition system is formally defined as follows:

**Definition 3.24 (SR-Swap transition system).**
A Shift-Reduce-Swap transition system for an alphabet of terminals $T$ and an alphabet of nonterminals $N$ is a 5-tuple $\mathfrak{S}^{\text{SRS}} = \langle \mathcal{C}^{\text{SRS}}, \mathcal{T}^{\text{SRS}}, \sigma, F, d \rangle$ where

1. $\mathcal{C}^{\text{SRS}}$ is the set of all configurations $\langle S, B \rangle$ composed of partial $\varepsilon$-free binary range-labelled trees over $N$,
2. $\mathcal{T}^{\text{SRS}} = \{\textsc{Shift}, \textsc{Reduce-X}, \textsc{ReduceUnary-X}, \textsc{Swap}\}$ the set of transitions defined on $\mathcal{C}^{\text{SRS}}$,
3. $\sigma(w_{0:n}) = \langle [\ ], [\langle 0, 1 \rangle, ..., \langle n-1, n \rangle] \rangle$ for every $w_{0:n} \in T^+$,
4. $F = \{\langle [\text{ROOT}], [\ ] \rangle \mid \text{ROOT}$ is an $\varepsilon$-free binary range-labelled tree over $N\}$ and
5. $d$ retrieves the tree from the sole element in $S$ for every $\langle S, B \rangle \in F$.

**Example 3.25.**
Figure 3.9 shows an example transition sequence for the sentence *Darüber muß nachgedacht werden* and the derivation tree depicted in figure 1.3.

It is straightforward to see that one can apply different transitions to retrieve the same result. More specifically, in figure 3.9 a Swap at step 2 would have already brought the configuration into a form where reducing the tree-cut would be possible with the standard set of transitions. Indeed, the possibility to put items from the stack back onto the buffer leads to significant ambiguity and even allows for looping behaviour by alternating between Shift and Swap transitions. This poses a challenge for defining oracles that lead to an optimally short transition sequence.

| Step | Stack | Buffer | Action |
|---|---|---|---|
| 0 | | Darüber$_1$, muß$_2$, nachgedacht$_3$, werden$_4$ | Shift |
| 1 | Darüber$_1$, | muß$_2$, nachgedacht$_3$, werden$_4$ | Shift |
| 2 | Darüber$_1$, muß$_2$, | nachgedacht$_3$, werden$_4$ | Shift |
| 3 | Darüber$_1$, muß$_2$, nachgedacht$_3$, | werden$_4$ | Swap |
| 4 | Darüber$_1$, nachgedacht$_3$ | muß$_2$, werden$_4$ | Reduce-VP |
| 5 | $\langle$VP, $\{\bullet, \bullet\}\rangle$ | muß$_2$, werden$_4$ | Shift |
| 6 | $\langle$VP, $\{\bullet, \bullet\}\rangle$, muß$_2$ | werden$_4$ | Shift |
| 7 | $\langle$VP, $\{\bullet, \bullet\}\rangle$, muß$_2$, werden$_4$ | | Swap |
| 8 | $\langle$VP, $\{\bullet, \bullet\}\rangle$, werden$_4$ | muß$_2$ | Reduce-VP |
| 9 | $\langle$VP, $\{\bullet, \bullet\}\rangle$ | muß$_2$ | Shift |
| 10 | $\langle$VP, $\{\bullet, \bullet\}\rangle$, muß$_2$ | | Reduce-S |
| 11 | $\langle$S, $\{\bullet, \bullet\}\rangle$ | | |

Figure 3.9: Example parse using Swap transition for *Darüber muß nachgedacht werden* with tree shown in figure 1.3, POS tags omitted, indices given alongside tokens instead of ranges.

**Swap Loops** Several suggestions have been made for restricting the application of transitions to reduce ambiguity. A straightforward restriction is to prohibit Swap loops by ensuring that two elements cannot be swapped twice. For this, Maier and Lichte (2016) define the following order.

**Definition 3.26 (Indices of ranges).**

1. $Ind_{\text{range}} : Ranges \to \mathcal{P}(\mathbb{N})$ is defined as $\langle a, b\rangle \mapsto \{a+1, ..., b\}$.
2. Let $N$ be an alphabet. Let $D = \langle V, \triangleleft, \hat{v}, \lambda\rangle$ be a range-labelled tree over $N$. $Ind_D : V \to \mathcal{P}(\mathbb{N})$ gives the set of all indices dominated by a node:

$$v \mapsto \{i \mid \exists u \in V : v \triangleleft^* u, u \text{ is a leaf}, i \in Ind_{\text{range}}(\lambda(u))\}.$$

**Definition 3.27 (Leftmost index ordering).**
Let $D_{0:n}$ be a partial parse for $w$ with $D_i = \langle V_i, \triangleleft, \hat{v}_i, \lambda\rangle$ for every $i \in \{1, ..., n\}$. $<_{\text{ind}}$ is defined as a strict total order on the elements of $D_{0:n}$:

$$D_i <_{\text{ind}} D_j :\Leftrightarrow \min(Ind_{D_i}(\hat{v}_i)) < \min(Ind_{D_j}(\hat{v}_j))$$

Now, for two trees on the stack $s_1, s_0$ Swap is only allowed if they satisfy $s_1 <_{\text{ind}} s_0$ as shown in the modified deduction rule in figure 3.10. This is the case for all elements in the initial order. If Swap is performed, this changes the order of $s_1, s_0$ in the configuration. Since $s_0 <_{\text{ind}} s_1$ cannot be fulfilled, Swap cannot be performed a second time for these two elements.

$$\text{Swap} \quad \frac{\langle\ S|s_1|s_0,\ B\ \rangle}{\langle\ S|s_0,\ s_1|B\ \rangle}\ s_1 <_{\text{ind}} s_0$$

Figure 3.10: Swap transition standard condition.

**Eager Swap** Nivre (2009) and Nivre et al. (2009) outline two strategies for reducing transition sequence lengths for Swap in dependency parsing: *eager swap* and *lazy swap*. They were adapted for constituent parsing by Versley (2014). These are conditions concerning oracle extraction. The oracle has access to the gold tree and is constructed to yield only partial analyses of the gold tree.

Therefore, in the following, trees in the deduction rules are simply treated as their roots, i.e. as nodes in the gold tree.

The idea of the SWAP transition is to convert the input sequence into a projective order, or in other words, to an order where the target derivation tree can be recovered by reducing only adjacent elements. Therefore eager swap restricts the SWAP transition to cases where the two elements in consideration are not in a projective order. It prohibits swapping when a SWAP would be unnecessary to retrieve the tree. The quest of Versley (2014) is to define a desirable order for every tree-cut that allows to parse the elements projectively. For this, he defines the following property.

**Definition 3.28 ($\triangleleft$-compatibility).**
Let $D = \langle V, \triangleleft, \hat{v}\rangle$ be a tree. A node ordering $\prec \subseteq V \times V$ defined on all pairs of nodes that co-occur in some tree-cut of $D$ is called $\triangleleft$-*compatible* iff for $v, v' \in V$ with $v \triangleleft^* v'$ and $u, u' \in V$ with $u \triangleleft^* u'$ it holds that $u \prec v \Rightarrow u' \prec v'$.

Note that this is equivalent to conditions 3.(b) and 3.(c) for the linear precedence relation of an ordered tree given in definition 2.10.

Versley (2014) shows that sorting the leaves of a tree in a $\triangleleft$-compatible order $\prec$ makes it possible to find a derivation using standard transition sets for projective parsing. This stems from the observation that reducing a $\prec$-ordered tree-cut yields another $\prec$-ordered tree-cut. Thus, no crossing edges are encountered and no reductions of non-adjacent elements are necessary.

**Corollary 3.29.**
Let $\langle V, \triangleleft, \hat{v}\rangle$ be a tree, $\prec$ a $\triangleleft$-compatible order and $v_1, ..., v_n$ a $\prec$-ordered tree-cut. Reducing some elements $v_i, v_{i+1}$ with $i \in \{1, ..., n-1\}$ to $p \in V$ with $p \triangleleft v_i$ and $p \triangleleft v_{i+1}$ results in another $\prec$-ordered tree cut $v_1, ..., v_{i-1}, p, v_{i+2}, ..., v_n$.

*Proof.* Assumption: $v_1, ..., v_{i-1}, v_i, v_{i+1}, v_{i+2}, ..., v_n$ is $\prec$-ordered and there is $p \in V$ such that $p \triangleleft v_i, p \triangleleft v_{i+1}$. In particular this means that $v_{i-1} \prec v_i \prec v_{i+1} \prec v_{i+2}$. If $p \prec v_{i-1}$ then $\triangleleft$-compatibility of $\prec$ demands that also $v_i \prec v_{i-1}$ which contradicts the assumption. Therefore, it must hold that $v_{i-1} \prec v_i$. Likewise, if $v_{i+2} \prec p$ then per $\triangleleft$-compatibility of $\prec$ it would also hold that $v_{i+2} \prec v_{i+1}$ which contradicts the assumption. Therefore $p \prec v_{i+2}$. Thus, $v_1, ..., v_{i-1}, p, v_{i+2}, ..., v_n$ is $\prec$-ordered. □

A generalisation to non-binary trees is straightforward. The precedence order $<_{\text{range}}$ on leaves can be naturally extended to a partial $\triangleleft$-compatible order for internal nodes in the following way (Versley, 2014):

$$v_1 <_{\text{N}} v_2 :\Leftrightarrow \forall u_1, u_2 \in V : (v_1 \triangleleft^* u_1 \wedge v_2 \triangleleft^* u_2 \wedge u_1 \text{ and } u_2 \text{ are leaves}) \Rightarrow u_1 <_{\text{range}} u_2 \tag{3.3}$$

Note however that this does not suffice for nonterminals $v_1, v_2$ with intertwined leaves since in such a case neither $v_1 <_{\text{N}} v_2$ nor $v_2 <_{\text{N}} v_1$. To solve this issue, one can instead extend $<_{\text{range}}$ to a $\triangleleft$-compatible order $\prec$ by defining a standard descendent for internal sister nodes to base the order on. This way, $<_{\text{N}}$ is respected locally, i.e. for two sister nodes $v_1, v_2$ it holds that $v_1 <_{\text{N}} v_2$ entails $v_1 \prec v_2$.

Versley (2014) works with a lexicalised model. His trees provide a function $h : V \to V$ that gives the head daughter for each node. Therefore he suggests recursively assigning the head leaf to an internal node:

$$h^*(v) := \begin{cases} v & \text{if } v \text{ is a leaf,} \\ h^*(h(v)) & \text{otherwise.} \end{cases} \tag{3.4}$$

Now, the order $<_h$ defined on all tree-cuts can be characterised as follows:

$$v_1 <_h v_2 \quad :\Leftrightarrow \begin{cases} \text{if } \exists p \in V : p \lhd v_1, p \lhd v_2 \text{ and } h^*(v_1) <_{\text{range}} h^*(v_2) \text{ or} \\ \text{if } \exists v_1', v_2' \in V : v_1' \lhd^* v_1, v_2' \lhd^* v_2 \text{ and } v_1' <_h v_2'. \end{cases} \tag{3.5}$$

The actual choice of leaf is arbitrary as Versley (2014) notes. One could likewise use leftmost index ordering as the base:

$$v_1 <_{\text{G}} v_2 \quad :\Leftrightarrow \begin{cases} \text{if } \exists p : p \lhd v_1, p \lhd v_2 \text{ and } v_1 <_{\text{ind}} v_2 \text{ or} \\ \text{if } \exists v_1', v_2' : v_1' \lhd^* v_1, v_2' \lhd^* v_2 \text{ and } v_1' <_{\text{G}} v_2'. \end{cases} \tag{3.6}$$

Maier (2015) extracts a projective ordering by use of *post-order traversal* on the tree. Post-order traversal on an ordered binary tree retrieves a sequence of all nodes by starting with the root node and for every node recursively traversing 1) through the left daughter, 2) through the right daughter and 3) appending the node itself to the sequence. In an ordered tree, the local order of node children is provided with the tree. Here, leftmost index ordering $<_{\text{ind}}$ is used instead.

Algorithm 3.4 shows how to extract a post-order sequence from a tree. $<_{\text{G}}$ is then defined by precedence in the extracted sequence. It can be seen that $<_{\text{G}}$ as defined by equation 3.6 entails the relation extracted from algorithm 3.4. Note however that algorithm 3.4 also defines an order for nodes that stand in the dominance relation. This does not influence parsing since two such nodes never co-occur in a valid parsing configuration.

**Algorithm 3.4** PostOrder($r$)

**Input:** a binary tree identified by its root node $r$
**Output:** a sequence of nodes

```
if r is leaf then
    return [r]
end if
v1, v2 ← children(r)
if v1 <ind v2 then
    return PostOrder(v1) ∘ PostOrder(v2) ∘ [r]
else
    return PostOrder(v2) ∘ PostOrder(v1) ∘ [r]
end if
```

Maier (2015) uses a strict eager strategy based on binary trees that demands that reductions take place according to the elements' post-ordering. For the topmost element $s_0$ of the stack, the oracle checks if the first element of the buffer is the direct projective successor. If not, Shift gets predicted until the direct successor $b_i$ is at the top of the stack. Then, all the elements that were shifted onto the stack located between $s_0$ and $b_i$ get swapped back to the buffer. This forces a left-to-right completion of the projective tree. Since this is the only allowed case for a Swap transition, it follows that the buffer consists only of leaves.

**Example 3.30.**
Figure 3.11 visualises the ordering retrieved by a post-order traversal on a discontinuous tree. Swapping $w_1$ and $w_2$ would be permissible by condition $<_{\text{ind}}$ but not by $<_{\text{G}}$. Figure 3.12 shows part of a transition sequence for the tree using the strict eager strategy of Maier (2015).

Although this approach is straightforward, one can see that it is not optimal. It predicts Swap transitions in cases where completing constituents first would result in shorter derivations. Here, reducing the intermediary subtree rooted at B would decrease the transition count.

A more flexible approach is taken by Stanojević and Alhama (2017) who perform online re-ordering. This is permissible since the $<_{\text{G}}$-relationships of nodes are preserved under Reduce and UNARAYREDUCE transitions due to $\lhd$-compatibility. Algorithm 3.5 extracts the oracle.[16] It demands that nonterminals be allowed to be swapped onto the buffer.

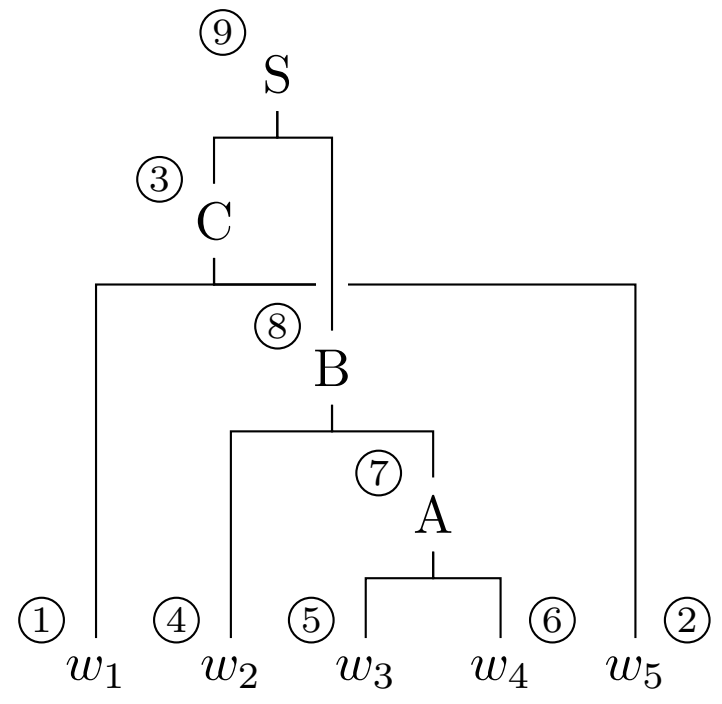

Figure 3.11: Example for projective ordering by post-order traversal.

| Step | Stack | Buffer | Action |
|---|---|---|---|
| 0 | | $w_1, w_2, w_3, w_4, w_5$ | SHIFT |
| 1 | $w_1$ | $w_2, w_3, w_4, w_5$ | SHIFT |
| 2 | $w_1, w_2$ | $w_3, w_4, w_5$ | SHIFT |
| 3 | $w_1, w_2, w_3$ | $w_4, w_5$ | SHIFT |
| 4 | $w_1, w_2, w_3, w_4$ | $w_5$ | SHIFT |
| 5 | $w_1, w_2, w_3, w_4, w_5$ | | SWAP |
| 6 | $w_1, w_2, w_3, w_5$ | $w_4$ | SWAP |
| 7 | $w_1, w_2, w_5$ | $w_3, w_4$ | SWAP |
| 8 | $w_1, w_5$ | $w_2, w_3, w_4$ | REDUCE-C |
| 9 | $\langle \mathrm{C}, \{\bullet, \bullet\}\rangle$ | $w_2, w_3, w_4$ | ... |
| ... | ... | ... | ... |

Figure 3.12: Example parse using the eager swap strategy outlined in Maier (2015) for the tree in figure 3.11.

Now, the oracle function returns the immediate successor of a configuration $c$ in $O$. This oracle prioritises the completion of substructures that are already in projective order. The swapAllowed condition is given in equation 3.7.

$$\text{swapAllowed}(\langle S|s_1|s_0, B\rangle) :\Leftrightarrow s_0 <_{\mathrm{G}} s_1 \tag{3.7}$$

Note however that the condition does not benefit parsing of figure 3.13. In this case, $<_{\mathrm{G}}$ blocks the more efficient SWAP of $\text{Darüber}_1$ and $\text{muß}_2$ since $\text{Darüber}_1 <_{\mathrm{G}} \text{muß}_2$. This pinpoints at the fact that the projective ordering that leads to a minimal number of transitions is dependent on the form of the individual tree. Here, a rightmost index ordering would lead to a smaller transition count. Generally, ordering by median leaf index or by start index of the largest consecutive span of the constituent could be used to make the parser complete constituents in an optimal place to reduce the number of SWAP transitions. To my knowledge, this has not been explored yet.

**Lazy Swap** Eager swap can lead to suboptimal transition sequence lengths when it swaps items into groups of nodes that could first be reduced projectively without the need for further swapping. This leads to chains of SWAP applications. Take for instance the tree in figure 3.14. Equation 3.8

[16] I adapted the algorithm to the unlexicalised transition set defined for binary trees in section 3.2.3.

**Algorithm 3.5** Oracle extraction

```
Input: a word w_1, ..., w_n, a tree D over w
Output: an oracle O
c ← ⟨ [ ], [w_1, ...w_n] ⟩
O ← List[ ]
while c ≠ goal do
    if Reduce-X(c) ∈ D then
        c ←Reduce-X(c)
        O.append(Reduce-X)
    else if ReduceUnary-X(c) ∈ D then
        c ←ReduceUnary-X(c)
        O.append(ReduceUnary-X)
    else if swapAllowed(c) = True then
        c ←Swap(c)
        O.append(Swap)
    else
        c ←Shift(c)
        O.append(Shift)
    end if
end while
return O
```

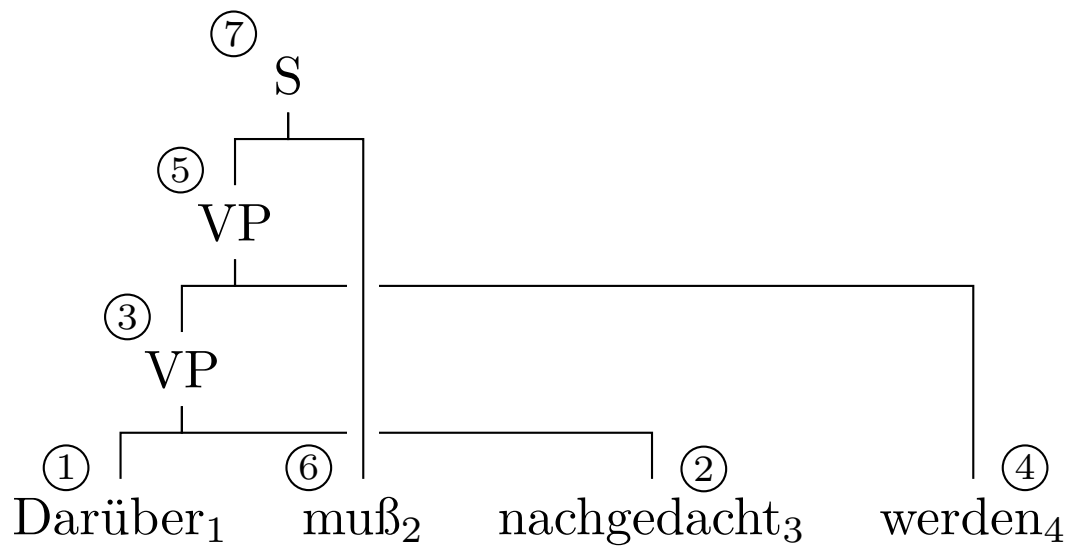

Figure 3.13: Projective ordering for the tree over *Darüber muß nachgedacht werden.*

shows the transition sequence that would be induced by algorithm 3.5.[17]

$$\begin{aligned}&||w_1, w_2, w_3, w_4 \underset{\text{Shift}}{\Rightarrow} w_1||w_2, w_3, w_4 \underset{\text{Shift}}{\Rightarrow} w_1, w_2||w_3, w_4 \underset{\text{Shift}}{\Rightarrow}\\ &w_1, w_2, w_3||w_4 \underset{\text{Swap}}{\Rightarrow} w_1, w_3||w_2, w_4 \underset{\text{Shift}}{\Rightarrow} w_1, w_3, w_2||w_4 \underset{\text{Shift}}{\Rightarrow}\\ &w_1, w_3, w_2, w_4|| \underset{\text{Swap}}{\Rightarrow} w_1, w_3, w_4||w_2 \underset{\text{Redcuce-A}}{\Rightarrow} w_1, A||w_2\end{aligned} \tag{3.8}$$

$w_2$ and $w_3$ get swapped since $w_3 <_{\mathrm{G}} w_2$. But this blocks the reduction of $w_3$ and $w_4$ to A so that a second Swap is necessary changing the order of $w_2$ and $w_4$. Then, $w_3$ and $w_4$ are adjacent again and can be reduced. Postponing the Swap of $w_2$ and $w_3$ and instead shifting, reducing A and then swapping $w_2$ and A would lower the transition count and reduce the number of Swaps by one. Equation 3.9 shows this shorter transition sequence leading to the same configuration.

$$\begin{aligned}&||w_1, w_2, w_3, w_4 \underset{\text{Shift}}{\Rightarrow} w_1||w_2, w_3, w_4 \underset{\text{Shift}}{\Rightarrow} w_1, w_2||w_3, w_4 \underset{\text{Shift}}{\Rightarrow}\\ &w_1, w_2, w_3||w_4 \underset{\text{Shift}}{\Rightarrow} w_1, w_2, w_3, w_4|| \underset{\text{Reduce-A}}{\Rightarrow} w_1, w_2, A|| \underset{\text{Swap}}{\Rightarrow}\\ &w_1, A||w_2\end{aligned} \tag{3.9}$$

Motivated by this observation, *lazy swap* adds a second condition to the Swap transition: the two topmost stack items are not swapped if the adjacent element to the right (i.e. the first position in the buffer) has the same closest maximal fully projective constituent as the stack top. This prevents the parser from swapping elements into groups of nodes that can be reduced to a

[17] || represents the border between stack and buffer in this shorthand notation.

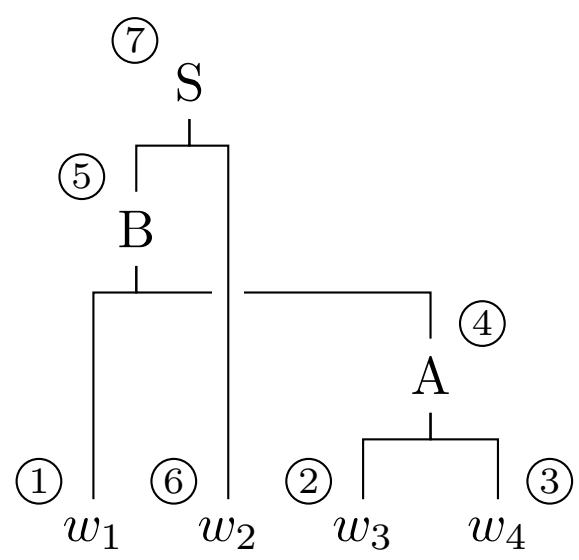

Figure 3.14: Minimal example for the motivation of lazy swap.

projective constituent. To formally define this notion, one needs the following (Stanojević and Alhama, 2017):

**Definition 3.31 (Projective Constituent).**
Let $w = w_1, ..., w_n$ be a word and $D = \langle V, \triangleleft, \hat{v} \rangle$ a tree for $w$. Let $v \in V$ be a node.

1. $v$ is called a *projective constituent* iff $v$ dominates a span of leafs without a gap.
2. $v$ is called a *fully projective constituent* iff $v$ is a projective constituent and if all descendants of $v$ are projective constituents.
3. $v$ is called a *maximal fully projective constituent* iff $v$ is a fully projective constituent and the parent of $v$ is not a fully projective constituent.

**Example 3.32.**
Figure 3.15 shows a tree where all projective, fully projective and maximal fully projective constituents are marked. Note that a projective constituent can have children that are not projective constituents, here for instance S. The property does not tell anything about the order of the nodes dominated by the constituent.

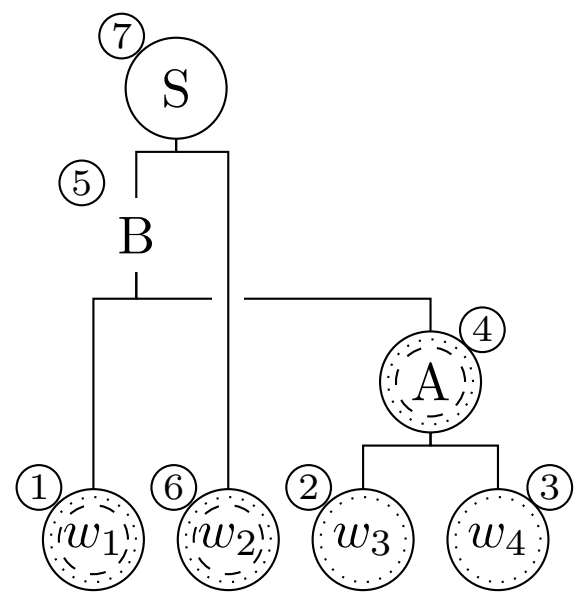

Figure 3.15: Tree with maximal fully projective node A. Projective constituents are circled in solid, fully projective constituents dotted and maximal fully projective constituents dashed.

Now, one can define a function that returns the closest ancestor that is a maximally projective constituent for a given constituent:

**Definition 3.33 (Closest maximal projective constituent).**
Let $w$ be a word, $D = \langle V, \triangleleft, \hat{v} \rangle$ a tree over $w$. $MPC : V \to V$ is defined as follows:

$$v \mapsto \begin{cases} \arg\min_{m \in M} |Ind_D(m)| & \text{if } M \neq \emptyset, \\ v & \text{otherwise,} \end{cases}$$

where $M$ denotes the set of all maximally fully projective constituents that are acendants of $v$ defined as follows:

$$M = \{p \in V \mid p \text{ is a maximally fully projective constituent and } p \triangleleft^+ v\}$$

Figure 3.10 shows the modified swapAllowed condition for lazy swap. Versley (2014) found that a lazy approach outperforms an eager one, presumably because a large number of SWAP transitions is harder to predict.

$$\text{swapAllowed}_{\text{lazy}}(\langle S|s_1|s_0, b_0|B\rangle) :\Leftrightarrow s_0 <_{\text{G}} s_1 \wedge MPC(s_0) \neq MPC(s_0) \tag{3.10}$$

**Lazier Swap** Stanojević and Alhama (2017) took the idea of lazy swap and introduced an even *lazier swap* strategy. They noticed that lazy swap does not help when the substructure that would be more efficient to complete first does not exhibit a maximal fully projective constituent but could be reduced to a projective constituent nonetheless.

**Example 3.34.**
Figure 3.16 shows a tree where only the terminals are maximally fully projective constituents. Therefore a lazy oracle would predict a SWAP for $w_2$ and $w_3$. $w_2$ would be swapped into the subtree rooted at B which would make two additional SWAP transitions necessary to allow for the treatment of the substructure. First completing the subtree rooted at B by resolving its lower-level discontinuity would lead to a smaller step count.

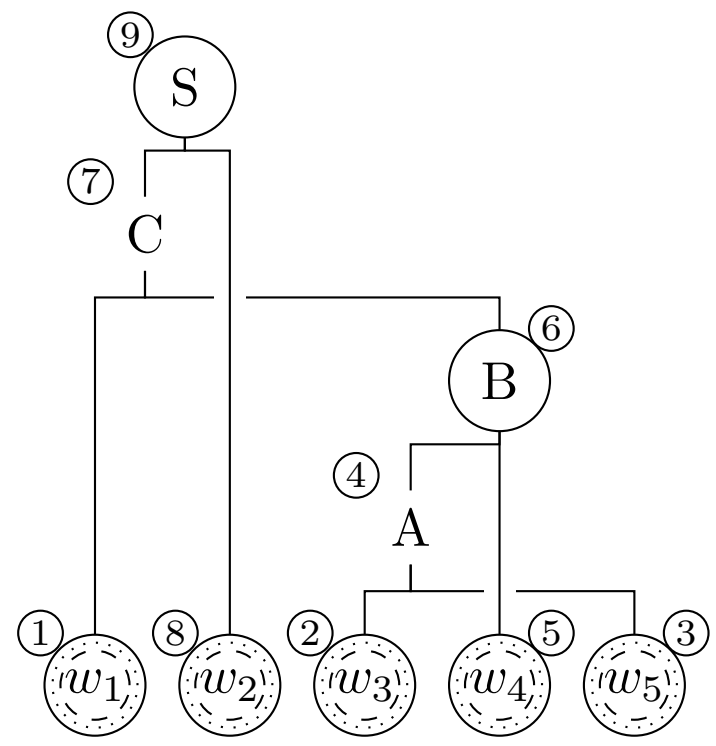

Figure 3.16: Tree without maximal fully projective nodes (except for terminals). Projective constituents are circled in solid, fully projective constituents dotted and maximal fully projective constituents dashed.

To achieve this, Stanojević and Alhama (2017) introduce a function that returns the lowest projective acendant of a given node:

**Definition 3.35 (Closest Projective Constituent).**
Let $w$ be a word and $D = \langle V, \triangleleft, \hat{v}\rangle$ a tree over $w$. The function $CPC : V \to V$ is defined as follows:

$$v \mapsto \underset{m \in M}{\arg\min}\, |Ind_D(m)|$$

where $M$ denotes the set of all maximal fully projective constituents that are ascendants of $v$ as established in definition 3.33.

$$\text{swapAllowed}_{\text{lazier}}(\langle S|v_1|s_0, s_0|B\rangle) :\Leftrightarrow s_0 <_{\text{G}} s_1 \wedge CPC(s_1) = CPC(s_0) \tag{3.11}$$

A SWAP is only allowed if the closest projective constituent of the topmost stack element and the second topmost stack element are the same. If they are not the same, one of the elements belongs to a substructure that can be reduced first (potentially by bringing it first internally into projective order). If they are the same there is no projective subtree for which the analysis would be blocked by swapping the second topmost element. Then, a SWAP is necessary to proceed with the derivation. This version of the transition condition is given in equation 3.11. Stanojević and Alhama (2017) show that the lazier swap condition entails lazy swap:

**Corollary 3.36.**
Let $D = \langle V, \triangleleft, \hat{v} \rangle$ be a tree and $s_i, ..., s_1, s_0, b_0, ...b_j$ a tree-cut of $D$ where $s_0 <_G s_1$. It holds that $CPC(s_0) = CPC(s_1) \Rightarrow MPC(s_0) \neq MPC(b_0)$.

*Proof.* Assumption: $CPC(s_0) = CPC(s_1)$. Either 1. there is a non-empty sequence $v_1, ..., v_k$ of non-projective constituents such that $CPC(s_0) \triangleleft v_1, v_1 \triangleleft v_2, ..., v_k \triangleleft s_0$ or 2. $CPC(s_0) \triangleleft s_0$.

1. Since the parent of $s_0$ is non-projective, it follows that $s_0$ is a maximal projective constituent. By definition 3.33 it holds that $MPC(s_0) = s_0$. Since $s_0$ per definition of a tree-cut does not dominate $b_0$, it holds that $MPC(s_0) \neq MPC(b_0)$.

2. Either (a) $CPC(s_0)$ is fully projective or (b) $CPC(s_0)$ is not fully projective.

   (a) This contradicts $s_0 <_{\text{G}} s_1$. The relation demands that the two sister nodes $p_0$ with $p_0 \triangleleft^* s_0$ and $p_1$ with $p_1 \triangleleft^* s_1$ fulfil $p_0 <_{\text{ind}} p_1$. It must hold that $p_0 = s_0$ since $CPC(s_0) \triangleleft s_0$ and $CPC(s_0) \triangleleft^* s_1$. Therefore $s_0 <_{\text{ind}} p_1$. Since $s_0$ is fully projective, it follows that $s_0 <_{\text{N}} p_1$. But this means that $s_1$ must have been swapped from a position right of $s_0$ to the position left of $s_0$ which would not have been allowed by $<_{\text{G}}$ since the two elements already respected projective ordering. Therefore, $CPC(s_0)$ cannot be fully projective.

   (b) It follows that $MPC(s_0) = s_0$ and therefore $MPC(s_0) \neq MPC(b_0)$ (see 1.).

□

**Example 3.37.**
Figure 3.17 shows the transition sequence predicted by the oracle derived using algorithm 3.5 with the lazier swap condition given in figure 3.11 for the tree in equation 3.16. The lazy swap strategy would have predicted a SWAP at step 3 since $MPC(w_3) = w_3 \neq MPC(w_4) = w_4$. This would move $w_2$ into the subtree rooted at B, blocking its derivation.

| Step | Stack | Buffer | Action | check of $\text{SwapAllowed}_{\text{lazier}}$ |
|---|---|---|---|---|
| 0 | | $w_1, w_2, w_3, w_4, w_5$ | SHIFT | |
| 1 | $w_1$ | $w_2, w_3, w_4, w_5$ | SHIFT | |
| 2 | $w_1, w_2$ | $w_3, w_4, w_5$ | SHIFT | $w_2 \not<_G w_1$ |
| 3 | $w_1, w_2, w_3$ | $w_4, w_5$ | SHIFT | $CPC(w_2) = \text{S} \neq \text{B} = CPC(w_3)$ |
| 4 | $w_1, w_2, w_3, w_4$ | $w_5$ | SHIFT | $w_4 \not<_G w_3$ |
| 5 | $w_1, w_2, w_3, w_4, w_5$ | | SWAP | $w_5 <_G w_4 \wedge CPC(w_4) = \text{B} = CPC(w_5)$ |
| 6 | $w_1, w_2, w_3, w_5$ | $w_4$ | REDUCE-A | |
| 7 | $w_1, w_2$, A | $w_4$ | SHIFT | $CPC(w_2) = \text{S} \neq \text{B} = CPC(\text{A})$ |
| 8 | $w_1, w_2$, A, $w_4$ | | REDUCE-B | |
| 9 | $w_1, w_2$, B | | SWAP | $\text{B} <_G w_2 \wedge CPC(w_2) = \text{S} = CPC(\text{B})$ |
| 10 | $w_1$, B | $w_2$ | REDUCE-C | |
| 11 | C | $w_2$ | SHIFT | |
| 12 | C, $w_2$ | | REDUCE-S | |
| 13 | S | | | |

Figure 3.17: Example parse using lazier swap strategy for tree in figure 3.16. Pointers are omitted since the nodes are uniquely labelled.

**Complexity** For an input of length $n$ the number of REDUCE-X transitions to recover a binary tree is $n - 1$ (one for every internal node). The number of REDUCEUNARY-X transitions is potentially unbounded. Let $m$ denote the maximum unary chain length encountered during training used to hard-code a limit into the parser. Then the maximum number of REDUCEUNARY-X transitions is $m(2n - 1)$ (one chain for every leaf and for every internal node).

Every leaf must be shifted once. The number of Swap transitions and further Shift transitions is dependent on the choice of oracle. The worst case (not considering loops) is the strict eager oracle used by Maier (2015) since it features the maximum Swap number before a reduction. When establishing a discontinuous constituents, no intermediate elements are merged to reduce the number of elements to swap.

Figure 3.18 shows the worst case tree for length 5 given by Coavoux and Crabbé (2017a). With 1 and 2 being on top of the stack, the parser needs to perform 3 intermediate Shift transitions and 3 Swap transitions to achieve projective order for 1 and 5. After Reduce-A, the same procedure follows with 2 intermediate Shifts and 2 Swaps. For a sentence with length $n$, where $k = n - 2$ is the number of intermediate elements, this leads to

$$2(k + (k-1) + (k-2) + ... + 1) = 2\left(\frac{k(k+1)}{2}\right) = (n-2)(n-1) = n^2 - 3n + 2 \qquad (3.12)$$

transitions for projective reordering. This results in a total worst-case transition count of $n^2 - n + 1 + m(2n-1)$.

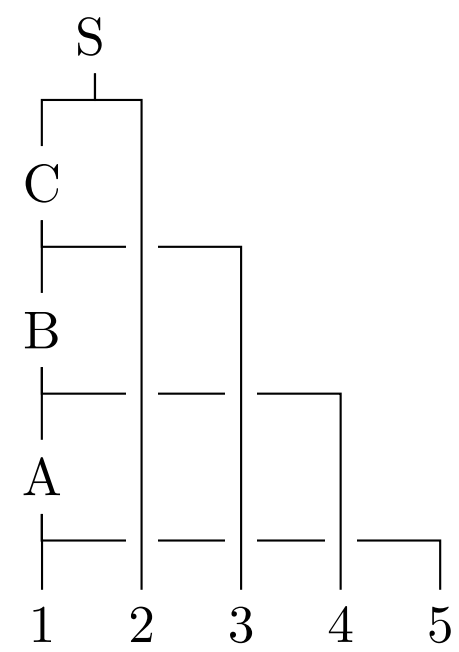

Figure 3.18: Example for worst case tree of length 5 with respect to Swap transition, adapted from Coavoux and Crabbé (2017a).

**Correctness** To show the correctness of the transition system for the set of all $\varepsilon$-free range-labelled binary trees, completeness and soundness must be shown.

For completeness, it has to be shown that every tree can be retrieved from some transition sequence. Any two nodes in a tree-cut found in $S$ and $B$ can be moved into an adjacent position via Shift and Swap and assigned a parent via Reduce-X. Furthermore, any single node can be assigned a unary parent using ReduceUnary-X after shifting or composing it via Reduce-X. Thus, from the initial configuration (tree-cut with leaves $\langle 0, 1\rangle, ..., \langle |w| - 1, |w|\rangle$ for some word $w$), any $\varepsilon$-free range-labelled binary tree for $w$ can be built.

For soundness, it must be shown that any transition sequence produces a tree. First, it can be seen that a node cannot have several parents. This follows from the fact that the daughter(s) are removed from $S$ when applying Reduce-X or ReduceUnary-X. Furthermore, by definition, a final configuration consists of an empty buffer and only one constituent on the stack. Since the number of constituents can only be reduced via Reduce-X, it follows that this one constituent spans all input elements. Lastly, by definition of the initialisation function, the leaves are labelled with one-length ranges that cover all indices of the input word. Thus, an $\varepsilon$-free range-labelled binary tree can be retrieved.

The design of the transitions guarantees that at every step at least one transition is possible. When prohibiting swap-loops via the $<_{\text{ind}}$-condition and setting a maximum number of unary chains, the maximum number of transitions is final. This follows directly from the fact that if all possible Swaps were performed, the end of the buffer was reached and the maximum number of unary chains were assigned to the topmost element of the stack only Reduce-X could be performed, reducing the number of elements. Eventually, only one root element is left. Therefore, with these restrictions, the algorithm always reaches a final configuration.

### 3.2.5 Reordering with a Split Stack and the Gap Action

Coavoux and Crabbé (2017a) introduce a new transition called Gap. Instead of altering the order of two elements as the Swap transition does, they give up the restriction that only the two topmost elements of the stack can combine. Instead, any element of the stack is allowed to combine with the topmost element.

They achieve this by splitting up the stack into two parts: the stack $S$ and a *deque* $\Delta$.[18] A dequeue is a sequence that can be accessed from both ends. Reductions can take place between the tops of $S$ and $\Delta$. The Gap transition takes an element from the top of $S$ and adds it to the bottom of $\Delta$. The Shift transition is modified so that it pushes all elements from $\Delta$ onto $S$ and then pops an element from the buffer onto $\Delta$.

In their original paper, Coavoux and Crabbé (2017a) use a lexicalised model that assigns heads to constituents. In later work they present an unlexicalised approach that will be the basis for this section (Coavoux et al., 2019).

The traditional Reduce transition is split into a Label-X and a Merge transition. Thus, they call their approach ML-Gap (Merge-Label-Gap). Furthermore, by introducing a No-Label transition, they can drop the requirement that the trees must be binary. Several nodes can be grouped together via Merge before assigning a common parent via Label.

Inspired by span-based parsing for projective trees (Cross and Huang, 2016b), ML-Gap represents nodes by referencing the indices they dominate. In order to account for discontinuities, a set of indices is used instead of a single range. The labelling of the nodes is stored in a separate set $K$. The final elements of $K$ can be arranged into a hierarchy from which the parse tree can be restored. The following definitions extend the formalisation of Coavoux et al. (2019).

**Definition 3.38 (Constituent candidate).**

1. A set of integers $s \subseteq \{1, ..., n\}$ with $n \in \mathbb{N}$ is called a *constituent candidate (in length n).*
2. The set of all constituent candidates in length $n \in \mathbb{N}$ is referred to by $ConstCand(n) := \mathcal{P}(\{1, ..., n\})$.
3. The set of all constituent candidates is referred to by $ConstCand := \mathcal{P}_{fin}(\mathbb{N})$.
4. Two constituent candidates $s_1, s_2$ are called *compatible* iff $(s_1 \subset s_2) \vee (s_2 \subset s_1) \vee (s_1 \cap s_2 = \emptyset)$

**Definition 3.39 (Instantiated constituent).**
Let $N$ be an alphabet of nonterminal labels.

1. A tuple $\langle A, s\rangle$ is called an *instantiated constituent wrt. N (in length n)* iff $A \in N$ and $s$ is a constituent candidate in length $n$.
2. The set of all instantiated constituents wrt. $N$ in length $n \in \mathbb{N}$ is called $Const(N, n) := N \times ConstCand(n)$.
3. The set of all instantiated constituents wrt. $N$ is called $Const := N \times ConstCand$.
4. $\langle A_1, s_1\rangle, \langle A_2, s_2\rangle \in Const(N, n)$ are called *compatible* iff $s_1$ and $s_2$ are compatible.

**Definition 3.40 (Sets of compatible constituents).**
Let $N$ be an alphabet of nonterminals and $n \in \mathbb{N}$. Let $S \subseteq ConstCand(n)$ be a set of constituent candidates in length $n$.

[18] Coavoux and Crabbé (2017a) use $D$ to refer to the deque which I changed to $\Delta$ to prevent confusion since I frequently use $D$ to refer to a tree.

1. $S$ is called *consistent* iff for all $s_1, s_2 \in S$, $s_1$ and $s_2$ are compatible.
2. $S$ is called *rooted* iff there exists $s \in S$ such that $s' \subset s$ for all $s' \in S$ with $s \neq s'$. $s$ is called the *root of* $S$ and denoted by $root(S) = s$.
3. $S$ is said to *span length* $n$ iff for all $i \in \{1, ..., n\}$ there exists some element $s \in S$ such that $i \in s$.
4. $S$ is called *complete* for length $n$ iff it is consistent, rooted and spans length $n$.
5. The terms *consistent, rooted, spanning length* $n$ and *complete* naturally extend to sets of instantiated constituents.

**Definition 3.41 (Maximal subset).**
Let $S \subseteq ConstCand(n)$ be a consistent set of constituent candidates for some $n \in \mathbb{N}$. The relation $\underset{\max}{\subset} \subset S \times S$ is defined as follows:

$$s_1 \underset{\max}{\subset} s_2 :\Leftrightarrow s_1 \subset s_2 \wedge \nexists s_3 \in S : s_1 \subset s_3 \subset s_2, \qquad \text{for all } s_1, s_2 \in S$$

The relation is naturally extended to consistent sets of constituents.

**Corollary 3.42.**
Let $n \in \mathbb{N}$ and $S \subseteq ConstCand(n)$ be a consistent set of constituent candidates. For every $s_1 \in S$ there exists at most one element $s_2$ such that $s_1 \underset{max}{\subset} s_2$.

*Proof.* Let us assume, there are $s_1, s_2, s_3 \in S$ such that $s_1 \underset{\max}{\subset} s_2$ and $s_1 \underset{\max}{\subset} s_3$. Since $S$ is a compatible constituent candidate set, it must hold that either $s_2 \subset s_3$ or $s_3 \subset s_2$ or $s_2 \cap s_3 = \emptyset$. Take some index $i \in s_1$. By assumption also $i \in s_2$ and $i \in s_3$. Now, it cannot be that $s_2 \cap s_3 = \emptyset$. Assume that $s_2 \subset s_3$. This contradicts the assumption that $s_1 \underset{\max}{\subset} s_3$. The reverse also holds true. □

**Definition 3.43 (Unary-free tree).**
Let $D = \langle V, \triangleleft, \hat{v} \rangle$ be a tree. $D$ is called *unary-free* iff for all internal nodes $v \in V$ either $v$ has a daughter that is a leaf or $v$ has at least two daughters.

It can be seen that in a unary-free tree, there are no two internal nodes (i.e. nonterminals) that dominate exactly the same set of leaves. The parent of an internal node must always enlarge the indices covered by the yield and a daughter must always narrow them down. Since the leaves are labelled with ranges, a unary relationship is allowed for pre-terminals to grant the possibility of assigning a nonterminal label spanning a single symbol.

**Corollary 3.44 (Equivalence of trees and constituent sets).**

1. Let $n \in \mathbb{N}$ and $D = \langle V, \triangleleft, \hat{v}, \lambda \rangle$ a unary-free $\varepsilon$-free range-labelled tree spanning length $n$. One can define a function $\psi$ converting nodes to instantiated constituents as follows:

$$\psi : V \to Const(N, n), \qquad v \mapsto \langle \lambda(v), Ind_D(v) \rangle,$$

   where it holds that:

   (a) $\{\psi(v) \mid v \in V \text{ and } v \text{ is internal}\}$ is complete for length $n$.

   (b) $v_1 \triangleleft v_2 \Leftrightarrow \psi(v_2) \underset{\max}{\subset} \psi(v_1)$ for all $v_1, v_2 \in V$ where $v_1, v_2$ are internal.

2. Let $C \subseteq Const(N, n)$ be complete for length $n$. One can construct a unary-free $\varepsilon$-free range-labelled tree $D = \langle V, \triangleleft, \hat{v}, \lambda \rangle$ spanning length $n$ in the following fashion:

   (a) $V := root(K) \cup K$,

(b) $\hat{v} := root(K)$,

(c) $\lhd := \underset{\max}{\subset}^{T} \cup \{\langle c, i\rangle \mid i \in \hat{v}, \begin{pmatrix} \{i\} \in K, c = \{i\} \text{ or} \\ \{i\} \notin K, \{i\} \subset c, c \in K, \nexists c' \in K : (c \neq c' \wedge \{i\} \subset c' \subset c) \end{pmatrix} \}$[19]

(d) $\lambda(\langle A, s\rangle) := A$ for all $\langle A, s\rangle \in K$,
$\lambda(i) := \langle i-1, i\rangle$ for all $i \in \hat{v}$.

Now it holds that $c_1 \underset{\max}{\subset} c_2 \Leftrightarrow c_2 \lhd c_2$.

It can be seen that the two notations are equivalent and that the conversions preserve the dominance relation. Set notation is useful for transition-based parsers since they do not rely on the range tuples that arise from LCFRS rules. Note that set notation does not allow for two constituents to span the same indices since dominance is expressed via the subset relation. Therefore, trees are not allowed to exhibit chains of unary relations. It is possible to convert any tree to a unary-free tree by collapsing every path of internal unaries $A_0, A_1, ..., A_n$ to a single node $A_0@A_1@...@A_n$ and to convert it back by expanding the chain.[20]

Differentiating between constituent candidates and instantiated constituents makes it possible for a transition system to construct constituents with more than two daughters using multiple Merge actions. Every Merge enlarges the constituent candidate by one additional daughter. A full constituent is only asserted when performing Label-X. This is comparable to *implicit binarisation* as described by Gómez-Rodríguez (2014) for *Earley* and *Left-Corner* parsers since the sequence of Merge transitions implicitly gives rise to a binary derivation. Constituent candidates can be viewed as unordered *dotted items* and instantiated constituents as asserted productions. This is visualised in figure 3.19.

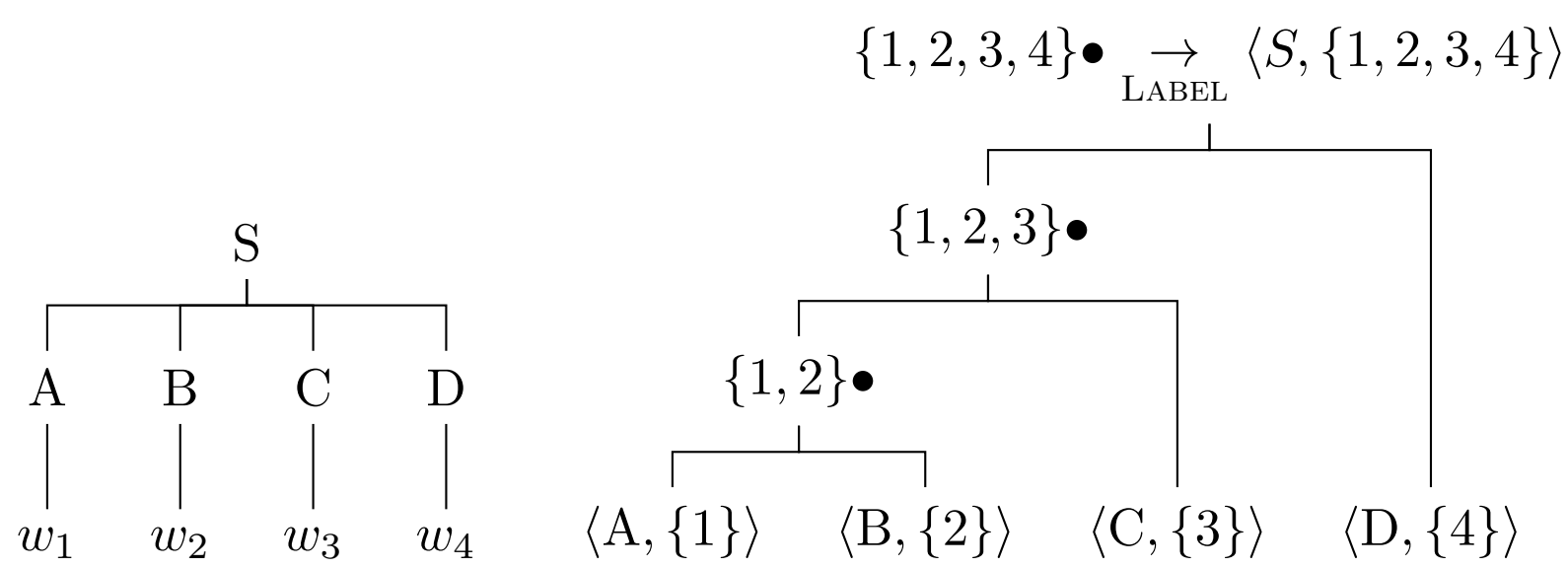

Figure 3.19: Implicit binarisation. The tree on the left exhibits a node with four children; on the right a derivation via Merge and Label-X is shown.

A configuration in ML-Gap comprises of a stack, a deque and a set of label assignments. Instead of a buffer as in SR-Swap, a simple index is used to keep track of the next element to shift. This is possible since — without a Swap transition — no element can be moved onto the buffer and thus its elements are guaranteed to respect linear precedence. The stack and the deque hold a partition of the set of indices shifted up to the current point which is equivalent to the tree-cut in SR-Swap (when concatenating the unread nodes). Figure 3.20 visualises a ML-Gap configuration.

**Definition 3.45 (ML-Gap configuration).**
Let $N$ be an alphabet of nonterminal labels and $Q$ a set of states. $\mathcal{C}^{\text{MLG}}$ is the set of all elements $\langle S, \Delta, i, j, K\rangle : q$ such that

1. $j \in \mathbb{N}$ marks the maximum index,[21]

[19] $R^T$ for a relation $R$ denotes the converse relation.

[20] Coavoux and Cohen (2019) perform this conversion in the implementation of their parser.

[21] Contrary to Coavoux and Crabbé (2017a) I include the right border index as a component of the configuration since I define the set of configurations for all possible input sequences. Therefore, a configuration must contain information as to the number of elements left to shift.

2. $S$ and $\Delta$ are sequences of constituent candidates in length $j$,

3. $i \in \{1, ..., j\}$ marks the current index,

4. $K \subseteq Const(N, j)$ is a set of instantiated constituents[22] and

5. $q \in Q$ is a state.

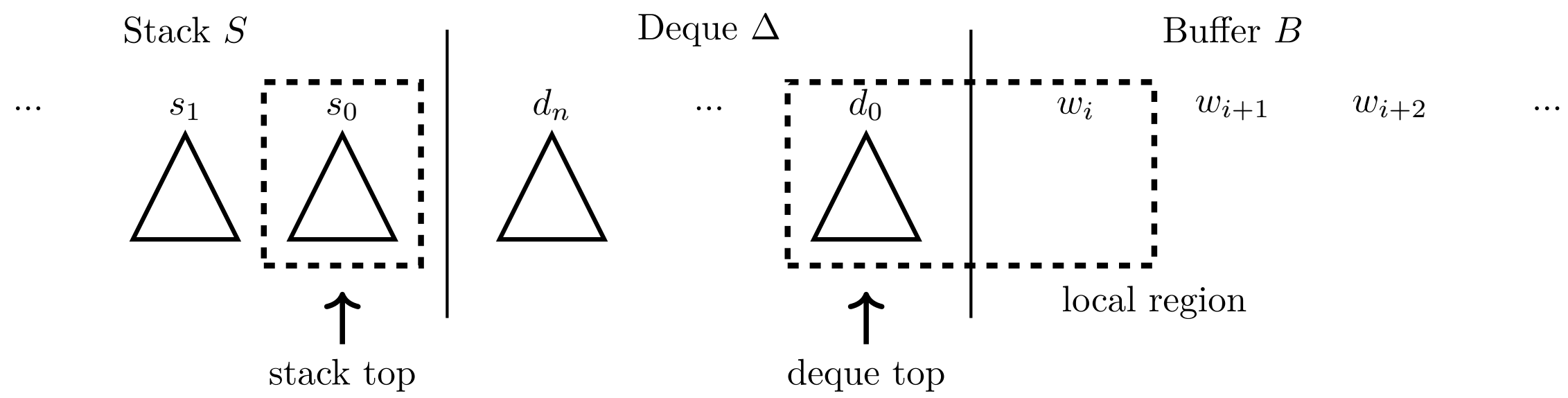

Figure 3.20: Illustration of the ML-Gap configuration (without the set of constituents $K$). Illustration inspired by Coavoux and Cohen (2019).

Coavoux et al. (2019) adopt the strategy of differentiating between *structural* and *labelling* transitions from Cross and Huang (2016b). Depending on the previous action(s), only structural or labelling transitions are possible which narrows down the search space for the neural scorer making predictions easier. The model of Cross and Huang (2016b) simply alternates between structural and labelling actions. To account for this, one can include a step index into the configuration that counts up the total number of transitions applied and represents an even or an odd step.

Here, the Gap transition is utilised to enable merging the top element with an element anywhere on the stack. Therefore, multiple consecutive applications of Gap must be possible. Furthermore, a sequence of Gap transitions should be followed by a Merge to limit its application to useful cases. Coavoux et al. (2019) define which transitions can follow each other using a *finite state automaton.* For clarity, I integrate this restriction into the transitions and treat the state as part of the configuration. The following definition of *finite state automata* is taken from Hopcroft et al. (2007, chapter 2).

**Definition 3.46 (Finite state automaton).**
A *finite state automaton (FSA)* is a 5-tuple $\langle \Sigma, Q, q_0, \delta, F \rangle$ where:

1. $\Sigma$ is an alphabet,

2. $Q$ is a finite non-empty set of states,

3. $q_0 \in Q$ is called the *initial state*,

4. $\delta : Q \times \Sigma \to Q$ is called the *transition function*,

5. $F \subseteq Q$ is the *set of final states.*

**Definition 3.47 (MLG FSA).**
The FSA for ML-GAP action sequences is defined as $\mathfrak{A}^{\text{MLG}} = \langle \mathcal{T}^{\text{MLG}}, Q, \text{Struct}, \delta, \{\text{Struct}\} \rangle$ where

1. $\mathcal{T}^{\text{MLG}} = \mathcal{T}^{\text{MLG}}_{\text{struct}} \cup \mathcal{T}^{\text{MLG}}_{\text{label}}$ is the set of ML-Gap transitions with

   (a) $\mathcal{T}^{\text{MLG}}_{\text{struct}} = \{\textsc{Shift}, \textsc{Gap}, \text{Merge}\}$ and

   (b) $\mathcal{T}^{\text{MLG}}_{\text{label}} = \{\textsc{No-Label}, \textsc{Label-X}\}$,

[22] Coavoux et al. (2019) use $C$ to denote the set of instantiated constituents. I changed the symbol to prevent confusion with variables used for configurations.

2. $Q = \{\text{Struct}, \text{Struct'}, \text{Label}\}$,

3. $\delta$ is the transition function defined as follows:

$$\begin{aligned}
\delta(\text{Struct}, \textsc{Gap}) &= \text{Struct'}\\
\delta(\text{Struct}, \textsc{Shift}) &= \text{Label}\\
\delta(\text{Struct}, \textsc{Merge}) &= \text{Label}\\
\delta(\text{Struct'}, \textsc{Gap}) &= \text{Struct'}\\
\delta(\text{Struct'}, \textsc{Merge}) &= \text{Label}\\
\delta(\text{Label}, \textsc{Label-X}) &= \text{Struct}\\
\delta(\text{Label}, \textsc{No-Label}) &= \text{Struct}
\end{aligned}$$

The state diagram for $\mathfrak{A}^{\text{MLG}}$ is given in figure 3.21. The current state of the automaton is part of the configuration and a new configuration can only be deduced from a transition if the FSA accepts the transition as input at the given state.

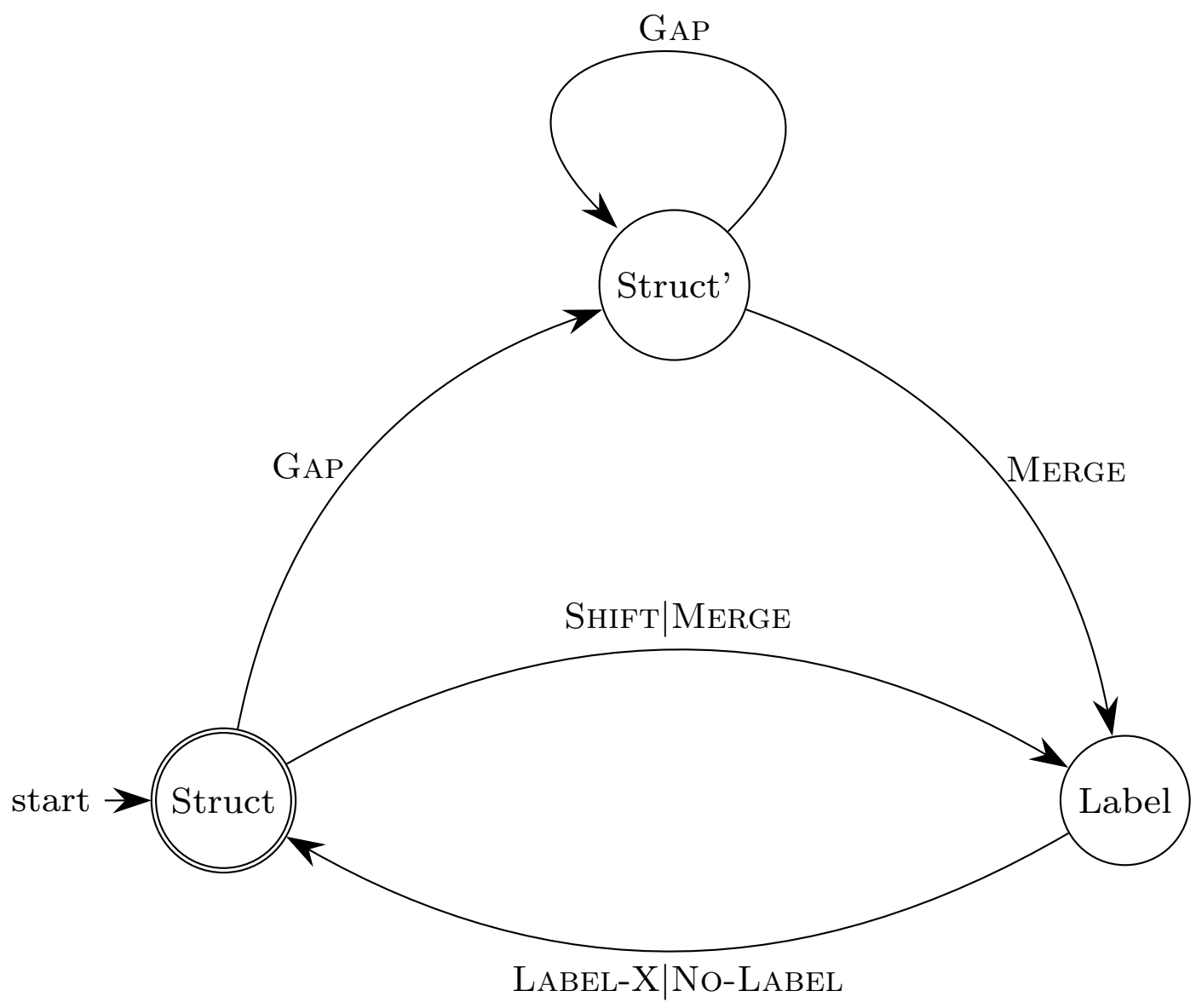

Figure 3.21: Illustration of the FSA for the action sequences allowed in ML-Gap; reproduction of a figure in Coavoux et al. (2019).

The full set of transitions for ML-Gap is given as deduction rules in figure 3.22. The formalisation of the transition system is given in definition 3.48.

**Definition 3.48 (Merge-Label-Gap transition system).**
A Merge-Label-Gap transition system for an alphabet $T$ and an alphabet of nonterminals $N$ is a 5-tuple $\mathfrak{S}^{\text{MLG}} = \langle \mathcal{C}^{\text{MLG}}, \mathcal{T}^{\text{MLG}}, \sigma, F, d \rangle$ where

1. $\mathcal{T}^{\text{MLG}}$ is defined for $\mathcal{C}^{\text{MLG}}$ as described in figure 3.22,

2. $\sigma(w) = \sigma'(|w|) = \langle\ [\ ], [\ ], 0, |w|, \emptyset\ \rangle : \text{Struct}$ for every $w \in T^+$,

3. $F = \{\langle\ S, \Delta, i, j, K\ \rangle : q \in \mathcal{C}^{\text{MLG}} \mid i = j \wedge \exists A \in N : \langle A, \{1, ..., j\}\rangle \in K\ \}$ and

4. $d$ retrieves the tree from $K$ for every $\langle\ S, \Delta, i, j, K\ \rangle : q \in F$.

The definition of constituent sets allows to abstract away from symbols of an alphabet of terminals when examining the properties of the transition system. Structurally, all words with the same length are equivalent. It is only for the action scorer that indices must be used as references to individual input items. In the following, when assuming a valid configuration $c = \langle\ S, \Delta, i, j, K\ \rangle :$

$q$ the input sequence for which it is valid is omitted since the maximal index $j$ already uniquely determines that $\sigma'(j) \overset{*}{\Rightarrow} c$ and therefore the statement holds for all words with length $j$ over the alphabet the transition system is defined on.

**Corollary 3.49.**
For every valid configuration $c = \langle\ S, \Delta|d_0, i, j, K\ \rangle : q$ it holds that

$$\forall s \in S \cup \Delta : \max(s) \leq \max(d_0).$$

*Proof.* Let $c = \langle\ S, \Delta|d_0, i, j, K\ \rangle : q$ be a valid configuration. $S$ and $\Delta$ can only contain those elements that were added to the memory by Shift. Therefore, all elements are smaller than $i$ or equal. $i$ must have been moved as a single-element set $\{i\}$ to the top of the deque. Gap, Label-X and No-Label do not alter the top of the deque. Merge performs set union on the top of $S$ and the top of $\Delta$ but since all those elements are smaller than $i$, $i$ remains the largest element. Only Shift can move a larger element onto the deque top but it also enlarges the buffer index. Therefore $\forall s \in S \cup \Delta : \max(s) \leq i = \max(d_0)$. □

**Corollary 3.50.**
Let $N$ be an alphabet of nonterminals. Every valid configuration $c \in F$ has the form $c = \langle\ [\ ], [\{1, ..., n\}], n, n, K\ \rangle$ : Struct where $K$ is a rooted set of instantiated constituents with $root(K) = \langle A, \{1, ..., n\}\rangle$ for some $A \in N$.

*Proof.* Let us assume that $c = \langle\ S, \Delta, i, j, K\ \rangle : q \in F$ is a valid configuration. From the definition of $F$ already follows that $i = j$ and $\exists A \in N : \langle A, \{1, ..., j\}\rangle \in K$. Since $j$ does not change for any transition, it must hold that $\sigma'(j) = \langle\ [\ ], [\ ], 0, j, \emptyset\ \rangle : \text{Struct} \overset{*}{\Rightarrow} \langle\ S, \Delta, j, j, K\ \rangle : q$ and therefore all instantiated constituents in $K$ must be composed from elements in the range from 1 to $j$ i.e. it holds that $K \subseteq Const(N, j)$. There must exist $c' = \langle\ S', [\{1, ..., n\}], i', j, K'\ \rangle : \text{Label} \in \mathcal{C}^{\text{MLG}}$ such that $\sigma'(j) \overset{*}{\Rightarrow} c' \underset{\text{Label-X}}{\Rightarrow} \langle\ S', [\{1, ..., n\}], i', j, K' \cup \{\langle \text{A}, \{1, ..., n\}\rangle\}\ \rangle : \text{Struct} \overset{*}{\Rightarrow} c$. Since the top of $\Delta$ always contains the maximum index of $S$ and $\Delta$ (cf. corollary 3.49), it must hold that $S' = [\ ]$ and $i' = j$. Therefore, the result of the labelling is already the goal element. Hence, $q = \text{Struct}$ and there cannot be another instantiated constituent in $K$ with indices $\{1, ..., n\}$. Now it holds that for all $\langle A', s'\rangle \in (K \setminus \{\langle A, \{1, ..., n\}\rangle\}) : s' \subset \{1, ..., n\}$ and thus $root(K) = \langle A, \{1, ..., n\}\rangle$. □

**Example 3.51.**
Figure 3.24 shows an example parse for the tree in figure 3.23.

**Oracle** Nodes are necessarily created according to a rightmost index ordering. As Coavoux et al. (2019) note, this stems from the property described by corollary 3.49.[23] From the deque top element, new nodes are established via Label-X. For two elements with different rightmost indices, the one with the lower rightmost index is created first. For two nodes in the tree that have the same rightmost index, one necessarily contains less indices than the other and is thus labelled first before applying Merge once or more and then labelling the other. The following definition is adopted from Coavoux and Cohen (2019):

**Definition 3.52 (Rightmost index ordering).**
The order $\leq_{\text{right}}$ is defined for constituent candidates as follows:

$$s_1 \leq_{\text{right}} s_2 :\Leftrightarrow \begin{cases} \max(s_1) < \max(s_2) \text{ or} \\ \max(s_1) = \max(s_2), s_1 \subseteq s_2. \end{cases}$$

[23]A more thorough proof of this property is given for the stack-free successor formalism in corollary 4.8.

$$\textit{axiom} \quad \frac{}{\langle\ [\,], [\,], 0, |w|, \emptyset\ \rangle} : \text{Struct}$$

$$\textit{goal} \quad \langle\ S, \Delta, j, j, K\ \rangle : \text{Struct} \quad \exists A \in N : \langle A, \{1, ..., j\}\rangle \in K$$

Structural actions $\mathcal{T}^{\text{MLG}}_{\text{struct}}$

$$\textsc{Shift} \quad \frac{\langle\ S, \Delta, i, j, K\ \rangle : q}{\langle\ S|\Delta, [\{i+1\}], i+1, j, K\ \rangle : \delta(q, \textsc{Shift})} \quad \delta(q, \textsc{Shift}) \text{ is defined}$$

$$\textsc{Merge} \quad \frac{\langle\ S|I_{s_0}, \Delta|I_{d_0}, i, j, K\ \rangle : q}{\langle\ S|\Delta, [I_{s_0} \cup I_{d_0}], i, j, K\ \rangle : \delta(q, \textsc{Merge})} \quad \delta(q, \textsc{Merge}) \text{ is defined}$$

$$\textsc{Gap} \quad \frac{\langle\ S|I_{s_0}, \Delta, i, j, K\ \rangle : q}{\langle\ S, I_{s_0}|\Delta, i, j, K\ \rangle : \delta(q, \textsc{Gap})} \quad \delta(q, \textsc{Gap}) \text{ is defined}$$

Labelling actions $\mathcal{T}^{\text{MLG}}_{\text{label}}$

$$\textsc{Label-X} \quad \frac{\langle\ S, [I_{d_0}], i, j, K\ \rangle : q}{\langle\ S, [I_{d_0}], i, j, K \cup \{\langle X, I_{d_0}\rangle\}\ \rangle : \delta(q, \textsc{Label-X})} \quad \delta(q, \textsc{Label-X}) \text{ is defined}$$

$$\textsc{No-Label} \quad \frac{\langle\ S, [I_{d_0}], i, j, K\ \rangle : q}{\langle\ S, [I_{d_0}], i, j, K\ \rangle : \delta(q, \textsc{No-Label})} \quad \delta(q, \textsc{No-Label}) \text{ is defined}$$

Figure 3.22: Deduction Transitions for GAP-ML with input word $w$ and FSA $\mathfrak{A}^{\text{MLG}}$.

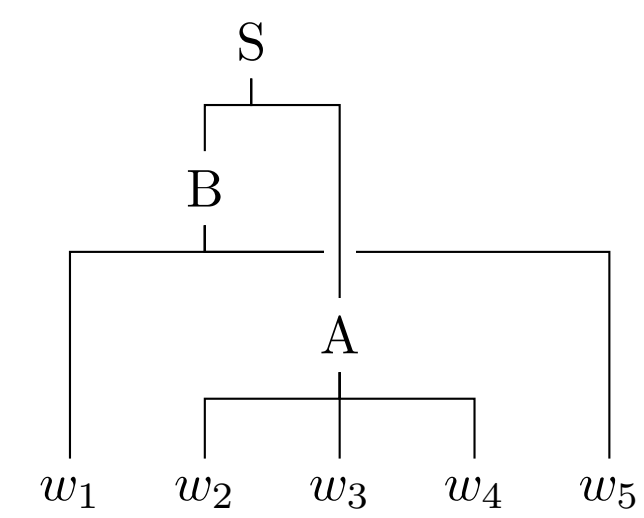

Figure 3.23: Example non-binary tree.

The order extends naturally to instantiated constituents:

$$\langle A_1, s_1\rangle \leq_{\text{right}} \langle A_2, s_2\rangle :\Leftrightarrow s_1 \leq_{\text{right}} s_2$$

For binary trees, there exists a bijection between derivations and trees (Coavoux and Crabbé, 2017a). For nodes with more than two daughters in non-binary trees, there exists more than one implicit binarisation. See for instance steps 4 to 9 in figure 3.24 where instead indices 3 and 4 could have been merged first. Therefore, a standard binarisation must be fixed to define an oracle. Coavoux and Crabbé (2017a) define an eager oracle that performs a leftmost binarisation by prioritizing MERGE in all possible cases. Algorithm 3.6 shows their oracle extraction in full detail.

At each step, there is at most one element on the stack that can be retrieved by use of GAP to merge with the top element. Therefore, the index used in the algorithm is unique.

**Corollary 3.53.**
Let $T$ be an alphabet and $w \in T^+$ a word. Let $K_g$ be a set of instantiated constituents complete for length $|w|$ and $c = \langle\ S, \Delta|d_0, i, |w|, K\ \rangle$ : Label a configuration in a transition sequence for $w$ licensed by the oracle $o$ extracted by algorithm 3.6 for $K_g$. Let $p \in K_g$ with $d_0 \underset{\max}{\subset} p$ be the parent of $d_0$. For any two indices $m, n \in \{1, ..., |S|\}$ such that $S^{[m]} \underset{\max}{\subset} p$ and $S^{[n]} \underset{\max}{\subset} p$ it holds

| step | Stack | Dequeue | index | current state | new constituent | Action |
|---|---|---|---|---|---|---|
| 0 | | | 0 | Struct | | Shift |
| 1 | | $\{1\}$ | 1 | Label | | No-Label |
| 2 | $\{1\}$ | $\{2\}$ | 2 | Struct | | Shift |
| 3 | $\{1\}$ | $\{2\}$ | 2 | Label | | No-Label |
| 4 | $\{1\}, \{2\}$ | $\{3\}$ | 3 | Struct | | Merge |
| 5 | $\{1\}$ | $\{2,3\}$ | 3 | Label | | No-Label |
| 6 | $\{1\}$ | $\{2,3\}$ | 3 | Struct | | Shift |
| 7 | $\{1\}, \{2,3\}$ | $\{4\}$ | 4 | Label | | No-Label |
| 8 | $\{1\}, \{2,3\}$ | $\{4\}$ | 4 | Struct | | Merge |
| 9 | $\{1\}$ | $\{2,3,4\}$ | 4 | Label | | Label-A |
| 10 | $\{1\}$ | $\{2,3,4\}$ | 4 | Struct | $\langle$A, $\{2,3,4\}\rangle$ | Shift |
| 11 | $\{1\}, \{2,3,4\}$ | $\{5\}$ | 5 | Struct | | Gap |
| 12 | $\{1\}$ | $\{2,3,4\}, \{5\}$ | 5 | Struct' | | Merge |
| 13 | $\{2,3,4\}$ | $\{1,5\}$ | 5 | Label | | Label-B |
| 14 | $\{2,3,4\}$ | $\{1,5\}$ | 5 | Struct | $\langle$B, $\{1,5\}\rangle$ | Merge |
| 15 | | $\{1,2,3,4,5\}$ | 5 | Label | | Label-S |
| 16 | | $\{1,2,3,4,5\}$ | 5 | Struct | $\langle$S, $\{1,2,3,4,5\}\rangle$ | |

Figure 3.24: Example parse using the Gap transition for the tree in figure 3.14. Pointers are omitted since the nodes are uniquely labelled.

that $m = n$.

*Proof.* Let us assume, without loss of generality, that $m \leq n$. There must be some configuration $c' = \langle\ S', \Delta'|S^{[m]}, i', |w|, K'\ \rangle : q'$ such that $\sigma(w) \overset{*}{\Rightarrow} c' \overset{*}{\Rightarrow} c$. If $m \neq n$, then $\exists t \in \{1, ..., |S|\}$ : $S^{[n]} = S'^{[t]}$. But since $S^{[m]}$ and $S^{[n]}$ have a common parent in $K_g$, $o(c')$ predicts to merge $S^{[n]}$ and $S^{[m]}$ via $t-1$ Gaps and one Merge transition. Then, neither $S^{[n]}$ nor $S^{[m]}$ can be in $S$. Therefore, it must hold that $m = n$. □

**Complexity** As Coavoux and Crabbé (2017a) note, for an input of length $n$ the number of Shift transitions is $n$ and the number of Merge operations $n-1$. Furthermore, the number of labelling actions is $2n-1$ (one for every Shift and one after each Merge transition). This results in a minimum number of $4n-2$ transitions.

The number of Gap transitions is variable. In the worst case, the first and the last token have a common parent and the nodes in between incrementally form higher-level constituents as visualised in figure 3.25 (Coavoux and Crabbé, 2017a). Let $k = n-2$ be the number of intermediate elements, then the number of Gap actions needed is

$$k + (k-1) + (k-2) + ... + 1 = \frac{k(k+1)}{2} = \frac{(n-2)(n-1)}{2} = \frac{n^2-3n+2}{2} \tag{3.13}$$

Therefore, the total maximum number of transitions is $\frac{n^2+5n-2}{2}$.

Coavoux and Crabbé (2017a) suggest using CompoundGap transitions that perform $i$ consecutive Gaps inspired by the CompoundSwap transition introduced by Maier (2015). Every Merge transition must be preceded by a CompoundGap. $\text{CompoundGap}_0$ has no effect.

When following this proposal, the number of CompoundGaps is $n-1$. Then the total number of actions is exactly $5n-3$. Ensuring that the number of transitions is equal for every input with length $n$ is a way to counteract the bias of the scorer towards long transition sequences as observed by Crabbé (2014).

**Algorithm 3.6** ML-Gap oracle extraction

**Input:** a sequence $w$, a gold set of instantiated constituents $K_g$ complete for length $|w|$
**Output:** an oracle $O$

$c \leftarrow \sigma(w)$
$O \leftarrow List[\,]$
**while** $c \notin F$ **do**
  **if** $c$.state = Struct **then**
    **if** $c$.stack.top $\underset{\max}{\subset} p$ with $c$.dequeue.top $\underset{\max}{\subset} p \in K_g$ **then**
      $c \leftarrow$ MERGE($c$)
      $O$.append(MERGE)
    **else if** $\exists i \in \{1, ..., |c.\text{stack}|\} : c.\text{stack}^{[i]} \underset{\max}{\subset} p$ with $c$.dequeue.top $\underset{\max}{\subset} p \in K_g$ **then**
      **while** $i \neq 1$ **do**
        $c \leftarrow$ GAP($c$)
        $O$.append(GAP)
        $i \leftarrow i - 1$
      **end while**
    **else**
      $c \leftarrow$ SHIFT($c$)
      $O$.append(SHIFT)
    **end if**
  **else**
    **if** $\exists \langle A, s \rangle \in K_g : s = c$.dequeue.top **then**
      $c \leftarrow$ LABEL-A($c$)
      $O$.append(LABEL-A)
    **else**
      $c \leftarrow$ NO-LABEL($c$)
      $O$.append(NO-LABEL)
    **end if**
  **end if**
**end while**
**return** $O$

**Correctness** Coavoux and Crabbé (2017a) proof the correctness of the transition system for the set of discontinuous binary trees. This result can be generalised for all discontinuous trees. Completeness follows from the correctness of the oracle as outlined above. For soundness, it must be shown that any transition sequence produces a discontinuous tree. First, it can be seen that a node cannot have several parents. This follows directly from the fact that a LABEL-X transition must be followed by a structural action which either enlarges the element or shifts it onto the stack from which it can only be retrieved to unify with another non-empty constituent candidate. Therefore, for any valid configuration, $K$ is a consistent constituent set.

Per design of the transition system, at least one action can be performed at any step. The number of transitions has an upper bound. Thus, the algorithm always reaches a final configuration. The set of instantiated constituents in the final configuration spans the length of the input by definition. Furthermore, from corollary 3.50 it follows that $K$ is rooted and thus $K$ is a complete constituent set wrt. the input.

**Comparing with Swap** The main benefit of ML-GAP lies in the extended local region in comparison to shift-reduce parsing with the SWAP transition (compare figures 3.5 and 3.20). The model uses three data structures instead of two and places items onto the deque when reordering. In a SWAP model items are pushed back onto the buffer which is why Coavoux et al. (2019) assume that a SWAP parser may have difficulties differentiating between swapped elements and tokens that have not been shifted yet.

Coavoux and Crabbé (2017a) assume SWAP or GAP transitions are the hardest to predict for the respective system and that they are the driving factor for derivation length. The SWAP transition doubles the number of necessary transitions associated with reordering because it comes with the necessity to shift the swapped element(s) back from the buffer onto the stack in some future step. Shorter derivations are desirable since this reduces the overall uncertainty of the system.

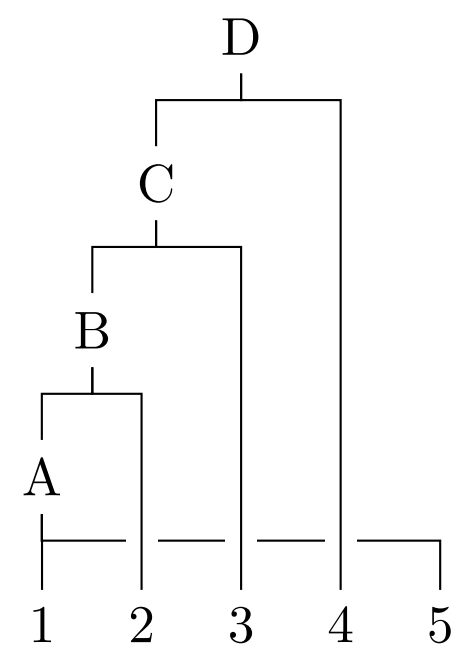

Figure 3.25: Example of worst case tree of length 5 with respect to the Gap transition as given by Coavoux and Crabbé (2017a).

Coavoux and Crabbé (2017a) show that their lexicalised ML-Gap model for binary trees approaches derivation lengths half as long as those of the SR-Swap model for the worst case longest derivation. This result is confirmed empirically: for the TIGER corpus the average derivation length with respect to the sentence length $n$ is $3.09n$ for SR-Swap and $2.03n$ for ML-Gap which amounts to a 50 % derivation length reduction.

Surprisingly the average length of Gap derivation sequences roughly equals that of a standard shift-reduce parser $(2n-1)$ (Coavoux and Crabbé, 2017a).

### 3.2.6 Lexicalisation

Lexicalised models are defined by Coavoux et al. (2019) as models that (a) assign a lexical head to each constituent and (b) use heads of constituents as features to score parsing transitions. They are motivated by lexicalised probabilistic context-free grammars where a nonterminal is annotated with a terminal element representing its lexical head as shown in equation 3.14 (Coavoux et al., 2019).

$$r = \text{VP[saw]} \rightarrow \text{VP[saw] PP[telescope]} \tag{3.14}$$

$q(r)$ then gives the probability of *telescope* modifying *saw*. The general assumption is that information about a constituent's head might help the parser to resolve ambiguities. Following this idea, the original transition-based Shift-Reduce constituent parser of Sagae and Lavie (2005) features two binary Reduce transitions instead of one:

- ReduceLeft-X: the head of the left element is assigned as head of the new constituent
- ReduceRight-X: the head of the right element is assigned as head of the new constituent

While this was a long-time standard approach for transition-based constituent parsers, Coavoux et al. (2019) empirically show that unlexicalised models consistently outperform their lexicalised counterparts. Unlexicalised ML-Gap shows an improvement of 0.1 absolute points on the DPTB and as much as 1.1 and 1.3 absolute points on NeGra and TIGER when compared to ML-Gap with lexicalised Merge transitions. The improvement on discontinuous constituents alone is even greater. Providing the scorer with the predicted heads of the constituents in its domain as a feature has varying effects and in some cases harms the score.

Coavoux et al. (2019) suggest that part of the reason for this result may lie in the head-driven oracle of lexicalised models (cf. the $<_h$-order for SR-Swap in equation 3.5). They hypothesise that derivations given by eager oracles are easier to learn. In particular, they observe that lexicalised models require more complicated chains of transitions when constructing constituents with more than two daughters. Unlexicalised models can use implicit left-branching-binarisation and directly start merging elements as soon as the two topmost items have the same parent while lexicalised models wait until the head of a constituent is shifted and then construct the constituent leading to longer chains of transitions. Additionally, Coavoux et al. (2019) show that this leads to a larger average size of stack and deque. They suggest that a smaller stack size is easer to represent as a feature for the scorer and thus benefits the results of the unlexicalised model.

They also find that the overall number of GAP actions, which they assume to be most difficult to predict, reduces for unlexicalised models. This is due to the fact that the lexicalised model waits for a lexical head if it is discontinuous with respect to its $k$ sisters. Then, it needs to perform $k$ chains of GAP transitions to move its sisters to an adjacent position while the unlexicalised model can immediately merge these $k$ elements and then only performs one chain of GAPs. The number of consecutive GAP actions also reduces since intermediate elements might also belong to a constituent with a discontinuous head. The unlexicalised model can reduce them first to a constituent candidate which lowers the length of the GAP chains necessary for the first constituent. Figure 3.26 illustrates this case.

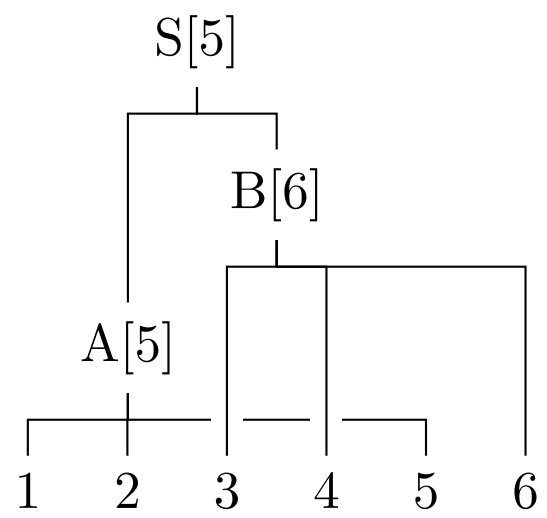

Figure 3.26: Lexicalised discontinuous tree; a head-driven GAP oracle would predict 4 GAP actions to construct A while in an eager unlexicalised approach only 1 GAP action would be needed.

Finally, unlexicalised models are simpler to evaluate for the scorer since they feature fewer transition types. This is in line with the findings of Cross and Huang (2016b) regarding alternating structural and labelling steps to reduce the number of actions for the scorer to choose from at a given step. Additionally, lexicalisation might increase the issue of error propagation. A wrong head assignment that is used as a feature in subsequent steps might encourage further false classifications (Coavoux et al., 2019).

Coavoux et al. (2019) point out that most recent unlexicalised transition-based parsing approaches (Coavoux and Crabbé, 2017b; Cross and Huang, 2016b; Coavoux et al., 2019, among others) are based on a *long short-term memory network (LSTM)*. LSTMs create context-aware representations that enrich individual token representations with information of their surroundings. Coavoux et al. (2019) hypothesise that the LSTM transducer might implicitly learn lexicalised information, which coincides with the result of Kuncoro et al. (2017) that suggests that LSTMs learn head rules when faced with the training objective of jointly learning surface sequences and syntactic trees.

### 3.2.7 Scorers

Transition-based parsers apply actions based on the predictions of a scorer. The scorer receives features associated with certain elements in the configuration. It usually takes the form of a machine learning model that is trained to predict the optimal sequence of transitions given by the oracle.

Versley (2014) use a linear classifier with a set of features that include (among others) POS tags, word forms, lemmas and morphological tags in a two-token window around the two nodes in consideration, as well as the category and POS tag of the leftmost and rightmost dependent of the nodes themselves. Coavoux et al. (2019) successfully use the constituents in the domain of locality of their transition system (stack top, deque top, buffer top) as well as the second topmost element of both stack and deque.

Recent proposals employ a concatenation of word embeddings and character-aware embeddings to represent leaves (Coavoux et al., 2019). The vectors are then enriched with contextual information over the whole input sequence prior to parsing using a recurrent neural network. Constituents are represented using their leftmost and their rightmost leaf representation. Stanojević and Alhama (2017) pioneered using neural networks as a classifier. Most importantly, they utilise the whole configuration as a feature to a recurrent neural network for deciding the next parsing action instead of limiting themselves on the elements in the local region of the shift-reduce framework.

# 4. Parsing with a Stack-Free Transition System

Coavoux and Cohen (2019) present a novel approach to transition-based parsing by replacing the configuration stack with an unordered random-access set. The parser is able to consider every constituent in this memory regardless of its position in the sentence with a single action to construct a new constituent. This omits the use of a SWAP or a GAP transition for reordering and allows for the derivation of a discontinuous tree over an input sequence of length $n$ in $4n-2$ steps.

The role of the stack top in the traditional shift-reduce approach is performed by a separate *focus item*. As in the GAP approach, a buffer of unread tokens is implemented in the form of an index that is counted up when a new token gets retrieved.

The model is a direct advancement of the ML-GAP model of Coavoux et al. (2019) and likewise driven by the idea to enable merging a focus element with any item in the memory without the need to move intermediate items back onto the buffer. It continues the trend of enlarging the domain of locality a transition has access to.

## 4.1. Configurations

**Definition 4.1 (Stack-free configuration).**
Let $N$ be an alphabet of nonterminal labels. $\mathcal{C}^{\mathrm{SC}}$ denotes the set of all elements of the form $\langle\ S, s_f, i, j, K\ \rangle : n$ such that

1. $j \in \mathbb{N}$ with $j > 1$,[24]
2. $S \subseteq ConstCand(j-1)$ is called the *memory of* $c$,
3. $s_f \in ConstCand(j-1)$ is called the *focus item of* $c$,
4. $i \in \{1, ..., j\}$, marks the next index,[25]
5. $K \subseteq Const(N, j-1)$ is a set of instantiated constituents and
6. $n \in \mathbb{N}_0$ marks the step count.

$S$ is the set that replaces the traditional stack while $i$ represents the index of the next token in the buffer. It differs from ML-GAP where it represented the most recent index shifted onto the memory. $s_f$ is a separate focus element. $K$ is the set of constituents labelled so far. A counter $n$ encodes the current step count. Figure 3.5 visualises the configuration. Note the enlarged local region when compared to the SR-SHIFT and the ML-GAP configurations in figures 3.5 and 3.20.

## 4.2. Transitions

A set of deduction rules dictates the construction of new items. SHIFT takes the next item off the buffer and makes the one-item set containing it the new focus item. The old focus item is added to the memory set. COMBINE-$s$ removes $s$ from the set and forms a union with the focus item creating a new focus item. The LABEL-X transition is used to declare the focus item $s_f$ an instantiated constituent $\langle X, s_f\rangle$ with some label X. NO-LABEL has no effect. The full list of deductions is given in figure 4.2.

The unordered set allows for the focus item to be combined with any possibly non-adjacent constituent candidate present in the memory. Therefore, there is no need for a reordering transition.

[24] Just like in definition 3.45 I include the desired maximal index as a component of the configuration since I define the set of configurations for all possible input sequences. Therefore, a configuration must contain information as to the number of elements left to shift.

[25] Coavoux and Cohen (2019) start indexing sequences with 0. I adjusted the definition to the convention of 1 in this work.

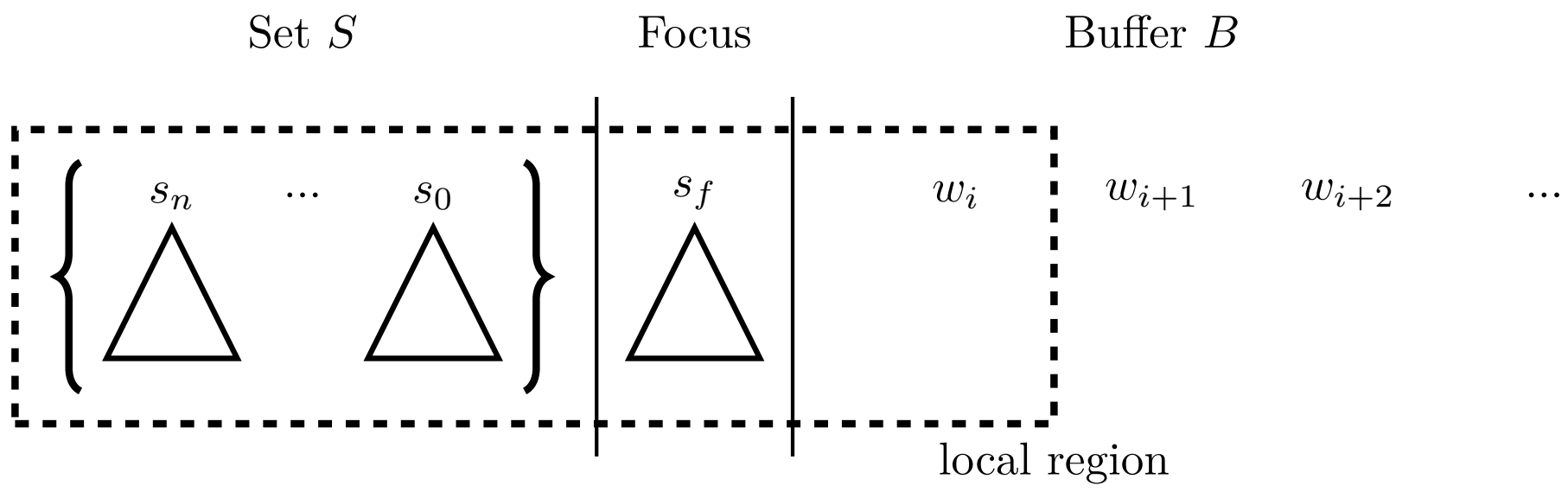

Figure 4.1: Illustration of the stack-free configuration (without the set of initiated constituents). Reproduction of an illustration in Coavoux and Cohen (2019).

$$\textit{axiom} \quad \frac{}{\langle\ \emptyset, \emptyset, 1, |w|+1, \emptyset\ \rangle : 0}$$

$$\textit{goal} \quad \langle\ S, s_f, j, j, K\ \rangle : n \qquad \exists A \in N : \langle A, \{1, ..., j\}\rangle \in K$$

Structural actions $\mathcal{T}^{\text{SC}}_{\text{struct}}$

$$\textsc{Shift} \quad \frac{\langle\ S, s_f, i, j, K\ \rangle : n}{\langle\ S \cup \{s_f\}, \{i\}, i+1, j, K\ \rangle : n+1} \quad i < j,\ n \text{ is even}$$

$$\textsc{Combine-s} \quad \frac{\langle\ S, s_f, i, j, K\ \rangle : n}{\langle\ S \setminus \{s\}, s_f \cup s, i, j, K\ \rangle : n+1} \quad s \in S,\ n \text{ is even}$$

Labelling actions $\mathcal{T}^{\text{SC}}_{\text{label}}$

$$\textsc{Label-X} \quad \frac{\langle\ S, s_f, i, j, K\ \rangle : n}{\langle\ S, s_f, i, j, K \cup \{\langle\ X, s_f\ \rangle\}\ \rangle : n+1} \quad n \text{ is odd}$$

$$\textsc{No-Label} \quad \frac{\langle\ S, s_f, i, j, K\ \rangle : n}{\langle\ S, s_f, i, j, K\ \rangle : n+1} \quad (i \neq j \vee S \neq \emptyset), n \text{ is odd}$$

Figure 4.2: Transitions for the set-free system, $\mathcal{T}^{\text{SC}} = \mathcal{T}^{\text{SC}}_{\text{struct}} \cup \mathcal{T}^{\text{SC}}_{\text{label}}$.

The need for a No-Label-transition arises from the fact that Coavoux and Cohen (2019) divide the transitions into *structural* (Shift, Combine-$s$) and *labelling* (Label-X, No-Label) actions. Contrary to the Gap proposal described in section 3.2.5 and true to the original proposal by Cross and Huang (2016b), a structural action is always followed by a labelling action and vice versa. Therefore, a configuration carries a step counter that, if it is even, allows for a structural transition to follow, or if it is odd, allows for a labelling transition.

**Definition 4.2 (Stack-free transition system).**
A stack-free or Shift-Combine transition system for an alphabet $T$ and an alphabet of nonterminals $N$ is a 5-tuple $\langle \mathcal{C}^{\text{SC}}, \mathcal{T}^{\text{SC}}, \sigma, F, d\rangle$ where

1. $\mathcal{T}^{\text{SC}}$ is defined on $\mathcal{C}^{\text{SC}}$ as shown in figure 4.2,
2. $\sigma(w) = \sigma'(|w|) = \langle\ \emptyset, \emptyset, 1, |w|+1, \emptyset\ \rangle : 0$ for every $w \in T^+$,
3. $F = \{\langle\ S, s_f, i, j, K\ \rangle : n \in \mathcal{C}^{\text{SC}} \mid i = j \wedge \exists A \in N : \langle A, \{1, ..., i\}\rangle \in K\}$[26] and
4. $d$ retrieves the tree from $K$ for every $\langle\ S, s_f, i, j, K\ \rangle : n \in F$.

**Corollary 4.3.**
Let $c = \langle S, s_f, i, j, K\rangle : n$ be a valid configuration.

[26] Coavoux and Cohen (2019) do not require that $K$ contain an instantiated constituent spanning $\{1, ..., i\}$ but instead that $j = 4n - 2$. One follows from the other as can be seen in section 4.4.

1. It holds that $\max(s_f) = i - 1$.[27]
2. Let $c' = \langle S', s'_f, i', j', K' \rangle : n' \in \mathcal{C}^{\text{SC}}$ be a configuration such that $c \Rightarrow c'$. It holds that $\max(s_f) \leq \max(s'_f)$.

*Proof.* Part 1. follows from the fact that the focus item cannot be swapped with a set item. It can only be enlarged with items from the set or shifted. Therefore, given the first buffer item $i$, $i-1$ is always a member of the focus item, and therefore $\max(s_f) = i - 1$. Now, if Shift is performed, the new maximum focus element is larger than the preceding one. If another action is performed, it remains equal. Hence, 2. is true. □

**Corollary 4.4.**
For every valid configuration $c = \langle S, s_f, i, j, K \rangle : n$ it holds that $\forall s \in S : \max(s) < s_f$.

*Proof.* For every $s \in S$ there must exists a configuration $c' = \langle S', s, i', j, K' \rangle : n' \in \mathcal{C}^{\text{SC}}$ such that $c' \underset{\text{Shift}}{\Rightarrow} \ldots \overset{*}{\Rightarrow} c$. Therefore, following corollary 4.3, $\max(s) \leq \max(s_f)$. Furthermore, since Shift was applied, $i' < i$ and thus $\max(s) < \max(s_f)$. □

**Corollary 4.5.**
Given an alphabet of nonterminals $N$ and valid configurations $c_1 = \langle\ S_1, s_f, i_1, j_1, K_1\ \rangle : n_1,\ c'_1 = \langle\ S_1, s_f, i_1, j, K_1 \cup \{\langle A, s_f \rangle\}\ \rangle : n_1 + 1$ with $A \in N$ such that $c_1 \underset{\text{Label-A}}{\Rightarrow} c'_1$, there are no configurations $c_2 = \langle\ S_2, s_f, i_2, j, K_2\ \rangle : n_2,\ c'_2 = \langle\ S_2, s_f, i_2, j, K_2 \cup \{\langle A', s_f \rangle\}\ \rangle : n_2 + 1$ with $A' \in N$ such that $c'_1 \overset{*}{\Rightarrow} c_2 \underset{\text{Label-A'}}{\Rightarrow} c'_2$, i.e. no set can be assigned a label twice.

*Proof.* Let us assume that $c \underset{\text{Label-A}}{\Rightarrow} c' \overset{*}{\Rightarrow} c_2 \underset{\text{Label-A'}}{\Rightarrow} c'_2$. Label-$A$ must be followed by a structural transition. If Merge-$s$ follows, then $s_f$ gets merged with some non-empty $s$ and Label-$A'$ cannot be applied to $s_f$ any more. If Shift follows, the maximum focus index gets increased by one and for any following configuration, it does not decrease to $\max(s_f)$ (corollary 4.3). Therefore, $c_2$ is not derivable from $c'$. □

**Example 4.6.**
Figure 4.3 shows an example parse. Note the ambiguity: the constituent A could have also been constructed starting with $\{3, 4\}$, i.e. by performing the sequence Shift, No-Label, Combine-$\{3\}$, No-Label, Combine-$\{2\}$ starting at step 6.

## 4.3. Oracle

Coavoux and Cohen (2019) present a static oracle as well as a dynamic oracle. The latter represents the first dynamic oracle suggested for transition-based discontinuous constituent parsing.

### 4.3.1 Static Oracle

Coavoux and Cohen (2019) design an eager static oracle with a priority on applying Combine-$s$ to keep the number of elements in the set as small as possible. A smaller memory means fewer choices for Combine-X actions in the future which facilitates decisions and makes scoring faster.

Algorithm 4.1 describes the static oracle extraction for the stack-free-transition system. It is a direct successor of the ML-Gap oracle given in algorithm 3.6. Instead of a chain of Gap transitions followed by a Merge it simply predicts one Combine-$s$ transition. Corollary 3.53 translates to the stack-free approach: there is at most one sister to combine with the focus item present in the set at any step.

[27] Coavoux and Cohen (2019) make this a condition on all stack-free configurations.

| step | set | focus | index | new constituent | action |
|---|---|---|---|---|---|
| 0 | | | 1 | | SHIFT |
| 1 | | 1 | 2 | | NO-LABEL |
| 2 | | 1 | 2 | | SHIFT |
| 3 | $\{1\}$ | 2 | 3 | | NO-LABEL |
| 4 | $\{1\}$ | 2 | 3 | | SHIFT |
| 5 | $\{1\}, \{2\}$ | 3 | 4 | | NO-LABEL |
| 6 | $\{1\}, \{2\}$ | 3 | 4 | | COMBINE-$\{2\}$ |
| 7 | $\{1\}$ | $2, 3$ | 4 | | NO-LABEL |
| 8 | $\{1\}$ | $2, 3$ | 4 | | SHIFT |
| 9 | $\{1\}, \{2, 3\}$ | 4 | 5 | | NO-LABEL |
| 10 | $\{1\}, \{2, 3\}$ | 4 | 5 | | COMBINE-$\{2, 3\}$ |
| 11 | $\{1\}$ | $2, 3, 4$ | 5 | | LABEL-A |
| 12 | $\{1\}$ | $2, 3, 4$ | 5 | $\langle \mathrm{A}, \{2, 3, 4\}\rangle$ | SHIFT |
| 13 | $\{1\}, \{2, 3, 4\}$ | 5 | 6 | | NO-LABEL |
| 14 | $\{1\}, \{2, 3, 4\}$ | 5 | 6 | | COMBINE-$\{1\}$ |
| 15 | $\{2, 3, 4\}$ | $1, 5$ | 6 | | LABEL-B |
| 16 | $\{2, 3, 4\}$ | $1, 5$ | 6 | $\langle \mathrm{B}, \{1, 5\}\rangle$ | COMBINE-$\{2, 3, 4\}$ |
| 17 | | $1, 2, 3, 4, 5$ | 6 | | LABEL-S |
| 18 | | $1, 2, 3, 4, 5$ | 6 | $\langle S, \{1, 2, 3, 4, 5\}\rangle$ | |

Figure 4.3: Example parse using the stack-free transition system for the tree in figure 3.14. The columns *set* and *focus* represent sets with the outer brackets being are omitted.

### 4.3.2 Dynamic Oracle

A dynamic oracle predicts the best actions to perform for any valid configuration — not only the configurations part of the gold path to the desired tree. This is motivated by the fact that training a neural classifier only on the gold scenario limits the quality of information provided at training. When taking a wrong turn by choosing a bad transition the parser should still be able to construct the best possible parse tree even if it is no longer on the path to the perfect prediction.

This directly addresses the issue of *error propagation*. The term refers to a common problem with incremental models where predictions are dependent on previous decisions. It describes the tendency of an initial erroneous assignment to evoke more errors in subsequent steps. McDonald and Nivre (2007) first observed this tendency in transition-based parsers.

The idea of a dynamic oracle was first brought forward by Goldberg and Nivre (2012b) for dependency parsing. Their oracle outputs a set of best actions to perform and is therefore non-deterministic. Dynamic oracles for projective constituent parsing were proposed by a number of authors, for instance Cross and Huang (2016b). The dynamic oracle for discontinuous constituent parsing proposed by Coavoux and Cohen (2019) is an extension of this work. The following prerequisites are needed.

**Definition 4.7 (Reachability).**
Let $N$ be an alphabet of nonterminal labels.

1. Let $c = \langle\ S, s_f, i, j, K\ \rangle : n \in \mathcal{C}^{\mathrm{SC}}$ be a configuration. An instantiated constituent $\langle A, s\rangle \in$ *Const*$(N)$ such that $\langle A, s\rangle \notin K$ is said to be *reachable* from $c$ iff there exists a configuration $c' = \langle\ S', s'_f, i', j, K'\ \rangle$ such that $\langle A, s\rangle \in K'$ and $c \overset{*}{\Rightarrow} c'$.[28]

[28]In contrast to Coavoux and Cohen (2019) I explicitly require that $c \notin K$. This facilitates the following definitions.

**Algorithm 4.1** Stack-free static oracle extraction

```
Input: a word w, a gold set of instantiated constituents K_g complete for length |w|
Output: an oracle O
c ← σ(w)
O ← List[ ]
while c ∉ F do
    if c.step is even then
        if ∃s ∈ c.memory : s ⊂_max p with c.focus ⊂_max p ∈ K_g then
            c ← COMBINE-s(c)
            O.append(COMBINE-s)
        else
            c ← SHIFT(c)
            O.append(SHIFT)
        end if
    else
        if ∃⟨A, s⟩ ∈ K_g : c.focus = s then
            c ← LABEL-A(c)
            O.append(LABEL-A)
        else
            c ← NO-LABEL(c)
            O.append(NO-LABEL)
        end if
    end if
end while
return O
```

2. The function *reach* is defined as follows:

$$reach : \mathcal{C}^{\text{SC}} \times \mathcal{P}(Const(N)) \to \mathcal{P}(Const(N)),$$
$$\langle c, K \rangle \mapsto \{k \in K \mid k \text{ is reachable from } c\}.$$

In other words, for a set of instantiated constituents $K$ wrt. $N$ in any length, *reach* gives all of the constituents in $K$ that are reachable from a configuration $\langle\ S, s_f, i, j, K\ \rangle : n$. It is straightforward to see that only constituents in length $j-1$ are reachable.

**Corollary 4.8.**
Let $N$ be an alphabet of nonterminal labels, $\langle A, s\rangle, \langle A', s'\rangle \in Const(N, j-1)$ for some $j \in \mathbb{N}$ instantiated constituents with $\langle A, s\rangle \leq_{\text{right}} \langle A', s'\rangle$ and $s \neq s'$ and $c_b = \langle\ S_b, s_b, i_b, j, K_b\ \rangle : n_b \in \mathcal{C}^{\text{SC}}$ a valid configuration with $\langle A, s\rangle, \langle A', s'\rangle \in K_b$. Then it follows that there exist valid configurations $\langle\ S, s, i, j, K\ \rangle : n, \langle\ S', s', i', j, K'\ \rangle : n' \in \mathcal{C}^{\text{SC}}$ such that $\langle\ S, s, i, j+1, K\ \rangle : n \underset{\text{LABEL-A}}{\Rightarrow} \langle\ S, s, i, j+1, K \cup \{\langle A, s\rangle\}\ \rangle : n+1 \overset{*}{\Rightarrow} \langle\ S', s', i', j+1, K'\ \rangle : n' \underset{\text{LABEL-A'}}{\Rightarrow} \langle\ S', s', i', j+1, K' \cup \{\langle A', s'\rangle\}\ \rangle : n'+1 \overset{*}{\Rightarrow} c_b$. In other words: $\langle A, s\rangle$ must be constructed before $\langle A', s'\rangle$.

*Proof.* Let us assume the contrary: there is a transition sequence that contains $c_b$, that does not fulfil this property. By definition of the transitions in figure 4.2: if $\langle A, s\rangle, \langle A', s'\rangle \in K_b$, there must exist $c = \langle\ S, s, i, j+1, K\ \rangle : n$, $c_l = \langle\ S, s, i, j+1, K \cup \{\langle A, s\rangle\}\rangle : n+1$, $c' = \langle\ S', s', i', j+1, K'\ \rangle : n'$ and $c'_l = \langle\ S', s', i', j, K' \cup \{\langle A' s'\rangle\}\ \rangle : n'+1$ such that $\sigma'(j-1) \overset{*}{\Rightarrow} c \underset{\text{LABEL-A}}{\Rightarrow} c_l \overset{*}{\Rightarrow} c_b$ and $\sigma'(j-1) \overset{*}{\Rightarrow} c' \underset{\text{LABEL-A'}}{\Rightarrow} c'_l \overset{*}{\Rightarrow} c_b$. By assumption, it cannot be that $c_l \overset{*}{\Rightarrow} c'$, so either 1. the variables denote the same transition or 2. $\sigma'(j-1) \overset{*}{\Rightarrow} c' \underset{\text{LABEL-}A'}{\Rightarrow} c'_l \overset{*}{\Rightarrow} c \underset{\text{LABEL-A}}{\Rightarrow} c_l \overset{*}{\Rightarrow} c_b$.

1. This contradicts the corollary assumption that $s \neq s'$.
2. Following corollary 4.3 it must hold that $\max(s') \leq \max(s)$. Now, there are two cases: (a) $\max(s') < \max(s)$ or (b) $\max(s') = \max(s)$

(a) This contradicts the assumption that $\langle A, s\rangle \leq_{\text{right}} \langle A', s'\rangle$.

(b) First, $i = i'$ (corollary 4.3). By definition 3.52 it must hold that $s \subseteq s'$. Since $s \neq s'$ by assumption it follows that $s \subset s'$. Thus, there exists $x \in s'$ such that $x \notin s$. This contradicts with $c' \overset{*}{\Rightarrow} c$ since there is no structural action to achieve this. Thus, this also contradicts the assumptions and the corollary is true.

□

In their statement about the property in corollary 4.8, Coavoux and Cohen (2019) do not require that $s \neq s'$. But $s = s'$ would trivially mean that the variables denote the same nonterminal and the same configuration since no set can be assigned a label twice (corollary 4.5).

Coavoux and Cohen (2019) note the following:

**Corollary 4.9.**
Given an alphabet of nonterminals $N$, a valid configuration $c = \langle\ S, s_f, i, j, K\ \rangle : n \in \mathcal{C}^{\text{SC}}$ where $n$ is odd and an instantiated constituent $\langle A, s_g\rangle \in \mathit{Const}(N, j-1)$ with $\langle A, s_g\rangle \notin K$. $\langle A, s_g\rangle$ is reachable from $c$ iff:

1. $\max(s_f) \leq \max(s_g)$ and
2. $\forall s \in S \cup \{s_f\} : (s \subseteq s_g)$ or $(s \cap s_g = \emptyset)$.

*Proof.*

(⇒) From the reachability of $\langle A, s_g\rangle$ follows that there exists a configuration $c' = \langle\ S', s'_f, i', j', K'\ \rangle : n'$ such that $\langle A, s_g\rangle \in K'$ and $c \overset{*}{\Rightarrow} c'$. Since $\langle A, s_g\rangle \notin K$, it follows that there exist configurations $c_g = \langle\ S_g, s_g, i_g, j, K_g\ \rangle : n_g$ and $c_{g'} = \langle\ S_g, s_g, i_g, j, K_g \cup \{s_g\}\ \rangle : n_g + 1$ such that $c \overset{*}{\Rightarrow} c_g \underset{\text{LABEL-A}}{\Rightarrow} c_{g'}$. Condition 1. now follows directly from corollary 4.3. Furthermore, if there was an element $s \in S \cup \{s_f\}$ such that there was an index $x \in s \cap s_g$ and an index $y \in s \setminus s_g$, $\langle A, s_g\rangle$ would not be reachable from $c$, since MERGE-$s$ would be necessary to build $s_g$ but MERGE-$s$ would also make $y$ an element in the focus set. Therefore condition 2 must also hold.

(⇐) Either (a) $\max(s_f) = \max(s_g)$ or (b) $\max(s_f) < \max(s_g)$.

(a) By condition 2 it must hold that $s_f \subseteq s_g$. Let $\{y_1, ..., y_m\} = s_g \setminus s_f$. No element in $s_g \setminus s_f$ was introduced by a SHIFT following $c$ since $\max(s_f) = \max(s_g)$. Therefore it holds that $\exists Y_z \in S : y_z \in Y_z$ for all $z \in \{1, ..., m\}$. Due to condition 2, it must hold that $Y_z \subseteq s_g$ for every $z \in \{1, ..., m\}$. It holds that $s_f \cup Y_1 \cup ... \cup Y_m = s_g$. One can construct a path: $c \underset{\lambda_1}{\Rightarrow} ... \underset{\text{COMBINE-}Y_1}{\Rightarrow} \langle\ S \setminus \{Y_1\}, s_f \cup Y_1, i, j, K'\ \rangle : n+2 \underset{\lambda_2}{\Rightarrow} ... \underset{\text{COMBINE-}Y_m}{\Rightarrow}$ $\langle\ S \setminus \{Y_1, ..., Y_m\}, s_g, i, j, K''\ \rangle : n + 2m \underset{\text{LABEL-}A}{\Rightarrow} \langle\ S \setminus \{Y_1, ..., Y_m\}, s_g, i, j, K \cup \langle A, s_g\rangle\ \rangle : n + 2m + 1$ with $\lambda_1, ..., \lambda_m \in \mathcal{T}^{\text{SC}}_{\text{label}}$. Therefore $\langle A, s_g\rangle$ is reachable from $c$. If $s_f = s_g$, then simply $m = 0$.

(b) For all $s \in S$ it must hold that $\max(s) < \max(s_f)$ (corollary 4.4). Therefore, all elements $i, i+1, ..., \max(s_g)$ must be shifted from the buffer to construct $s_g$. Now construct a path: $c \underset{\lambda_1}{\Rightarrow} c' \underset{\text{SHIFT}}{\Rightarrow} \langle\ S \cup \{s_f\}, \{i\}, i+1, j, K'\ \rangle : n+2 \underset{\lambda_2}{\Rightarrow} ... \underset{\text{SHIFT}}{\Rightarrow}$ $\langle\ S \cup \{s_f, \{i\}, ..., \{\max(s_g) - 1\}\}, \{\max(s_g)\}, \max(s_g) + 1, j, K''\ \rangle : n + 2(\max(s_g) - \max(s_f)) = c'$ with $\lambda_1, ..., \lambda_{\max(s_g)-\max(s_f)} \in \mathcal{T}^{\text{SC}}_{\text{label}}$. It holds that $\forall s \in \{\{i\}, ..., \{\max(s_g) - 1\}\} : (s \subseteq s_g)$ or $s \cap s_g = \emptyset$ since $s$ has only one element. Now, for $c'$ it follows by case (a), that $\langle A, s_g\rangle$ is reachable from $c'$. Therefore $\langle A, s_g\rangle$ is also reachable from $c$.

□

**Corollary 4.10.**
Let $N$ be an alphabet of nonterminals, and $c = \langle\ S, s_f, i, j, K\ \rangle : n \in \mathcal{C}^{\mathrm{SC}}$ a valid configuration where $n$ is even. Let $\langle A, s_g\rangle \in \mathit{Const}(N, j-1)$ be an instantiated constituent with $\langle A, s_g\rangle \notin K$. $\langle A, s_g\rangle$ is reachable from $c$ iff:

1. $\max(s_f) \leq \max(s_g)$,
2. $\forall s \in S \cup \{s_f\} : (s \subseteq s_g)$ or $(s \cap s_g = \emptyset)$ and
3. $s_f \neq s_g$.

*Proof.* This is straightforward to see following corollary 4.9 in combination with the fact that the current focus item cannot be labelled since only Shift and Combine are allowed. ☐

Following Cross and Huang (2016b), Coavoux and Cohen (2019) define a function to retrieve the smallest (with respect to $\leq_{\mathrm{right}}$) reachable gold constituent. I adapted the function to the notation used here:

**Definition 4.11 (Next function).**
Let $N$ be an alphabet of nonterminal labels. The function *next* is defined as follows:

$$\{\langle c, K_g\rangle \mid \exists m \in \mathbb{N} : \sigma'(m) \overset{*}{\Rightarrow} c, c \notin F, K_g \subseteq \mathit{Const}(N, m), K_g \text{ is complete for } m\} \to \mathit{Const}(N)$$

$$\langle c, K_g\rangle \mapsto \underset{\leq_{\mathrm{right}}}{\arg\min}\ \mathit{reach}(c, K_g)$$

Restricting the function to valid but not final configurations for an input length $m$ and to sets of constituents complete for length $m$ guarantees that at least $\{1, ..., m\} \in \mathit{reach}(c, K_g)$. Furthermore, since $K_g$ is consistent, the elements in $\mathit{reach}(c, K_g)$ are pairwise-compatible and therefore a minimal element exists. Thus, *next* is well-defined.

Now, the oracle is defined separately for odd and for even steps:

**Definition 4.12 (Dynamic oracle odd).**
Let $N$ be an alphabet of nonterminals, $m \in \mathbb{N}$ an integer, $K_g$ a complete set of instantiated constituents wrt. $N$ for length $m$ and $c = \langle\ S, s_f, i, m+1, K\ \rangle : n$ a configuration such that $n$ is odd and $\sigma'(m) \overset{*}{\Rightarrow} c$.

$$o_{odd}(c, K_g) = \begin{cases} \{\textsc{Label-}A\}, & \text{if } \exists\langle A, s_f\rangle \in K_g, \\ \{\textsc{No-Label}\}, & \text{otherwise.} \end{cases}$$

**Definition 4.13 (Dynamic oracle even).**
Let $N$ be an alphabet of nonterminals, $m \in \mathbb{N}$ an integer, $K_g$ a complete set of instantiated constituents wrt. $N$ for length $m$ and $c = \langle\ S, s_f, i, m+1, K\ \rangle : n$ a configuration such that $n$ is even and $\sigma'(m) \overset{*}{\Rightarrow} c$.

$$o_{even}(c, K_g) = \begin{cases} \{\textsc{Combine-}s \mid (s_f \cup s) \subseteq s'_f\}, & \text{if } \max(s'_f) = \max(s_f), \\ \{\textsc{Combine-}s \mid (s_f \cup s) \subseteq s'_f\} \cup \{\textsc{Shift}\}, & \text{if } \max(s'_f) > \max(s_f), \end{cases}$$

where $\mathit{next}(c, K_g) = \langle A, s'_f\rangle$.

For training, Coavoux and Cohen (2019) make the oracle deterministic by choosing Combine at an even step if the oracle allows for it. This is done to keep the stack as small as possible which reduces the number of elements future Combine actions need to evaluate. If several Combine actions are possible, the element with the highest right index is merged. This is done to reduce the number of non-adjacent parts in the new set which could potentially be impossible to represent uniquely by the four index constituent embedding representation (see section 4.5.2).

### 4.3.3 Oracle Correctness

Coavoux and Cohen (2019) proof that the oracle described in definitions 4.12 and 4.13 always leads to the reachable tree with the highest *F-score* with respect to a gold tree for any configuration. They give the following definitions for *precision, recall* and *F-score*:

**Definition 4.14 (F, precision, recall).**
Let $T$ be an alphabet of terminals, $N$ an alphabet of nonterminals, $w \in T^+$ a word and $K_g, \hat{K}$ sets of instantiated constituents wrt. $N$ complete for length $|w|$. $K_g$ is the gold tree and $\hat{K}$ the predicted tree.

$$\text{precision}(\hat{K}, K_g) = \frac{|\hat{K} \cap K_g|}{|\hat{K}|}$$

$$\text{recall}(\hat{K}, K_g) = \frac{|\hat{K} \cap K_g|}{|K_g|}$$

$$\text{F}_1(\hat{K}, K_g) = \frac{2 \cdot \text{precision}(\hat{K}, K_g) \cdot \text{recall}(\hat{K}, K_g)}{\text{precision}(\hat{K}, K_g) + \text{recall}(\hat{K}, K_g)}$$

Precision, recall and $\text{F}_1$-score are common evaluation metrics for information retrieval. Precision measures the percentage of identified constituents that are actually correct (i.e. that are gold constituents). In other words: it rises when the number of *false positives* goes down. Recall measures the percentage of gold constituents that were identified by the parser. It increases when the number of *false negatives* reduces. $\text{F}_1$-score is a means to combine the two metrics into one measure (Jurafsky and Martin, 2009, chapter 13.5.3).

These concepts are related to the notions of soundness, completeness and correctness introduced in section 3.1.5. Proving that the dynamic oracle is optimal for precision and recall also entails that the oracle is correct in a static scenario. A proof can be performed by showing optimality both for odd and even steps.

Coavoux and Cohen (2019) give the following proof: By definition, for a configuration $c = \langle S, s_f, i, j, K \rangle : n$ and a gold tree $K_g$, $o_{odd}(c, K_g)$ is optimal for precision because it only predicts a LABEL-$A$-action if the constituent $\langle A, s_f \rangle$ is actually in the gold tree $K_g$. It does not add to the number of false positives. Furthermore, it is optimal for recall since, if the current focus is a gold constituent, it will assign a label to it. It does not add to the number of false negatives by omitting a label.

$o_{even}(c, K_g)$ is optimal for precision, since it does not label new constituents at all, so it cannot add to the number of false positives. For its relationship with recall, the question stands whether its prediction can have the effect of prohibiting reaching a gold constituent. Coavoux and Cohen (2019) show the following:

**Corollary 4.15.**
Given an alphabet of nonterminals $N$, a valid configuration $c = \langle\ S, s_f, i, j, K\ \rangle : n$ such that $n$ is even and a set of instantiated constituents complete for length $j-1$ called $K_g$, it holds that $\forall k_g \in K_g : k_g \in \text{reach}(c, K_g) \Rightarrow (\forall c' \in o_{even}(c, K_g) : k_g \in \mathit{reach}(c', K_g))$, i.e. after any transition in the oracle's even prediction, $k_g$ will still be reachable.

*Proof.* Assume that $\text{next}(c, K_g) = \langle A', s_h \rangle$ and that $\langle A, s_g \rangle \in K_g$ is a constituent reachable from $c$. There are two cases to explore, based on the definition of $o_{even}$: 1. SHIFT or 2. COMBINE-$s$.

1. If SHIFT $\in o_{even}(c, K_g)$, then $\max(s_h) > \max(s_f)$. Following corollary 4.10, it holds that $\max(s_f) \leq \max(s_g)$ as well as $\forall s \in S \cup \{s_f\} : (s \subseteq s_g)$ or $(s \cap s_g = \emptyset)$. Furthermore $s_g \neq s_f$. Let us assume that $\langle A, s_g \rangle$ is not reachable from SHIFT$(c) = \langle\ S \cup \{s_f\}, \{i\}, i+1, j, K\ \rangle : n+1$. Then either (a) $\max(\{i\}) > \max(s_g)$ or (b) $\exists s \in S \cup \{s_f, \{i\}\} : (s \nsubseteq s_g)$ and $(s \cap s_g \neq \emptyset)$ (corollary 4.9).

(a) This entails that $\max(s_g) = \max(s_f) < \max(s_h)$. But then $s_g \leq_{\text{right}} s_h$ which contradicts the assumption that $\text{next}(c, K_g) = \langle A', s_h \rangle$.

(b) Per assumption this can only be true for $\{i\}$. From $\{i\} \cap s_g \neq \emptyset$ follows $\{i\} \subseteq s_g$. This contradicts $\{i\} \nsubseteq s_g$. Therefore, $\langle A, s_g \rangle$ must be reachable from Shift$(c)$.

2. For any Combine-$s_b \in o_{even}(c, K_g)$: per oracle definition it holds that $(s_f \cup s_b) \subseteq s_h$. Since $\langle A', s_h \rangle$ is also reachable from $c$, it must hold that $s_g$ and $s_h$ are compatible or $s_g = s_h$. Four cases:

(a) $s_g = s_h$. Then $\langle A, s_g \rangle$ is trivially also reachable from Combine-$s_b(c)$.

(b) $s_h \subset s_g$. Therefore, also $\forall s \in (S \setminus \{s_b\}) \cup \{s_f \cup s_b\} : (s \subseteq s_g)$ or $(s \cap s_g = \emptyset)$. Therefore, by corollary 4.9, $\langle A, s_g \rangle$ is still reachable from Combine-$s_b(c)$.

(c) $s_g \subset s_h$. It follows that $s_g \leq_{\text{right}} s_h$. But then, $next(c, K_g) \neq \langle A', s_h \rangle$ since $s_g$ is smaller than $s_h$ which contradicts the definition of $s_h$.

(d) $s_g \cap s_h = \emptyset$. Then also $s_f \cap s_g = \emptyset$ and $s \cap s_g = \emptyset$. Therefore $\forall s \in (S \setminus \{s_b\}) \cup \{s_f \cup s_b\} : (s \subseteq s_g) \vee (s \cap s_g = \emptyset)$ and thus, by corollary 4.9, $\langle A, s_g \rangle$ is still reachable from Combine-$s_b(c)$.

□

## 4.4. Complexity

Given a word $w = w_0 ... w_n$ and a transition sequence, for every element in $\{1, ..., n\}$, four actions are applied: It is shifted to become the focus item, then a labelling transition is performed. Furthermore, each Shift puts the previous focus item into the memory from which it is retrieved at some future point by Merge. Then, another labelling step follows. Only for the focus element identified by $\max(s_f) = |w|$, the second pair of steps is not present, since it is never put into the memory. When all these actions have been performed, the focus element consists of $\{1, ..., n\}$ and a labelling action has established the root. Therefore one can construct a tree for a sentence of length $n$ in $4n - 2$ steps.

As Coavoux and Cohen (2019) note, for structural steps, $|S|+1$ different transitions are possible. In the worst case, $S$ contains $n-1$ items (i.e. when no Merge has been performed and all items in $\{1, ..., n\}$ have been shifted). For labelling steps, $|N| + 1$ different actions can be predicted. Thus, the time complexity is $\mathcal{O}(n(n-1) + n(|N|+1)) = \mathcal{O}(n(|N|+n))$ (Coavoux and Cohen, 2019). This is a slight downgrade in comparison to the time complexity of ML-Gap (cf. section 3.2.5). In practice, fast matrix-multiplication for one-step evaluation of all Combine actions with elements in the memory mitigates this disadvantage.

## 4.5. Implementation

Coavoux and Cohen (2019) implement their parsing algorithm using the Python programming language. The action scorer, trained to predict the optimal action according to the oracle, is a neural network model built with the Pytorch library (Paszke et al., 2017). From this arises the question of how to represent tokens and sets of tokens in a form usable by the neural model. Furthermore, an objective function must be set. Additionally, Coavoux and Cohen (2019) include a POS-tagging component.

### 4.5.1 Token Representations

Coavoux and Cohen (2019) follow previous proposals for constituency and dependency parsing like Cross and Huang (2016a) and Kiperwasser and Goldberg (2016) that suggest using a recurrent neural network (RNN) to construct context-aware representations for the input tokens. In the following, I will deviate slightly from Coavoux and Cohen (2019) in favour of the notation used in Goldberg (2022).

Given a sequence of input tokens $w_{1:n} = w_1, ..., w_n$, Coavoux and Cohen (2019) first compute a representation for each token using a concatenation of a character-aware embedding and a word embedding.

**Word Embedding** When encoding a set of $k$ distinct categorial features, it is common to associate each of these features with a dense vector representation called *embedding*. These vectors are trained and optimised as parameters of the model to achieve the optimal representation for each symbol (Goldberg, 2022, chapter 4.8).

**Definition 4.16 (Alphabet index).**
Let $V$ be an alphabet. $\text{idx}_V : V \to \{i \in \mathbb{N} \mid i \leq |V|\}$ denotes a bijective function from symbols in $V$ to unique indices.

**Definition 4.17 (Embedding matrix).**
For an alphabet $V$ and an embedding dimension $d$, $E \in \mathbb{R}^{|V| \times d}$ is called an *embedding matrix.* For a symbol $s \in V$, the $\text{idx}_V(s)$-th row of $E$ is defined as the *embedding of $s$.*

In many natural language processing tasks symbols represent natural language words. Thus, the term *word embedding* was coined. Traditionally, features of an alphabet $V$ are encoded as vectors with dimensionality $|V|$ filled with 0s except for one position with value 1 that corresponds to a distinct symbol. This is called a *one-hot encoding.* Dense feature representations allow it to represent a vocabulary in a much smaller dimensionality than $|V|$.[29]

For the vocabulary of observed natural language words $V$, an embedding matrix $\mathbf{E_{word}}$ for $V$ with dimension $d_{word\text{-}emb}$ and an input sequence $w = w_1...w_n$, Coavoux and Cohen (2019) use $\mathbf{e}(w_i)$ to denote the embedding for $w_i$ with $i \in \{1, ..., n\}$.

$$\mathbf{e}(w_i) := \mathbf{E}_{\mathbf{word}}^{[\text{idx}_V(w_i)]} \tag{4.1}$$

$$\mathbf{E_{word}} \in \mathbb{R}^{|V| \times d_{word\text{-}emb}}, \ \mathbf{e}(w_i) \in \mathbb{R}^{d_{word\text{-}emb}}$$

When training on the DPTB corpus, Coavoux and Cohen (2019) extract 44530 unique words, including special <PAD> and <UNK> symbols. They embed them into a 32-dimensional space.

**Character-Aware Embedding** While word embeddings are a very capable way to represent discrete categories, they disregard the internal structure of words. Prefixes and suffixes may give information about the word's role in a sentence. For instance, the suffix *-ing* for English verbs indicates an infinitive. In our context, it usually means that it is dominated by another (main) verb in the constituent tree.

An embedding matrix that assigns distinct embeddings to different word surface forms can be assumed to capture the similarities and differences that are related to suffixes or prefixes from their distribution. However, it has been shown that combining word embeddings with a layer based on sub-word information improves POS-tagging results (Plank et al., 2016). This has two reasons. First, treating every word discretely generally requires more training data. Secondly, looking deeper allows to compute representations for words not seen at all during training (Gaddy et al., 2018).

Coavoux and Cohen (2019) use a character-aware-embedding as the second foundation for their token representations. It is generated using a *bidirectional long short-term memory (biLSTM)*

[29] Please refer to Goldberg (2022, chapter 8.4) for more information on the relation between one-hot and dense encoding.

*encoder*. Viewed abstractly, an LSTM (Hochreiter and Schmidhuber, 1997) is a function that receives as input an arbitrarily long sequence of vectors $\mathbf{x}_{0:n} = \mathbf{x}_1, ..., \mathbf{x}_n$ with $\mathbf{x}_i \in \mathbb{R}^{d_{in}}$ for all $i \in \{1, ..., n\}$ and returns as output a vector $\mathbf{y}_n \in \mathbb{R}^{d_{out}}$ with $d_{in}$ and $d_{out}$ being pre-defined input and output dimensionalities (Goldberg, 2022).

When used as a *transducer*, each input vector $\mathbf{x}_i$ is mapped to the output generated for the sequence up until $\mathbf{x}_i$, i.e. for $\mathbf{x}_{0:i}$. The internal computation function of the LSTM makes it possible for information from the previous entries in the sequence to influence the output for $\mathbf{x}_i$. When used as an *encoder*, only the final output is used. It represents a context-aware embedding for the whole input sequence.

The following full definition of an LSTM is based on Goldberg (2022, chapter 15). $\odot$ denotes element-wise vector multiplication.

**Definition 4.18 (LSTM).**
A *long short-term memory (LSTM)* model is defined as a function with the following properties:

$$\begin{aligned}
LSTM^*(\mathbf{x}_{0:n}) &= \mathbf{y}_{0:n} \\
\mathbf{y}_i &= LSTM(\mathbf{x}_{1:i}) \\
LSTM(\mathbf{x}_{0:t}) &= \mathbf{h}_t \\
\mathbf{h}_t &= \mathbf{o} \odot \tanh(\mathbf{c}_t) \\
\mathbf{c}_t &= \mathbf{f} \odot \mathbf{c}_{t-1} + \mathbf{i} \odot \mathbf{g} \\
\mathbf{i} &= \mathrm{sigm}(\mathbf{h}_{t-1}\mathbf{U_i} + \mathbf{x}_t\mathbf{W_i}) \\
\mathbf{f} &= \mathrm{sigm}(\mathbf{h}_{t-1}\mathbf{U_f} + \mathbf{x}_t\mathbf{W_f}) \\
\mathbf{o} &= \mathrm{sigm}(\mathbf{h}_{t-1}\mathbf{U_o} + \mathbf{x}_t\mathbf{W_o}) \\
\mathbf{g} &= \tanh(\mathbf{h}_{t-1}\mathbf{U_g} + \mathbf{x}_t\mathbf{W_g})
\end{aligned}$$

$$\mathbf{x}_i \in \mathbb{R}^{d_{in}},\ \mathbf{y}_i \in \mathbb{R}^{d_{out}},\ \mathbf{U_i}, \mathbf{U_f}, \mathbf{U_o}, \mathbf{U_g} \in \mathbb{R}^{d_{out} \times d_{out}},$$
$$\mathbf{W_i}, \mathbf{W_f}, \mathbf{W_o}, \mathbf{W_g} \in \mathbb{R}^{d_{in} \times d_{out}},\ \mathbf{i}, \mathbf{f}, \mathbf{o}, \mathbf{g} \in \mathbb{R}^{d_{out}}$$

A bidirectional LSTM model (Graves and Schmidhuber, 2005) consists of two separate LSTMs: one for the forward and one for the backward direction. When considering an input sequence $\mathbf{x}_{0:n}$ the desired output for word $i \in \mathbb{N}$ with $i \in \{1, ..., n\}$ is constructed by using the standard forward sequence $\mathbf{x}_{0:i} = \mathbf{x}_1, \mathbf{x}_2, ..., \mathbf{x}_i$ as the input for the first LSTM and the backward sequence $\mathbf{x}_{n:i-1} = \mathbf{x}_n, \mathbf{x}_{n-1}, ..., \mathbf{x}_i$ for the second LSTM. The output is a concatenation of both results. This enables contextual influence from elements preceding and elements following $\mathbf{x}_i$.

**Definition 4.19 (bi-LSTM notation).**
A bi-LSTM is defined as a function where $LSTM_f$ and $LSTM_b$ denote LSTMs of input dimension $d_{in}$ and output dimension $d_{out}$:

$$\begin{aligned}
biLSTM^*(\mathbf{x}_{0:n}) &= LSTM(\mathbf{x}_{0:n}, 1), ..., LSTM^*(\mathbf{x}_{0:n}, n) \\
biLSTM(\mathbf{x}_{0:n}, i) &= LSTM_f(\mathbf{x}_{0:i}) \circ LSTM_b(\mathbf{x}_{n:i-1})
\end{aligned}$$

$$\mathbf{x}_i \in \mathbb{R}^{d_{in}},\ \mathbf{y}_i \in \mathbb{R}^{2 \cdot d_{out}}$$

To construct character-aware embeddings, Coavoux and Cohen (2019) deconstruct tokens into sequences of individual characters. $L$ denotes the alphabet of all character symbols. Every symbol gets assigned a character-embeddings:

$$\mathbf{c_e}(l) := \mathbf{E}_{\mathbf{char}}^{[\mathrm{idx}_L(l)]} \tag{4.2}$$

$$\mathbf{E_{char}} \in \mathbb{R}^{|L| \times d_{char\text{-}emb}},\ \mathbf{c_e}(l) \in \mathbb{R}^{d_{char\text{-}emb}}$$

Then, for each token $w_i$ comprised of character sequence $w_{i,0:m}$, a biLSTM is used to encode a context-aware embedding. The backward network focuses on prefixes while the forward network is more sensitive to suffixes.

$$\mathbf{c}(w_i) := biLSTM_l(\mathbf{c_e}(w_{i,1}), ..., \mathbf{c_e}(w_{i,m})) \tag{4.3}$$
$$\mathbf{c}(w_i) \in \mathbb{R}^{d_{char}},\ \mathbf{c_e}(w_{i,j}) \in \mathbb{R}^{d_{char\text{-}emb}}$$

**Token Representation** Coavoux and Cohen (2019) construct a token representation $\mathbf{w}_i$ as the concatenation of the character-aware embedding and the word embedding:

$$\mathbf{w}_i = \mathbf{c}(w_i) \circ \mathbf{e}(w_i) \tag{4.4}$$
$$\mathbf{w}_i \in \mathbb{R}^{d_{token}},\ d_{token} = d_{char} + d_{word\text{-}emb}$$

Coavoux and Cohen (2019) build context-aware token embeddings by using a 2-layer biLSTM as a transducer. In this way, the network can capture contextual information regarding preceding and following tokens in each token vector. The first biLSTM has an input dimension denoted by $d_{token}$ and an output dimension denoted by $d_{hid}$. The second layer biLSTM has input dimension $d_{hid}$ and output dimension $d_{hid}$. The representations for $w_{0:n}$ are computed in the following way:[30]

$$\begin{aligned} \mathbf{m}_{1,0:n} &= biLSTM_1^*(\mathbf{w}_{0:n}) \\ \mathbf{m}_{2,0:n} &= biLSTM_2^*(\mathbf{m}_{1,0:n}) + \mathbf{m}_{1,0:n} \end{aligned} \tag{4.5}$$
$$\mathbf{w}_i \in \mathbb{R}^{d_{token}},\ \mathbf{m}_{1,i}, \mathbf{m}_{2,i} \in \mathbb{R}^{d_{hid}}$$

The first layer output is added to the second layer output as a *residual connection*. Residual mappings facilitate the learning of the identity function. Furthermore, they counteract the problem of exploding and vanishing gradients encountered in deep neural networks (He et al., 2016).[31]

As shall be seen, the output of the first layer biLSTM is used for the prediction of part-of-speech tags while the second layer is used to predict parsing actions. The architecture described so far is depicted in figure 4.4.

### 4.5.2 Set Representations

There are different possibilities of constructing constituent representations using token vectors. In projective constituency parsing it is common to use the leftmost leaf and the rightmost leaf of a constituent (Hall et al., 2014; Crabbé, 2015; Durrett and Klein, 2015). Using the notation established above such a constituent representation could be defined as follows for a constituent candidate $s$:

$$\mathbf{r_{proj}}(s) = \mathbf{m}_{2,\min(s)} \circ \mathbf{m}_{2,\max(s)} \tag{4.6}$$
$$\mathbf{m}_{2,t} \in \mathbb{R}^{d_{hid}}, \mathbf{r_{proj}}(s) \in \mathbb{R}^{2 \cdot d_{hid}}$$

In the view of overlapping constituents in discontinuous parsing, Coavoux and Cohen (2019) suggest enriching the constituent representation with information about gaps in the constituent. First, they define the following notation for a constituent candidate $s$, which yields all indices not in $s$ located between the minimum and maximum of $s$.

**Definition 4.20 (Gap set).**
Let $s$ be a constituent candidate.

$$\overline{s} := \{i \mid \min(s) < i < \max(s), i \notin s\}$$

[30]Coavoux and Cohen (2019) do not mention the use of a residual connections in their report, however it is included in the released code. Therefore I introduce an explicit variable $\mathbf{m}_{l,t}$ for each layer output (LSTM plus optional residual connection).

[31]A more thorough exploration of residual connections is given in section 6.2.2.

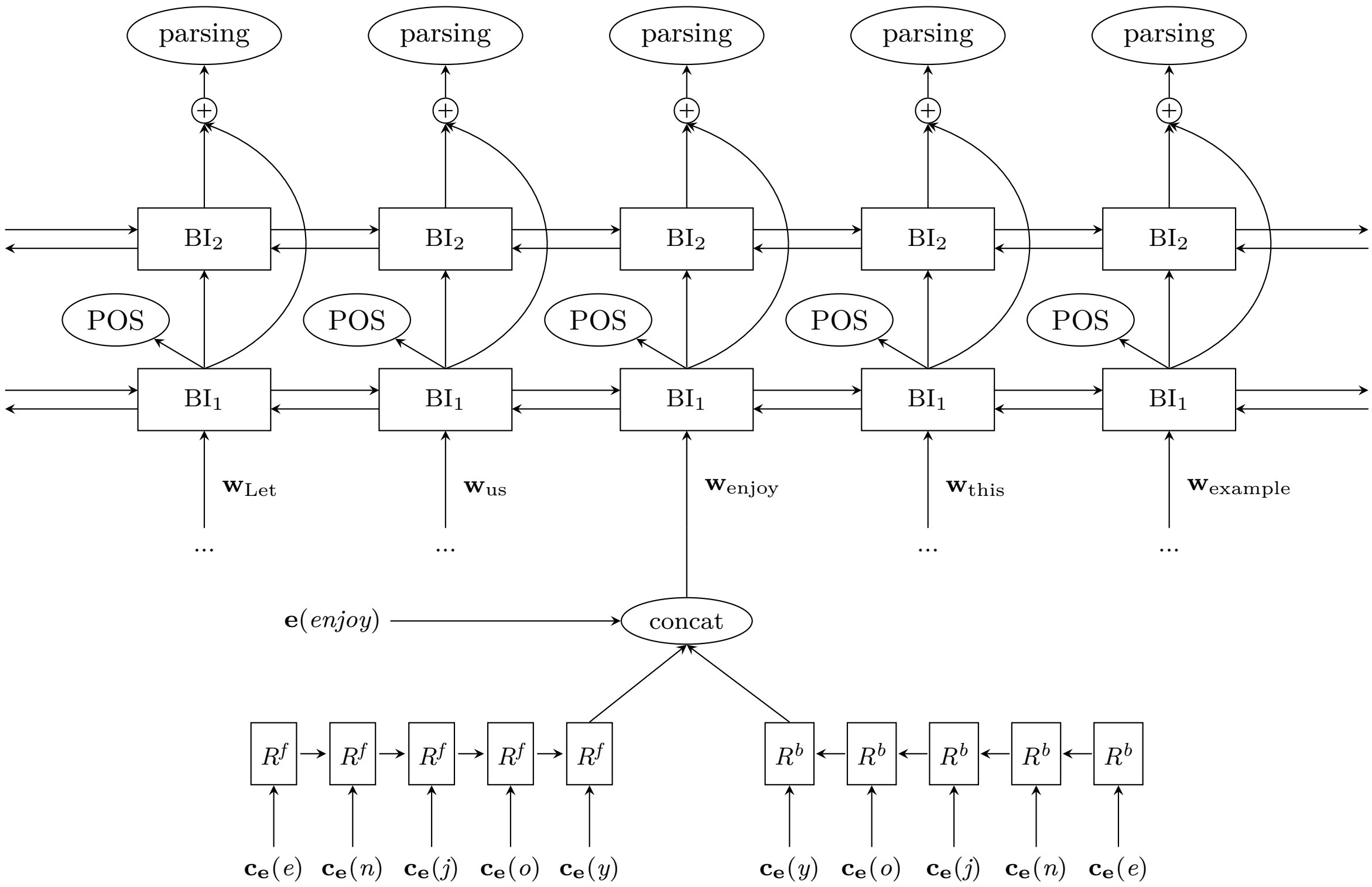

Figure 4.4: Stack-free neural parser with shared POS-tagging task architecture. For clarity the token representations are subscripted with words instead of indices. Figure inspired by Goldberg (2022, chapter 20.2).

The constituent is now represented using the set boundaries as well as the minimum gap element and the maximum gap element. A special vector $\mathbf{m_{nil}}$ that is randomly initialised and learned as part of the model is used if the set does not contain a gap.

$$\mathbf{r}(s) = \begin{cases} \mathbf{m}_{2,\min(s)} \circ \mathbf{m}_{2,max(s)} \circ \mathbf{m}_{2,\min(\overline{s})} \circ \mathbf{m}_{2,\max(\overline{s})} & \text{if } \overline{s} \neq \emptyset, \\ \mathbf{m}_{2,\min(s)} \circ \mathbf{m}_{2,\max(s)} \circ \mathbf{m_{nil}} \circ \mathbf{m_{nil}} & \text{else} \end{cases} \tag{4.7}$$

$$\mathbf{m_{nil}}, \mathbf{m}_{2,t} \in \mathbb{R}^{d_{hid}}, \mathbf{r}(s) \in \mathbb{R}^{4 \cdot d_{hid}}$$

This modification enables the model to differentiate between constituents with maximally two consecutive parts. It is not enough to distinguish constituents with more than one gap, i.e. with three or more consecutive parts. See the following two equations. The 4-index formalisation does not consider any intermediate continuous blocks lying between the two outermost gap indices.

$$\begin{aligned} s_1 &= \{1, 3, 5\} \\ \overline{s_1} &= \{2, 4\} \\ \langle \min(s_1), \max(s_1), \min(\overline{s_1}), \max(\overline{s_1}) \rangle &= \langle 1, 5, 2, 4 \rangle \end{aligned} \tag{4.8}$$

$$\begin{aligned} s_2 &= \{1, 5\} \\ \overline{s_2} &= \{2, 3, 4\} \\ \langle \min(s_2), \max(s_2), \min(\overline{s_2}), \max(\overline{s_2}) \rangle &= \langle 1, 5, 2, 4 \rangle \end{aligned} \tag{4.9}$$

Despite this limitation, one could argue that this notation should be capable of providing the scorer with adequate information about the form of a constituent since in the DPTB, which features at most 3 discontinuous blocks, the overwhelming majority of 3 block occurrences is caused by punctuation (Maier et al., 2012).[32]

[32] cf. section 3.1.7

### 4.5.3 Action Scorer

Differentiating between structural and labelling actions as odd and even steps allows using two separate feed-forward networks for action prediction.

**Structural actions** At a structural step for a configuration $c = \langle\ S, s_f, i, j, K\ \rangle : n \in \mathcal{C}^{\mathrm{SC}}$, the model has to decide between a COMBINE action with some element in $S$ and SHIFT. Coavoux and Cohen (2019) compute the score of a COMBINE-$s$ action using only the representation of $s_f$ and $s$ as input. The score of a SHIFT action is computed using $s_f$ and $\{i\}$, i.e. the element to shift.

These $|S| + 1$ scores can be computed in parallel using the following matrix. On the right side I note the action the row corresponds to.

$$\mathbf{M} = \begin{pmatrix} \mathbf{r}(s_1) & \circ & \mathbf{r}(s_f) \\ ... & ; & ... \\ \mathbf{r}(s_n) & \circ & \mathbf{r}(s_f) \\ \mathbf{r}(\{i\}) & \circ & \mathbf{r}(s_f) \end{pmatrix} \begin{matrix} \textsc{Combine-}s_1 \\ ... \\ \textsc{Combine-}s_n \\ \textsc{Shift} \end{matrix} \tag{4.10}$$

$$\mathbf{M} \in \mathbb{R}^{(|S|+1)\times 8\cdot d_{hid}},\ \mathbf{r}(s) \in \mathbb{R}^{4\cdot d_{hid}}$$

The score of an action is predicted using a feed-forward network with initial dropout.[33] For a pair of constituent representations, it computes a single scalar. tanh is used as the *activation function.* The *weight matrices* $\mathbf{W_{1struct}}, \mathbf{W_{2struct}} \mathbf{W_{3struct}}$ and the *bias vectors* $\mathbf{b_{1struct}}, \mathbf{b_{2struct}}$ are trained by the model.

$$\begin{aligned} FF_{struct}(\mathbf{a}) &= \mathbf{h_2 W_{3struct}} \\ \mathbf{h_2} &= \tanh(\mathbf{h_1 W_{2struct}} + \mathbf{b_{2struct}}) \\ \mathbf{h_1} &= \tanh(\mathbf{a' W_{1struct}} + \mathbf{b_{1struct}}) \\ \mathbf{a'} &= \mathbf{d} \odot \mathbf{a} \\ \mathbf{d} &\sim \text{Bernoulli}(drop_{action}) \end{aligned} \tag{4.11}$$

$$\mathbf{a}, \mathbf{a'}, \mathbf{d} \in \mathbb{R}^{8\cdot d_{hid}},\ \mathbf{W_{1struct}} \in \mathbb{R}^{8\cdot d_{hid} \times d_{action}}, \mathbf{W_{2struct}} \in \mathbb{R}^{d_{action} \times d_{action}},$$
$$\mathbf{W_{3struct}} \in \mathbb{R}^{d_{action} \times 1},\ \mathbf{b_{1struct}}, \mathbf{b_{2struct}} \in \mathbb{R}^{d_{action}},\ drop_{action} \in [0, 1]$$

A probability distribution over the possible structural actions is generated by applying the Softmax function on the $|S|+1$ dimensional output of $FF_{struct}$ on $\mathbf{M}$. Finally, the model performs the action corresponding to the highest score.

$$P(\cdot|c) = \text{Softmax}(FF_{struct}(\mathbf{M})) \tag{4.12}$$

$$P(\cdot|c) \in \mathbb{R}^{|S|+1},\ \mathbf{M} \in \mathbb{R}^{(|S|+1)\times 8\cdot d_{hid}}$$

**Labelling Action** At a labelling step with configuration $c = \langle\ S, s_f, i, j, K\ \rangle : n \in \mathcal{C}^{\mathrm{SC}}$, $|N| + 1$ different actions can be performed, where $N$ is the alphabet of nonterminals. $s_f$ can be labelled with some nonterminal $A \in N$ or NO-LABEL can be performed.

An action is predicted using only $s_f$ as input for a feed-forward network with an architecture that corresponds to that of $FF_{struct}$ with the exception of the input and the output dimensionality.

[33] A detailed description of the dropout technique is given in Goldberg (2022, chapter 4.6).

Here, the output is a vector of length $|N| + 1$.

$$\begin{aligned} FF_{label}(\mathbf{a}) &= \mathbf{h_2 W_{3label}} \\ \mathbf{h_2} &= \tanh(\mathbf{h_1 W_{2label}} + \mathbf{b_{2label}}) \\ \mathbf{h_1} &= \tanh(\mathbf{s' W_{1label}} + \mathbf{b_{1label}}) \\ \mathbf{a'} &= \mathbf{d} \odot \mathbf{a} \\ \mathbf{d} &\sim \text{Bernoulli}(drop_{action}) \end{aligned} \tag{4.13}$$

$$\mathbf{a}, \mathbf{a'}, \mathbf{d} \in \mathbb{R}^{4 \cdot d_{hid}},\ \mathbf{W_{1label}} \in \mathbb{R}^{4 \cdot d_{hid} \times d_{action}},\ \mathbf{W_{2label}} \in \mathbb{R}^{d_{action} \times d_{action}},$$
$$\mathbf{W_{3label}} \in \mathbb{R}^{d_{action} \times (|N|+1)},\ \mathbf{b_{1label}}, \mathbf{b_{a2label}} \in \mathbb{R}^{d_{action}},\ drop_{action} \in [0,1]$$

A probability distribution over the possible labelling actions is generated by applying the Softmax function on the output of $FF_{label}$ on $\mathbf{r}(s_f)$ Finally, the model performs the action corresponding to the highest score.

$$P(\cdot|c) = \text{Softmax}(FF_{label}(\mathbf{r}(s_f))) \tag{4.14}$$

$$P(\cdot|c) \in \mathbb{R}^{|N|+1},\ \mathbf{r}(s_f) \in \mathbb{R}^{4 \cdot d_{hid}}$$

### 4.5.4 POS Tagger

Following Søgaard and Goldberg (2016), POS tagging is performed as an auxiliary task using the output of the first layer LSTM transducer. A linear transformation is used to predict the POS tag probabilities for a token. Let $A$ be the alphabet of all POS-tags.

$$\begin{aligned} FF_{POS}(\mathbf{x}) &= \mathbf{x' W_{POS}} \\ \mathbf{x'} &= \mathbf{d} \odot \mathbf{x} \\ \mathbf{d} &\sim \text{Bernoulli}(drop_{POS}) \end{aligned} \tag{4.15}$$

$$\mathbf{W_{POS}} \in \mathbb{R}^{dim_{hid} \times |A|},\ drop_{POS} \in [0,1]$$

For an input sequence $w_{0:n}$ the probability of a sequence of POS tags $t_{0:n} = t_1, ..., t_n$ is then computed as follows:

$$P(t_{0:n}|w_{0:n}) = \prod_{i=1}^{n} \text{Softmax}(FF_{POS}(\mathbf{m}_{1,i}))^{[\text{idx}_A(t_i)]} \tag{4.16}$$

$$\mathbf{m}_{1,i} \in \mathbb{R}^{d_{hid}},\ FF_{POS}(\mathbf{m}_{1,i}) \in \mathbb{R}^{|A|}$$

### 4.5.5 Objective Function

The word and character embeddings, the LSTM parameters and the parameters of the feed-forward networks introduced above are optimised during training. The following objective functions determine the loss that the model is trained to minimise.

**Static Oracle** For the static oracle given an input sequence $w_{0:n}$ with gold POS-tags $t_{0:n}$ and gold tree $K_g$, the sum of the log-likelihood of $t_{0:n}$ and the log-likelihood of gold transition actions $a_{0:4n-2}$ given by the oracle with respect to $K_g$ is optimised:

$$\begin{aligned} \mathcal{L} &= \mathcal{L}_t + \mathcal{L}_p \\ \mathcal{L}_t &= -\sum_{i=1}^{n} \log\ P(t_i|w_{0:n}) \\ \mathcal{L}_p &= -\sum_{i=1}^{4n-2} \log\ P(a_i|(a_{i-1}(...(a_1(\sigma(w_{1:n})))))) \end{aligned} \tag{4.17}$$

In practice, Coavoux and Cohen (2019) alternate between optimizing parsing and POS-tagging. This has been shown by Caruana (1997) to be effective for multitask-learning. At each step, they provide a sentence from the train corpus and optimise either for parsing or for POS-tagging.

**Dynamic Oracle** Instead of optimizing the likelihood of the gold actions, in the dynamic setting Coavoux and Cohen (2019) sample a configuration from the set of all valid configurations. Then the likelihood of the best action with respect to this configuration given by the dynamic oracle is optimised.

Before a training epoch, each sentence from the training set is sampled with probability $p$. The current parameters are used to parse and retrieve a sequence of configurations. Inspired by (Ballesteros et al., 2016) at each step an action is chosen randomly based on the softmax distribution instead of taking the best-scoring action.

Let *Choice* sample an action randomly based on a tuple of likelihoods. Then the sequence of (suboptimal) configurations is denoted by $d_{0:4n-2}$ and generated for an input $w_{0:n}$ as follows.

$$\begin{aligned} d_1 &= Choice(P(\cdot|\sigma(w_{0:n}))) \\ d_i &= Choice(P(\cdot|d_{i-1})) \qquad \text{for } i \in \{2, ..., 4n-2\} \end{aligned} \tag{4.18}$$

Finally, the likelihood of the best action determined by the oracle for each step is optimised.

$$\mathcal{L}_p = -\sum_{i=1}^{4n-2} \log P(o(d_i)|d_i) \tag{4.19}$$

When the oracle predicts more than one best action, one is chosen deterministically as discribed in section 4.3.2.

### 4.5.6 Dealing with Unknown Words

Coavoux and Cohen (2019) assign a special *UNK* word embedding to words that were not observed during training. *UNK* is integrated as part of the vocabulary $V$. To learn a suitable representation for unknown words, before each training step they replace embeddings that belong to the $\frac{2}{3}$ least frequent words with a probability of 0.3 by *UNK*.

### 4.5.7 Training

Coavoux and Cohen (2019) train the model with the averaged stochastic gradient descent (ASGD) algorithm (Polyak and Juditsky, 1992) for 100 epochs. Every 4 epochs, the model is evaluated using the development set and saved if the validation F-score surpasses the results of previous evaluations. This way, degradation by overfitting does not affect the final model.

Coavoux and Cohen (2019) train the model on the DPTB[34] using sections 2-21 for training (39832 sentences), 22 for development (1700 sentences) and 23 for testing (2416 sentences). Furthermore, they report results for TIGER and NeGra. Table 4.1 gives the total list of hyperparameters.

### 4.5.8 Evaluation

The model is evaluated by Coavoux and Cohen (2019) using a dedicated module of `disco-dop`[37] (van Cranenburgh et al., 2016). `disco-dop` (*Discontinuous Data-Oriented Parsing*) is an LCFRS

[34] DPTB7 which sections 5.1 and 5.2 in Evang and Kallmeyer (2011) are based on.

[35] Xavier refers to a weight initialisation scheme proposed by Glorot and Bengio (2010) that has shown to allow faster convergence in deep neural networks.

[36] Gradient clipping is a common technique to prevent the problem of exploding gradients in deep (recurrent) networks (Pascanu et al., 2013).

[37] `https://github.com/andreasvc/disco-dop`

| Architecture hyperparameters | | |
|---|---|---|
| Dimension of word embeddings | $d_{word\text{-}emb}$ | 32 |
| Dimension of character embeddings | $d_{char\text{-}emb}$ | 100 |
| Dimension of character biLSTM state | $d_{char}$ | 100 (50 per direction) |
| Dimension of sentence biLSTM | $d_{hid}$ | 400 (200 per direction) |
| Dimension of hidden layers for action scorer | $d_{action}$ | 200 |
| Number of action scorer FF layers | | 3 |
| Number of tagger FF layers | | 1 |
| Final bias in action scorer FF networks | | False |
| Final bias in tagger FF network | | False |
| **Optimisation hyperparameters** | | |
| Initial learning rate | $l_0$ | 0.01 |
| Learning rate decay | $l_t$ | $\frac{l_0}{1+t\cdot 10^{-7}}$, ($t$ is step number) |
| Dropout for tagger input | $drop_{POS}$ | 0.5 |
| Dropout for parser input | $drop_{action}$ | 0.2 |
| Training epochs | | 100 |
| Batch size | | 1 sentence |
| Optimisation algorithm | | ASGD |
| Word and character embedding initialisation | | $\mathcal{U}([-0.1, 0.1])$ |
| Other parameters initialisation (incl. LSTMs) | | Xavier[35] |
| Gradient clipping (norm)[36] | | 100 |
| Dynamic oracle | $p$ | 0.15 |

Table 4.1: Hyperparameters of the model given in Coavoux and Cohen (2019).

parser that comes with tools that aid in the evaluation of parsers for discontinuous treebanks.

Coavoux and Cohen (2019) ignore punctuation and root symbols for evaulation (as set in the standard evaluation parameters in `proper.prm`). The evaluation metrics are F-score (F) and F-score limited to only the discontinuous constituents (Disc. F). A detailed comparison of results is given in section 6.3.

# 5. Explaining Supertags

In the most successful transition-based discontinuous constituent parsing approaches, POS-tags are jointly learned as an auxiliary task (Coavoux and Crabbé, 2017b; Coavoux et al., 2019). This strategy was first explored by Søgaard and Goldberg (2016) as an auxiliary task for chunking[38] and CCG supertagging[39] and adopted for parsing by Coavoux and Crabbé (2017b).

POS-tags very coarsely indicate the relations their carriers can stand in (Jurafsky and Martin, 2009, chapter 5) (e.g. *bark* marked as a verb gives rise to a different constituent structure than *bark* marked as a noun). A hierarchical LSTM network seems to be able to learn useful common representations for performing both POS tagging and parsing decisions which suggests a strong relationship between these tasks. I assume that training to predict POS tags from an intermediate layer leads the model to associate similar vectors to tokens that bear the same POS tag at that level. This implicitly evokes a distribution over constituent structures. The LSTM networks serve as a means of lexicalizing this distribution, giving more weight to those structures that occur more frequently given the surroundings.

Coavoux et al. (2019) note that the unlexicalised GAP-transition parser resolves discontinuities especially well when a lexical trigger like *why* in *wh*-extractions is present. Most unrecognised discontinuities occur where no token in itself is indicative of a discontinuous structure. In such cases, the POS-tag assignments are usually not helpful either. See for instance (5.2) where a. is a discontinuous extraposition taken from the DPTB (*evidence* and *that commissions were paid* belong to one constituent; example given by Coavoux et al. (2019)) and b. a projective sentence using the same words. The POS tag assignments for each word are the same.

(5.1) a.

| | | | | | | | | | | |
|---|---|---|---|---|---|---|---|---|---|---|
| In | April | 1987 | , | evidence | surfaced | that | commissions | were | paid | . |
| IN | NNP | CD | , | NN | VBD | IN | NNS | VBD | VBN | . |

b.

| | | | | | | | | | | |
|---|---|---|---|---|---|---|---|---|---|---|
| In | April | 1987 | , | evidence | that | commissions | were | paid | surfaced | . |
| IN | NNP | CD | , | NN | IN | NNS | VBD | VBN | VBD | . |

Nudging the parser into building more distinct vector representations in such cases by training to predict fine-grained tag assignments that differentiate the projective and the discontinuous use of certain words might be beneficial for the ability of the parser to correctly resolve discontinuities. Supertags could fulfil this role - especially since they were shown to be compatible with POS-tags in an auxiliary-scenario (Søgaard and Goldberg, 2016).

Furthermore, they could help the parser to recognise types of discontinuous constituents which occur with a very low frequency. Maier (2015) observe that grammar-based parsers perform well on low frequency discontinuities while transition-based neural parsers perform poorly. This is not surprising considering that a multi-layer neural approach needs a lot of data to learn successfully. The resources available for phenomena that are only sparsely attested in the DPTB simply do not suffice. Thus, the statistical model, having no restrictions beyond the transition system design, resorts to erroneous actions. Meanwhile, an exact grammar extraction algorithm builds rules for all cases that occur in a given corpus. However low their frequency may be, the parser eventually will apply them if no other more frequent rules yield a successful parse.

Using subparts of structural supertag information to learn representations for certain aspects of discontinuity may aid in dealing with rare constituent types much akin to the effect that character-aware word embeddings have on parsing. The idea is to increase the similarity of the vector representation of items whose syntactic impact is alike in some aspect and reduce the similarity where it is different. This could help to improve representations for rare combinations of phenomena and thereby reduce the problem of data availability for rare discontinuities.

[38] Please refer to Jurafsky and Martin (2009, chapter 13.5) for an introduction to chunking.
[39] See section 5.2 for an account of CCG supertags.

In the following, I will first introduce the concept of supertags in the context of combinatory categorial grammar (CCG). Then I will examine several categories of discontinuous phenomena in more detail and analyse how CCG supertags may help a discontinuous constituent parser to correctly recognise them. Finally, I will test my hypothesis by integrating supertags both as an input feature and as an auxiliary task into the transition-based parser of Coavoux and Cohen (2019) and explore ways to give a hierarchical model more control over the use of auxiliary information for the top level task. A thorough analysis of the results will be provided as well.

## 5.1. Lexicalised Grammar Formalisms

Broadly speaking, grammar formalisms fall into two groups. In the traditional approach, a set of very simple primitive components is used as a foundation. Operations of varying complexity and usually large in quantity are then introduced to construct complex structures out of the primitives. CFG and LCFRS fall into this category where productions serve as the combining operations and where the set of lexical category types (POS-tags) is very small. Contrasting that, strongly lexicalised formalisms like lexicalised tree adjoining grammar (LTAG) (Bangalore and Joshi, 2010a) and combinatory categorial grammar (CCG) (Steedman, 1989, 1996, 2000) follow an approach called *complicate locally, simplify globally* (CLSG) (Bangalore and Joshi, 2010a). They associate the lexical items (i.e. words in a sentence) with informative primitive structures from which a complex syntactical description of the phrase can be retrieved using a minimal and language-independent set of combining operations (Bangalore and Joshi, 2010a).

The primitive structures associated with lexical items are called *supertags*. A supertag localises information about (a) the number of arguments (b) (type) constraints for the arguments and/or (c) the positions of the arguments relative to the position of the item bearing the supertag. There may be several possible supertags associated with one lexical item. In CFGs, this information is distributed over more than one primitive structure (Bangalore and Joshi, 2010a)

As a result, non-local syntactic dependencies are localised into the primitive structure in CLSG formalisms. Thus, knowledge of supertags may be beneficial for parsing discontinuous constituents. In the following, I will give an overview of the CCG formalism and the resulting supertags.

## 5.2. Combinatory Categorial Grammar

Combinatory categorial grammar (CCG), developed by Steedman (1989, 1996, 2000), is an extension to *categorial grammars* (Adjukiewicz, 1935; Bar-Hillel, 1953). The central assumption of categorial grammars is that constituents are composed using function application. They associate each word in a sentence with a (complex) function that can receive a word to its left or to its right as an argument.

CCG extends this formalism by introducing directed versions of *combinators* originally found in combinatory logic (Curry, 1930). This approach can be described as a *deduction-based logical framework* (Kallmeyer, 2010, chapter 1). It produces constituent-like descriptions and features a novel strategy to deal with discontinuous phenomena. Its efficient parsibility and its straightforward compositional semantic interpretation makes the formalism of computational interest for covering natural language (Bozsahin, 2013). The following overview of the CCG formalism is based on Jurafsky and Martin (2023).

In the CCG formalism categories can be *atomic elements* or *functional categories*:

**Definition 5.1 (Categories).**
Let $\mathscr{A}$ be a set of atomic elements. The set of categories $\mathscr{C}$ is defined as:

- $\mathscr{A} \subseteq \mathscr{C}$,

- $(\mathsf{X}/\mathsf{Y}), (\mathsf{X}\backslash\mathsf{Y}) \in \mathscr{C}$, if $\mathsf{X}, \mathsf{Y} \in \mathscr{C}$,
- nothing else is in $\mathscr{C}$.

Functional categories represent functions of a single argument category to a desired category. In $(\mathsf{X}/\mathsf{Y})$, the symbol $\mathsf{Y}$ is called the *argument* and $\mathsf{X}$ the *functor*. It means that an element of type $\mathsf{Y}$ is expected to the right of category $(\mathsf{X}/\mathsf{Y})$. On the other hand $(\mathsf{X}\backslash\mathsf{Y})$ expects an element of type $\mathsf{Y}$ to its left. $\mathsf{X}$ is the result of application. The outermost brackets of functional categories are conventionally omitted.

Usually, the set of atomic elements is small (e.g. sentences and noun-phrases). Verb phrases, conjunctions and other complex types are realised though complex functional categories (Jurafsky and Martin, 2023). A lexicon contains assignments from words to sets of possible categories. Due to ambiguity, the assignments often contain several choices.

**Example 5.2.**
A minimal example lexicon is given in figure 5.1. Note that *lie* has two possible categories: $\mathsf{N}$ marks its use as a noun while $\mathsf{S}\backslash\mathsf{NP}$ allows for a predicative use like in *You lie.* $\mathsf{S}\backslash\mathsf{NP}$ is a function that expects an argument of type $\mathsf{NP}$ to its left and returns a category of type $\mathsf{S}$.

| | |
|---|---|
| *you:* | $\mathsf{NP}$ |
| *that:* | $\mathsf{NP}$ |
| *is:* | $(\mathsf{S}\backslash\mathsf{NP})/\mathsf{NP}$ |
| *a:* | $\mathsf{NP}/\mathsf{N}$ |
| *lie:* | $\mathsf{S}\backslash\mathsf{NP},\ \mathsf{N}$ |

Figure 5.1: Minimal example CCG lexicon.

The rules in figure 5.2 define the standard combining operations of most forms of CCG. ($>$) and ($<$) are the rules for forward and backward function application respectively. Forward and backward composition ($>\mathbf{B}$, $<\mathbf{B}$) allow combining two functions where one asks for an argument of type $\mathsf{Y}$ and the second gives $\mathsf{Y}$ as a result. The resulting function has the return type of the first function and the argument of the second. Type raising ($>\mathbf{T}$, $<\mathbf{T}$) converts a category into a function that receives as argument a function that seeks said category as its argument. Finally the conjunction operator ($<\mathbf{\Phi}>$) allows combining the $\mathsf{CONJ}$ category and two elements of the same type on both sides of the conjunction into a single element of this type.

Rule application for sequences of lexical items is performed on their associated categories in the style of natural deduction with the restriction that only adjacent elements can be used in deductions. This way, the deduction process adheres to the *principle of adjacency* (Steedman, 2000, chapter 4.1) implicitly yielding an ordered tree. The goal is to use up every input element and to retrieve a single category at the end. The resulting category is the type of the whole sequence.

A strong argument in favour of the CCG formalism is the compositional semantics that naturally emerges from it. Interpretations of constituents can be formalised as terms of the $\lambda$-calculus (Church, 1933).[40] Every lexical entry is assigned with a $\lambda$-expression and each combinator is given an interpretation within the $\lambda$-calculus, e.g. function application for forward and backward application. Through the replacement of bound variables, the terms of a sentence are brought into an interpretable logical form and order.

**Example 5.3.**
Figure 5.3 shows an example CCG derivation for the sentence *that is a lie* using the lexicon given in figure 5.1. To the right, a tree visualisation of the constituents in the CCG derivation is given. As can be seen, this analysis amounts to the canonical CFG derivation tree for this sentence.

[40] An interested reader is referred to Rojas (2015) for an introduction to the $\lambda$-calculus

| | | |
|---|---|---|
| Application | $\dfrac{X/Y \quad Y}{X}>$ | $\dfrac{Y \quad X\backslash Y}{X}<$ |
| Composition | $\dfrac{X/Y \quad Y/Z}{X/Z}>\mathbf{B}$ | $\dfrac{Y\backslash Z \quad X\backslash Y}{X\backslash Z}<\mathbf{B}$ |
| Type raising | $\dfrac{X}{T/(T\backslash X)}>\mathbf{T}$ | $\dfrac{X}{T\backslash(T/X)}<\mathbf{T}$ |
| Conjunction | $\dfrac{X \quad CONJ \quad X}{X}<\Phi>$ | |

Figure 5.2: Rules for CCG. X, Y, Z and T are category variables.

Function application allows to fill in the argument slot of a category. Take for instance *is* and its category (S\NP)/NP. It requires an NP category to the right. Applying (>) on (S\NP)/NP and on category NP to the right yields the functor S\NP.

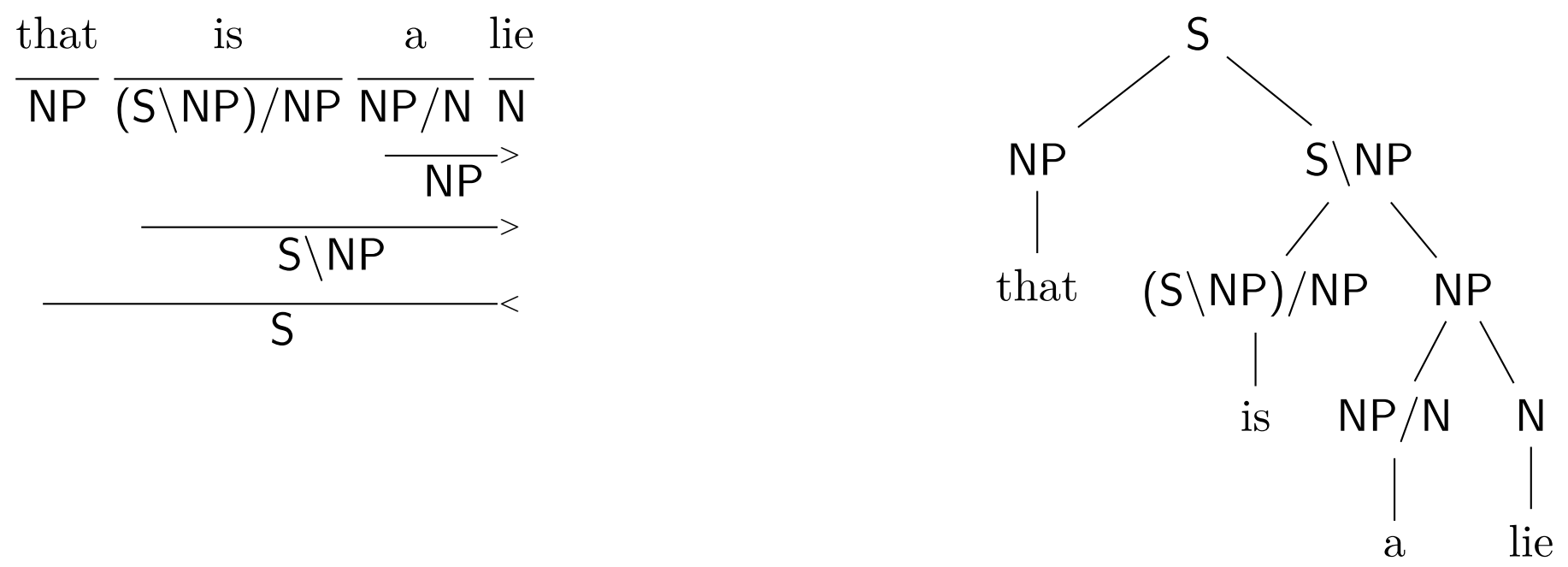

Figure 5.3: Example for CCG function application.

Figure 5.4 shows an example for forward composition (>**B**) and forward type-raising (>**T**) using the same supertag assignments for the sentence *that is a lie.* The category (S\NP)/NP is combined with NP/N via composition. (S\NP)/NP requires an argument of type NP while NP/N provides category NP if provided with N. Composition allows it to combine both items before filling missing arguments. The argument requirement of the category on the right is inherited by the resulting category (S\NP)/N.

NP is type-raised to a complex function yielding S and expecting a category that looks for an NP to its left with functor S. This way, an argument takes the syntactic role of a functor (Koller and Kuhlmann, 2009). In this way, forward composition can be used to fill the left argument slot of a VP before the right one. Here, *that* is combined before *lie*.

Note the equivalent tree structure to the right in figure 5.4. Grouping together a verb and a determiner or attaching the subject at a lower level than the object contradicts the canonical constituent analysis found in S-bar-inspired constituency trees. Both CCG tree descriptions are licensed by the same CCG lexicon in combination with the minimal set of combinatory rules. It becomes apparent that the formalism produces constituents that are not comparable with our traditional notion of constituency.

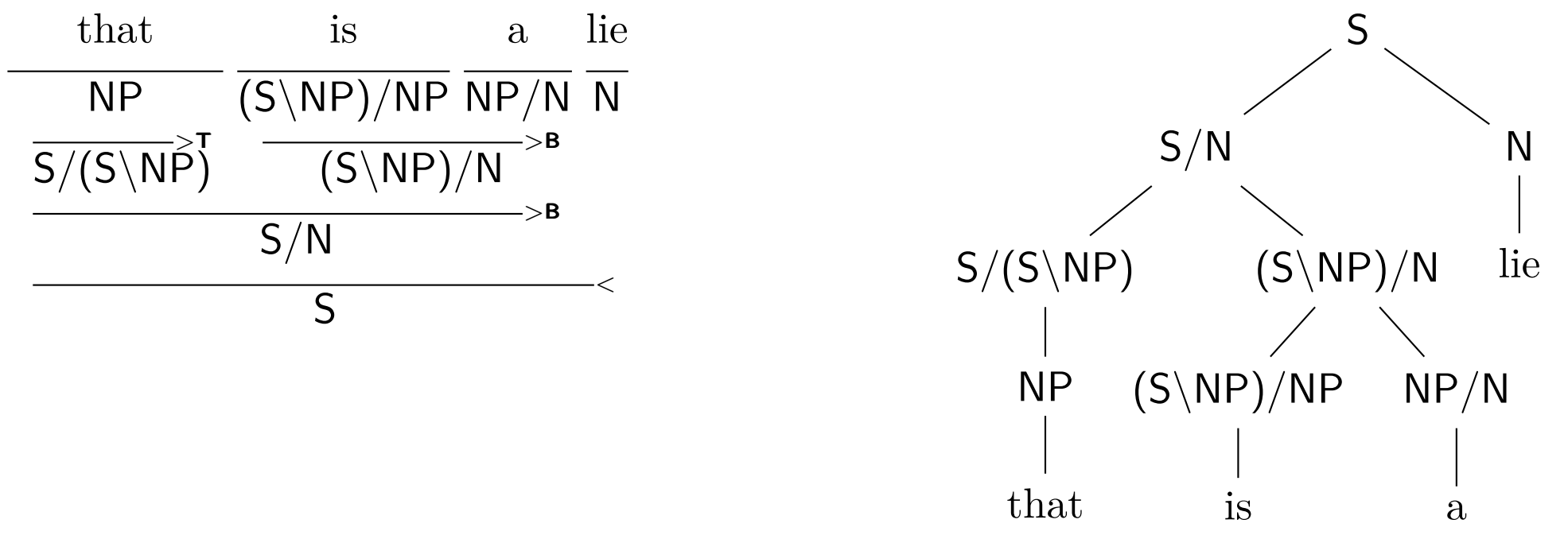

Figure 5.4: Example for CCG composition and type raising.

**Example 5.4.**
Composition and type-raising are not necessary to analyse the sentence in example 5.3. In other cases, like in coordination of non-canonical constituents they are argued to be required tools (Steedman, 2000, chapter 3). Figure 5.5 shows an example. Here, *Laura* and *a gift* as well as *Tommy* and *an invoice*, in both cases two elements with type NP, combine via type-raising and composition. *Laura* is type-raised to expect a ditransitive VP to its left and to return a transitive VP. *a gift* is type-raised to expect a transitive VP to its left and to return an intransitive VP. Then, the two type-raised terms are combined via backward composition. Doing the same for *Tommy* and *an invoice* allows using the conjunction operator (<**Φ**>) for these non-constituents.

The two combinators play an important role in dealing with long-range dependencies which I will analyse in the following subsection.

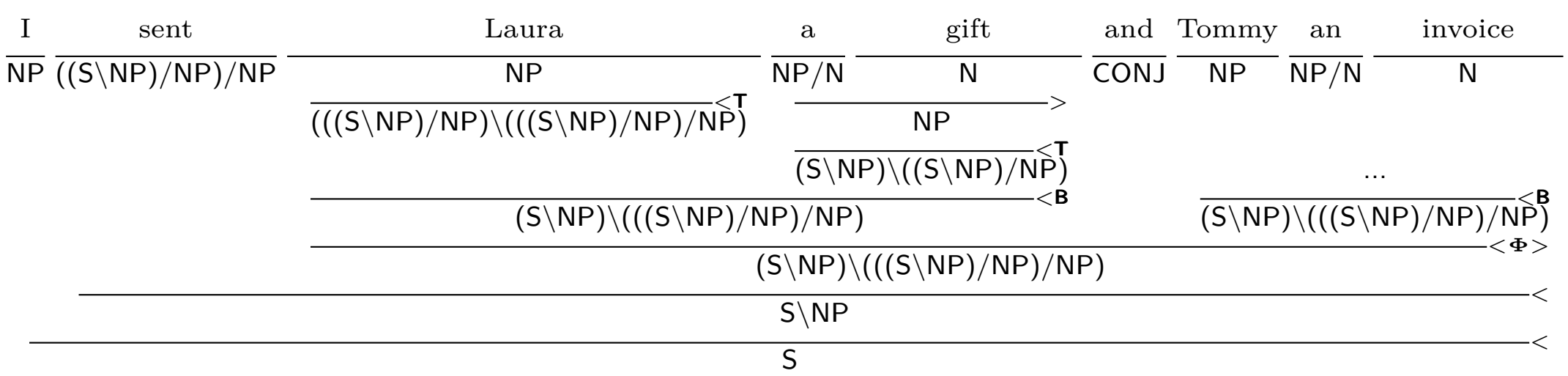

Figure 5.5: Coordinative use of type-raising and composition in CCG.

As became apparent, type raising and coordination allow any substring in a sentence to form a constituent (Houtman, 1994, 63–85).[41] This enables strict left-to-right derivation of sequences which is argued by Jurafsky and Martin (2023) to be a desirable feature from a psycholinguistic point of view since it resembles the way humans process language. Cases of derivational ambiguity where the same input sequence with the same supertag assignments can be reduced to the same category are claimed by (Lewis and Steedman, 2014) to license an identical same semantic composition in many cases.

A key argument of the proponents of CCG as a theory of grammar is that there is no necessity to model several layers of syntactic representation to account for a transformation from a deep form grouping together the words in a sentence that belong together in terms of our intuition of meaning, and from which a semantic representation can be extracted, to a surface form cluttered with discontinuities, scrambling or other phenomena inert to human utterance while adhering to the reflection of a narrow definition of constituency in the syntactic representation. Instead, syntax should be treated as a transparent interface between spoken language and logical form (Steedman, 2000, chapters 1 and 2). CCG is a *monostratal* grammar formalism in the sense that it features only one syntactic level of representation with the goal of assigning every syntactic rule with a semantic rule applied in parallel (Steedman, 2000, chapter 1). Since it aims at describing the competent composition of a semantic form, over-generation of spuriously ambiguous derivational structures is not considered problematic.

[41]This only holds if type-raising is not restricted to some limited set of operations as explained by Steedman (2000, chapter 4).

While in contrast the discontinuous constituent structure found in the DPTB is very strict and designed to exhibit one clear derivational analysis per phenomenon, its motivation for analysing constituents as discontinuous and not as instances of a separate movement process as employed by Chomsky (1975) and grammatical traditions alike is similar to that of CCG: constructing only one syntactic level of representation. Discontinuous tree descriptions achieve this by loosening the adjacency requirement of constituency while maintaining a narrow requirement for what can form a constituent based on semantic intuition. This precisely leads to the necessity of mild context-sensitive generative devices like LCFRS. In CCG the idea is executed by binding arguments into predicate-argument structure while combining adjacent words into syntactic constituents resulting in a semantic expression that does not necessarily correspond with the sentence word order (Steedman, 2000, Chapter 3).

My hope is that the strong correspondence of syntactic rules and semantic function application inherent to CCG categories as well as the fact that word order is projected entirely from the lexicon might enable a neural model to retrieve information about the predicate-argument structure from lexical category assignments that motivated the notion of constituency found in discontinuous treebanks.

### 5.2.1 CCG Treebanks

The *CCGbank* (Hockenmaier and Steedman, 2007) is a version of the Penn Treebank annotated with CCG supertags and derivations. It was automatically extracted from the Penn Treebank and contains 48,934 fully annotated sentences as well as a lexicon of 44,000 words. Sections 2-21 feature 1,286 distinct lexical category types.

CCGbank differs from the pure CCG presented above. It uses features to distinguish different types of primitive categories such as S[dcl] for declarative sentences and S[q] for yes-no questions. Furthermore, it introduces a range of special combinatory rules, some of which deal with punctuation, others allow unary category changes, e.g. from N to NP. Without further specification, I will use the notation (<**r**>) for all punctuation rules and (<**u**>) for all unary rules.

Honnibal et al. (2010) revised the original CCGbank by introducing a more precise analysis of compound nouns, verb argument structure, verb-particles and noun-arguments. Furthermore, they reintroduced quotation marks which were previously removed by Hockenmaier and Steedman (2007). The resulting treebank is known as *CCGrebank* and features 1,574 distinct lexical category types in sections 2-21. I will base my analysis of discontinuous phenomena in section 5.3 on a comparison of derivation examples in DPTB and in CCGrebank.

### 5.2.2 Supertaggers and Parsers

The lexicalised nature of combinatory categorial grammar allows for a unique approach to parsing for the formalism. Usually, a word is assigned more than one category in a CCG lexicon. This fact and the non-standard notion of constituency have the effect that a sentence can yield a large number of different parses. This problem can be approached by assigning lexical categories in a pre-parsing step using a statistical *supertagger*. Bangalore and Joshi (1999) call this *an approach to almost parsing*. Since the set of combinators is kept minimal and the syntactic information inherent in supertags is so rich most ambiguities are resolved through the assignment of adequate supertags.

The idea of using a dedicated supertagger before parsing to narrow down the space of possible supertag assignments for an input sequence was first explored by Bangalore and Joshi (1999) for LTAG and adapted by Clark and Curran (2010) for parsing CCG. Note, however, that some ambiguity remains that is not resolved through supertagging (cf. example 5.3).

Lewis and Steedman (2014) propose a new CCG parsing strategy that allows for an efficient A* search by assuming that the optimal CCG parse for a sentence is determined by the highest probability supertag assignment. Then, they use a heuristic to eliminate spurious derivational ambiguity. They use 511 unique supertags and a total of 23 combinatory rules. Their model is trained on sections 02-21 of CCGbank.[42]

Yoshikawa et al. (2017) propose an extension to this parser called `depccg`[43] that resolves spurious attachment ambiguity by jointly modelling a dependency structure. Their model successfully approaches cases where spurious ambiguity indeed leads to a difference in semantic representation. They illustrate this problem by help of the Japanese example reproduced in figure 5.6.

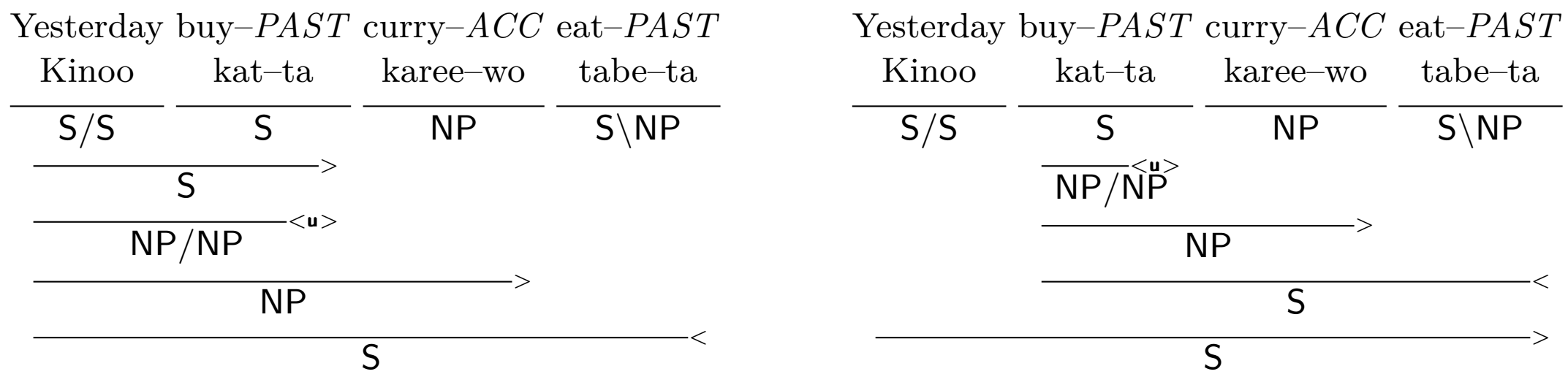

Figure 5.6: Spurious ambiguity in CCG derivations that leads to distinct semantic representations; adapted from Yoshikawa et al. (2017). The English translation is *I ate the curry I bought yesterday. yesterday* can refer to either *bought* or *ate*.

Alongside the standard parser trained on the original CCGbank which they describe in their paper, Yoshikawa et al. (2017) also release a model trained on the CCGrebank, which I will use as a supertagger in the pipeline approach presented in section 6.1.

## 5.3. Discontinuous Constituents and CCG

Due to its monostratal nature, CCG deals differently with phenomena that are traditionally analysed to exhibit discontinuous constituents. Being a lexicalised approach, the recipe for dealing with discontinuities is largely (with the exception of the combining rules) encoded in the assigned supertags. In the following, I will take a closer look at the way CCG resolves sentences in non-standard word order and explore the relationship with common discontinuous constituent analyses from the DPTB.[44]

Several suggestions were made to account for discontinuous constituents using the CCG framework. One example is the introduction of infixation points and *wrap* and *infix* operations, see Warstadt (2015) for a recent proposal. It introduces the notion of discontinuity by weakening the formalism's *principle of adjacency* (Steedman, 2000, chapter 4.1). Since annotated corpora for non-standard proposals are generally unavailable, I will only focus on the handling of discontinuities by CCG in the form described above.

Evang (2011) enumerates six classes of discontinuities found in the DPTB. Coavoux et al. (2019) revise these categories resulting in the following list and example phrases:

[42] `https://github.com/mikelewis0/easyccg`
[43] `https://github.com/masashi-y/depccg`
[44] In the following derivation examples pre-terminal nodes are regularly omitted for reasons of space. In most cases only extracts of the full trees are shown.

1. *wh*-extraction: What should I do?
2. fronted quotations: Absolutely, he said.
3. extraposed dependent: In April 1987, evidence surfaced that commissions were paid.
4. circumpositioned quotations: In general, they say, avoid takeover stocks.
5. *it*-extrapositions It's better to wait.
6. subject-verb inversion Said the spokeswoman: "The whole structure has changed."

Note that terms like *movement* and *extraction* are not meant to allude to a specific grammatical theory that uses the notion of movement as a syntactic operation but only to refer to syntactic phenomena metaphorically.

### 5.3.1 *Wh*-movement

*Wh*-movement denotes the non-canonical word-order correlated with the use of interrogative words like *what* and *why*. The *wh*-word is placed at the front of the sentence or clause instead of the position it would be expected in with respect to its grammatical function. Take for example (5.2b) where *this*, the direct object of *do*, is found in the standard position, following the predicate. Compare this with (5.2a) where *What*, corresponding to *do*, is positioned at the beginning of the sentence.

(5.2) a. What should I do?

b. I should do this.

Figure 5.7 shows the analysis for the sentence *What should I do?* in the DPTB as well as in CCGrebank. In the constituent tree, the fronted *wh*-word *What* is attached at the level of the VP that directly dominates the verb *do*. In the CCG analysis, the subpart *should I do* is derived first via forward application and composition. The resulting category S/NP expects an object NP in standard position to its right. The discontinuity is dealt with by assigning *what* a higher-order function: S/(S/NP). It fills the missing argument slot by simply "consuming" the incomplete sentence.

This supertag assignment could give a discontinuous constituent parser a hint: It could learn to identify the fact that the right NP argument slot of *do* is not removed by combination with the two items on its left. It could then conclude from the supertag assignment of the first word, S/(S/NP), that it should be used as a filler of this argument slot promoting a Merge operation between *What* and *do* before merging with *should* or *I*.

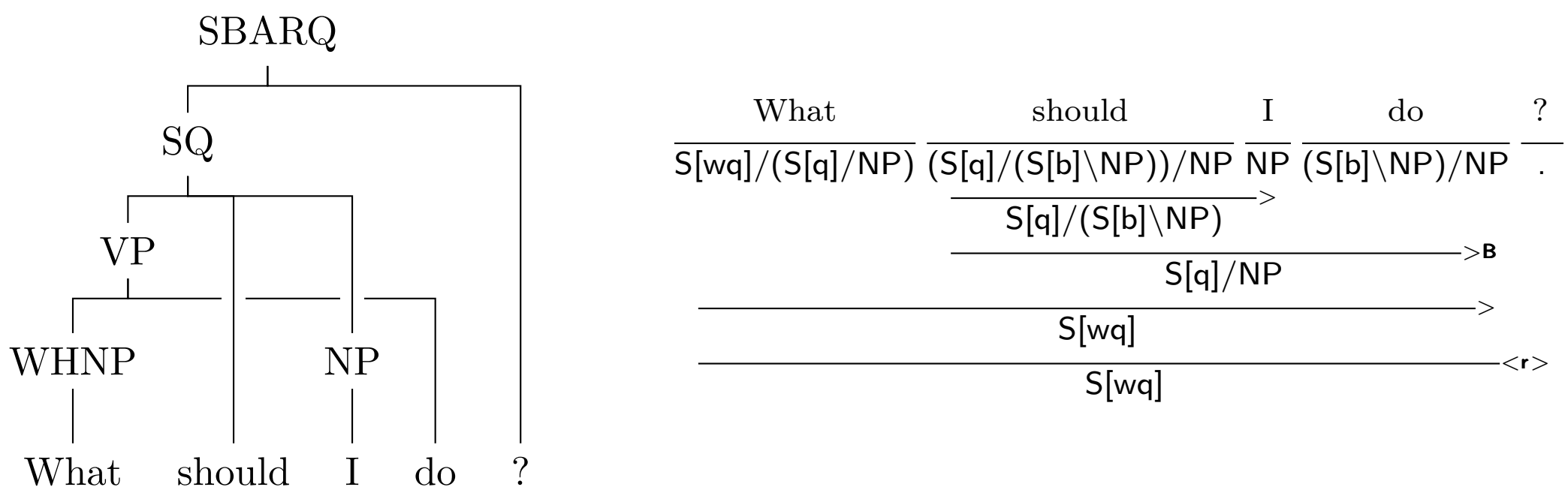

Figure 5.7: *Wh*-movement. Derivation in the DPTB (on the left) and in CCGrebank (on the right).

Sentences, where the *wh*-word asks for the subject, are usually projective. The first position corresponds with the subject position and so the supertag assignment differs compared to the previous case. This can be seen in figure 5.8. Here, *What* expects a sentence with a missing NP in left position. One can clearly differentiate the projective and the unprojective case on the basis of supertag assignments.

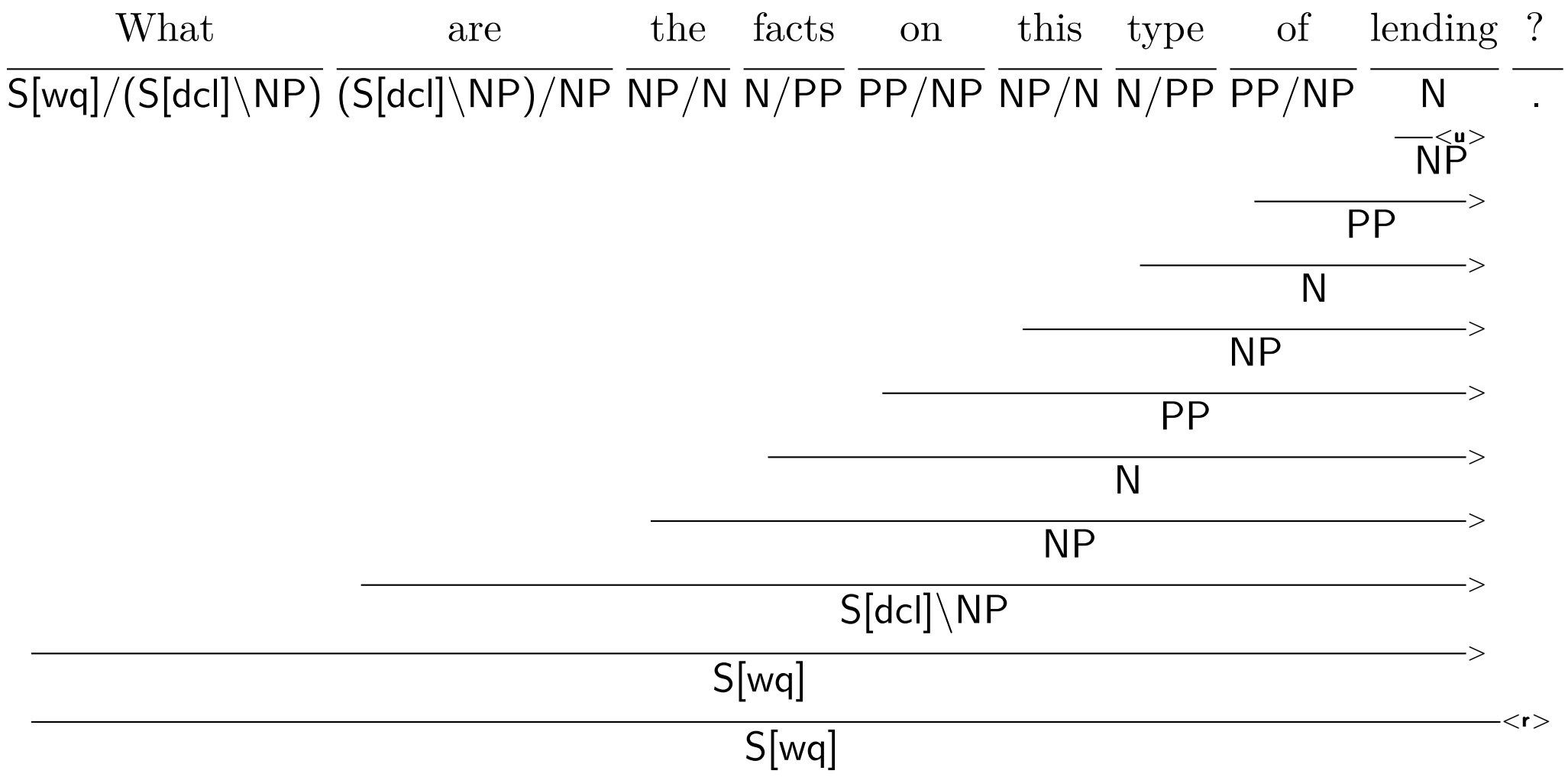

Figure 5.8: *Wh*-word that asks for the subject from CCGrebank.

*Wh*-movement also occurs in relative clauses. Figure 5.9 shows a DPTB derivation for the nominal phrase *federal statutes that the Supreme Court invalidated.* Here *that* acts as an object of the verb. The subject *the Supreme Court* blocks a projective analysis.

The CCGrebank analysis, given in figure 5.10, is similar to figure 5.7: The subject and the verb are first composed into a non-standard constituent by use of type-raising and composition that then serves as an argument for *that*. Interrogative words like *what* and *which* are also assigned the category (N\NP)/(S[dcl]/NP) when they occur in this kind of construction. It differs from the assignment in questions in that it does not return a sentence type but type N\N that makes it function as a modifier of a preceding noun.

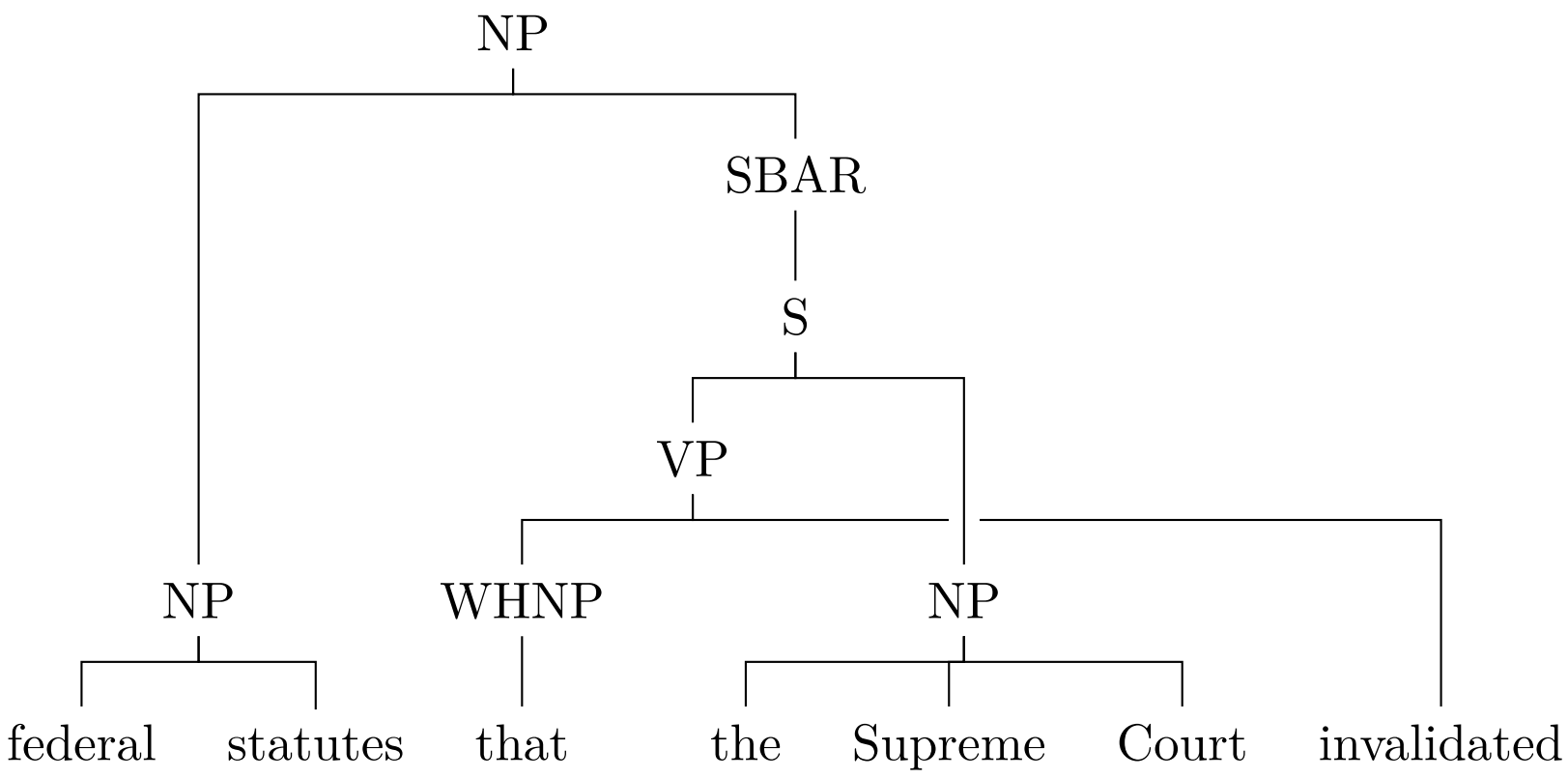

Figure 5.9: Relative clause with *that*-fronting in DPTB annotation.

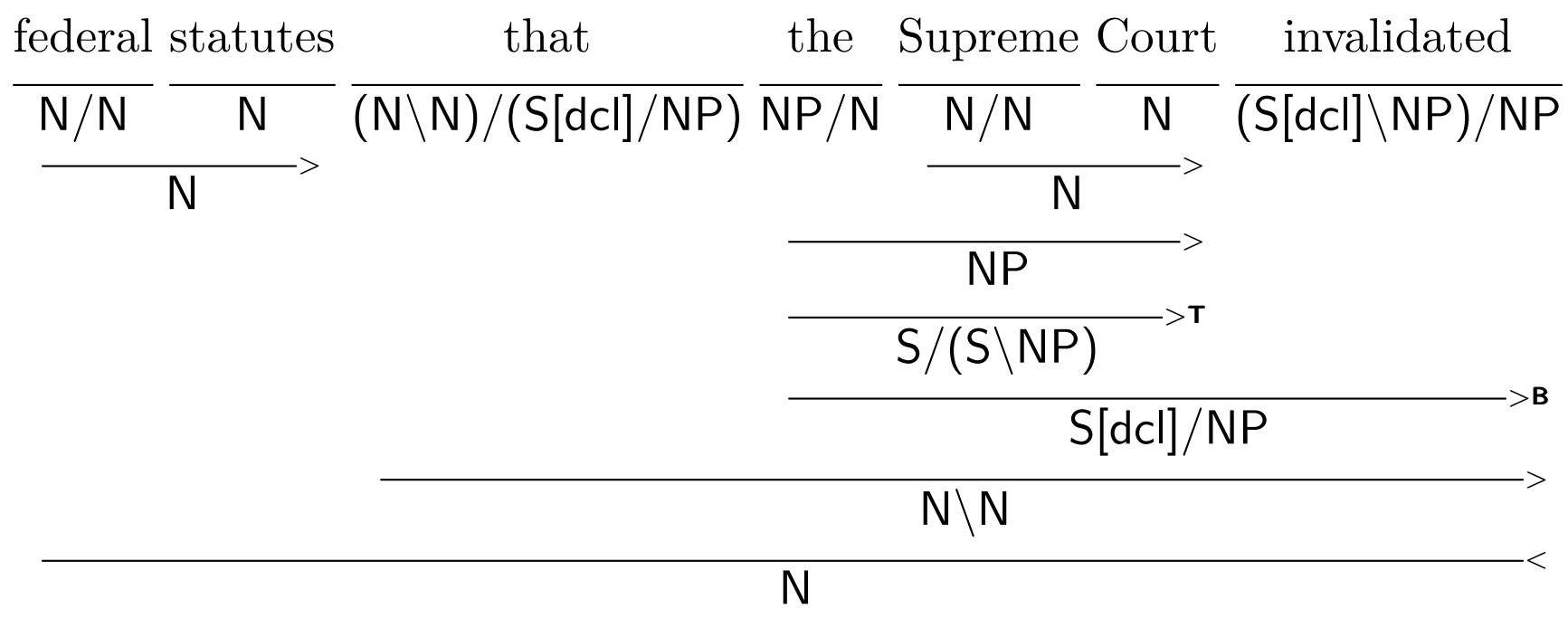

Figure 5.10: Relative clause with *that*-fronting in CCGrebank annotation.

*that* is given a different supertag when it does not accompany a relative clause but is used as a demonstrative pronoun (see figure 5.11) or as a conjunction in a complement clause (see figure 5.12). Coavoux et al. (2019) point out that the ambiguity of *that*-clauses is a notable source of error concerning *wh*-movement. Supertag information may enable the parser to better differentiate between the structurally distinct uses of *that*.

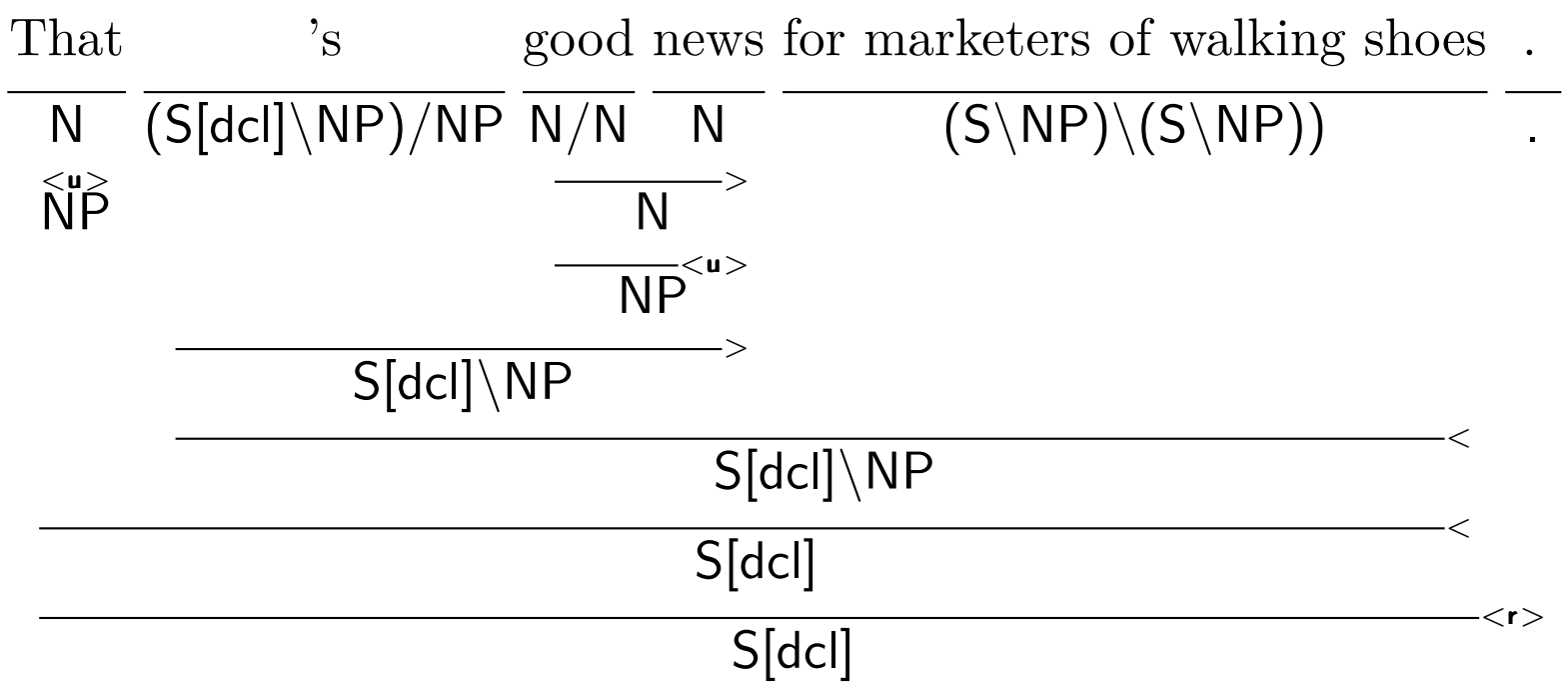

Figure 5.11: Example for the use of *that* as a demonstrative pronoun with the corresponding derivation in CCGrebank.

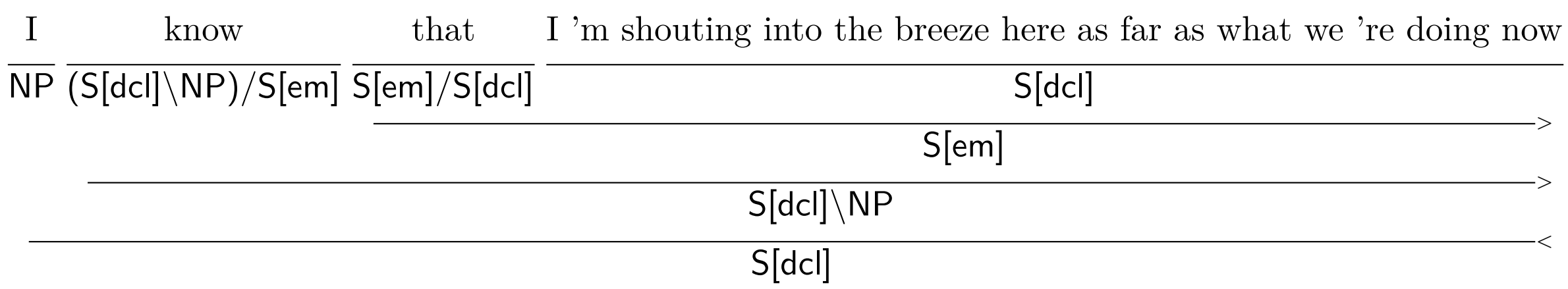

Figure 5.12: Example for the use of *that* as a conjunction in a complement clause including the derivation from CCGrebank.

Another import reason for parsing-errors in *wh*-extracted phrases, according to Coavoux et al. (2019), is ambiguity on the extraction site, i.e. the node the fronted question word should attach to to form a discontinuous constituent. As an example they mention the phrase *which many clients didn't know about* for which the DPTB analysis and its CCGrebank counterpart are given in figures 5.13 and 5.14 respectively.[45] Their parser commits the error of treating *which* as an argument of *know*. Given the correct CCG analysis, it would be immediately clear that the fronted phrase must attach to *about* since *know* does not posses an NP argument. Direct attachment of *What* to *know* would be licensed by a different supertag, namely (S\NP)/NP. Supertag information could therefore help to resolve extraction site ambiguities of this kind.

[45] The CCGrebank analysis uses the backward crossed composition combinator $<\mathbf{B}_{\times}$ here. Please refer to Steedman (2000, chapter 3) for more information on it.

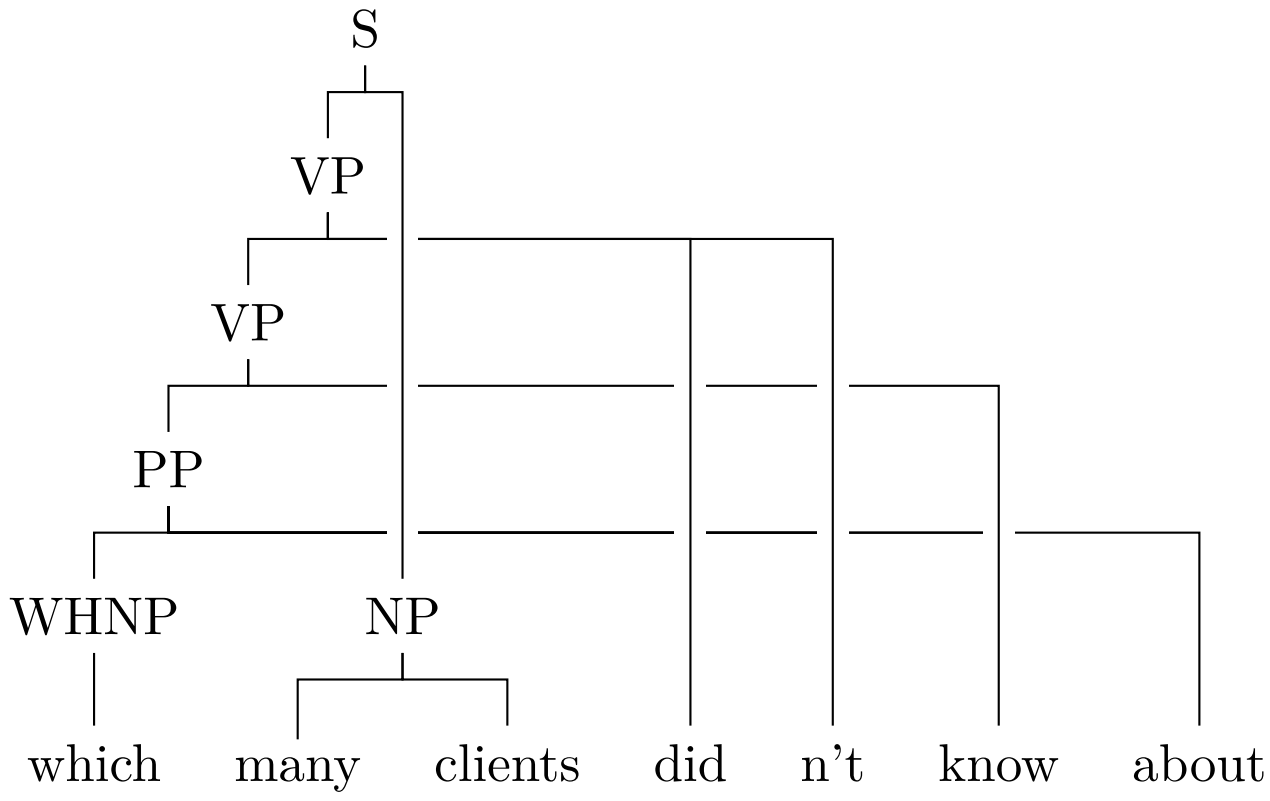

Figure 5.13: DPTB example for *wh*-fronting with attachment ambiguity (PP vs. VP)

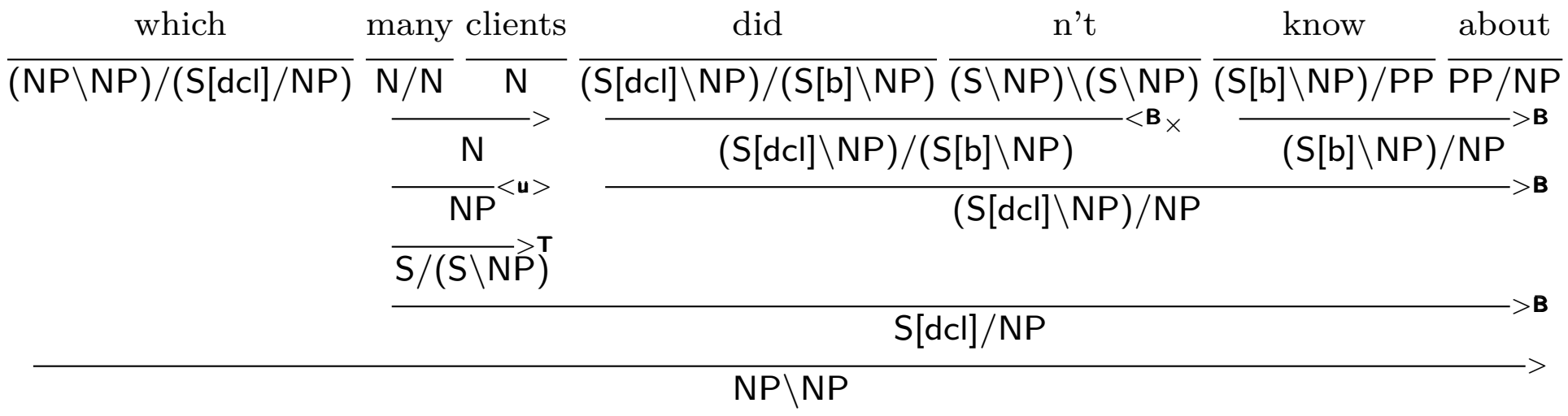

Figure 5.14: CCGrebank derivation corresponding to the phrase in figure 5.13.

Unfortunately, supertags are not always that informative for retrieving the exact discontinuous constituent structure. Figure 5.15 depicts a DPTB analysis containing the fronted adverbial phrase *How deeply*. It is, just like in the previous example, analysed as a daughter of the VP node. Compare this with the CCGrebank analysis in figure 5.16. Here, *How deeply* is simply treated as a sentential modifier for the rest of the phrase. Neither is it associated with the subject-less verbal phrase S[b]\NP nor is there an indicator for the fact that it is to be placed at a lower level since the phrase on its right does not show any missing arguments.

This also happens for other extractions of features realised as adjuncts (prepositional and adverbial phrases) instead of arguments of right-side components. I assume that the parser will be able to infer from the appearance of *do* in a sentence that the preceding sentential modifier most likely refers to a constituent to the right of *do*. Nonetheless, if there was more than one valid option for attachment beyond *do*, supertags would not help in resolving such an ambiguity.

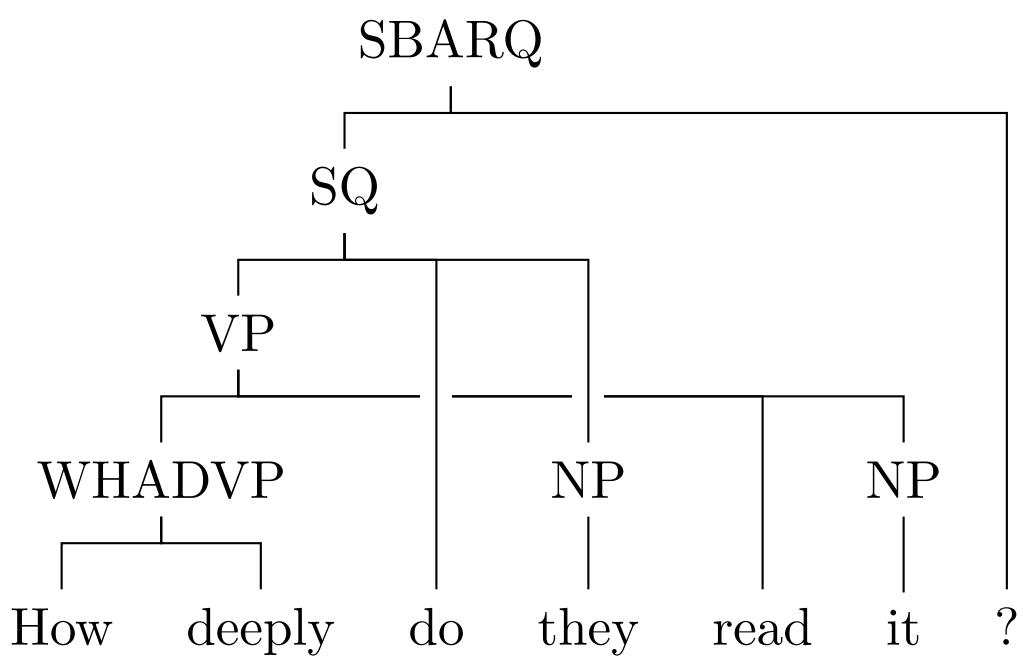

Figure 5.15: DPTB derivation of a sentence that contains a *wh*-word that asks for an adverbial phrase.

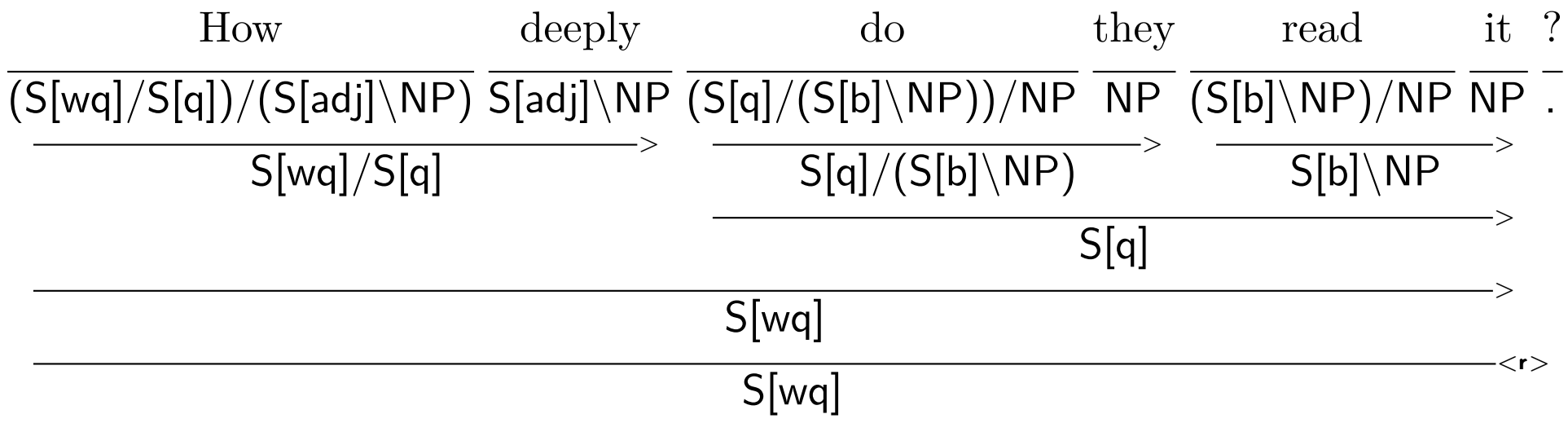

Figure 5.16: CCG derivation of the sentence *How deeply do they read it?* in the CCGrebank.

### 5.3.2 Fronted Quotations

Since the Penn Treebank is based on newspaper articles, it features a lot of quotations (Evang, 2011) - some of which result in discontinuous structures. Figure 5.17 shows an example of a fronted quotation from the DPTB. The phrase framed by quotation marks is analysed as a complement of *said*. The two elements are separated by a gap containing a comma and the closing quotation marks as well as the subject of *said* which are all attached one level higher in the tree.

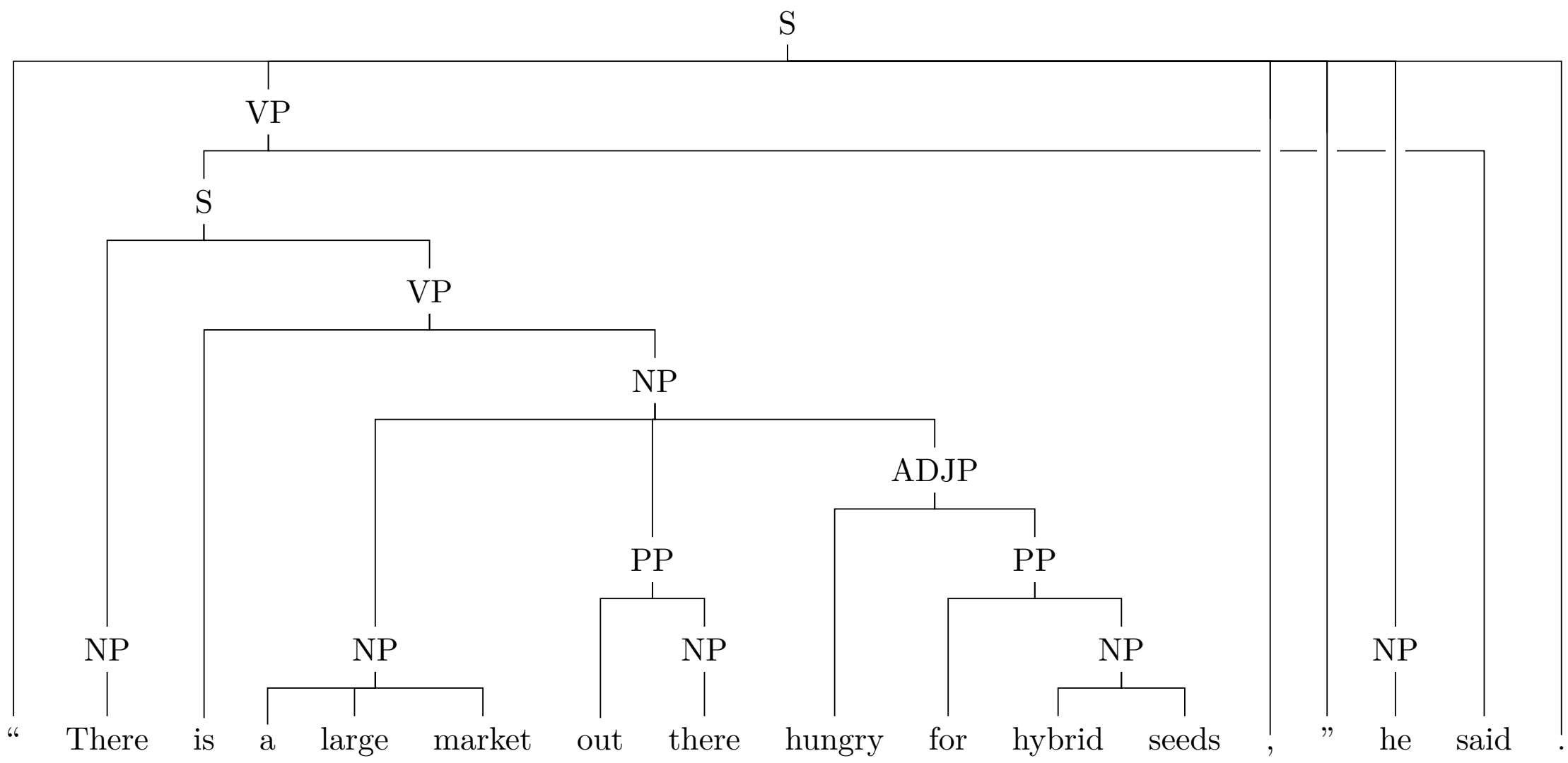

Figure 5.17: Discontinuous fronted quotation in the DPTB.

Figure 5.18 shows the CCGrebank analysis of the sentence. In the CCGrebank, opening and closing quotation marks are assigned special categories: LQU and RQU respectively. They are removed by use of (<r>) and thus cannot serve as an explicit indicator of the fronted quotation. The non-standard word-order is resolved through the category assignment of *said*. It expects as an inner argument a sentential category to its left. The first argument is the subject in standard position. This necessarily means that an intermediate constituent must be constructed by providing its subject *he* before merging with the quotation which is opposite to the order of the DPTB analysis.

Nevertheless, the supertag assignment can be an indicator for the constituent parser since it differs from the projective case where the quotation is positioned to the right as can be seen in the partial derivation in figure 5.19. Here the order of derivation still does not reflect the canonical analysis but the supertag assignment conforms to the fact that the quotation is expected directly to the left without intermediate arguments. Thus, supertag information could help to resolve discontinuities caused by fronted quotations.

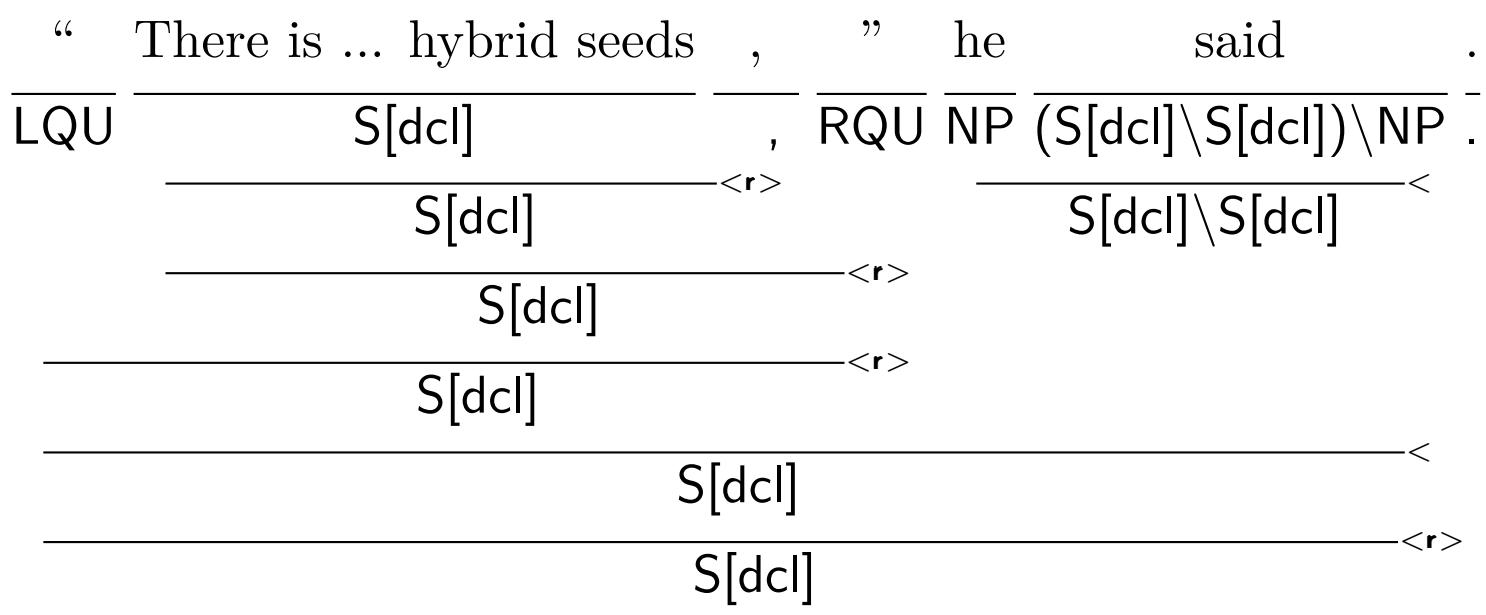

Figure 5.18: Fronted quotation in CCGrebank.

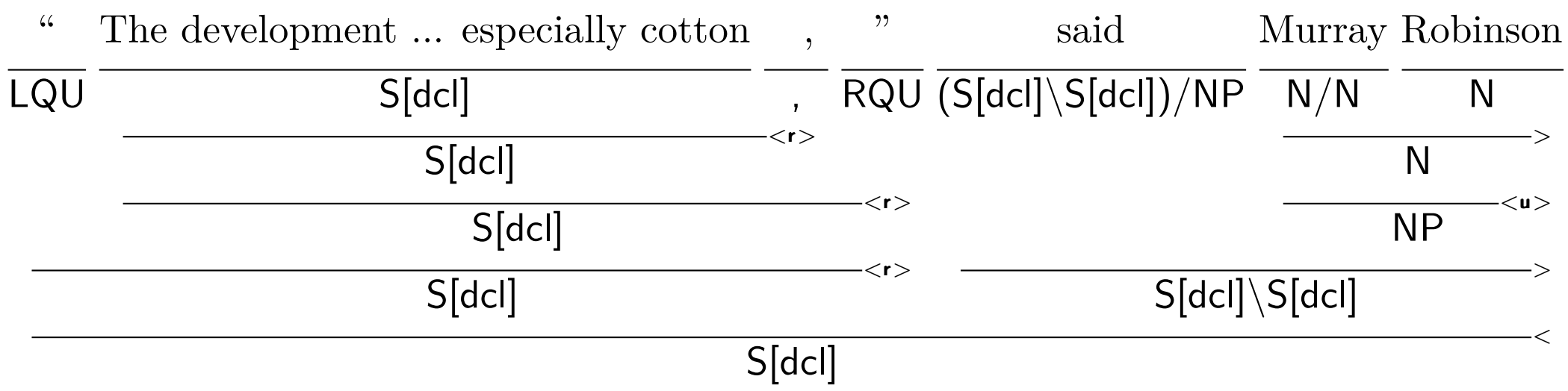

Figure 5.19: Projective fronted quotation with CCGrebank derivation.

### 5.3.3 Extraposed Dependents

The category of extraposed dependents mainly contains discontinuities that arise from appended clauses that refer to some preceding word. Contrary to *wh*-extractions and fronted quotations, no lexical trigger is present (Coavoux et al., 2019). Figure 5.20 shows an example where a subordinate clause is appended after the predicate of the main clause but relates to its subject.

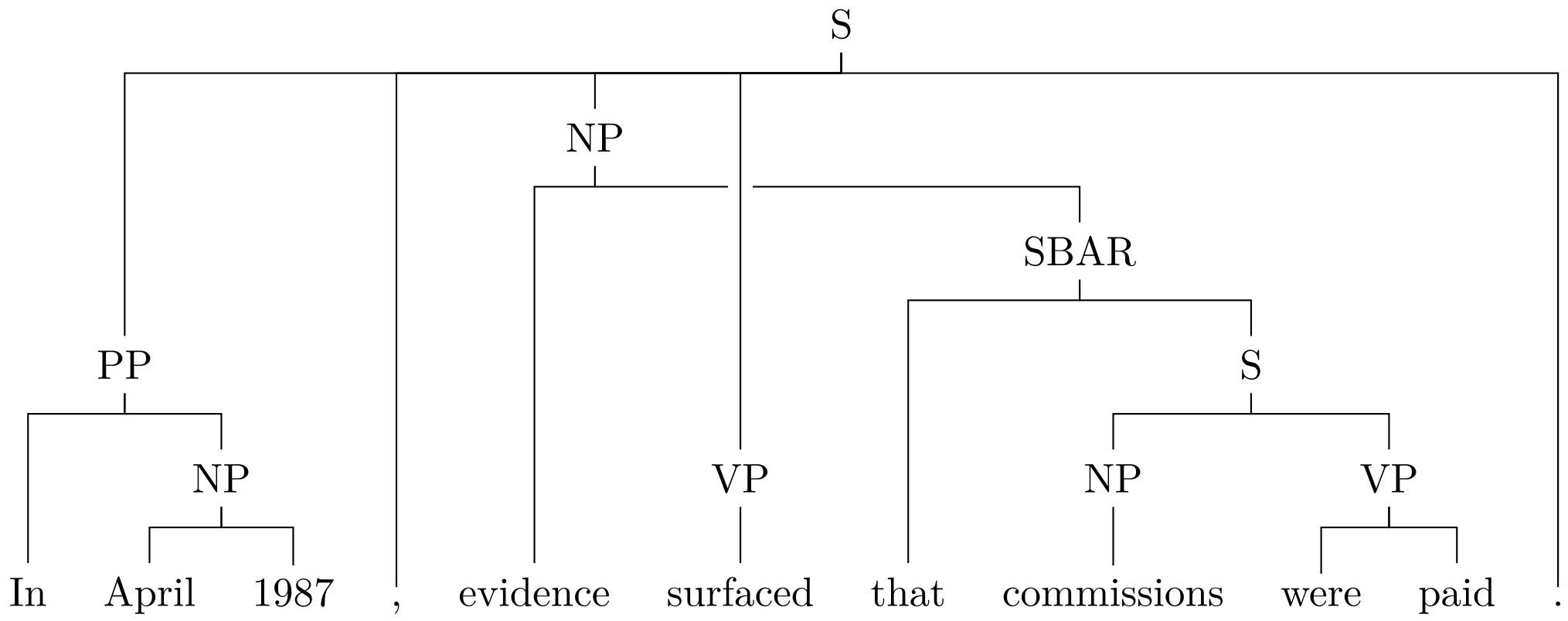

Figure 5.20: Extraposed subordinate clause from DPTB.

The CCGrebank analysis is shown in figure 5.21. Here, the subordinate clause *that commissions were paid* functions as an adjunct to the predicate *surfaced*. The lexical category of *that* expects a sentence to its right and as inner argument a VP (i.e. S[dcl]\NP) to its left, returning a VP. Then, the NP *evidence* is realised as a regular argument.

The attachment of the relative clause at the level of the NP cannot be recovered from this analysis. In fact, the same supertag assignment is given when the extraposed clause actually attaches at VP level in the DPTB analysis, see figure 5.22. Thus, the parser could be tempted to attach the extraposed clause incorrectly in one of the cases purely based on distributional properties of the corpus.

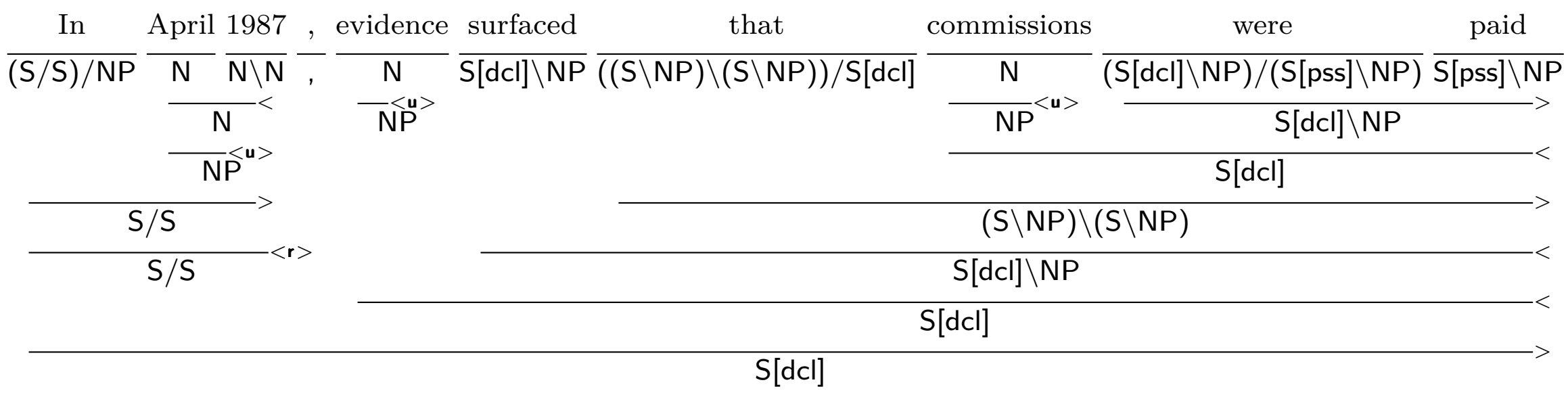

Figure 5.21: Extraposed subordinate clause in CCGrebank notation.

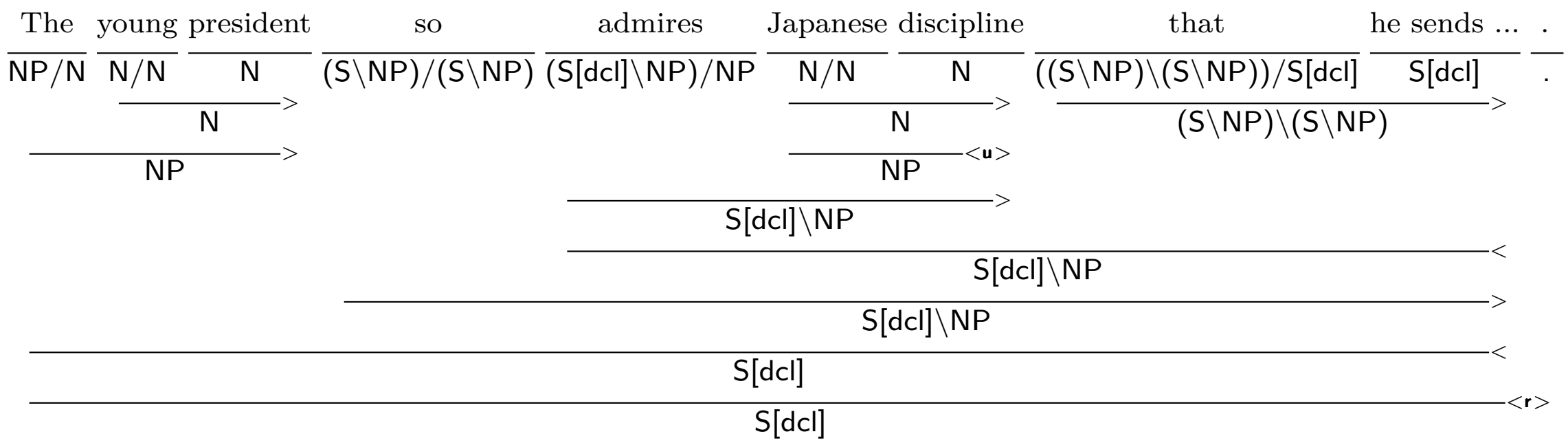

Figure 5.22: CCGrebank derivation where the subordinate clause *that he sends ...* is also a modifier to the VP in the corresponding DPTB annotation.

This unhelpful analysis is not limited to the conjunctive use of *that*. The same supertag is assigned to *than* in (5.3). The parser may be able to infer from the distribution of such cases that the extraposed clause should not form a direct constituent with the verb since *than* rarely modifies a VP. With that being said, the supertags are neither aiding in such cases nor do they help in resolving attachment ambiguity past the verb boundary. See for instance the CCG derivation in figure 5.23 that would correspond to three distinct DPTB analyses. The extraposed clause could be attached to *anything*, *the fundamentals* or *our business*.

(5.3) In any case, the firms are clearly moving faster to create new ads than they did in the fall.

Note that this issue does not arise from derivation ambiguity as explored in example 5.3 and in figure 5.6. It is either that the annotations in CCGrebank simply do not treat such phenomena properly or that the CCG formalism itself cannot appropriately account for discontinuities of this kind. While there is, by formulation of its objective, no necessity for CCG to make discontinuous constituents deducible from lexical category assignments; nonetheless, the observed behaviour does not live up to the paradigm of the formalism. Steedman (2000, chapter 2.1) postulates that every syntactic rule shall have a semantic representation. In our case, several different semantic representations arise from exactly the same syntactic operation of backward application. Applying ((S\NP)\(S\NP))/(S[dcl]\NP) would prescribe that the meaning of its right argument be associated with the object of the verbal phrase on its left in figure 5.23, with its missing left NP argument in figure 5.21 and with the verbal phrase as a whole in figure 5.22. Alternatively one could assume that in each of these cases, *that* refers to a distinct lexicon entry with a distinct semantic representation. While this is reasonable in the case of the VP modifier in figure 5.22, it is at least questionable for figures 5.21 and 5.23 that only differ in regards to the function the NP, modified by the extraposed clause, serves (i.e. which argument slot of the VP it fills).

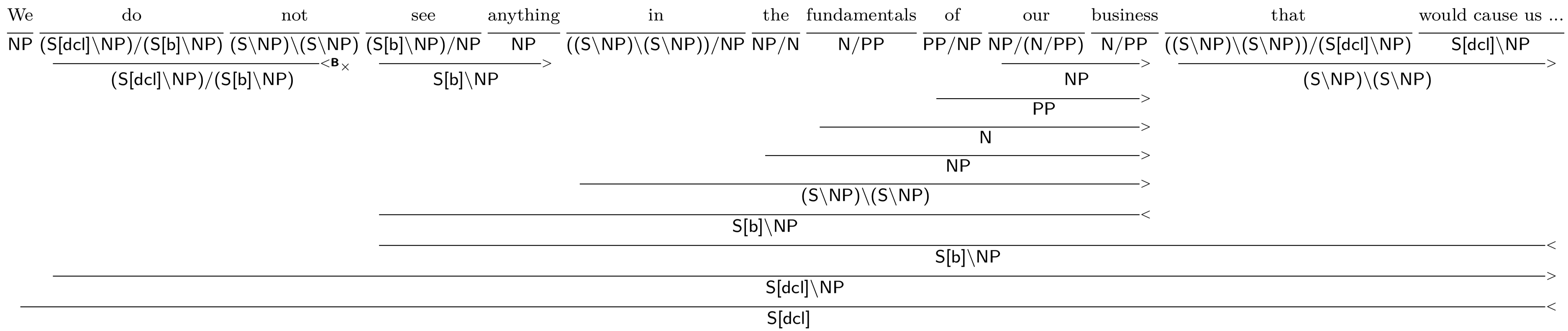

Figure 5.23: Extraposition together with its CCGrebank derivation where the corresponding DPTB annotation suggests attachment of the extraposition at the NP node directly above *anything*.

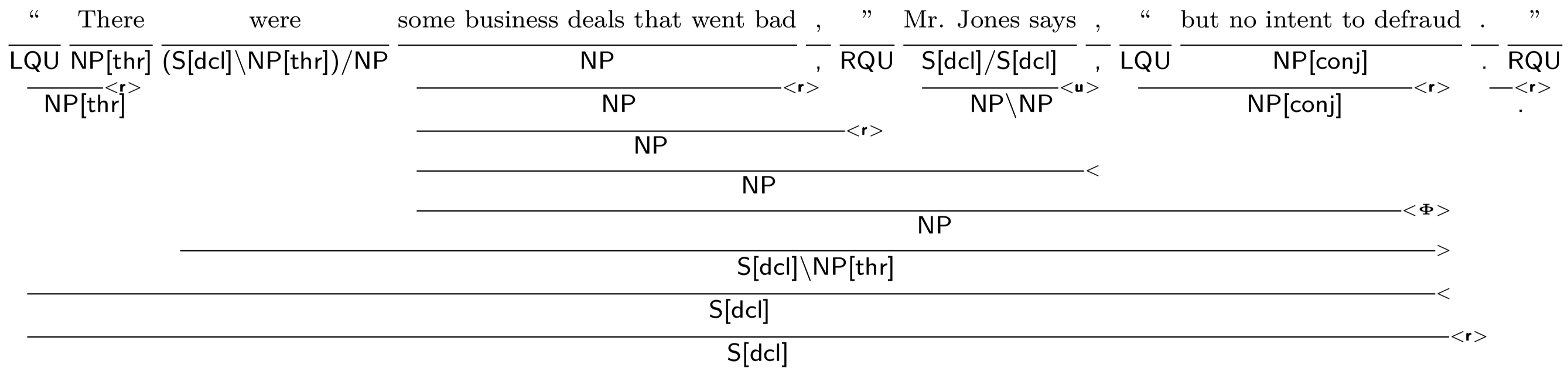

Figure 5.24: Circumpositioned quotation with an obscure CCGrebank analysis.

There is also a rare case of fronted extrapositions without a lexical indicator in the form of a *wh*-word. This happens when an argument or an adjunct appears before the subject and the verb and is attached at VP level instead of S (Evang, 2011). Figure 5.25 shows an example. Here, CCGrebank treats the extraposed part as a sentential modifier in the same way as *And* is treated. Since cases like this are very infrequent, I predict this shortcoming to have a negligible effect on the score of a supertag-enriched parser.

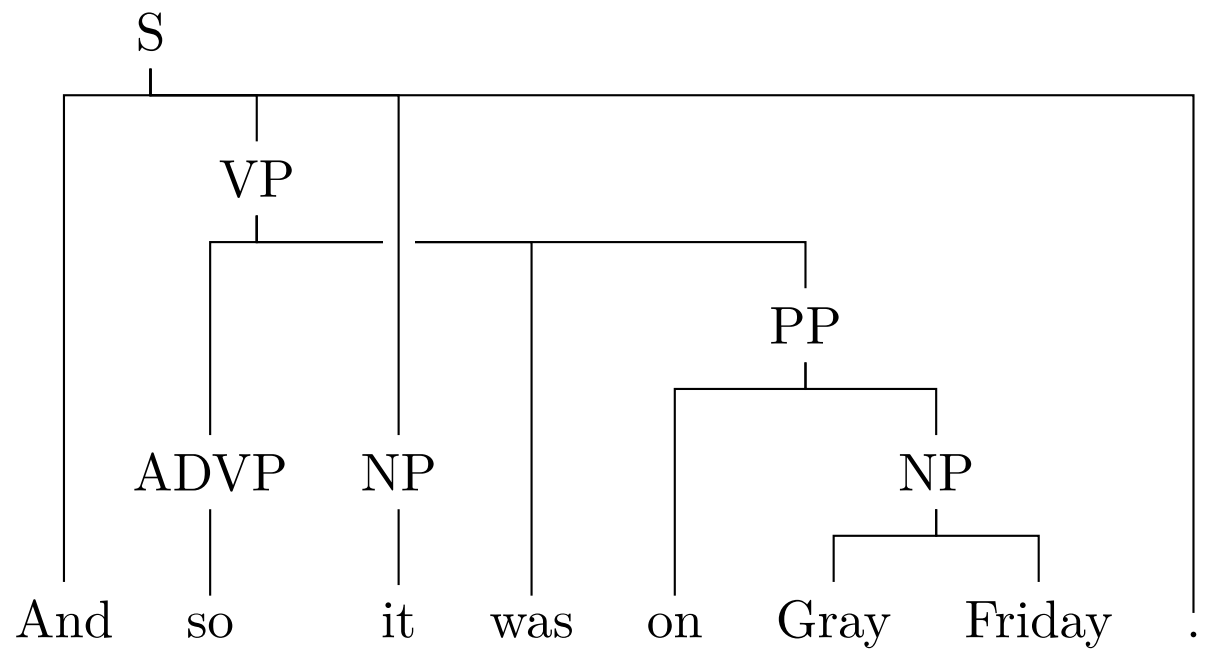

Figure 5.25: Fronted extraposition with the DPTB annotation.

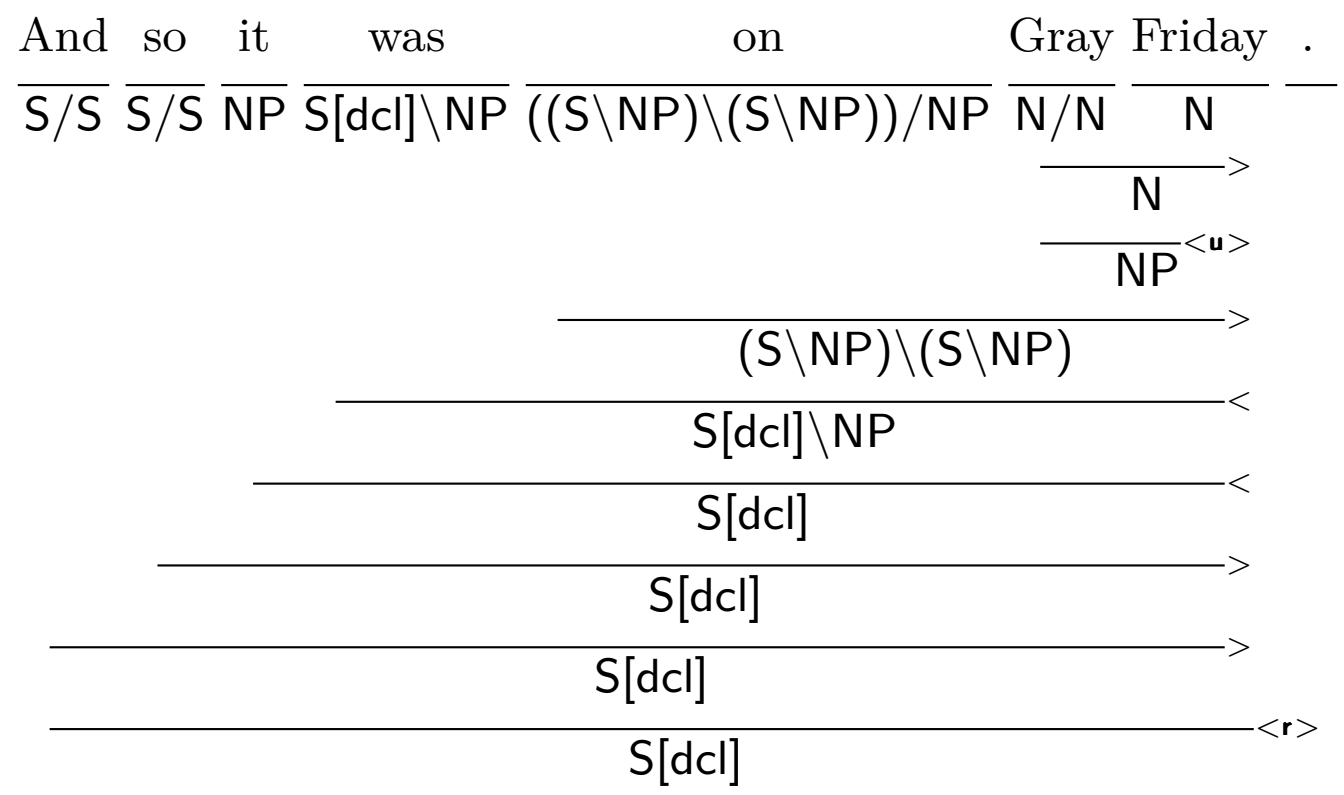

Figure 5.26: Fronted extraposition with the CCGrebank annotation.

#### 5.3.4 Circumpositioned Quotations

Figure 5.27 shows an example of a circumpositioned quotation from the DPTB. The two parts of the quotation *Working with lawyers* and *I need it* are dominated by a common parent S which is assigned as a complement to the intermediary verb *says* to form a VP. The structure is similar to the fronted quotation with the addition of a tail expression.

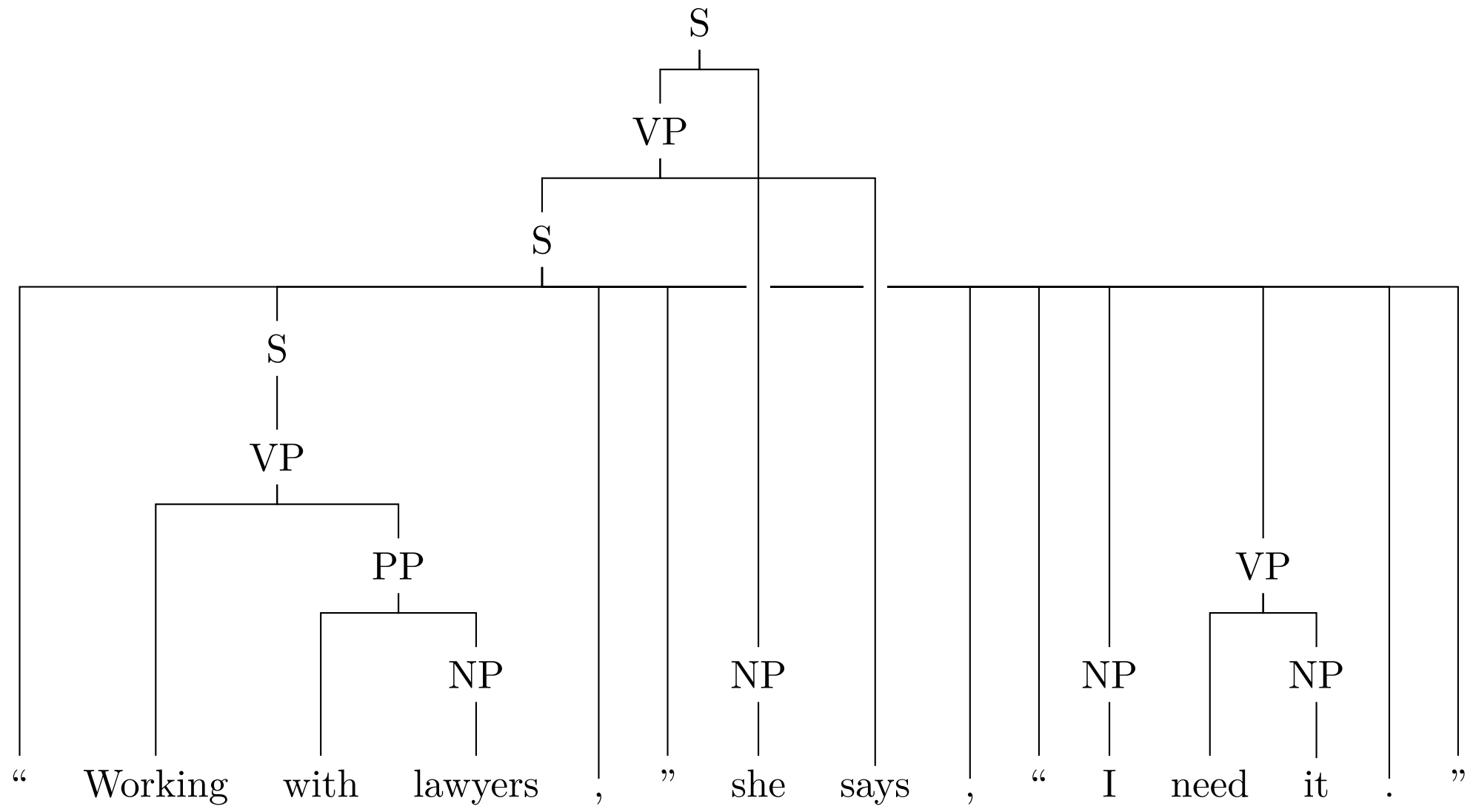

Figure 5.27: Circumpositioned qtoation and its DPTB analysis.

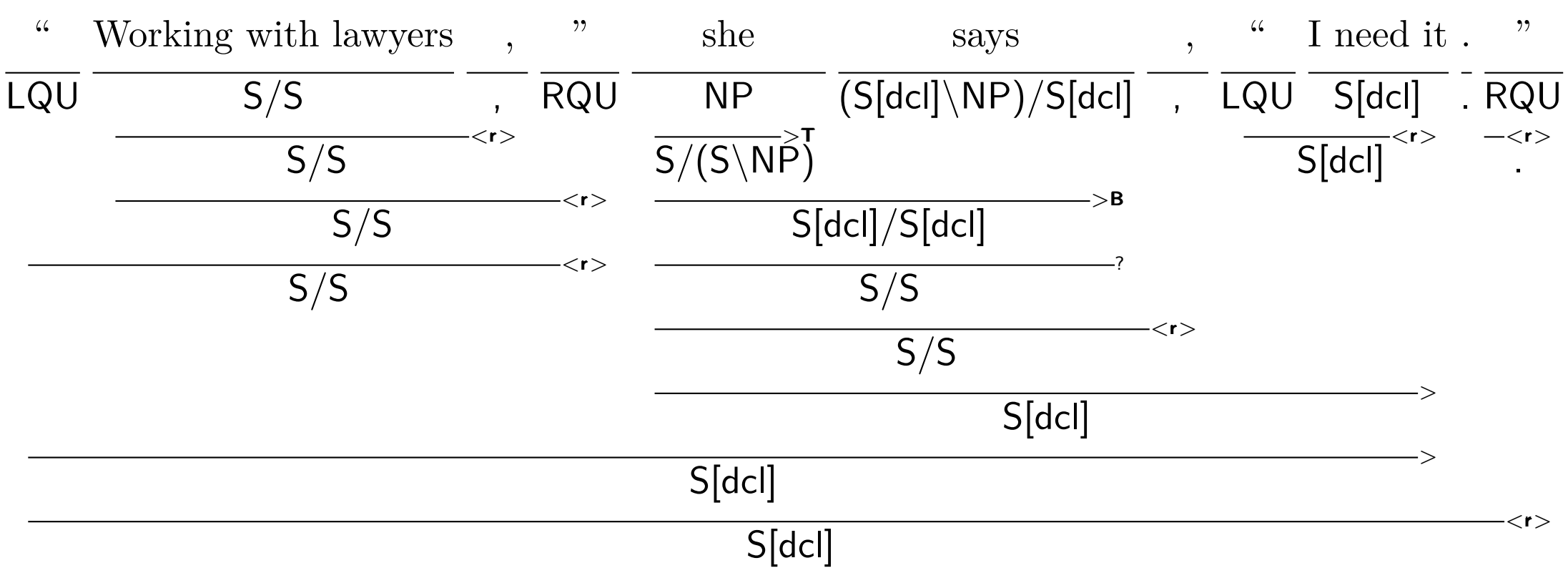

Figure 5.28: Circumpositioned quotation and its CCGrebank analysis.

The corresponding CCGrebank analysis shown in figure 5.28 is problematic. The substring *she says, “I need it.”* is analysed like a regular quotation in standard position yielding an S category. Thus, the prepended participle construction takes as argument not only the right sub-quote but also the quotative verb and its subject. No supertags give any indication of a circumpositioned arrangement.

The structure is, with the exception of an additional pair of quotation marks, indistinguishable from a sentence like *Reluctantly he said, “I want to go.”* where *Reluctantly* would receive category S/S. Therefore, I suspect that supertag information will not help the parser to resolve circumpositioned quotations. In the worst case, it might even mislead it.

The presence of quotation marks will probably counteract such an analysis. However, as Coavoux et al. (2019) points out, there are cases of circumpositioned quotations without quotation marks that are already hard to resolve by an unmodified neural model. Here, supertag information should be least beneficial.

In other cases, the CCG analysis is even more questionable. Figure 5.24 shows a case where the quotative verb and its subject are treated as a modifier of a preceding NP inside the quotation. This is done by converting the S[dcl]/S[dcl] category of *Mr. Jones says* into NP\NP by some unary rule. The supertag assignments alone suggest that *Mr. Jones says* should be followed by an S category. This could mislead the parser since the predicate can actually be found to the left. Overall, I assume supertag information on the basis of the CCGrebank relating to circumpositioned quotations to be of little informative value for the parser.

### 5.3.5 *It*-Extrapositions

*It*-extrapositions denote cases where an expletive, i.e. semantically empty, *it* is found in the standard-place of a clausal argument that is instead positioned at the end of the sentence (Evang, 2011). See (5.4) for an example. The extraposed clause could be regarded as specifying *it* semantically and is therefore analysed as one common discontinuous NP in the DPTB.

(5.4) a. It is easy to forget something.

b. To forget something is easy.

Figure 5.29 shows the DPTB analysis of part of the direct quotation in (5.5a) *to know what to expect at this point* and *It* are dominated by a common NP node.

(5.5) a. "It's hard to know what to expect at this point," said Peter Rogers, an analyst at Robertson Stephens & Co.

b. It is also likely to bolster fears that the Japanese will use their foothold in U.S. biotechnology to gain certain trade competitive advantages.

Figure 5.30 shows the CCGrebank counterpart. Here, *to know what to expect* has category S[to]\NP and thus expects an NP to its left. This is in line with the canonical analysis of *it* and the extraposed clause forming an NP constituent. However, due to the discontinuity, this argument slot cannot be filled. Instead, the lexical category assigned to *'s* deals with both elements. To account for *to know what ...* with category S\NP, *'s* has a second, inner right argument.

While the subject *it* functions as a regular, innermost NP argument of the predicate and while the non-finite clause is also realised as an argument of *'s*, the specialised supertag assignment in itself could still give the parser an indication to merge the preceding NP *it* when *to know what ...* is the focus item before merging *'s*. This clue is even stronger when including category features. The NP category of *It* is marked with the feature [expl], as is the left argument slot of *'s*. Thus, the parser could learn to associate this supertag with the necessity to find a discontinuous appended clause. The supertag assignments could further specify the boundaries between the projective parts and the extraposed clause in such a scenario.

This assumption is solidified by the fact that the projective use of infinite clauses as an adjectival modifier in combination with *It is* is tagged differently in the CCGrebank. Take for instance (5.5b) for which figure 5.31 shows part of the CCGrebank derivation. Here, the adjective *likely* receives a different supertag than *hard* in the last example. It expects a non-finite *to*-clause to its right as an additional argument. Consequently, *is* does not require a non-finite clause as a second right argument here. When one considers categorial feature the difference is even more pronounced since the NP and its corresponding argument slot lack the [expl] feature. Therefore, I expect the parser to be able to infer useful information for the treatment of *it*-extrapositions from supertag information.

### 5.3.6 Subject-Verb Inversion

Subject-verb inversion is an infrequent class of discontinuities where the subject can be found between the predicate on the left and one of the verb dependents on the right Evang (2011). It occurs mainly in combination with quotations and thus can be considered as the reverse case of fronted quotations. Figure 5.32 shows an example where the quotative verb *Said* takes the initial position in the sentence.

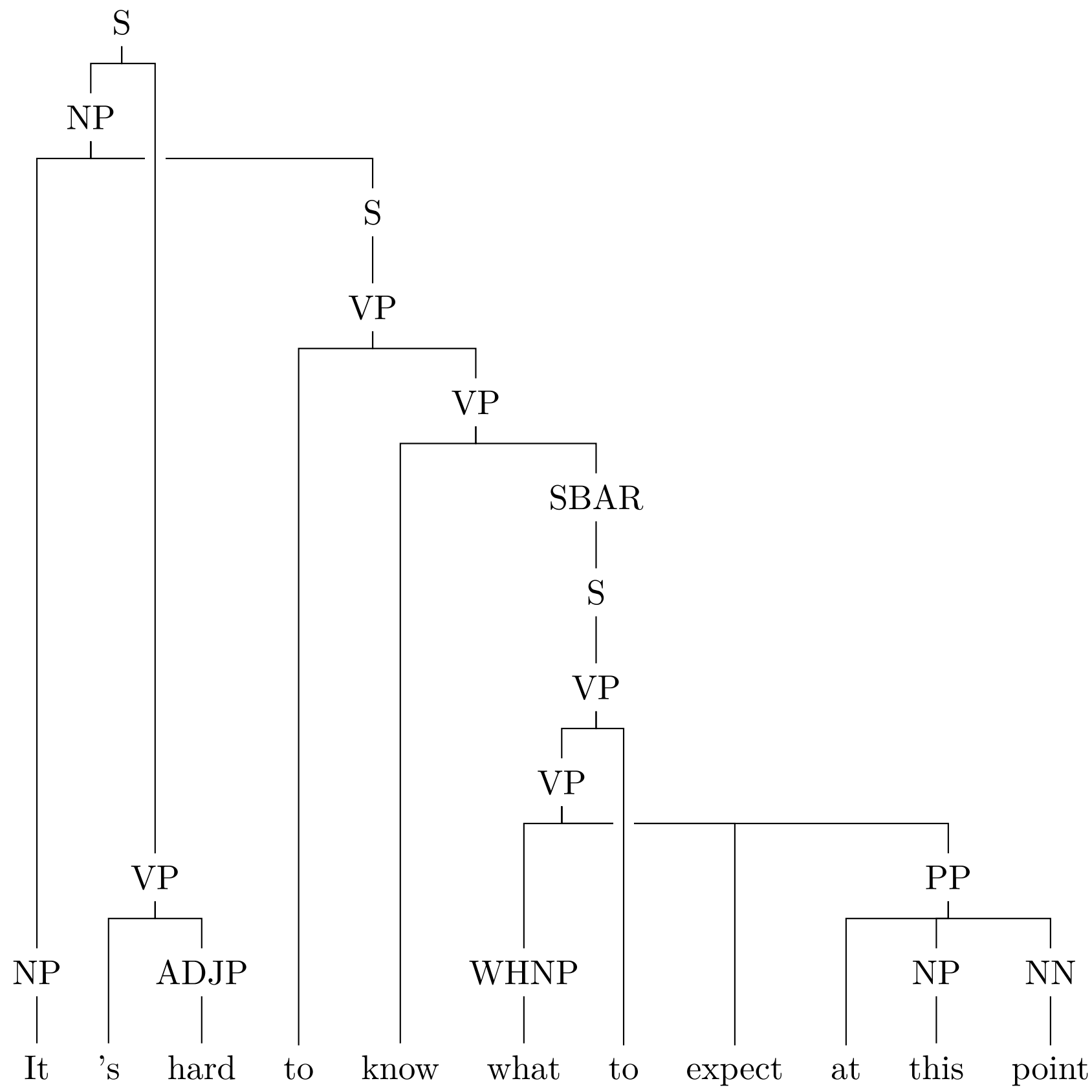

Figure 5.29: DPTB analysis of an *it*-extraposition.

The corresponding CCGrebank analysis is shown in figure 5.33. Here, the lexical category assignment of *Said* accounts for the reversed argument structure. A manual analysis showed that the category (S[dcl]/S[dcl])/NP is only assigned in cases of subject-verb inversion with quotations. It should therefore be a useful hint for the parser.

When disregarding the category feature [dcl], (S/S)/NP is also used for prepositions in front position sentential modifiers. The following sentence gives an example where *In* is assigned the category (S/S)/NP in the CCGrebank.

(5.6) In any case, supplies to patients won't be interrupted, the company added.

Following the category, *In* first accepts *any case* as an NP-argument and after that the S-argument *supplies to patients won't be interrupted*. This derivation order corresponds to the projective analysis found in DPTB. While the neural model might learn a correlation between the supertag and verbs of speech, I do not assume that supertag information will benefit the analysis of subject-verb inversion when the feature [dcl] is dropped. In the worst case, it could produce noise that pushes the parser towards analysing the quotative verb and the subject as a modifier of the quotation.

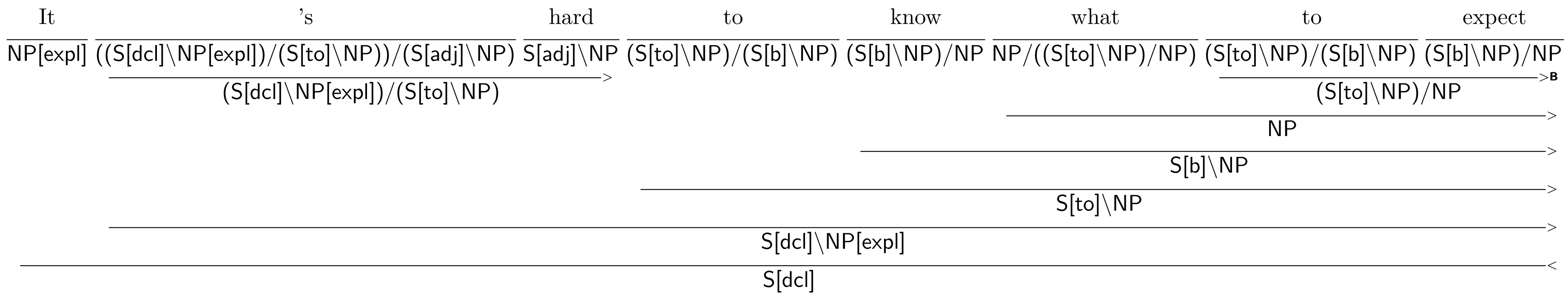

Figure 5.30: CCGrebank analysis of an *it*-extraposition.

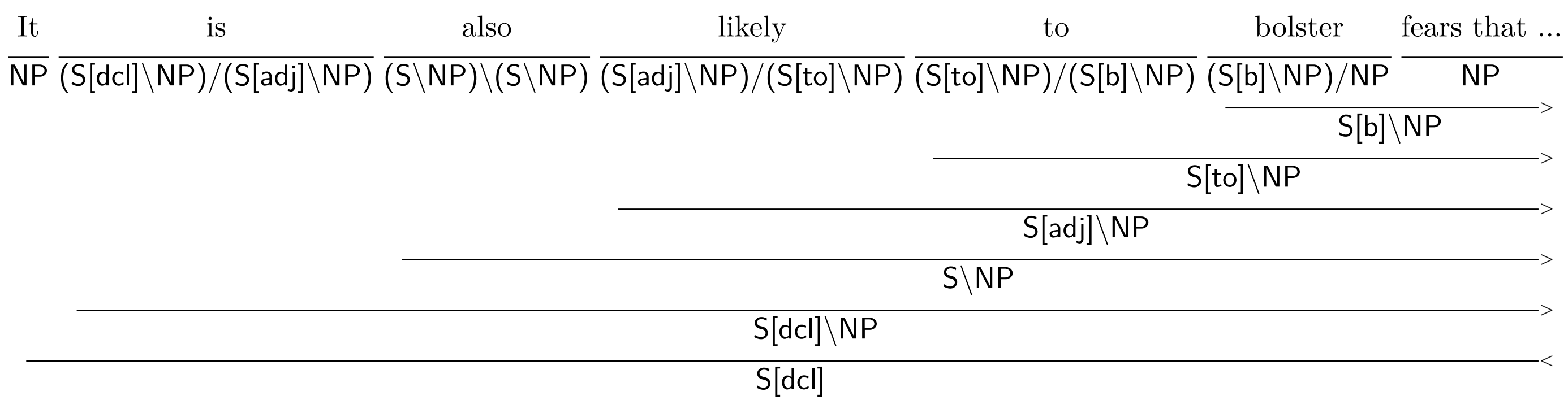

Figure 5.31: CCGrebank analysis of a projective sentence with a non-expletive use of *it*.

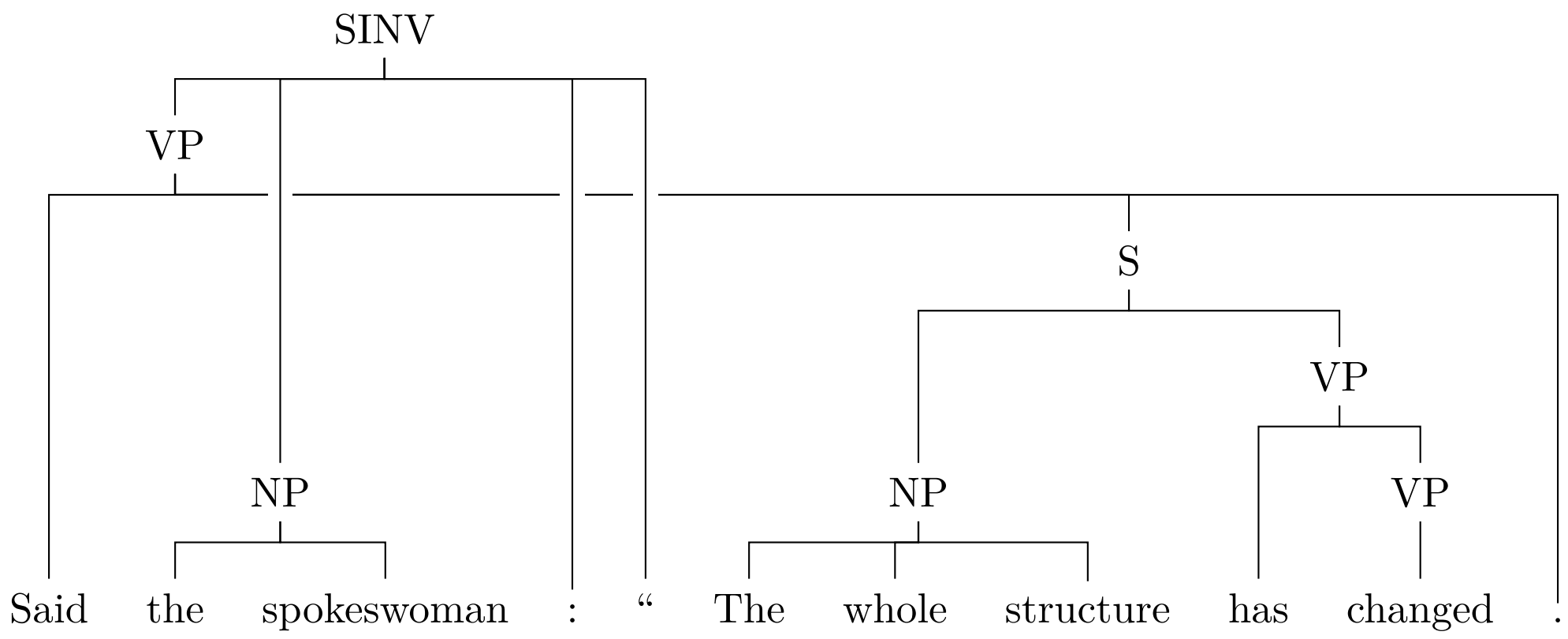

Figure 5.32: Example sentence for subject-verb inversion in the DPTB.

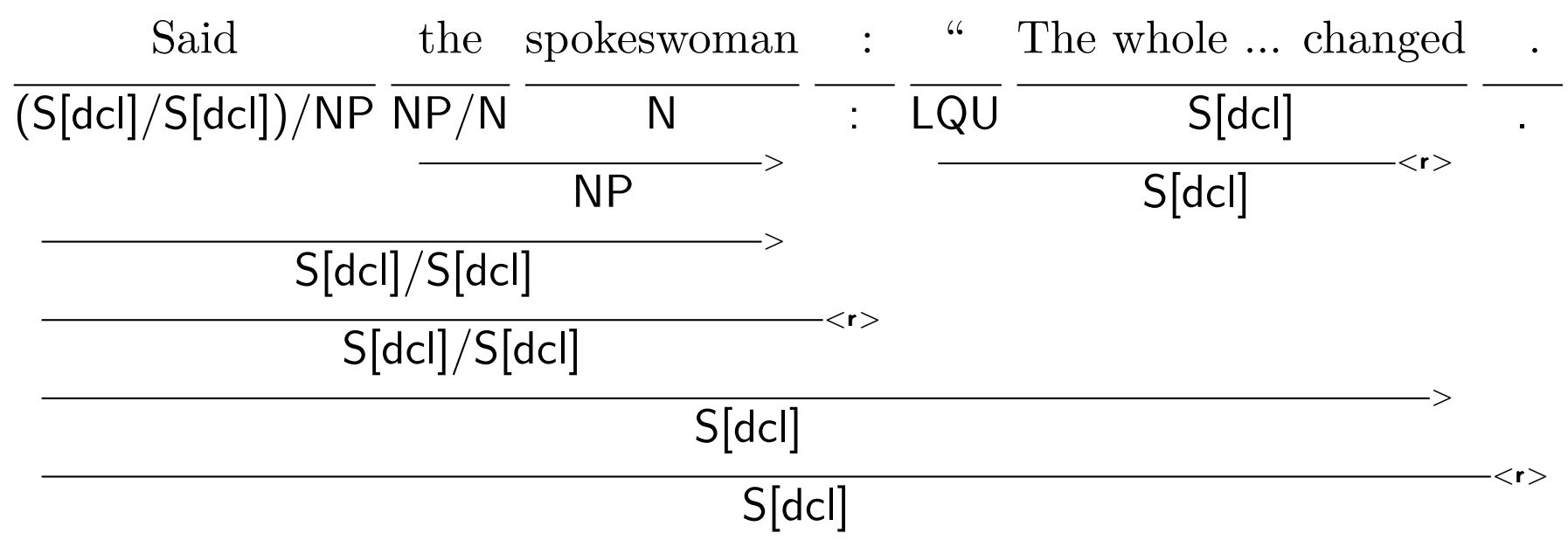

Figure 5.33: Example sentence for subject-verb inversion in the CCGrebank.

### 5.3.7 Summary

In the preceding paragraphs, I provided a detailed analysis of the treatment of discontinuous phenomena in the DPTB and in the CCGrebank. Furthermore, I explored the relationship between the two approaches concerning the information supertags may give a discontinuous constituent parser to construct correct parses for long-range phenomena. The analysis has shown that CCG supertags may be beneficial for resolving some types of discontinuities, namely *wh*-movement, fronted quotations and *it*-extrapositions. On the other hand, the analysis did not show a transparent correlation in the case of circumpositioned quotations, extraposed dependents and subject-verb inversion and in some cases even suggested a negative effect. Noteworthy is the result that quotations seem to be particularly difficult to resolve for the CCG formalism and that they are inconsistently analysed in the CCGrebank.

In their analysis of the ML-Gap parser, Coavoux et al. (2019) found that *wh*-extractions and *it*-extrapositions are, in absolute figures, the second and third largest sources of false negatives among the discontinuous phenomena. Due to the transition system and the neural network implementation being similar to the stack-free approach (Coavoux and Cohen, 2019) serving as the base for the following experiments, I expect the integration of supertag information to reduce these numbers.

Coavoux et al. (2019) report that extrapositions are the third largest class of discontinuities in the DPTB development split while they account for half of the false negatives their model produces. Unfortunately, deviating from my initial assumption, in many cases the supertags assigned in CCGrebank can not accurately account for attachment ambiguities with extraposed dependents. This class therefore most likely will remain to pose a challenge and as Coavoux et al. (2019) note, in many cases world-knowledge would be needed to resolve such cases.

It is important to point out that this prognosis is based on the assumption that the system is provided with gold supertags on the basis of CCGrebank. The actual performance will be dependent on the accuracy and style of implementation of the supertags and the variety of CCG, for instance whether categorial features are included.

On a general note not limited to discontinuities, supertag assignments alone can not be expected to help a constituent parser where they lead to spurious attachment ambiguity in their own right. Take for instance the case of figure 5.6 which was briefly discussed in the previous section. A constituent parser could not infer at what depth an attachment would be preferable based on the lexical category assignment. To deal with such cases, it might be beneficial to include information about gold parsing attachments in CCGrebank as a feature by enriching the supertags or training an additional auxiliary task.

A related area for exploration would be the integration of some notion of dependency structures into the model. While the `depccg` parser presented in section 5.2.2 has shown that a dependency factored approach can successfully treat attachment ambiguities inert to the CCG formalism itself, it would be worth to investigate if it could also complement a CCG-enriched constituent parser in those cases where CCG does not model discontinuous attachment ambiguities at all as in the case of extraposed dependents explored in section 5.3.3.

# 6. Integrating Supertags

In the following, two approaches for the integration of supertag information into discontinuous constituent parsing are explored. The *pipeline* model uses a pre-trained supertagger to assign the input sequences with lexical categories both at training and deployment. This assignment is then used as an additional input feature for the parser. In the *auxiliary* approach, the neural model is trained to predict both parsing actions and supertag assignments from a shared representation, exploiting statistical correlations between the two tasks.

## 6.1. Pipeline Approach

In natural language processing, it is common to process several tasks that build on each other in a sequential order. Using the outputs of one dedicated system as input for another model where the models have parameters that are independent of each other is called a *pipeline* approach (Goldberg, 2022, chapter 20).

The implementation of the stack-free discontinuous transition-based parser from Coavoux and Cohen (2019), presented in section 4, will serve as the basis for a supertagging pipeline approach. I use the supertag output of the `depccg` model trained on the CCGrebank (Yoshikawa et al., 2017, described in section 5.2.2) as an additional input to the parser. One option would be to supertag the input sentences and use a trainable embedding for the one-best supertag for each token as an input. Instead, I decided to use the probability distribution over the 511 lexical categories that `depccg` outputs for each token as an input feature. This should give the parser more useful information in cases where `depccg` is unsure of the correct assignment and assigns high probabilities to more than one choice.

Such an approach is called *cascading* (Goldberg, 2022, chapter 20). Instead of feeding the first model's final output to the next task, an intermediate representation that is trained to be informative for the lower-task prediction is used. Here, the last layer before the application of arg max is passed forward.

A common strategy is to *fine-tune* an upstream model when training the main task by including all or some of its parameters in the error gradient backpropagation (Jiang et al., 2020). I will not pursue such a strategy here since this would effectively enlarge the parsing model and skew a direct comparison with the baseline. Furthermore, re-optimizing the supertag predictor on the parsing task would contradict the goal of identifying what effect supertag features have on the parser.

### 6.1.1 Supertag Representation

On an abstract level, the `depccg` supertagger is a function *depccg* that takes a sequence of tokens $w_{0:n}$ and outputs a sequence of probability distributions over 511 categories. These include categorial features as in S[dcl] and NP[expl].

$$depccg(w_{0:n}) = \mathbf{d}_1, ..., \mathbf{d}_n \tag{6.1}$$

$$\mathbf{d}_i \in \mathbb{R}^{511}$$

$\mathbf{d}_i$ contain log probabilities. Thus, to obtain probabilities in the unit interval, one has to apply the exponential function to the entries. Prior to enriching the token representation, I pre-process the distribution using a two-layer feed-forward network with dropout before the second layer.

$$
\begin{aligned}
FF_{sup}(\mathbf{a}) &= \mathbf{h}'\mathbf{W_{2sup}} + \mathbf{b_{2sup}} \\
\mathbf{h}' &= \mathbf{d} \odot \mathbf{h} \\
\mathbf{d} &\sim \text{Bernoulli}(drop_{sup}) \\
\mathbf{h} &= \tanh(\mathbf{a}\mathbf{W_{1sup}} + \mathbf{b_{1sup}})
\end{aligned} \tag{6.2}
$$

$$\mathbf{a} \in \mathbb{R}^{511},\ \mathbf{W_{1sup}} \in \mathbb{R}^{511 \times d_{sup}}, \mathbf{W_{2sup}} \in \mathbb{R}^{d_{sup} \times d_{sup}},\ \mathbf{b_{1sup}}, \mathbf{b_{2sup}} \in \mathbb{R}^{d_{sup}}$$

Then I enrich the token representation $\mathbf{w}_i$ by concatenating the output of the feed-forward network to the word embedding and the character-aware encoding:

$$
\begin{aligned}
\mathbf{w}_i &= \mathbf{c}(w_i) \circ \mathbf{e}(w_i) \circ \mathbf{s}(w_i) \\
\mathbf{s}(w_i) &= FF_{sup}(\exp(\mathbf{d}_i))
\end{aligned} \tag{6.3}
$$

$$\mathbf{d}_i \in \mathbb{R}^{511},\ \mathbf{s}(w_i) \in \mathbb{R}^{d_{sup}},\ \mathbf{w}_i \in \mathbb{R}^{d_{token}},\ d_{token} = d_{char} + d_{word\text{-}emb} + d_{sup}$$

The 2-layer bi-LSTM transducer then operates over the enriched token representations constructing context-aware representations that encode character and word information as well as information about the supertag structure in the sentence for each token. The architecture of the model is visualised in figure 6.1.

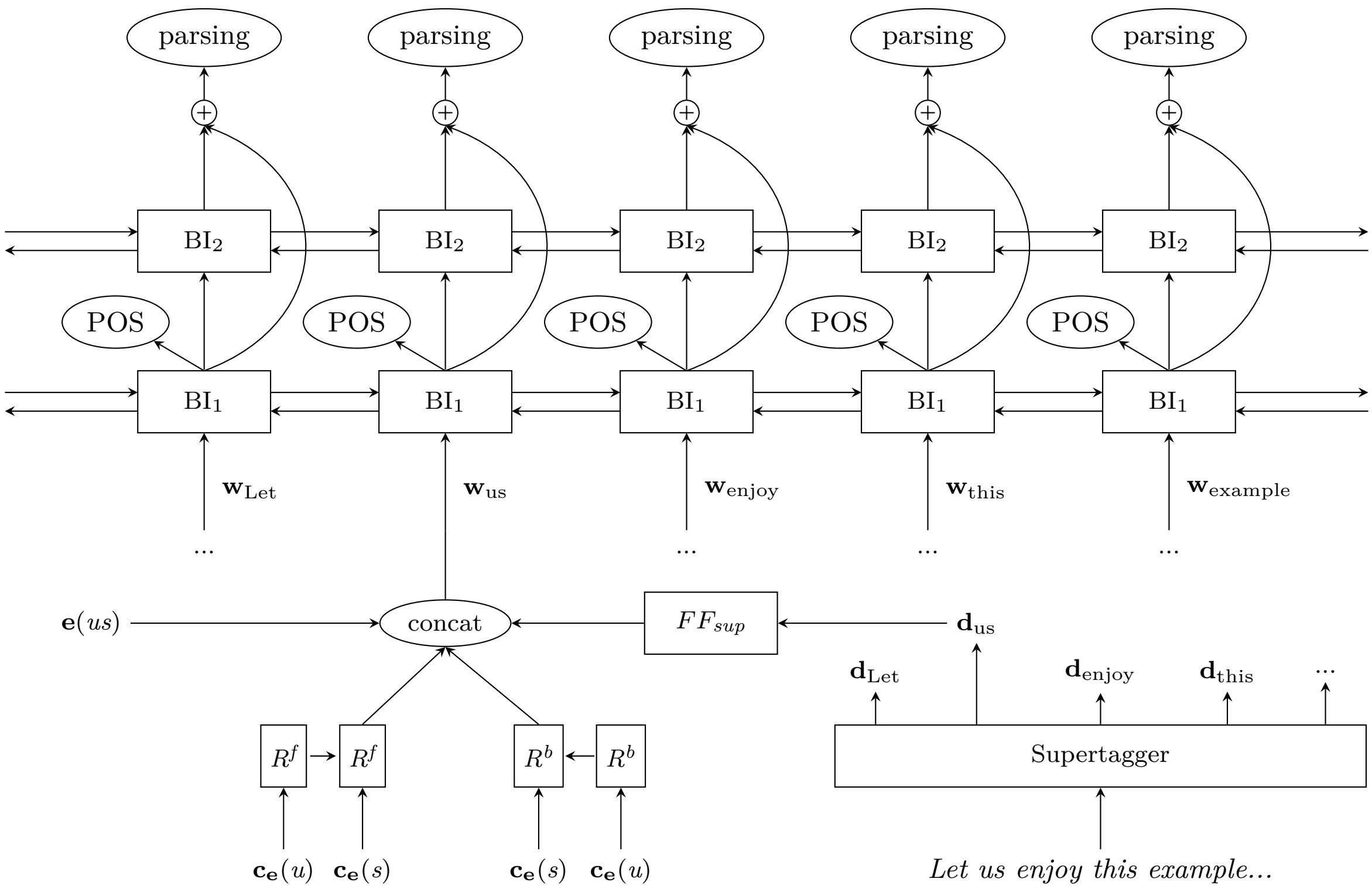

Figure 6.1: Stack-free neural parser with supertagging pipeline approach. For clarity the token representations and the supertag prediction vectors are subscripted with words instead of indices. Figure inspired by Goldberg (2022, chapter 20.2).

### 6.1.2 Hyperparameters

The rest of the model resembles the implementation of Coavoux and Cohen (2019). Their choice of hyperparameters given in table 4.1 remains unchanged. Approach-specific hyperparameters are listed in table 6.1. The dimensionality of the supertag feed-forward network is set to the same value as the size of the probability distribution it receives as input in order to rule out any loss of information that could impede model performance. In the comparison of results given in section 6.3 the model will be called $\textsc{Pipe}_{\text{rebank}}$.

| Architecture hyperparameters | | |
|---|---|---|
| Dimension of supertag input representation | $d_{sup}$ | 511 |
| Number of layers for supertag FF network | | 2 |
| Activation function for supertag FF network | | tanh |
| **Optimisation hyperparameters** | | |
| Dropout for supertag FF network | $drop_{sup}$ | 0.1 |

Table 6.1: Hyperparameters of the supertag pipeline model. The other architecture and optimisation parameters match those of the baseline listed in table 4.1.

## 6.2. Auxiliary Approach

A common technique related to fine-tuning in pipeline-systems is *auxiliary* or *multi-task learning* which was first explored by Caruana (1997). Instead of using the predictions of a specialised system as additional input for the main task model, the idea is to jointly train individual networks for both tasks that share some part of their structures (Goldberg, 2022, chapter 20). Such an architecture learns a shared core representation that is thought to leverage synergies between the tasks.

In deep neural network models, it is common practice to arrange easier tasks at lower levels, stipulating a hierarchy between tasks. Instead of sharing the entire deep neural network, some task-specific predictors are provided with vectors resulting from lower layers of the network. This strategy is known as *collective sharing* or *stack propagation* (Goldberg, 2022, chapter 20) and was first explored for natural language processing by Søgaard and Goldberg (2016).

Auxiliary tasks based on the main task dataset effectively allow to enlarge the available training data without the need to acquire additional corpora. Often it is possible to break down the main objective into lower-level auxiliary tasks (Zhu and Sarkar, 2019). Furthermore, when training in parallel, multi-task learning can have a regulatory effect on large models that tend to overfit (Goldberg, 2022, chapter 20). Introducing auxiliary objectives forces the shared substructures to be more general which in turn helps with the treatment of data not seen during training.

The following sections present several architectures for the inclusion of CCG supertagging as an auxiliary task in the discontinuous constituent transition-based parser of Coavoux and Cohen (2019).

### 6.2.1 Simple Model

The original stack-free discontinuous parser of Coavoux and Cohen (2019) features two biLSTM layers where POS tagging is supervised at the first layer and parsing actions are supervised at the second layer. Thus, the first biLSTM network is shared by both POS tag prediction and parsing while the second biLSTM network is only included in the computation of parsing transitions. I propose integrating CCG supertagging as an auxiliary task by enlarging the LSTM stack through the insertion of an intermediate layer. The output of this biLSTM module is then used as input for a task-specific feed-forward network for CCG supertagging and as input for the top biLSTM network that is exclusive to parsing action prediction.

This arrangement is motivated by two factors. Firstly, Søgaard and Goldberg (2016) show that using POS-tagging as an auxiliary lower-level task for CCG supertag prediction in a three-layer recurrent neural network setting yields improvements in supertagging accuracy suggesting a natural hierarchy between the two tasks. Secondly, supervising CCG supertagging at a lower level than parsing is in line with the traditional view on supertags as “almost parsing” (Bangalore and Joshi, 1999). Furthermore, the findings of the manual analysis in section 5.3 suggested that a constituent parser would need to be able to contextualise supertag assignments to benefit from them. This motivates maintaining a final biLSTM layer that is only included in parsing action prediction and operates over the vectors that are also provided to the supertagging feed-forward network.

The proposed architecture can be formalised as follows. Each layer $l$ consists of a biLSTM module receiving the last layer's output as input.

$$\mathbf{m}_{l,0:n} = biLSTM_l^*(\mathbf{m}_{l-1,0:n}) \tag{6.4}$$

For the first layer the token representations are used as input:

$$\mathbf{m}_{1:0:n} = biLSTM_1^*(\mathbf{w}_{0:n}) \tag{6.5}$$

In the baseline model, the output of the first LSTM layer is added to the output of the second LSTM layer (residual connection). The combined output is directed to the feed-forward networks for parsing action prediction. I opted to replicate the simple residual connection from the second LSTM's input to its output for the additional layer. The first LSTM does not feature a residual connection since its input and output dimensionalities do not match.

$$\mathbf{m}_{l,0:n} = biLSTM_l^*(\mathbf{m}_{l-1,0:n}) + \mathbf{m}_{l-1,0:n} \qquad \text{for } l \in \{2,3\} \tag{6.6}$$

Now, the feed-forward network for POS-tagging $FF_{POS}$ still receives $\mathbf{m}_1$ as input while the parsing action scorers $FF_{label}$ and $FF_{struct}$ are provided with the last layer $\mathbf{m}_3$. For supertagging, a specialised feed-forward network with the same architecture as $FF_{POS}$ is introduced that receives $\mathbf{m}_2$ as input. Let $B$ be the alphabet of supertags.

$$\begin{aligned} FF_{sup}(\mathbf{x}) &= \mathbf{x}'\mathbf{W_{sup}} \\ \mathbf{x}' &= \mathbf{d} \odot \mathbf{x} \\ \mathbf{d} &\sim \text{Bernoulli}(drop_{sup}) \end{aligned} \tag{6.7}$$

$$\mathbf{W_{sup}} \in \mathbb{R}^{dim_{hid} \times |B|},\ drop_{sup} \in [0,1]$$

For an input sequence $w_{0:n}$ the probability of a sequence of supertags $s_{0:n} = s_1, ..., s_n$ is then computed as follows:

$$P(s_{0:n}|w_{0:n}) = \prod_{i=1}^{n} \text{Softmax}(FF_{sup}(\mathbf{m}_{2,i}))^{[\text{idx}_B(s_i)]} \tag{6.8}$$

$$\mathbf{m}_{2,i} \in \mathbb{R}^{d_{hid}},\ FF_{sup}(\mathbf{m}_{2,i}) \in \mathbb{R}^{|B|}$$

In the train split of the CCGrebank, 1574 distinct lexical category assignments occur, thus $|B| = 1574$. Only 511 supertags appear 10 or more times (Honnibal et al., 2010). The large number of supertags and the small amount of training data for two thirds of the categories may make it hard to achieve useful generalisations for rare phenomena.

An overview of the modified architecture of the model is given in figure 6.2.

**Objective Function** I perform supervised learning using the lexical category annotations provided by the CCGrebank. For training, sections 2-21 are used so that the auxiliary task does not extend the domain of the main task DPTB dataset.

The neural network parameters are optimised during training. The objective function introduced in section 4.5 now includes the supertagging loss.

$$\begin{aligned} \mathcal{L} &= \mathcal{L}_t + \mathcal{L}_p + \mathcal{L}_{sup} \\ \mathcal{L}_{sup} &= -\sum_{i=1}^{n} \log P(s_i|w_{0:n}) \end{aligned} \tag{6.9}$$

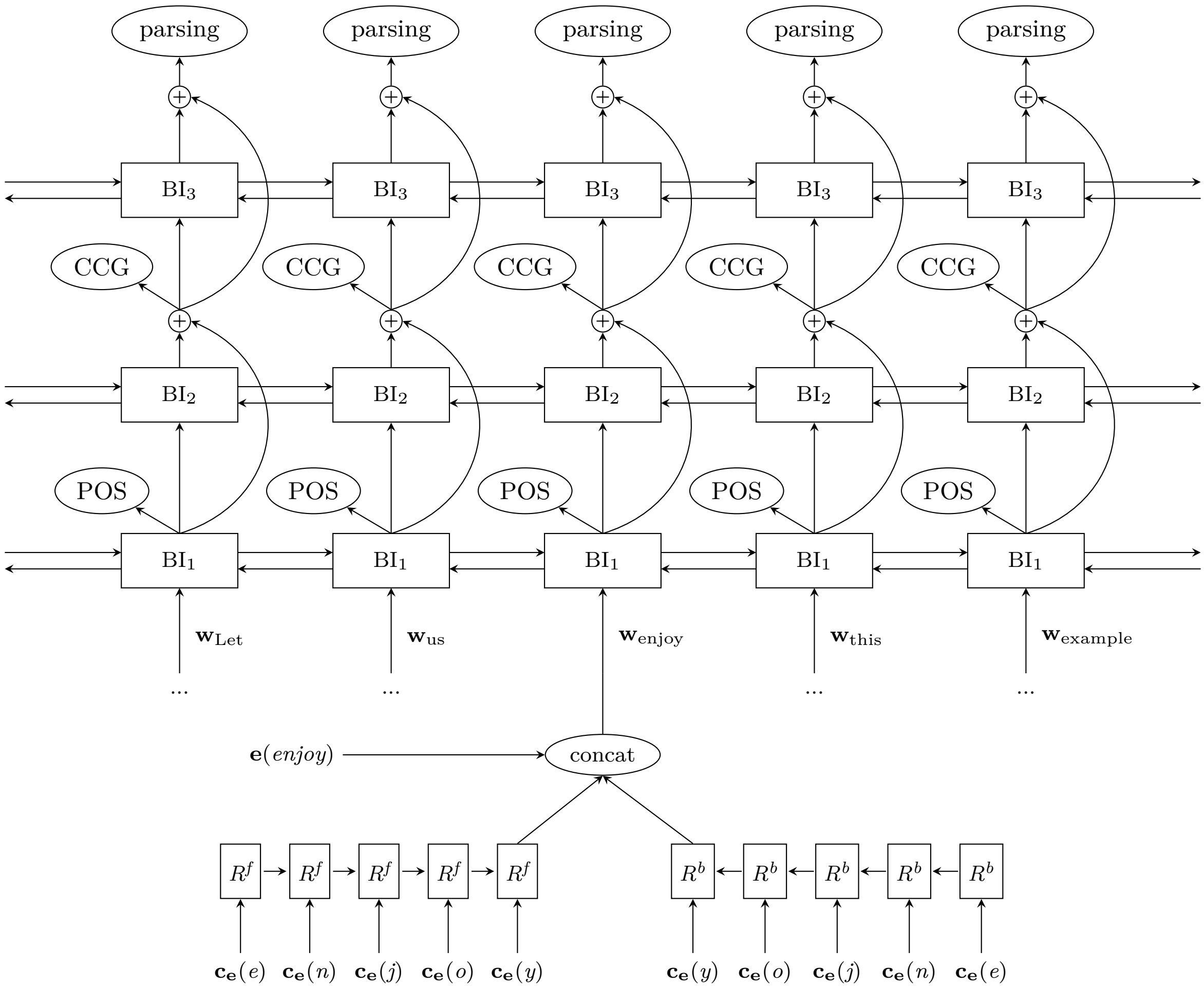

Figure 6.2: Stack-free neural parser with a three-layer biLSTM network and supertagging as an auxiliary task. For clarity the token representations and the supertag prediction vectors are subscripted with words instead of indices. Figure inspired by Goldberg (2022, chapter 20.2).

In practice, instead of strictly alternating between the three tasks, a strategy called *uniform sampling* (Sanh et al., 2019), I iterate through the randomly shuffled list of all available pairs of task and annotated sentence. At each step, the network performs an optimisation for the chosen pair. This is known as *proportional sampling* where the probability of a task is proportional to its corpus size and was first proposed by Sanh et al. (2019). Each epoch iterates through the full list. The pseudo-code for this training procedure is given in algorithm 6.1.

This modification to the model was necessary since aligning the CCGrebank with the DPTB is not possible due to a few missing sentences in the CCGrebank (0.56%, i.e. 274 of the 49.208 sentences in the Penn Treebank). These sentences were already left out in the original CCGbank since they could not be converted to CCG by the translation algorithm utilised by Hockenmaier and Steedman (2007). They note that 107 of these cases include sentential gapping and 66 non-sentential gapping while only a minor number of the remaining sentences was excluded because of an erroneous treatment of long-range dependencies. According to Hockenmaier and Steedman (2007) the translation algorithm incorrectly identified a complement as an adjunct. Despite their small number and without investigating these exclusions any closer, it could of course be the case that they would provide important information for the treatment of certain discontinuous phenomena that already occur very infrequently in the corpus (cf. section 5.3). However, manually reintegrating these sentences would be very labour-intensive.

**Algorithm 6.1** Proportional task sampling. Inspired by (Goldberg, 2022, chapter 5).

**Input:** a list of task-dataset pairs $M$, a number of training epochs $n$

$D \leftarrow [\,]$
**for all** $\langle task, dataset \rangle$ in $M$ **do**
  **for all** training example $\langle x, y \rangle$ in $dataset$ **do**
    $D$.append($\langle task, x, y \rangle$)
  **end for**
**end for**
**for all** $epoch = 1$ to $n$ **do**
  **for all** training example $\langle task, x, y \rangle$ in $D$ **do**
    loss_node ← build_computation_graph$_{task}$(x, y)
    loss_node.forward()
    gradients ← loss_node.backward()
    update_parameters$_{task}$(gradients)
  **end for**
  shuffle($D$)
**end for**

### 6.2.2 Residual Connections

In deep multi-layer neural network models trained with gradient descent the effective backpropagation of errors can be difficult to achieve. A central challenge stems from the problem of vanishing gradients where the gradient being backpropagated through the network decays with each additional layer, hardly affecting early parameters of the system, or grows exponentially due to repeated scaling (Bengio et al., 1994; Hochreiter et al., 2001; Glorot and Bengio, 2010). Thus, with increasing depth, a model tends to be harder to train effectively.

*Residual*, *skip* or *shortcut connections* are used to mitigate this effect by allowing for the direct flow of gradients from the final loss computation to lower parts of the system through the establishment of connections that leave out certain components or layers of the network. Motivated by the *ResNet* model of He et al. (2016), this is typically. This strategy has the additional benefit of enabling the model to easily learn the identity function which tends to be difficult for models with non-linear activation functions (He et al., 2016).

Stacked recurrent neural networks are connected both horizontally, with the hidden units in the same layer representing a time-step in the sequence, and vertically, as a feed-forward process through the layers at the same time-step. The architecture of LSTMs is designed to prevent the problem of vanishing gradients for repeated horizontal application by enforcing constant error flow along the sequence using a special cell unit controlled by gates that perform element-wise multiplication to *forget* information in the cell or add new (transformed) *input* (Hochreiter and Schmidhuber, 1997; Sundermeyer et al., 2012) (cf. definition 4.18). However, when stacking several LSTM models vertically, the problem reoccurs and leads to slow and difficult training (Wu et al., 2016b).

Therefore, it is common practice to include residual connections in stacked LSTM models. They were first proposed for stacked recurrent neural networks by Raiko et al. (2012) and explored for LSTM stacks specifically by Wu et al. (2016b). As explained in section 4.5, Coavoux and Cohen (2019) use a simple skip connection that adds the output vector of the first LSTM layer to the output of the second layer element-wise. This allows information to bypass the second LSTM network.

With larger vertical LSTM stacks, the question of where to add residual connections becomes non-trivial. While simply using an additive residual skip connection for every LSTM layer or blocks of LSTM layers appears to be the standard approach, several alternatives have been suggested, one of which motivates a modification to the three-layer auxiliary approach at hand. Inspired by the gate design of the LSTM, Wu et al. (2016a) propose integrating a multiplicative gate for the residual connection directly into the LSTM hidden unit. Their best scoring approach is to connect the output of layer $l-2$ to the cell output of layer $l$. This entails the following change to the standard LSTM equation given in definition 4.18.

$$\mathbf{h}_{l,t} = \mathbf{o}_{l,t} \odot \tanh(\mathbf{c}_{l,t-1}) + \mathbf{r}_{l,t} \odot \mathbf{h}_{l-2,t} \tag{6.10}$$

$$\mathbf{h}_{l-2,t}, \mathbf{o}_{l,t}, \mathbf{c}_{l,t-1}, \mathbf{r}_{l,t}, \mathbf{h}_{l-1,t} \in \mathbb{R}^{d_{out}}$$

where $t$ represents the time step in the sequence and $\mathbf{r}_{l,t}$ is a gate computed alongside the other gates in the LSTM:

$$\mathbf{r}_{l,t} = \text{sigm}(\mathbf{h}_{l,t-1}\mathbf{W}_{\mathbf{r},l} + \mathbf{h}_{l-1,t}\mathbf{U}_{\mathbf{r},l}) \tag{6.11}$$

$$\mathbf{r}_{l,t} \in \mathbb{R}^{d_{out}},\ \mathbf{W}_{\mathbf{r},l} \in \mathbb{R}^{d_{out} \times d_{out}},\ \mathbf{U}_{\mathbf{r},l} \in \mathbb{R}^{d_{in} \times dim_{out}}$$

When $\mathbf{r}_{l,t}$ equals 1 at some position, the output of $\mathbf{h}_{l-,t}$ at that position gets passed through unobstructed. When it equals 0, the network behaves like a traditional stacked feed-forward network (Wu et al., 2016a). This strategy has the benefit of allowing the network to choose the scales with which the individual entries in the tensor should be directed forward to improve the result depending on the sequential context encoded in the internal states. It should lend itself well to a multitask approach that tries to build a general feature representation benefitting several objectives. An LSTM layer can fully specialise on retrieving context-dependent information for a certain task while being able to directly retrieve information already encoded by a lower level task depending on input and context. Wu et al. (2016a) report that their proposal improves results for CCG supertagging over a model simply adding the penultimate layer output to $\mathbf{h}_{l,t}$.

I would like to propose a complementary perspective on skip connections in the context of hierarchical multi-task models. Direct additive connections from early layers to the final layer allow for the unobstructed backpropagation of errors from the main task's feed-forward network to lower levels from which auxiliary tasks are supervised. In a fully connected architecture, with shortcut connections in each layer, I assume that for layer $l$ this significantly works towards the construction of mutually beneficial representations with respect to tasks supervised on levels $\geq l$. This, however, presupposes closely related tasks which is a key condition for effective multi-task learning as explained above. When auxiliary tasks are expected in some systematic way to be synergistic in a certain number of classes but noisy and unhelpful for the main task in others, using gates to enable the model to dampen or to increase the intensity of values in skip connections dynamically and individually depending on context may allow the model to benefit from less closely related auxiliary tasks, as in the case of CCG supertagging for discontinuous constituent parsing. This might be a balanced compromise between an ungated residual LSTM stack based on element-wise summation and a computationally expensive but more informative residual connection realised through concatenation with free access to all preceding outputs.

Unfortunately, using a custom LSTM cell architecture in pytorch would mean a significant increase in computational cost. The baseline model of Coavoux and Cohen (2019) uses the fast native LSTM implementation written in the C++ programming language. To maintain this efficiency and to better compare the effect of a gated residual connection to the baseline implementation of Coavoux and Cohen (2019) that skips only a single LSTM layer at a time instead of two, I explore simply using the current and the last layer output as inputs to the gate. The residual connection is established from the output of the preceding layer and does not alter the hidden unit of the LSTM but is connected thereafter. The exploration of Wu et al. (2016a)'s approach is left for future work. I expect the LSTM hidden state of the current layer to be able to encode the contextual information provided by the preceding time step hidden state $\mathbf{h}_{l,t-1}$ used in Wu et al. (2016a). My approach can also be seen as a variation of the concept of *highway connections* (Srivastava et al., 2015).

**Definition 6.1 (Gated Residual).**
A gated residual module can be defined as a function that takes two vectors $\mathbf{x}, \mathbf{y} \in \mathbb{R}^{dim}$ and has learnable parameters $\mathbf{b} \in \mathbb{R}^{dim}$, $\mathbf{W} \in \mathbb{R}^{2dim \times dim}$ for some dimensionality $dim$:

$$Res(\mathbf{x}, \mathbf{y}) := \mathbf{x} + \mathbf{r} \odot \mathbf{y}$$
$$\mathbf{r} = \text{sigm}((\mathbf{x} \circ \mathbf{y})\mathbf{W} + \mathbf{b})$$

A visualisation of the gated residual module is given in figure 6.3.

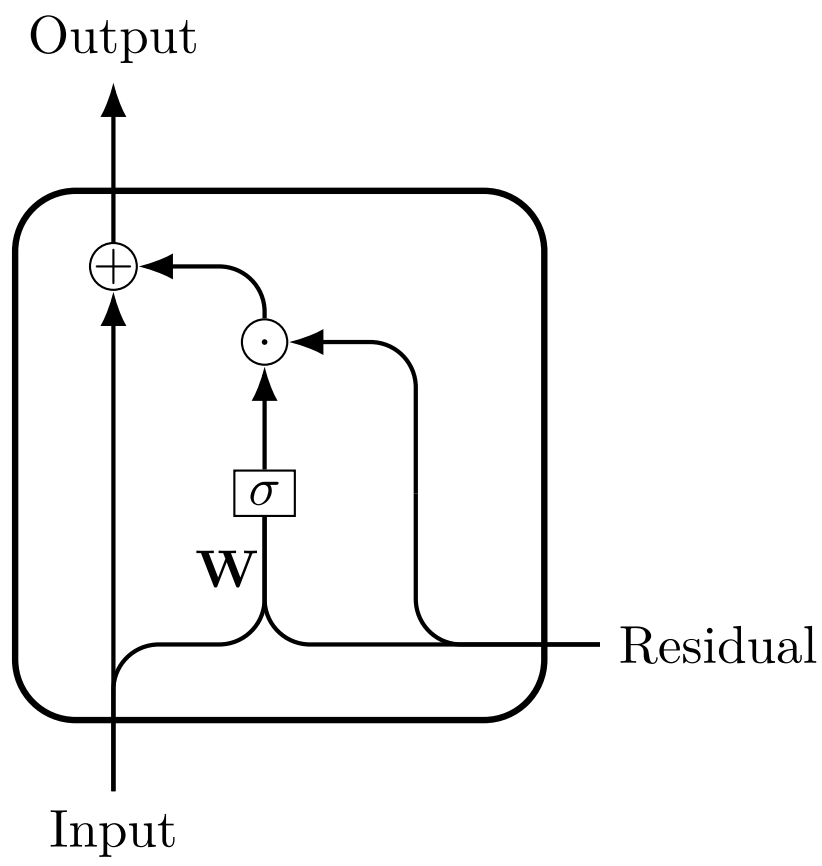

Figure 6.3: Design of the gated residual connection.

For each LSTM layer, the output and the previous layer output are combined via a layer-specific gate. The result of this computation is then forwarded to the specific task feed-forward network(s) supervised at this level and to the next LSTM recurrent layer as well as the following gated residual module.

$$\mathbf{m}_{l,t} = Res_l(biLSTM_l^*(\mathbf{m}_{l-1,0:n})^{[t]}, \mathbf{m}_{l-1,t}) \tag{6.12}$$

Furthermore, I suggest adding a shortcut connection from the LSTM stack input, i.e. from the token representations, by performing an initial linear transformation that converts this tensor into the LSTM hidden layer dimensionality $d_{hid}$. This way, the first LSTM layer can also carry a residual module. The full modified architecture is visualised in figure 6.4.

### 6.2.3 Increasing Model Width

To allow for a fair comparison with the original model of Coavoux and Cohen (2019), the simple CCG approach maintains the LSTM hidden state dimension of 400. However, against the backdrop of the manual analysis in section 5.3 which suggested that CCG lexical category assignments might statistically correlate with DPTB descriptions for some phenomena while being unhelpful for resolving others, one can expect a multi-task model optimised for both tasks to require a higher dimensionality to establish a representation that can encode the additional information in uncorrelated cases.

At the same time, Goldberg (2022, chapter 20.2.6) points out that if an auxiliary model with $k$ tasks only started to show improvements if the dimensionality was simply enlarged $k$-fold, the network likely did not established shared knowledge for several tasks and the improvement was only built on increased neural capacity. Therefore, in order to identify if the model demands increased capacity to support the additional task in a shared representation, I also perform training with the LSTM hidden dimension $d_{hid}$ enlarged by factor 1.5 to 600.

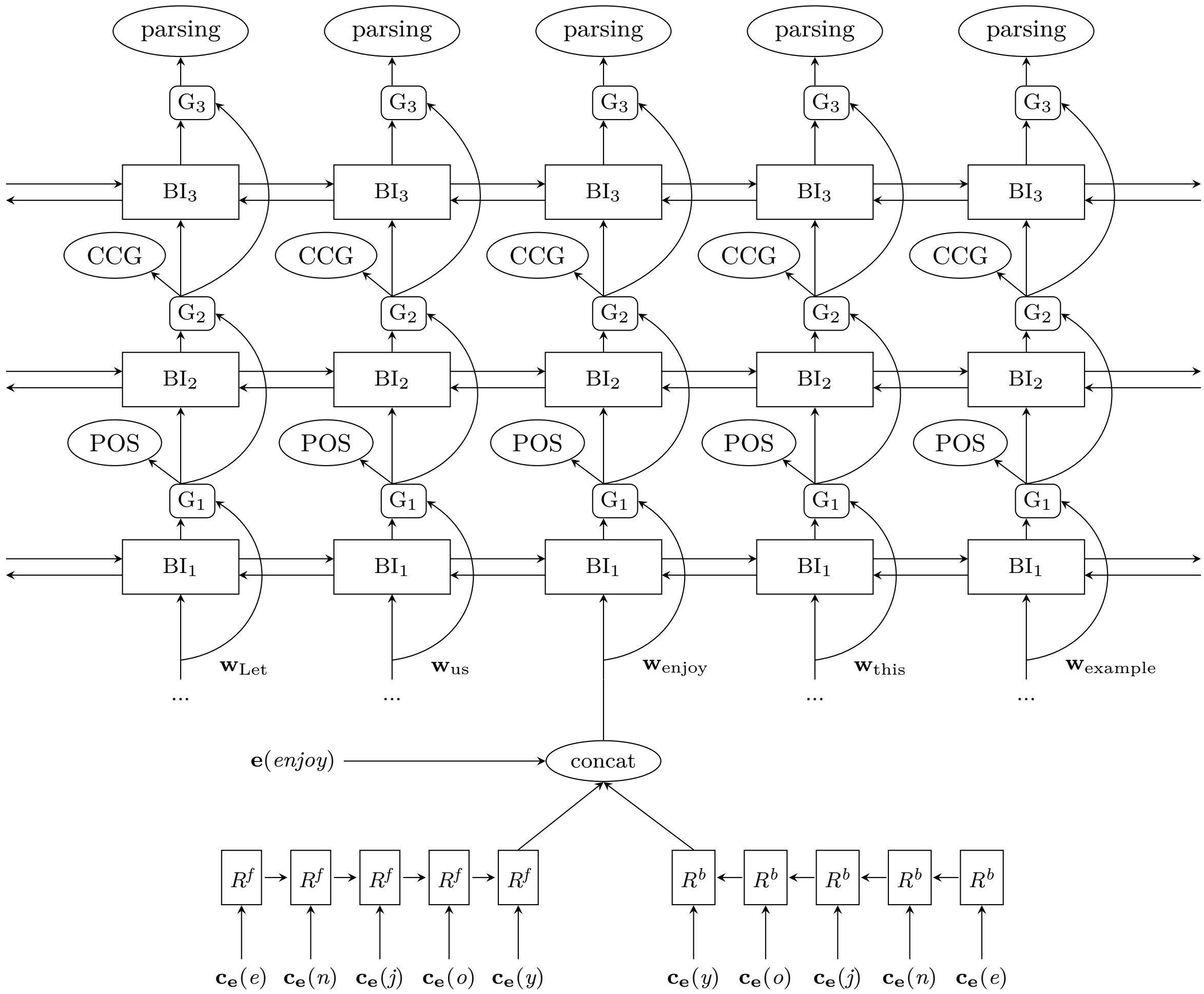

Figure 6.4: Stack-free neural parser with a three-layer biLSTM network, supertagging as an auxiliary task and gated residual connections to cell outputs. For clarity the token representations and the supertag prediction vectors are subscripted with words instead of indices. Figure inspired by Goldberg (2022, chapter 20.2).

### 6.2.4 Feature Bootstrapping

Zhu and Sarkar (2019) perform multi-task learning for lexicalised tree adjoining grammar (LTAG) supertagging by deconstructing complex TAG categories into classes of substructures like *root*, which naturally refers to the label of the root node of the lexical tree, or *sketch*, which captures only the structure of the lexical category without labels. Six objectives, including supertag prediction, are trained using separate biLSTM models. After training, the supertagger is used to generate a certain number of best supertags. The auxiliary models then compute the probability of these assignments' substructures from which the best supertag among the predictions is determined. Zhu and Sarkar (2019) report consistent improvements in a range of settings compared to a single-task network for supertagging. By reusing the training set for the supertagging task the authors effectively generate more data instances without enlarging the dataset.

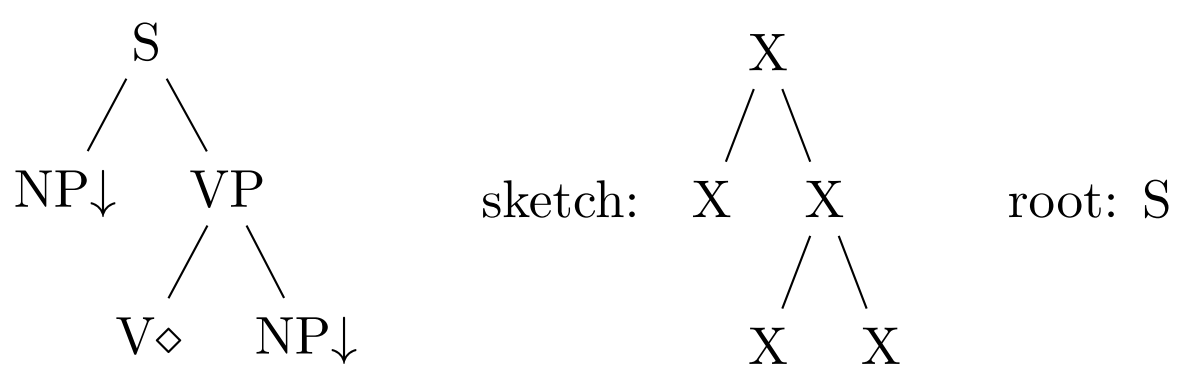

Figure 6.5: Example for root and sketch subtasks for an LTAG lexical tree as proposed by Zhu and Sarkar (2019).

Figure 6.5 shows an example for the decomposition of lexical trees in LTAG as proposed by Zhu and Sarkar (2019). In the following, I borrow from this idea and present the extraction of subtasks for CCG supertags.

**Argument** For each supertag the rightmost/outermost argument as well as its directionality symbolised by "-" for left and "+" for right are concatenated. For categories without arguments, "+" is used as a dedicated label. (6.1) gives an example of this schema. 148 distinct arguments are found in the training split.

(6.1) a.

| | | | |
|---|---|---|---|
| I | like | this | train |
| NP | (S[dcl]\NP)/NP | NP/N | N |
| + | +NP | +N | + |

b.

| | | |
|---|---|---|
| You | sing | beautifully |
| NP | S[dcl]\NP | (S\NP)\(S\NP) |
| + | -NP | -(S\NP) |

**Head** The *head* subtask predicts the innermost return type of a CCG category. It is also known as the *range* of a category (Steedman, 2000, chapter 3). There are 26 distinct head tags.

(6.2) a.

| | | | |
|---|---|---|---|
| NP | (S[dcl]\NP)/NP | NP/N | N |
| NP | S[dcl] | NP | N |

b.

| | | |
|---|---|---|
| NP | S[dcl]\NP | (S\NP)\(S\NP) |
| NP | S[dcl] | S |

**Sketch** The *sketch* ignores atomic categories in function category descriptions and only represents the overall argument structure of the supertag. It is inspired by the sketch task of Zhu and Sarkar (2019) who found that it helps with disambiguation for TAG. To extract the sketch feature from a CCG supertag, all atomic categories are replaced by an X. The number of distinct sketch labels is 212.

(6.3) a.

| | | | |
|---|---|---|---|
| NP | (S[dcl]\NP)/NP | NP/N | N |
| X | (X\X)/X | X/X | X |

b.

| | | |
|---|---|---|
| NP | S[dcl]\NP | (S\NP)\(S\NP) |
| X | X\X | (X\X)\(X\X) |

Given the hierarchical framework of the approach at hand, subtasks will be supervised at a dedicated additional layer between POS-tagging and supertagging. The three subtasks are supervised from the same layer since there is no obvious hierarchy regarding their difficulty and inter-relationship. An illustration of the full modified architecture is given in figure 6.6.

Each of the three subtasks has a dedicated feed-forward network that outputs a distribution over the distinct labels found in the training dataset. It is optimised in parallel according to algorithm 6.1. The inclusion of these three tasks leads to twice the amount of training instances per epoch compared with the simple auxiliary model (237,940 sentences in total). In case the approach shows to be successful, the extraction of other subtasks like functor or full argument structure could be performed.

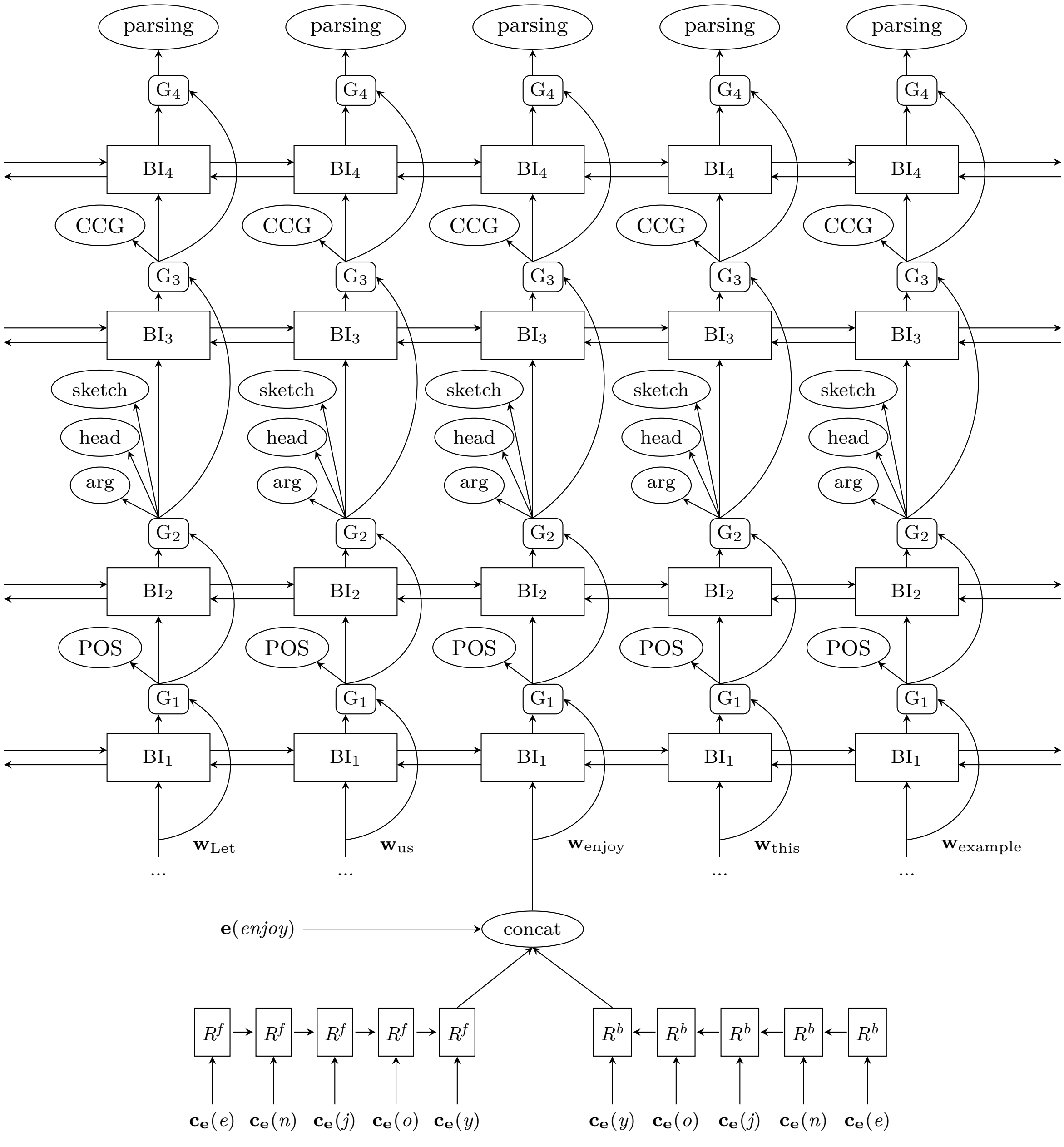

Figure 6.6: Stack-free neural parser with a multi-task four-layer biLSTM network, including gated residual connections. Supertag-component tagging (sketch, head, arg) and supertagging (CCG) are added as auxiliary tasks, with supertagging being supervised one layer higher than the subcomponents. For clarity the token representations and the supertag prediction vectors are subscripted with words instead of indices. Figure inspired by Goldberg (2022, chapter 20.2)

### 6.2.5 Head-Dependency Structure

The CCGbank and its derivative, the CCGrebank, provide head-dependency structures that encode the saturation of category arguments. The dependencies hold between a lexical functor and the heads of the constituents that fill the functor's argument slots (Hockenmaier and Steedman, 2007).

Figure 6.7 shows an example CCG derivation from the CCGrebank as well as its head-dependency structure depicted using directed labelled edges. The edge labels refer to the respective argument slot that is filled in the category the edge originates in. For transparency, I added subscripts to the supertag assignments that make explicit the argument number. As can be seen in the case of *does*, a word can have two incoming edges. This stems from the fact that adjuncts are not treated as arguments of their governor, but instead take it as their argument, as prescribed by combinatory categorial grammar.

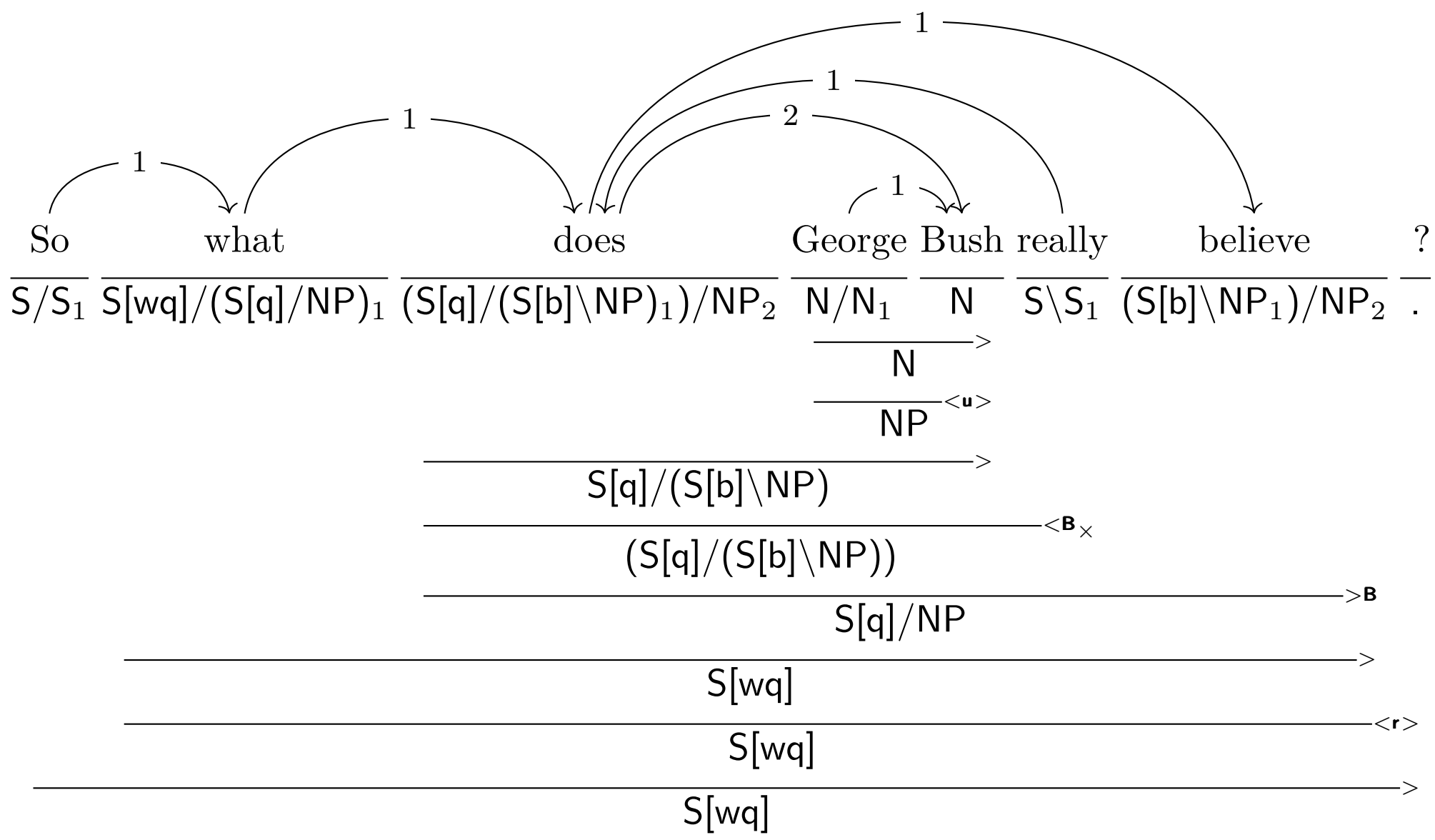

Figure 6.7: Derivation from the CCGrebank alongside its head-dependency structure visualised as labelled directed edges above the tokens.

The head-argument information used here closely correlates with the CCGrebank gold derivation and is less akin to conventional dependency descriptions. This can be seen in figure 6.7 where *what* is not encoded as an argument of *believe* but as taking a constituent with *does* as the lexical head as an argument. Therefore, while this information might be useful to solve projective attachment ambiguities (cf. figure 5.6) its effect on difficult discontinuous cases that arise from extraposed dependents where the CCGrebank unhelpfully analyses constituents as adjuncts of higher-level clauses (cf. figure 5.23) remains to be seen.

To test the usefulness of head-dependency structure prediction as an additional auxiliary task, I encode this information into two distinct sequence labelling tasks encompassing left dependencies and right dependencies respectively. For each position $i$, the relative position of the words right of $i$ connected with an outgoing edge as well as the corresponding argument number is predicted (*right action*). The same is done for left dependents (*left action*). (6.4) shows the tags resulting from the annotation in figure 6.7 where each outgoing edge takes the form <argument_number>:<relative_position>. 0 represents the absence of outgoing edges. In cases of multiple arguments they are separated by an underscore symbol. To keep the number of distinct categories manageable, relative distances above absolute value 5 are all set to 5 or -5 respectively.

(6.4)

| | So | what | does | George | Bush | really | believe | ? |
|---|---|---|---|---|---|---|---|---|
| right: | 1:1 | 1:1 | 2:2_1:4 | 1:1 | 0 | 0 | 0 | 0 |
| left: | 0 | 0 | 0 | 0 | 0 | 1:-3 | 0 | 0 |

In the train split, 351 distinct left action labels and 955 right action labels are predicted. This corresponds with the right-branching tendency of the English language (Kiparsky, 1996). The two sequence tagging tasks are treated as higher-level tasks compared to CCG supertagging since they encode relationships between different supertag assignments in the sentence, and therefore, in a four-biLSTM-layer network the tasks are supervised in the following order: POS-tags, CCG supertags, action labels, parsing (from bottom to top).

### 6.2.6 Other Auxiliary Tasks

In addition to auxiliary tasks based on the CCGrebank, I perform a number of experiments to assess the suitability of other syntactically motivated sequence labelling tasks for discontinuous constituent parsing. Due to space limitations, the representations and formalisms which the auxiliary tasks are derived from will not be introduced in detail and an interested reader will be referred to more comprehensive introductions and formal accounts.

**LCFRS** Ruprecht and Mörbitz (2021); Ivliev (2020) present an algorithm for the extraction of supertags in the LCFRS formalism from discontinuous constituent treebanks. For LCFRSs have shown to be able to adequately capture the syntactic descriptions in the DPTB (Evang, 2011), LCFRS supertagging immediately suggests itself as an auxiliary task for neural parsing. Using the `prepare_data` script of the `LCFRS Supertag Parser` released by Ruprecht and Mörbitz (2021)[46], I converted the DPTB into a format annotated with LCFRS supertags. The model is trained to predict these from an intermediate layer between POS tagging and parsing in the same fashion as suggested for CCG supertags in section 6.2.1.

Unfortunately, the extraction process results in a large number of distinct supertags. In total 4504 lexical categories are assigned in sections 2–21. Since this number is three times larger than the 1574 distinct CCG categories found in the CCGrebank train split, LCFRS-supertag prediction is likely to be more difficult to learn for the neural model. This is an important thing to keep in mind when comparing the results of different auxiliary tasks to draw conclusions about their informativeness for discontinuous constituent parsing.

**Chunking** Chunking separates a sentence into simple constituent types like NP or VP and is sometimes called *shallow parsing* (Collobert and Weston, 2008) or *partial parsing* (Jurafsky and Martin, 2009, chapter 13.5). Each word is only assigned to a single constituent. Using the so-called IBO notation (Jurafsky and Martin, 2009, chapter 13.5), chunking can be treated as sequence labelling by introducing special symbols where the beginning word of a constituent is marked with B and internal parts of a chunk are marked with I. Furthermore, if elements are outside of any chunk they are simply marked with O. Constituent type assignment and I, B and O features are treated as a single task, which doubles the total number of tags. (6.5) shows an example sentence chunked with brackets as well as the equivalent IBO-encoding taken from Jurafsky and Martin (2009, chapter 13.5).

(6.5)

| | | | | | | | | | | | | |
|---|---|---|---|---|---|---|---|---|---|---|---|---|
| $[_{NP}$ | The | morning | flight | ] | $[_{PP}$ | from | ] | $[_{NP}$ | Denver | ] | $[_{VP}$ | has | arrived | ] |
| | B-NP | I-NP | I-NP | | | B-PP | | | B-NP | | | B-VP | I-VP | |

46 `https://github.com/truprecht/lcfrs-supertagger`

In order to integrate chunking as an auxiliary task, I use the dataset made available as part of the CoNLL-2000 shared chunking task (Tjong Kim Sang and Buchholz, 2000). It is based on sections 15–18 (train split, 8936 sentences) and section 20 (test split, 2012 sentences) of the Penn Treebank. Therefore, the inclusion of this task does not enlarge the domain beyond the sentences used to train the main parsing task (PTB sections 2–21). Since no development split is defined, I use the last 399 sentences of the train split for development and remove them from the training set.

The model is trained to predict the chunking labels from an intermediate biLSTM layer between POS tagging and parsing in the same fashion as suggested for CCG supertagging in section 6.2.1. There are 22 distinct labels in the test split. Since task sampling is proportional to the dataset size, chunking only accounts for 9.7% of total training instances.[47] For all other models explored in this work, the individual task datasets are roughly equal in size.

**Dependency Parsing** Dependency grammars are a class of grammatical theories centred around the notion of dependency as opposed to constituency. While constituent phrase structure grammars arrange words into hierarchical constituents, dependency grammars exhibit word-to-word syntactic relations expressed through labelled directed edges (Jurafsky and Martin, 2009, chapter 12.7).

A dependency descriptions can be formalised as a tree with labelled edges over the tokens in a sentence. Each token has one incoming edge and zero or more outgoing edges. A single token is connected via an incoming edge with a special *root* symbol. For each token, the source of its incoming edge is called its *head*. This notion closely correlates with the concept of a constituent head discussed in section 3.2.6 (Jurafsky and Martin, 2009, chapter 12.7.1).

Several suggestions for the joint learning of constituent and dependency parsing have been made. Strzyz et al. (2019a) perform sequence labelling-based projective constituent and dependency parsing in a multi-task framework that yields improvements over comparable single-task models for both tasks. Zhou et al. (2020a) explore dependency parsing and span-based constituent parsing together with semantic parsing. To the best of my knowledge, no exploration of synergistic parsing of dependency relations and discontinuous constituents has been performed to this date.

Strzyz et al. (2019b) compare several approaches for the conversion of dependency parsing into a sequence labelling task. Among these approaches, so-called *relative POS-based encoding*, which was first used by Spoustová and Spousta (2010), leads to the best parsing score for single task dependency parsing. A dependency tree for a sentence $w_1, ..., w_n$ is linearised by assigning each token $w_i$ the label of its incoming edge $l_i$, the POS tag $t_j$ of its head $w_j$ and an integer $o_i$ such that if $i < j$, $w_j$ is the $o_i$th closest of the words to the right of $w_i$ that bear POS tag $t_j$ and if $j < i$, $w_j$ is the $-o_i$th closest of the words to the left of $w_i$ that have POS tag $t_j$. This results in a triple $(t_j, o_i, l_i)$ that is treated as a distinct tag. As Strzyz et al. (2019b) note, such a scheme is closer to the notion of valency, i.e. argument structure, when compared to strictly encoding relative position.

Figure 6.8 shows an example dependency description together with the corresponding relative POS-based encoding. For instance, the label VB_1_aux of the token *does* encodes an edge with label aux that connects the first token to the right of *does* that has POS tag VB with *does*. In case of a root relation, the distinct label root_-1_root is used as a standard.

I convert the dependency PTB (de Marneffe et al., 2006) into a dataset tagged with relative POS-based labels. The PTB POS tags are taken as the base. Subtypes for syntactic relations like :poss in nmod:poss are removed to reduce the number of distinct labels. Furthermore, as suggested by Spoustová and Spousta (2010), the maximum relative position is capped at -3 and 3. The conversion results in 1634 distinct labels for the train split of sections 2–22 of the PTB.

[47] Parsing and POS tagging amount to 39,832 instances each.

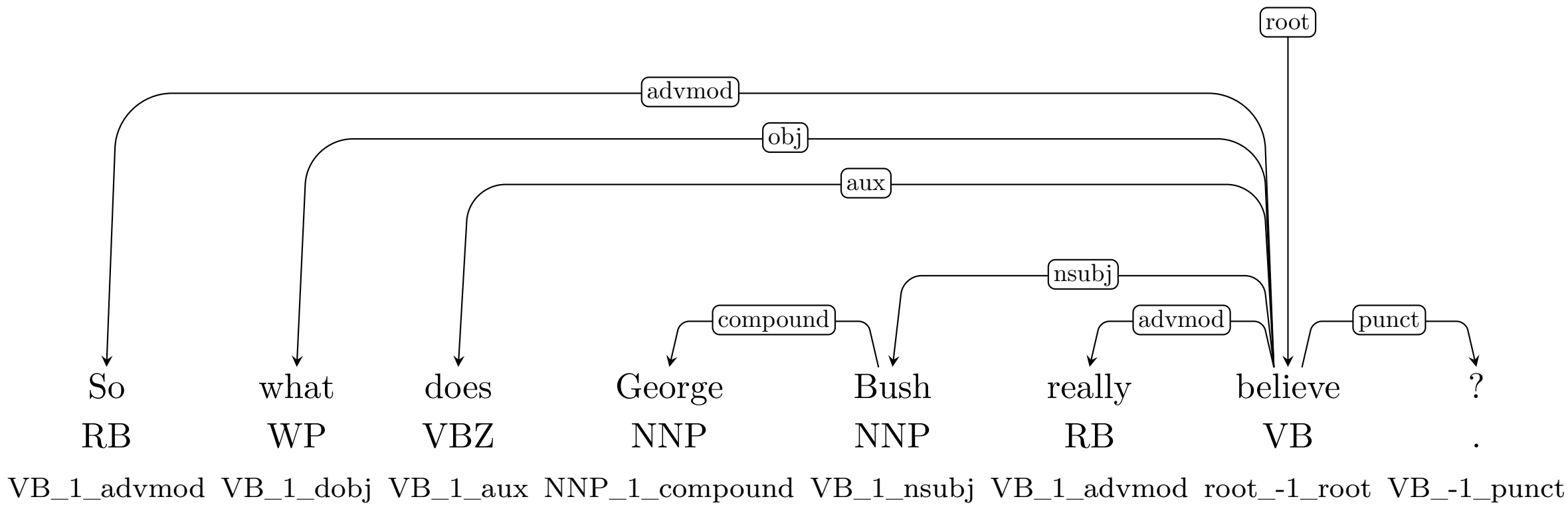

Figure 6.8: Dependency description using directed arcs from the dependency PTB as well as the corresponding relative POS-based encoding underneath.

In the experiment, dependency parsing is treated as a parallel task to constituent parsing and also supervised from the final layer. In order to be able to compare the model with the other auxiliary task experiments, the network maintains a 3-layer biLSTM stack. This amounts to a similar architecture as in Strzyz et al. (2019a) with the notable difference that they deconstruct the parsing tasks into 3 constituent labelling tasks and 2 dependency labelling tasks whereas I employ only one task per paradigm. Furthermore, the transition-based approach used here allows for discontinuous constituents while the sequence labelling encoding of constituent trees used by Strzyz et al. (2019a) does not.

**LTAG** Tree adjoining grammar (TAG) is a mildly context-sensitive formalism that was first introduced by Joshi et al. (1975). A lexicalised variant of TAG was proposed by Schabes and Joshi (1991) which led to the formalisation of lexicalised tree adjoining grammar (LTAG). An illustrative example of an LTAG elementary tree has been given in figure 6.5. Much of the work on supertags like the extraction of supertagged corpora from treebanks and the development of parsers that use supertaggers was initially conducted on LTAG. Bangalore and Joshi (2010b) give a comprehensive overview of important research in this regard.

The TAG extracted PTB of Chen et al. (2006) is commonly used as a basis for training LTAG supertaggers (Zhu and Sarkar, 2019; Kasai et al., 2018). Unfortunately, I was not able to acquire this corpus. Instead, I used the publicly available[48] LTAG-spinal treebank developed by Shen et al. (2007). LTAG-spinal omits subcategorisation information and the argument-adjunct distinction. This reduces the overall number of distinct supertags and the average number of elementary trees associated with a lexical item.

I train a discontinuous constituent parser with LTAG-spinal supertag prediction as an intermediate auxiliary task using the same 3-layer architecture as in the CCG model. The LTAG-spinal treebank is based on the PTB and split into the same training, development and testing datasets. 1202 distinct tag symbols occur in the training set.

### 6.2.7 Hyperparameters

The hyperparameters for the multi-task models introduced in the preceding section are given in tables 6.2 and 6.3. All remaining parameters are equal to those in the baseline as listed in table 4.1. The task specific feed-forward networks have the same architecture and are assigned the same input dropout as the POS tag feed-forward network.

[48] Available at: `https://www.cis.upenn.edu/~xtag/spinal`

| Architecture hyperparameters | | CCG | $\text{CCG}^{\text{gate}}$ | $\text{CCG}^{\text{gate}}_{600}$ | $\text{LCFRS}^{\text{gate}}$ | $\text{Chunk}^{\text{gate}}$ | $\text{Dep}^{\text{gate}}$ | $\text{LTAG}^{\text{gate}}$ |
|---|---|---|---|---|---|---|---|---|
| Dimension of sentence biLSTM | $d_{hid}$ | 400 | 400 | 600 | 400 | 400 | 400 | 400 |
| Number of biLSTM layers | | 3 | 3 | 3 | 3 | 3 | 3 | 3 |
| Number of auxiliary tasks labels | | 1574 | 1574 | 1574 | 4504 | 22 | 1634 | 1202 |
| Type of residual connection | | add | gate | gate | gate | gate | gate | gate |

Table 6.2: Hyperparameters of the supertag models with one auxiliary task. The other architecture and optimisation parameters match those of the baseline listed in table 4.1.

| Architecture hyperparameters | | $\text{CCG}^{\text{gate}}_{\text{multi}}$ | $\text{CCG}^{\text{gate}}_{\text{dep}}$ |
|---|---|---|---|
| Dimension of sentence biLSTM | $d_{hid}$ | 400 | 400 |
| Number of biLSTM layers | | 4 | 4 |
| Number of auxiliary tasks | | 4 | 3 |
| Number of auxiliary task labels | | 148, 26, 212; 1574 | 1574; 351, 955 |
| Type of residual connection | | gate | gate |

Table 6.3: Hyperparameters of the supertag models with more than one auxiliary task. The task label numbers are listed according to supervision depth from bottom to top. A semicolon indicates the next layer. The other architecture and optimisation parameters match those of the baseline listed in table 4.1.

Due to the promising results of the gated residual model $\text{CCG}^{\text{gate}}$ on the development dataset compared to the ungated residual model CCG (cf. section 6.3), the other experiments adhere to the gated residual approach.

Since all auxiliary models enlarge the baseline approach in some way, it is important to isolate the contribution of this change in architecture on the model performance. Therefore, for every combination of biLSTM stack depth, biLSTM hidden dimensionality and type of residual connection used in the experiments, I also train a control model that does not include auxiliary tasks at the intermediate biLSTMs. These are called $\text{Ctr}_3$, $\text{CCG}^{\text{gate}}_2$, $\text{Ctr}^{\text{gate}}_3$, $\text{Ctr}^{\text{gate}}_{3,600}$ and $\text{Ctr}^{\text{gate}}_4$ where the subscript indicates the number of biLSTM stacks. The auxiliary models will be compared with the respective control model to reliably assess the effect of training with auxiliary tasks.

## 6.3. Experiments

I carried out experiments that investigate the effect of integrating supertags into discontinuous constituent parsing by training the systems proposed in the preceding sections. In section 6.3.1, I present the experimental protocol, then I discuss the results of the experiments in section 6.3.2. Section 6.3.3 gives a per-phenomenon analysis of the pipeline model and the best scoring auxiliary-task model on discontinuous structures and section 6.3.4 evaluates these models against several error types. Section 6.3.5 investigates noticeable effects of supertag integration in more detail via a sample comparison of gold and predicted trees. Finally, I give a comparison with previously published discontinuous constituent parsing results in section 6.3.6.

All approaches are implemented as modifications of the discontinuous stack-free transition-based parser created by Coavoux and Cohen (2019) which I have presented in section 4. The code including all trained models and results will be made available at `https://github.com/filemon11/discoparset-supertag`.

The modified codebase includes a dynamic multi-task framework. Via a hyperparameter one can choose the number of layers for the biLSTM stack and what tasks should be shared up to which layer. Combining several tasks based on different corpora is possible without manual code changes. This allows for a straightforward exploration of a multitude of architectures. Furthermore, I included the optional evaluation of auxiliary task accuracy on the respective development and test corpora in the main evaluation function. This way, a multi-task-model's capabilities on an auxiliary task can be easily compared with previously reported scores of other sequence labelling systems.

### 6.3.1 Experimental Protocol

In order to use the dynamic oracle model of Coavoux and Cohen (2019) as a baseline and to allow a reliable comparison with their results the protocol remains unchanged. The models here also use the dynamic oracle and are trained with the ASGD algorithm (Polyak and Juditsky, 1992) for 100 epochs. Every 4 epochs the model is evaluated on the development set and selected if it surpasses the previous best scoring instance in parsing F-score.

Training took approximately two days for the $\text{PIPE}_{\text{rebank}}$ model and three days for the simple auxiliary CCG model using an NVIDIA GeForce RTX 2060 SUPER 8GB GPU. For the model with the highest number of parameters, $\text{CCG}^{\text{gate}}_{\text{multi}}$, training took roughly a week. Evaluation is performed using the standard evaluator `discodop` (van Cranenburgh et al., 2016) and follows the standard procedure of ignoring punctuation and root symbols. `discodop` retrieves the number of matching brackets in the parsing prediction compared to the gold annotation using precision, recall and F-score which has been agreed upon as the common metric for constituency parser evaluation (Black et al., 1991).

### 6.3.2 General Effects of Supertag Integration

Table 6.4 shows the results of the experiments on the development set of the DPTB corpus (section 22) as well as the baseline score as reported by Coavoux and Cohen (2019) and a rerun with the baseline model architecture and parameters using the modified codebase. The evaluation includes precision, recall and F-score as well as precision, recall and F-score on discontinuous constituents only. Furthermore, accuracy is reported for all auxiliary tasks. All models achieve convergence within the 100 training epochs.

**Baseline** The differences between Coavoux and Cohen (2019) and the rerun with the modified codebase can be attributed to the proportional task sampling used here that does not strictly alternate between tasks but chooses them at random (cf. section 6.2) and to other non-deterministic factors when training the network (e.g. dropout). Since the experiments here are all performed using the modified code, the rerun is taken as a baseline in the comparisons that follow.

**Pipeline** The $\text{PIPE}_{\text{rebank}}$ model is unable to make improvements over the baseline parser. Slight decreases in general precision (-0.35 in absolute score) and discontinuous precision (-5.01) result in degradations of F-score. Providing supertag distributions generated by the pre-trained rebank `depccg` model to the parser seems to decrease the model's syntactic capabilities questioning the helpfulness of CCG supertags for resolving discontinuous constituency. Interestingly, recall slightly increases in the general (+0.08) and in the discontinuous case (+1.79).

A possible cause for the poor results of the $\text{PIPE}_{\text{rebank}}$ model might be the quality of supertag distributions generated by the upstream supertagger. False lexical category assignments used as input to the parser could lead the transition system astray, especially in cases of only sparsely attested phenomena where the amount of training data cannot teach the parser successfully to counteract wrong information at input level. I test this hypothesis by evaluating the `depccg` rebank model on the CCGrebank development split which yields an accuracy of 81.61 when taking the 1-best supertag — much lower than the CCG auxiliary model's supertagging score, which amounts to 92.64. For 89% of tokens the best supertag is amongst the 3 highest scoring tags.

`depccg` rebank was trained on the 511 most frequent supertags in the CCGrebank training dataset. 1% of the tokens in the development split are not covered by this set. Strikingly, for those tags that can be predicted, the gold tag has an average rank of 19.3 in the distributions which the model outputs. Thus, there are some strong outliers with very low score in the prediction. A more detailed analysis showed that in 67.7% of the cases where the gold supertag is not amongst the best scoring 20 labels, the supertag is "," while the 1-best prediction is "conj". Perhaps there was a mix-up of datasets when the `depccg` rebank model was trained.

| Model | P | R | F | DP | DR | DF | POS | SUP | HEA | ARG | SKE | LEF | RIG |
|---|---|---|---|---|---|---|---|---|---|---|---|---|---|
| original | 91.5 | 91.3 | 91.4 | 76.1 | 66.4 | 70.9 | 97.2 | – | – | – | – | – | – |
| baseline | 91.45 | 91.15 | 91.30 | 79.95 | 65.25 | 71.85 | 97.25 | – | – | – | – | – | – |
| PIPE$_{\text{rebank}}$ | 91.1 | 91.23 | 91.16 | 74.94 | 67.04 | 70.77 | 97.25 | – | – | – | – | – | – |
| PIPE$_{\text{bank}}$ | 91.52 | 91.55 | 91.54 | 75.65 | 65.47 | 70.19 | 97.30 | – | – | – | – | – | – |
| CCG | 91.32 | 91.31 | 91.32 | 78.92 | 62.11 | 69.51 | 97.24 | 92.64 | – | – | – | – | – |
| CTR$_3$ | 91.34 | 91.14 | 91.24 | 72.25 | 61.88 | 66.67 | 97.18 | – | – | – | – | – | – |
| CTR$_2^{\text{gate}}$ | 91.67 | 91.2 | 91.43 | 81.23 | 65.02 | 72.23 | 97.31 | – | – | – | – | – | – |
| CCG$^{\text{gate}}$ | 91.86 | 91.67 | 91.77 | 82.48 | 68.61 | **74.91** | 97.22 | **92.85** | – | – | – | – | – |
| CTR$_3^{\text{gate}}$ | 91.82 | 91.56 | 91.69 | 78.2 | **69.96** | 73.85 | 97.23 | – | – | – | – | – | – |
| CCG$_{600}^{\text{gate}}$ | 91.85 | 91.57 | 91.71 | 78.36 | 66.59 | 72.0 | 97.21 | 92.66 | – | – | – | – | – |
| CTR$_{3,600}^{\text{gate}}$ | **91.99** | **91.76** | **91.88** | 80.63 | 69.06 | 74.4 | 97.26 | – | – | – | – | – | – |
| CCG$_{\text{multi}}^{\text{gate}}$ | 91.61 | 91.4 | 91.51 | 79.63 | 68.39 | 73.58 | 97.18 | 92.49 | 95.31 | 95.51 | 93.05 | – | – |
| CCG$_{\text{dep}}^{\text{gate}}$ | 91.57 | 91.24 | 91.4 | **83.33** | 65.02 | 73.05 | 97.20 | 92.67 | – | – | – | 95.64 | 92.76 |
| CTR$_4^{\text{gate}}$ | 91.82 | 91.71 | 91.77 | 79.53 | 67.94 | 73.28 | 97.16 | – | – | – | – | – | – |

Table 6.4: Results of the experiments on the DPTB development set. P, R and F are precision, recall and F-score. DP, DR and DF represent discontinuous precision, recall and F-score. POS and SUP are POS-tagging and supertagging accuracy respectively. HEA, ARG and SKE refer to the supertag subcomponents head, argument and sketch while LEF and RIG are left and right action labels. The best score for each metric across all models is set in bold if the metric is reported for more than one model. original refers to the dynamic oracle results of Coavoux and Cohen (2019) while baseline refers to the rerun using a modified codebase and proportional task sampling.

In order to check whether this problem is specific to this supertagger variant or if the pipeline approach is simply unsuitable, I decided to train an additional model (PIPE$_{\text{bank}}$) with the standard `depccg` trained on the original CCGbank as the supertagger. This model performs better for standard F-score (+0.24 compared to the baseline) but unfortunately even lower for discontinuities (-1.66). Using this model, for 98.4% of tokens in the DPTB development split the gold supertag is among the 3-best predictions which is why I conclude that there is indeed some kind of inconsistency in the rebank supertagger. For this reason further comparisons of the experiments in regards to discontinuous constituent analysis in sections 6.3.3 and 6.3.4 will be based on the results of PIPE$_{\text{bank}}$. Unfortunately, when comparing PIPE$_{\text{bank}}$ and the CCG auxiliary approach differences in results might be caused by changes in the analysis of some phenomena that CCGrebank introduced (Honnibal et al., 2010) and not only by the style of implementation (pipeline or auxiliary task) which I tried to evaluate. Furthermore, the results of the manual analysis of correlations between the DPTB and CCGrebank in section 5.3 may not always be applicable to PIPE$_{\text{bank}}$.

**Simple Auxiliary** It can be seen that the simple 3-layer control model with summative residual connections CTR$_3$ produces worse results than the baseline (-0.06 F, -5.18 DF) despite the fact that enlarging the LSTM stack depth gives the model more computational capacity. The simple CCG model is able to improve these results with slight increases in F-score (+0.08) and discontinuous F-score (+2.84) compared to the control model. However, it only leads to a marginal improvement in F-score and even a decrease in discontinuous F-score compared to the baseline. Both results suggest that the architecture is suboptimal for harnessing the increase in computational capacity provided by the additional LSTM transducer.

**Gated Residual Connection** Using gated residual connections in the 3-layer control model $\text{CTR}_3^{\text{gate}}$ leads to solid improvements in all metrics compared to the baseline (+0.39 F, +2.0 DF), with the notable exception of discontinuous precision (-1.75). The increase in discontinuous F-score therefore stems from better recall which is why I assume that the gate-controlled residual connections make it easier for the model to recognise long-range dependencies by having more control over the composition of the relevant contextual features across the vertical pass through the network. Improvements are not limited to networks with 3 biLSTM layers. They also show when training the 2-layer baseline-model in a gated residual setting ($\text{CTR}_2^{\text{gate}}$; +0.13 F, +0.38 DF).

The inclusion of supertags as an auxiliary task at the intermediate LSTM ($\text{CCG}^{\text{gate}}$) results in solid improvements both for general F-score and for discontinuous F-score which confirms the compatibility of gated residual connections with a hierarchical multi-task approach. Figure 6.9 depicts the progression of F-score over the epochs for the simple CCG model and for $\text{CCG}^{\text{gate}}$. The results for both the development split and a random selection of 425 sentences from the train split are included. It shows that the gain in F-score for the gated model is not simply a result of increased regularisation since both training and development results improve.

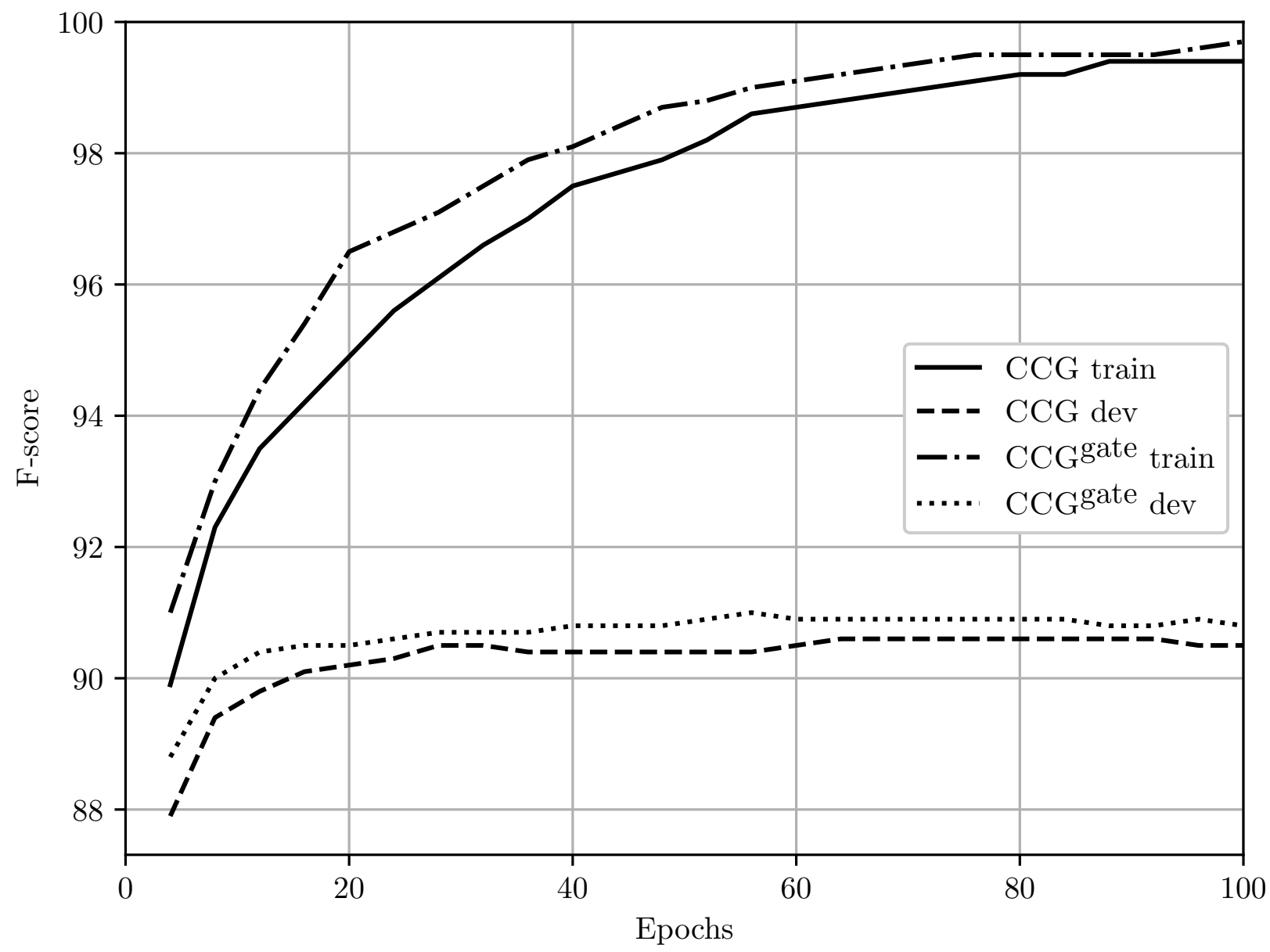

Figure 6.9: Progression of F-score for CCG and $\text{CCG}^{\text{gate}}$ on the development split and a random subset of the training split.

F-score for $\text{CCG}^{\text{gate}}$ increases by 0.08 absolute points and discontinuous F-score by 1.06 against the control model $\text{CTR}_3^{\text{gate}}$ as well by 0.47 for F-score and 3.06 for discontinuous F-score compared to the baseline. The former isolates the contribution resulting from CCG supertagging as an auxiliary task. $\text{CCG}^{\text{gate}}$ strongly corrects discontinuous precision from 78.2 for the control model to 82.48. This suggests that the network is able to acquire increased syntactic knowledge and competence for the analysis of long-range dependencies.

This finding confirms that CCG supertags can be helpful in discontinuous constituent parsing despite the seemingly incompatible nature of DPTB trees and CCG categories. The multi-layer neural auxiliary-task model is able to generate beneficial representations based on complex correlations which the pipeline model is apparently unable to acquire from the supertag distributions it is provided with at input level.

The increase in discontinuous precision comes with the cost of a slight decrease in discontinuous recall compared to the control model (-1.35). A large gap between precision (higher) and recall (lower) for discontinuities is often observed in discontinuous constituent parsers (Coavoux, 2021). The baseline model is no exception to this rule (79.95 precision, 65.25 recall). The fact that the pipeline approach does not add to this issue is coherent with the observations of Coavoux (2021) who enrich the input features of their parser with BERT-contextual embeddings (Devlin et al., 2019) which close the gap between recall and precision.

Learning a joint representation for both CCG supertagging and parsing apparently makes it harder for the parser to recognise long-range dependencies. One possible cause is that conditioning an intermediate vector representation both on supertagging and parsing as opposed to a pipeline approach impedes differentiating between informative supertag assignments and incompatible cases where a supertag corresponds to both projective and unprojective structures. This can be expected since the shared architecture up to the second biLSTM output is trained to provide useful features for CCG supertag prediction independent of supertag or context. As mentioned above, I suggest that this effect is mitigated through the adaptivity of the gated residual connection.

A second possibility may lie in the sparseness of training data for certain types of discontinuities (including their distinct CCG categories) coupled with the fact that auxiliary task training is performed on the same dataset. Therefore, the model might overfit for triggers of certain phenomena. This results in worse ability to recognise discontinuities not seen during training but when they are recognised, the syntactic capabilities induced by CCG supertag prediction allow the parser to competently reconstruct them. Enriching the input features using a pipeline approach naturally does not lead to this phenomenon since the number of training instances remains the same.

**Increasing Width** Contrary to the expectations, enlarging the LSTM hidden dimensionality from 400 to 600 for $\text{CCG}^{\text{gate}}_{600}$ does not increase F-score (-0.06) or discontinuous F-score (-2.91) compared to $\text{CCG}^{\text{gate}}$. It also does not result in increased supertagging accuracy (-0.19). This indicates that the model started to overfit.

When comparing with the very well performing single-task control model $\text{Ctr}^{\text{gate}}_{3,600}$ one could assume that the decrease in F-score (-0.17) and discontinuous F-score (-2.4) is caused by the integration of supertagging and be tempted to write off the auxiliary approach as ineffective. However, given the fact that CCG supertagging showed to increase F-score in the other settings (CCG and $\text{CCG}^{\text{gate}}$) I argue that this observation should be taken as another indicator for overfitting. This is supported by figure 6.10 which shows that the F-score on training instances significantly increases while the development score is slightly lower than $\text{Ctr}^{\text{gate}}_{3,600}$ from epoch 20 onwards. Apparently the model adapts too closely to the training data due to a combination of increased network width and the fact that the number of training instances on the same dataset is enlarged by factor 1.5.

As usual in cases of overfitting, one could either enlarge the training dataset, which is not an option here, or experiment with forms of regularisation like increased dropout (Goldberg, 2022, chapter 4.6) or the introduction of variational dropout for recurrent neural networks (Gal and Ghahramani, 2016). An exploration of beneficial dropout application is left for future work.

**Feature Bootstrapping** The $\text{CCG}^{\text{gate}}_{\text{multi}}$ multi-task model with four biLSTM-layers and subcomponent supertag features performs worse than the three-layer model $\text{CCG}^{\text{gate}}$ without bootstrapped sequence labelling tasks. General F-score is lower by 0.26 absolute points and discontinuous F-score by 1.33 points. When comparing with the 4-layer control model $\text{Ctr}^{\text{gate}}_{4}$ one can assess that while the approach leads to a degradation of default precision, recall and F-score, these metrics on only discontinuous constituents actually slightly increase. This is especially surprising for discontinuous recall (+0.45), which distinguishes this approach and the simple CCG model as the only two multi-task models that do not lead to a decline in discontinuous recall relative to their control model. Deconstructing supertags into distinct sequence labelling tasks may be a way to counteract the previously observed tendency of neural multi-task approaches to worsen in the model's ability to recognise discontinuities. Given the fact that $\text{CCG}^{\text{gate}}_{\text{multi}}$ still underperforms compared to the single-auxiliary task model $\text{CCG}^{\text{gate}}$, further research regarding viable subcomponent tasks and suitable architectures is necessary.

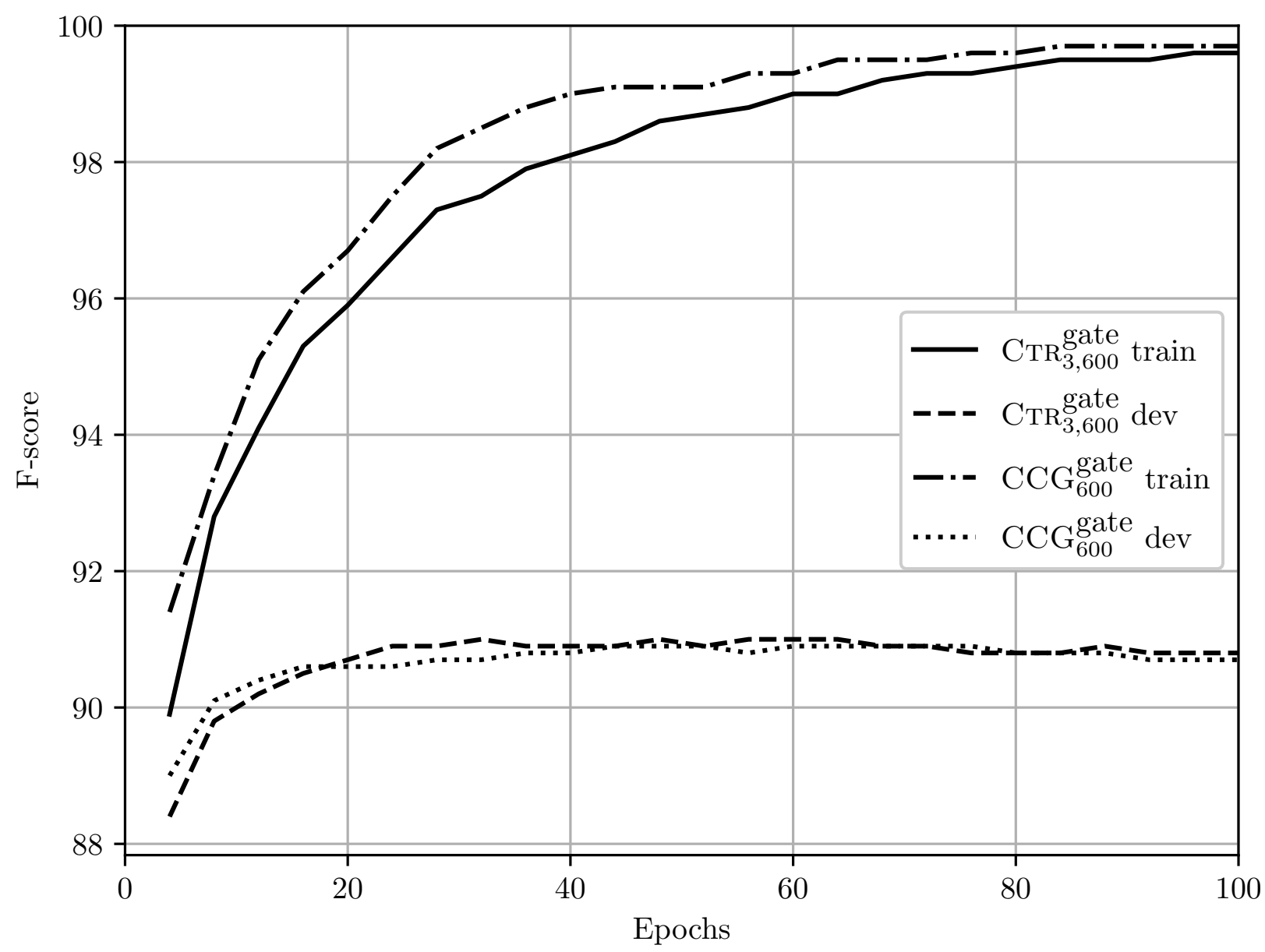

Figure 6.10: Progression of F-score for $\text{CCG}^{\text{gate}}_{600}$ and $\text{CTR}^{\text{gate}}_{3,600}$ on the development split and a random subset of the training split.

**Head-Dependency** As can be seen in table 6.4, using head-dependency information provided by the CCGrebank in a four-layer multi-task approach alongside CCG supertagging does not lead to improvements. The strongest drop in score occurs for discontinuous constituents, especially recall: -3.59 compared to $\text{CCG}^{\text{gate}}$. This observation cannot be explained by the choice of architecture since the control model $\text{CTR}^{\text{gate}}_{4}$ outperforms $\text{CCG}^{\text{gate}}_{\text{dep}}$ in most metrics.

Therefore, one can assess that head-dependency relations do not provide useful information for constituent parsing. My assumption is that the inclusion of this auxiliary task induces a representation for relationships between tokens that is close to the gold combinatory categorial grammar derivations found in the CCGrebank but incompatible with the DPTB constituent analysis. For instance, in figure 6.7 the *wh*-word *what* would receive the same right action label 1:1 as *George*, which indicates that its right argument is filled by a constituent that has its lexical head one position to the right in the sentence. In the case of *George*, this corresponds to the DPTB analysis of *George* and *Bush* as a nominal constituent. *what*, however, does not produce a constituent with its successor *does* but instead forms a discontinuity with *believe* in the DPTB. Clearly, the model should be able to distinguish these two cases in its representation.

On the basis of supertags only, which constitute discrete categories (here: S[wq]/(S[q]/NP) and N/N), the parser appears to learn distinct contextual token representations that lead to more competence in the prediction of constituent parsing transitions, as the results of $\text{CCG}^{\text{gate}}$ suggest. The neural model has no information as to the workings of the CCG formalism or the internal patterns of categories and only benefits from the fact that the category distinctions rooted in differences in argument and return structure used in the CCGrebank seem to correspond statisti-cally in some capacity with distinctions in constituent structure analysis. But when supervising head-dependency information at a higher level than CCG supertagging, these useful distinctions between supertag assigned tokens seem to be partially overridden harming the model's competence. Naturally, this particularly affects discontinuous constituents and the parser's ability to recognise them, as indicated by the decline in discontinuous recall.

**Other Auxiliary Tasks** Table 6.5 shows the results of the experiments with other auxiliary tasks as well as CCG$^{\text{gate}}$, the baseline and the 3-layer gated control model for comparison. The scores may serve as a first examination of the potential synergies between discontinuous constituent parsing and different kinds of syntactic sequence labelling tasks.

| Model | P | R | F | DP | DR | DF | POS | AUX | Auxiliary label count |
|---|---|---|---|---|---|---|---|---|---|
| original | 91.5 | 91.3 | 91.4 | 76.1 | 66.4 | 70.9 | 97.2 | – | – |
| baseline | 91.45 | 91.15 | 91.30 | 79.95 | 65.25 | 71.85 | 97.25 | – | – |
| CCG$^{\text{gate}}$ | 91.86 | **91.67** | 91.77 | **82.48** | 68.61 | **74.91** | 97.22 | 92.85 | 1574 |
| LCFRS$^{\text{gate}}$ | 91.5 | 91.38 | 91.44 | 78.42 | 66.82 | 72.15 | **97.30** | 87.0 | 4504 |
| CHUNK$^{\text{gate}}$ | **91.94** | 91.66 | **91.8** | 82.22 | 66.37 | 73.45 | 97.25 | 99.29 | 22 |
| DEP$^{\text{gate}}$ | 91.73 | 91.39 | 91.56 | 80.34 | 63.23 | 70.77 | 97.25 | 87.67 | 1634 |
| LTAG$^{\text{gate}}$ | 91.55 | 91.35 | 91.45 | 80.38 | 67.04 | 73.11 | 97.13 | 94.72 | 1202 |
| CTR$_3^{\text{gate}}$ | 91.82 | 91.56 | 91.69 | 78.2 | **69.96** | 73.85 | 97.23 | – | – |

Table 6.5: Results of additional experiments with a variation of auxiliary tasks on the DPTB development set. P, R and F are precision, recall and F-score. DP, DR and DF are discontinuous precision, recall and F-score. POS is POS-tagging accuracy while AUX refers to the respective auxiliary task accuracy. The best score for each metric across all models is set in bold, except for AUX which is incomparable.

Despite the success of Ruprecht and Mörbitz (2021) with grammar-based discontinuous constituent parsing using LCFRS-based supertags, the LCFRS$^{\text{gate}}$ model leads to a decline in both F-score (-0.25) and discontinuous F-score (-1.7) compared to the control model. Similar to the CCG models, discontinuous precision slightly increases while recall decreases which suggests that this phenomenon is not rooted in the kind of supertags used but has more to do with overfitting and data sparseness for distinct rare labels assigned to discontinuous structures.

The generally poor results of LCFRS supertagging as an auxiliary task might be explained with the large number of distinct labels (4504) which aggrevates the problem of data sparseness and makes it more difficult to build shared common representations in a multi-task environment.

Chunking as an auxiliary task improves general precision (+0.12), recall (+0.1) and F-score (+0.11) when compared to CTR$_3^{\text{gate}}$. For discontinuous constituents only, precision is improved significantly (+4.02) while recall worsens (-3.59) leading to a slight decline in F-scoree (-0.4). It is not surprising that the integration of chunking as an auxiliary task makes the recognition of long-range dependencies more difficult since the task does not correlate discontinuous chunks with each other.

Performing joint dependency parsing and constituent parsing leads to lower F-score (-0.13) and discontinuous F-score (-3.08) compared to the control model CTR$_3^{\text{gate}}$. Furthermore, the model experiences the lowest discontinuous recall amongst the auxiliary models compared in table 6.5. This is surprising since I expected long-distance syntactic dependencies to be a good indicator of discontinuous constituency. In section 3.2.6 I discussed the results of Coavoux et al. (2019) who show that lexicalisation as a component of transition systems decreases parsing score due to duplicating some transition types and leading to more complicated oracles. The results of DEP$^{\text{gate}}$ show that softly reintroducing the notion of constituent heads as a dependency prediction objective from the underlying contextualised token representation does not have a favourable effect either.

This may be an indicator that dependency parsing, or at least the relative POS-based encoding used here, conflict with the lexicalised information and the head rules that Coavoux et al. (2019) assume LSTMs to learn implicitly.

Using LTAG-spinal supertags decreases F-score (-0.24) and discontinuous F-score (-0.74) relative to the control model. LTAG$^{gate}$ performs only slightly better than the LCFRS$^{gate}$ model despite not being hampered by an unsuitably large number of distinct labels. I think that this result can in part be explained by the fact that Shen et al. (2007) remove subcategorisation information from their supertags. For instance, elementary trees of transitive and ditransitive verbs are the same in LTAG-spinal (Shen et al., 2007). But these distinctions would benefit parsing if they were induced in the network's contextualised token representations. Predicting subcategorisation structure from a number of possible choices requires a model to learn to retrieve contextual information in the input sequence effectively. Furthermore, it is a form of pre-disambiguation of parsing actions. Predicting LTAG-spinal supertags retains too much ambiguity in terms of the constituent structure patterns that can match a lexical category assignment.

The experimental results show that the CCG model is the best-performing auxiliary model in terms of discontinuous F-score among the five types of syntactic sequence labelling tasks explored. The Chunk$^{gate}$ model is unique in that the auxiliary task only covers 10% of the total number of training instances. It shows that auxiliary objectives can be benefitial even in cases of significant imbalance in terms of dataset sizes. The success of the chunking task might also be attributed to the small set of distinct labels (22). It may be a coincidence but is striking nonetheless that general F-score tends to behave inverse proportionally to the number of auxiliary task labels in table 6.5, an exception being LTAG-spinal which performs poorly despite being the supertag task with the smallest number of labels. For further research it might therefore be worthwhile to investigate whether models that dispense of supertag training entirely and only learn to predict deconstructed subcomponent tasks with smaller label set sizes can have a positive effect on parser performance.

### 6.3.3 Per-Phenomenon Evaluation

Using only F-score, precision and recall as evaluation metrics for discontinuous parsing does not allow for a detailed analysis and is hard to interpret. While the metrics of discontinuous F-score, discontinuous precision and discontinuous recall provided by `disco-dop` allow for a general evaluation of the performance on long-range dependencies, they do not give information about structurally distinct discontinuous phenomena like *wh*-extraction or circumpositioned quotations which leaves the syntactic capabilities of a parser unclear.

In order to achieve a finer-grained analysis several strategies have been explored. A comparison of the approaches suggested so far is presented by Coavoux (2021) and will be reproduced briefly in the following. Evang (2011) performed a manual analysis classifying sentences from the development set of the DPTB according to the types of discontinuity that appear in them and manually checked the resulting representations constructed by a PLCFRS parser. Coavoux et al. (2019) follow this strategy in the evaluation of their ML-Gap parser. As Coavoux (2021) note, manual evaluation is very time consuming and has to be repeated in full length for every newly developed parser.

Maier et al. (2014) created a testsuite for German called `discosuite` which allows the automatic evaluation of different types of discontinuous structures found in the TIGER treebank. To achieve this, they annotated sentences from the corpus with the long-range phenomena found in them and by this means makes a per-phenomenon evaluation possible.

Inspired by this idea Coavoux (2021) released a phenomenon-driven test suite for English.[49] They reuse the categories previously employed for manual error analysis by Evang (2011) and

[49] Found as supplementary material here: https://aclanthology.org/2021.findings-acl.288

Coavoux et al. (2019) and, ignoring unprojectivity resulting only from punctuation, categorise all discontinuous trees from the DPTB delevopment set according to one or several discontinuity types resulting in 266 tree instances (16% of the dataset). The per-phenomenon evaluation is achieved by retrieving the number of gold, correct and incorrect discontinuous constituents for each tree individually using `disco-dop` and then aggregating the results for each type. Two separate scores with and without labelling are included.

In the following, I use the testsuite provided by Coavoux (2021) to retrieve detailed and linguistically meaningful information about the syntactic capabilities of the supertag-enriched approaches presented in this work.

**Effects of the Pipeline Approach** Table 6.6 reports the per-phenomenon results of the $\text{Pipe}_{\text{bank}}$ model as well as the best-scoring auxiliary task model $\text{CCG}^{\text{gate}}$. For each metric, the absolute improvement/decline relative to the baseline is included.

For $\text{Pipe}_{\text{bank}}$ relevant increases in F-score can be observed for *it*-extraposition (+7.9), discontinuous dependency (+3.5) and circumpositioned quotation (+2.9). Improvements for *It*-extrapositions have been predicted by the manual comparison of the treatment of discontinuous phenomena in section 5.3. The increase in F-score for the other two categories is rather unexpected. Furthermore, contrary to the analysis in section 5.3 that suggested a beneficial correlation for *wh*-movement and fronted quotation, the F-score of these types actually decreases (-3.2 and -2.1 respectively). While both recall and precision drop, the decline in precision is larger (-5,6 extraction; -2.8 fronted quotation). These two phenomena are the most significant classes of discontinuity so that their relatively small decrease in F-score dominates the overall results.

The increase in F-score for *it*-extrapositions is caused by better recall (+8.3). I assume that the unique category NP[expl] assigned to expletive uses of *it* by the supertagger makes it easier for the parser to recognise discontinuities of this type.

For discontinuous dependency the increase in F-score is driven by a significant improvement in precision (+7.1). This effect is surprising since the manual analysis of phenomena did not suggest clear synergies in these cases due to extraposed dependents not encoding their correct attachment point in their category. Perhaps supertags helped to distinguish the boundaries of continuous subsections which allowed for a more precise analysis. I will investigate this result closer in section 6.3.5.

No change can be observed in the small class of subject inversion. The largest improvement in F-score occurs for sentences with two extractions (extraction+extraction; +15.3). While this class with only five instances is too small to draw statistically relevant conclusions, the result might indicate that lexical category assignments helped the scorer to differentiate between the two extractions through CCG category distinctions.

The unlabelled results are close to the results in the labelled case. A notable exception are all metrics for circumpositioned quotations which a noticeably higher when not including labels. This is coherent with the baseline and with previous results of Coavoux (2021). He traces the fall in labelled scoring back to cases where the discontinuous quotation exhibits an infrequent label like FRAG or SINV. Compared with the baseline, the unlabelled F-score for $\text{Pipe}_{\text{bank}}$ decreases slightly (-0.3) while the labelled score increases (+2.9), driven by recall (+8.0), which suggests that supertag information may have helped to recognise labels for such infrequent cases.

Overall, it can be seen that recall generally increases or only decreases slightly for the pipeline approach with the exception of combined discontinuous dependency and *wh*-extraction (-11.8). The network does not experience the drop in discontinuous recall inherent to the auxiliary models. Precision fluctuates more reflecting the mixed informativeness of CCG supertags for constituent parsing. CCG supertags as an input feature make analysing some types of discontinuity (*wh*-extraction, fronted quotation, circumpositioned quotation) correctly more difficult for the parser.

| | | Labelled | | | | | | | | | | | | | | |
|---|---|---|---|---|---|---|---|---|---|---|---|---|---|---|---|---|
| | | baseline | | | | | PIPE$_{\text{bank}}$ | | | | | CCG$^{\text{gate}}$ | | | | |
| Phenomenon | C | EM | PM | P | R | F | Exact match | Partial match | Precision | Recall | F | Exact match | Partial match | Precision | Recall | F |
| *Wh*-extraction | 91 | 76.9 | 85.7 | 90.0 | 78.8 | 84.0 | 73.6 (-3.3) | 84.6 (-1.1) | 84.4 (-5.6) | 77.5 (-1.3) | 80.8 (-3.2) | 78.0 (+1.1) | 87.9 (+2.2) | 89.9 (-0.1) | 83.1 (+4.3) | 86.4 (+2.4) |
| Fronted quotation | 71 | 93.0 | 93.0 | 95.7 | 93.0 | 94.3 | 90.1 (-2.9) | 91.5 (-1.5) | 92.9 (-2.8) | 91.5 (-1.5) | 92.2 (-2.1) | 93.0 (+0.0) | 93.0 (+0.0) | 94.3 (-1.4) | 93.0 (+0.0) | 93.6 (-0.7) |
| Discontinuous dependency | 37 | 24.3 | 29.7 | 78.6 | 25.0 | 37.9 | 29.7 (+5.4) | 32.4 (+2.7) | 85.7 (+7.1) | 27.3 (+2.3) | 41.4 (+3.5) | 27.0 (+2.7) | 29.7 (+0.0) | 92.3 (+13.7) | 27.3 (+2.3) | 42.1 (+4.2) |
| Circumpositioned quotation | 16 | 0.0 | 81.2 | 54.5 | 36.0 | 43.4 | 0.0 (+0.0) | 87.5 (+6.3) | 48.9 (-5.6) | 44.0 (+8.0) | 46.3 (+2.9) | 0.0 (+0.0) | 75.0 (-6.2) | 46.7 (-7.8) | 28.0 (-8.0) | 35.0 (-8.4) |
| *It*-extraposition | 12 | 41.7 | 41.7 | 100 | 41.7 | 58.8 | 50.0 (+8.3) | 50.0 (+8.3) | 100 (+0.0) | 50.0 (+8.3) | 66.7 (+7.9) | 50.0 (+8.3) | 50.0 (+8.3) | 100 (+0.0) | 50.0 (+8.3) | 66.7 (+7.9) |
| Extraction+fronted quotation | 7 | 71.4 | 100 | 84.2 | 84.2 | 84.2 | 71.4 (+0.0) | 100 (+0.0) | 84.2 (+0.0) | 84.2 (+0.0) | 84.2 (+0.0) | 100 (+28.6) | 100 (+0.0) | 100 (+15.8) | 100 (+15.8) | 100 (+15.8) |
| Discontinuous dependency+extraction | 5 | 20.0 | 80.0 | 61.1 | 64.7 | 62.9 | 0.0 (-20.0) | 80.0 (+0.0) | 75.0 (+13.9) | 52.9 (-11.8) | 62.1 (-0.8) | 0.0 (-20.0) | 100 (+20.0) | 73.3 (+12.2) | 64.7 (+0.0) | 68.8 (+5.9) |
| Extraction+extraction | 5 | 40.0 | 100 | 81.2 | 76.5 | 78.8 | 60.0 (+20.0) | 100 (+0.0) | 94.1 (+12.9) | 94.1 (+17.6) | 94.1 (+15.3) | 40.0 (+0.0) | 100 (+0.0) | 93.8 (+12.6) | 88.2 (+11.7) | 90.9 (+12.1) |
| Subject inversion | 5 | 60.0 | 60.0 | 100 | 60.0 | 75.0 | 60.0 (+0.0) | 60.0 (+0.0) | 100 (+0.0) | 60.0 (+0.0) | 75.0 (+0.0) | 60.0 (+0.0) | 60.0 (+0.0) | 100 (+0.0) | 60.0 (+0.0) | 75.0 (+0.0) |

| | | Unlabelled | | | | | | | | | | | | | | |
|---|---|---|---|---|---|---|---|---|---|---|---|---|---|---|---|---|
| | | baseline | | | | | PIPE$_{\text{bank}}$ | | | | | CCG$^{\text{gate}}$ | | | | |
| Phenomenon | C | EM | PM | P | R | F | Exact match | Partial match | Precision | Recall | F | Exact match | Partial match | Precision | Recall | F |
| *Wh*-extraction | 91 | 76.9 | 85.7 | 89.6 | 80.0 | 84.5 | 74.7 (-2.2) | 84.6 (-1.1) | 84.4 (-5.2) | 79.3 (-0.7) | 81.8 (-2.7) | 78.0 (+1.1) | 87.9 (+2.2) | 90.0 (+0.4) | 84.0 (+4.0) | 86.9 (+2.4) |
| Fronted quotation | 71 | 93.0 | 93.0 | 95.7 | 93.0 | 94.3 | 90.1 (-2.9) | 91.5 (-1.5) | 92.9 (-2.8) | 91.5 (-1.5) | 92.2 (-2.1) | 93.0 (+0.0) | 93.0 (+0.0) | 94.3 (-1.4) | 93.0 (+0.0) | 93.6 (-0.7) |
| Discontinuous dependency | 37 | 24.3 | 29.7 | 78.6 | 25.6 | 38.6 | 32.4 (+8.1) | 35.1 (+5.4) | 92.9 (+14.3) | 30.2 (+4.6) | 45.6 (+7.0) | 27.0 (+2.7) | 29.7 (+0.0) | 91.7 (+13.1) | 25.6 (+0.0) | 40.0 (+1.4) |
| Circumpositioned quotation | 16 | 75.0 | 87.5 | 93.9 | 81.6 | 87.3 | 81.2 (+6.2) | 87.5 (+0.0) | 96.8 (+2.9) | 78.9 (-2.7) | 87.0 (-0.3) | 75.0 (+0.0) | 87.5 (+0.0) | 96.7 (+2.8) | 76.3 (-5.3) | 85.3 (-2.0) |
| *It*-extraposition | 12 | 41.7 | 41.7 | 100 | 41.7 | 58.8 | 50.0 (+8.3) | 50.0 (+8.3) | 100 (+0.0) | 50.0 (+8.3) | 66.7 (+7.9) | 50.0 (+8.3) | 50.0 (+8.3) | 100 (+0.0) | 50.0 (+8.3) | 66.7 (+7.9) |
| Extraction+fronted quotation | 7 | 71.4 | 100 | 88.9 | 88.9 | 88.9 | 71.4 (+0.0) | 100 (+0.0) | 88.9 (+0.0) | 88.9 (+0.0) | 88.9 (+0.0) | 100 (+28.6) | 100 (+0.0) | 100 (+11.1) | 100 (+11.1) | 100 (+11.1) |
| Discontinuous dependency+extraction | 5 | 20.0 | 80.0 | 60.0 | 60.0 | 60.0 | 0.0 (-20.0) | 80.0 (+0.0) | 80.0 (+20.0) | 53.3 (-6.7) | 64.0 (+4.0) | 0.0 (-20.0) | 100 (+20.0) | 76.9 (+16.9) | 66.7 (+6.7) | 71.4 (+11.4) |
| Extraction+extraction | 5 | 40.0 | 100 | 80.0 | 75.0 | 77.4 | 60.0 (+20.0) | 100 (+0.0) | 93.8 (+13.8) | 93.8 (+18.8) | 93.8 (+16.4) | 40.0 (+0.0) | 100 (+0.0) | 93.3 (+13.3) | 87.5 (+12.5) | 90.3 (+12.9) |
| Subject inversion | 5 | 60.0 | 60.0 | 100 | 60.0 | 75.0 | 60.0 (+0.0) | 60.0 (+0.0) | 100 (+0.0) | 60.0 (+0.0) | 75.0 (+0.0) | 60.0 (+0.0) | 60.0 (+0.0) | 100 (+0.0) | 60.0 (+0.0) | 75.0 (+0.0) |

Table 6.6: Per-phenomenon results on the test suite from Coavoux (2021) for the baseline (rerun of Coavoux and Cohen (2019) with updated codebase), PIPE$_{\text{bank}}$ and CCG$^{\text{gate}}$ models. C, EM and PM are count, exact match and partial match. Sentences where several discontinuous phenomena occur are treated separately. Only combinations with at least 5 occurrences in the DPTB development split are included. Absolute changes relative to the baseline are included in parentheses.

It could be that the parser learned to identify functors, arguments and argument directions in CCG supertags. For instance it could have internalised the fact that supertags assigned to fronted *wh*-words like S[wq]/(S[q]/NP) predict an argument to the right and thus wrongfully merges the right neighbour constituent leading to less precision. A deeper investigation of this issue is provided in section 6.3.5.

**Effects of the Auxiliary Approach** For the auxiliary task model, improvements in the labelled case can be observed for *wh*-extractions (+2.4) and *it*-extrapositions (+7.9) which is true to the assumptions. However, like in the pipeline model, albeit to a smaller extent, F-score for fronted quotations decreases (-0.7).

For discontinuous dependency, circumpositioned quotations and subject inversion the manual analysis in section 5.3 did not suggest transparent correlations when including supertags in the parser. Indeed a significant drop for circumpositioned quotations can be observed (-8.4). On the other hand, discontinuous dependency F-score increases (+4.2) similar to the pipeline model. Precision even exhibits twice the amount of improvement as in $\text{CCG}_{\text{bank}}$ (+13.7). The integration of supertagging as an auxiliary task has no effect on subject inversion.

The decrease in F-score pertaining to circumpositioned quotations is caused by a degradation of both recall (-8.0) and precision (-7.8). This differs from the pipeline result where recall increased (+8.0) and may be due to the fact that circumpositioned quotations feature some rare categories that the auxiliary model is trained to predict while `depccg` rebank is only trained on the 511 most frequent supertags. Thus, this type of discontinuity might suffer from training data sparseness in the auxiliary approach.

As section 5.3 showed, the analysis of circumpositioned quotations generally seems to be inconsistent in the CCGrebank. Thus, conditioning a shared representation to predict supertags for such cases likely introduces a lot of noise. It may also suggest that the pipeline-model was able to identify uninformative assignments that it learned to ignore while the auxiliary model was conditioned to learn a shared representation for conflicting analyses.

The trends are roughly mirrored in the unlabelled case. Overall the improvements of the auxiliary approach compared to the pipeline model occur across most of the types of phenomena with the exception of circumpositioned quotiations. The phenomenon-specific analysis confirms that joint learning is more effective in building representations that can benefit from correlations between CCG lexical category assignments and discontinuous constituent parsing.

### 6.3.4 Error Analysis

In addition to the per-phenomenon analysis, the test suite released by Coavoux (2021) automatically classifies errors according to structural patterns. For this, the author builds on the standard `Berkeley Parser Analyser` (Kübler et al., 2009) and extends it to discontinuous trees. Errors in the predictions are classified according to manually defined patterns by analysing transformations that convert the predicted tree into the gold tree. Coavoux (2021) reports the following error types specifically for discontinuous nodes:

1. PP attachment
2. NP internal structure
3. Modifier Attachment
4. Unary constituent
5. Different label
6. Clause attachement

7. Coordination
8. NP attachment
9. VP attachment

After classification, for each category the number of error occurrences as well as the number of node errors involved is returned.

Table 6.7 contains the results of the error analysis on the models of this work. Changes relative to the baseline are included in parentheses. I will focus on comparing errors on discontinuous nodes specifically. The largest contributor to discontinuous errors in the baseline model is erroneous attachment of constituents. Both the pipeline and the auxiliary model significantly reduce the number of discontinuous PP attachment errors (-25.7%, -22.9%) and modifier attachment errors (-25.0%, -33.3%) as well as the number of affected nodes. This shows that CCG supertag information can be very helpful for a parser to resolve PP and modifier attachment ambiguities. I assume that this is a significant factor for the improvements in extraposed dependency score where appended discontinuous PP constituents are frequent.

| | baseline | | PIPE$_{\text{bank}}$ | | CCG$^{\text{gate}}$ | |
|---|---|---|---|---|---|---|
| Error type | Count | Nodes | Count | Nodes | Count | Nodes |
| PP attachment | 513 | 1076 | 446 (-13.1%) | 1045 (-2.9%) | 438 (-14.6%) | 952 (-11.5%) |
| ∟ discontinuous | 35 | 58 | 26 (-25.7%) | 37 (-36.2%) | 27 (-22.9%) | 41 (-29.3%) |
| Unclassified | 452 | 585 | 400 (-11.5%) | 518 (-11.5%) | 397 (-12.2%) | 541 (-7.5%) |
| Single word phrase | 386 | 453 | 353 (-8.5%) | 422 (-6.8%) | 352 (-8.8%) | 409 (-9.7%) |
| Unary | 325 | 325 | 339 (+4.3%) | 339 (+4.3%) | 321 (-1.2%) | 321 (-1.2%) |
| ∟ discontinuous | 0 | 0 | 2 (+∞%) | 2 (+∞%) | 0 (+0.0%) | 0 (+0.0%) |
| Different label | 268 | 538 | 214 (-20.1%) | 442 (-17.8%) | 255 (-4.9%) | 513 (-4.6%) |
| ∟ discontinuous | 17 | 36 | 19 (+11.8%) | 52 (+44.4%) | 19 (+11.8%) | 41 (+13.9%) |
| Modifier attachment | 261 | 449 | 258 (-1.1%) | 482 (+7.3%) | 229 (-12.3%) | 447 (-0.4%) |
| ∟ discontinuous | 12 | 17 | 9 (-25.0%) | 14 (-17.6%) | 8 (-33.3%) | 14 (-17.6%) |
| NP internal structure | 258 | 318 | 262 (+1.6%) | 339 (+6.6%) | 253 (-1.9%) | 317 (-0.3%) |
| ∟ discontinuous | 0 | 0 | 0 (+0.0%) | 0 (+0.0%) | 1 (+∞%) | 1 (+∞%) |
| Clause attachment | 192 | 457 | 187 (-2.6%) | 462 (+1.1%) | 192 (+0.0%) | 496 (+8.5%) |
| ∟ discontinuous | 26 | 44 | 27 (+3.8%) | 50 (+13.6%) | 27 (+3.8%) | 38 (-13.6%) |
| Co-ordination | 158 | 452 | 150 (-5.1%) | 456 (+0.9%) | 146 (-7.6%) | 413 (-8.6%) |
| ∟ discontinuous | 6 | 11 | 3 (-50.0%) | 8 (-27.3%) | 3 (-50.0%) | 8 (-27.3%) |
| NP attachment | 105 | 316 | 119 (+13.3%) | 363 (+14.9%) | 108 (+2.9%) | 354 (+12.0%) |
| ∟ discontinuous | 17 | 13 | 20 (+17.6%) | 29 (+123.1%) | 16 (-5.9%) | 18 (+38.5%) |
| VP attachment | 51 | 245 | 44 (-13.7%) | 189 (-22.9%) | 44 (-13.7%) | 186 (-24.1%) |
| ∟ discontinuous | 5 | 8 | 4 (-20.0%) | 7 (-12.5%) | 5 (+0.0%) | 8 (+0.0%) |
| XoverX Unary | 11 | 11 | 10 (-9.1%) | 10 (-9.1%) | 10 (-9.1%) | 10 (-9.1%) |

Table 6.7: Error types for the models including the baseline on the development set. Discontinuous counts are included in the respective main category. Changes relative to the baseline in percent are included in parentheses.

Discontinuous clausal attachment errors increase for PIPE$_{\text{bank}}$ (+3.8%) and for CCG$^{\text{gate}}$ (+3.8%). Discontinuous NP attachment errors only increase in count for the pipeline model (+17.6%) while they drop for the auxiliary model (-5.9%). Yet, the number of nodes affected increases for both models and the metrics also significantly increase for the general (including projective errors) case. The supertag-enriched models appear to experience some difficulties in the treatment of discontinuous NPs.

Errors due to wrong label assignments on discontinuous constituents increase for both models by 11.8% relative to the baseline. This is striking considering the fact that the general error count for different labels decreases both for the pipeline model (-20.1%) and for the auxiliary model (-4.9%). Recall that CCG supertags are designed as structural descriptions operating on a small number of atomic categories (NP, S, ...). Thus, some discontinuous phenomena that are assigned different labels in the DPTB due to linguistic motivation may be structurally similar in terms of their CCG supertag analysis which makes assigning the correct node label more difficult for the parser.

Overall, the effect of the inclusion of supertags on different error types does not follow a uniform trend. While the results of $\text{PIPE}_{\text{bank}}$ and $\text{CCG}^{\text{gate}}$ share some similar patterns (e.g. discontinuous PP attachment, discontinuous different label) some types of errors (e.g. general unary, general NP internal structure, discontinuous clause attachment) seem to be dependent on the kind of supertag integration. This may either be a hint at differences between the CCGbank and the CCGrebank analysis and/or further support for the ability of the auxiliary approach to leverage synergies that single-task models are not able to profit from.

### 6.3.5 Sample Analysis

In sections 6.3.3 and 6.3.4 several observations regarding the models' performance on *wh*-movement, fronted quotations and extraposed dependents were made that deserve closer investigation. In the following, I will conduct a more detailed analysis using samples from the models' predictions on the DPTB development split.

**Comparing Auxiliary and Pipeline: *wh*-movement** The per-phenomenon analysis in section 6.3.3 showed that F-score for *wh*-extraction is reduced relative to the baseline in the pipeline approach while it increases in the $\text{CCG}^{\text{gate}}$ model. To investigate the performance on *wh*-extraction in more detail, I computed separate scores for each trigger of frontation in the DPTB development split, listed in table 6.8. Sentences are assigned to the categories using annotations found in the supplementary material of Coavoux (2021).

| | | $\text{PIPE}_{\text{bank}}$ | | | | | $\text{CCG}^{\text{gate}}$ | | | | |
|---|---|---|---|---|---|---|---|---|---|---|---|
| Trigger | Count | Exact match | Partial match | Precision | Recall | F | Exact match | Partial match | Precision | Recall | F |
| when | 35 | 85.7 (+0.0) | 94.3 (+2.9) | 90.5 (+0.5) | 90.5 (+4.8) | 90.5 (+2.7) | 88.6 (+2.9) | 94.3 (+2.9) | 95.0 (+5.0) | 90.5 (+4.8) | 92.7 (+4.9) |
| where | 14 | 85.7 (+0.0) | 85.7 (+0.0) | 95.5 (+0.0) | 91.3 (+0.0) | 93.3 (+0.0) | 85.7 (+0.0) | 85.7 (+0.0) | 95.5 (+0.0) | 91.3 (+0.0) | 93.3 (+0.0) |
| which | 12 | 58.3 (-16.7) | 75.0 (-16.7) | 85.7 (-10.5) | 64.3 (-25.0) | 73.5 (-19.1) | 75.0 (+0.0) | 91.7 (+0.0) | 81.2 (-15.0) | 92.9 (+3.6) | 86.7 (-5.9) |
| how | 9 | 66.7 (+0.0) | 77.8 (+0.0) | 52.4 (-16.4) | 57.9 (+0.0) | 55.0 (-7.9) | 77.8 (+11.1) | 88.9 (+11.1) | 80.0 (+11.2) | 84.2 (+26.3) | 82.1 (+19.2) |
| that | 9 | 44.4 (+0.0) | 66.7 (+0.0) | 94.7 (+0.9) | 75.0 (+12.5) | 83.7 (+8.7) | 33.3 (-11.1) | 55.6 (-11.1) | 92.3 (-1.5) | 50.0 (-12.5) | 64.9 (-10.1) |
| what | 9 | 66.7 (-11.1) | 77.8 (+0.0) | 66.7 (-13.3) | 61.5 (+0.0) | 64.0 (-5.6) | 77.8 (+0.0) | 88.9 (+11.1) | 90.9 (+10.9) | 76.9 (+15.4) | 83.3 (+13.7) |

Table 6.8: Labelled evaluation results specific to lexical triggers of *wh*-extraction. Absolute improvements over the baseline are included in parentheses. Only cases with at least five occurrences are reported.

Section 5.3 suggested that the CCG analysis for *wh*-extraction is dependent on the syntactic function of the fronted word. More specifically: in cases where the *wh*-word represents an argument required by the predicate of the sentence it would be easier to resolve for the parser since the verb's supertag would provide it with the information where to attach the question word while adjunctive frontation where the *wh*-word does not match an argument slot in the verb would be more difficult for the parser, e.g. in the case of *how*.

The evaluation results in table 6.8 are not that clear-cut. F-score for *which*, thought to be encoded as a core-argument, drops for both models (pipeline: -19.1, auxiliary -5.9) while the score for *when*, being purely adjunctive, increases (+2.7, +4.9). The differences in F-score between PIPE$_{\text{bank}}$ and CCG$_{\text{gate}}$ result from extrapositions of *how* and *what* on which the pipeline model performs worse than the baseline while the auxiliary model significantly increases F-score. Likewise noteworthy is the improvement on *that*-extraposition in the pipeline model (+8.7) caused by better recall (+12.5) while recall (-12.5) causes a drop in F-score for this type in the auxiliary task model (-10.1). Interestingly, it is the only *wh*-word where the auxiliary model exhibits a degradation of recall.

Let us look at an example for the *wh*-word *how*. Figure 6.11 shows part of the gold DPTB derivation for the sentence in (6.6a) which is retrieved correctly by CCG$^{\text{gate}}$. The constituent *how long* is extraposed to the front and forms a discontinuous verbal phrase together with the VB take, its NP argument and the S constituent *to make a contribution*. The PIPE$_{\text{bank}}$ model returns the faulty analysis reproduced in figure 6.12. Here, the final S-constituent is not analysed as part of the VP but instead a discontinuous nominal phrase with *it* as an additional member is predicted, suggesting an occurrence of *it*-extraposition, and attached at the top-level S node.

(6.6) a. “ The question is how long it ’s going to take Barry Wright to make a contribution , ” says F. John Mirek , an analyst at Blunt Ellis Loewi in Milwaukee .

b. While reaching blockbuster proportions yesterday, the volume was still well within the 600 million-share capacity that the exchange has said it can handle daily since beefing up its computers after the October 1987 crash.

Figure 6.13 shows the CCGrebank supertag assignments which were used to train the auxiliary model as well as the corresponding derivation. The long-range dependency is realised through an additional NP argument for *take*. The *wh*-word *how* expects an N argument to its right, which is filled by *long*, and as a second argument a declarative sentence category with an unfilled NP slot. This way, it “consumes” the rest of the sentence. *how* is actually not realised as an adjunctive category but conforms with the way core *wh*-extractions were analysed in section 5.3. Furthermore, this coincides exactly with the orginal CCGbank assignments. The fact that *wh*-words that I thought to be purely adjunctive indeed match argument slots with their discontinuous counterpart in the CCG analysis may in part explain why the differences in performance between the triggers for *wh*-extraction in table 6.8 are not as clear-cut as predicted.

Figure 6.14 contains the 1-best supertags generated by standard `depccg` for the sentence in (6.6a). They deviate from the CCGbank which was used to train the supertagger in that *take* is only assigned one right NP argument. Furthermore, *how* does not expect a function with range S[dcl] and a missing NP argument to the right any more but only a complete S[dcl] type.

As figure 6.12 shows, the parser was still able to recognise the discontinuous relationship of *how long* and *take*. The error occurs thereafter. I assume that the parser learned that the tag ((S[b]\NP)/(S[to]\NP))/NP has only two arguments and is misled by this assignment provided by `depccg`. When the transition system combines *how long* and *take* it implicitly recognises one argument slot as filled and then fills the second slot with *Barry Wright*. The scorer notices that the following S constituent surpasses the arity of *take* so it predicts an alternative attachment point being led by the fact that *to make a contribution* can be combined to a category that looks for an NP to its left. Therefore, it merges with the NP *it* — a construction not uncommon for *it*-extrapositions.

The baseline model also commits the error of merging *to make a contribution* with *it*. These observations can be taken as an example for the help CCG predicate-argument structure can provide to the constituent parser and showcases how the auxiliary model is able to profit from synergies between DPTB constituents and CCG category assignments.

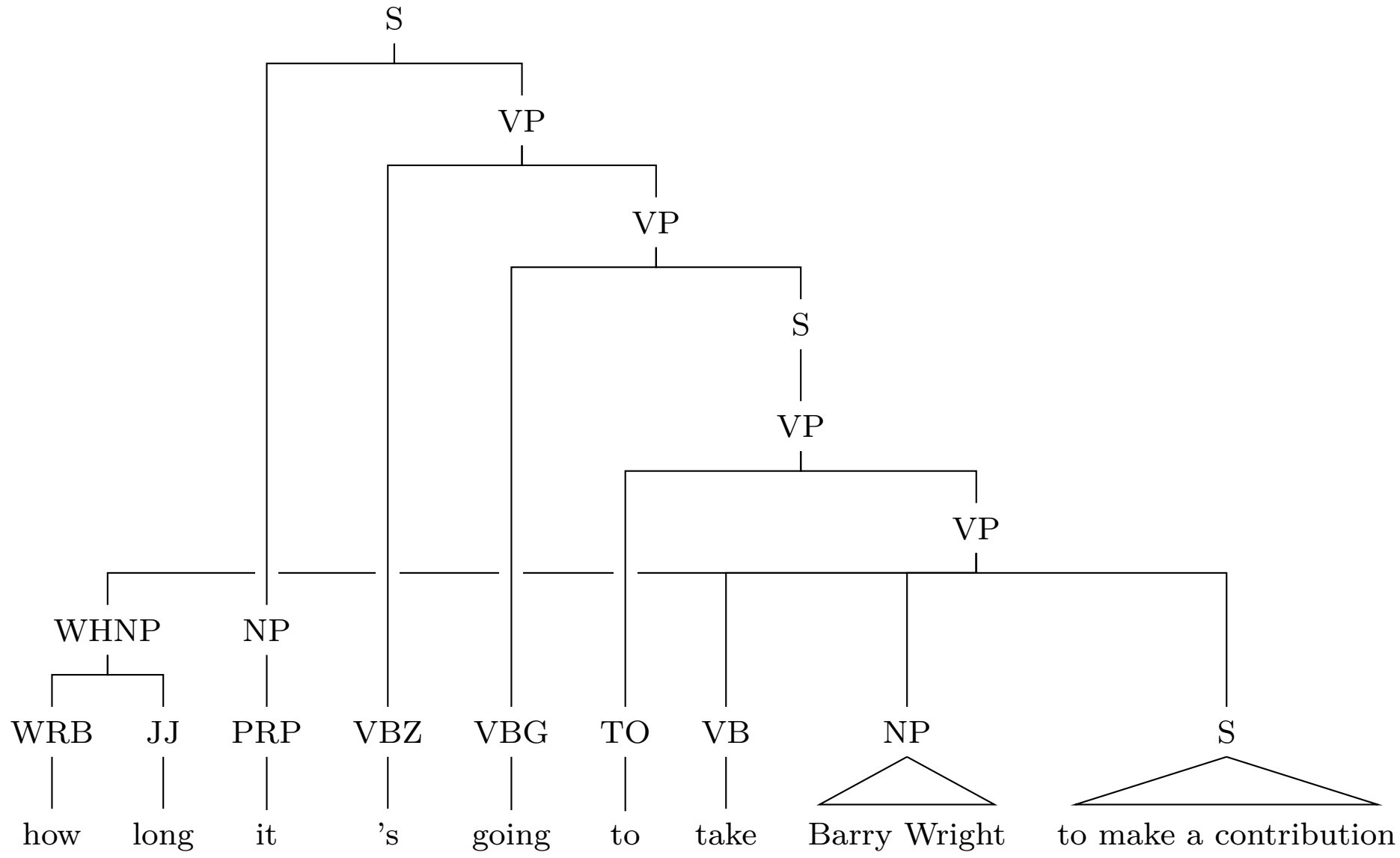

Figure 6.11: Gold derviation *wh*-extraction using *how* in the DPTB.

However, the picture is not always that clear. (6.6b) gives an example for *wh*-extraction caused by the relative pronoun *that*. Figure 6.15 shows the DPTB analysis of this sentence which the pipeline model is able to construct correctly. However, the auxiliary model attaches *that* at a higher level predicting an additional intermediary S-node that constitutes a discontinuity between *that* and *it can handle daily...* (figure 6.16). Thus, the fronted *wh*-word is analysed as a modifier at sentential level instead of as a verbal argument. This leads to lower recall since *that* is missing as part of the lower-level unprojective constituents SBAR, S, VP and VP.

Figure 6.17 depicts the CCG category assignments provided by the CCGrebank while figure 6.18 contains the 1-best predictions of the `depccg` supertagger which exactly coincide with the original CCGbank. Both supertag sequences are the same except for minor differences: *that* being an N-modifier in CCGrebank and an NP-modifier in CCGbank, the categorial type NP[nb] occurring only in the CCGbank and the particle *up* being an argument of *beefing* in the rebank but an adjunct in the original treebank. Differences in analysis can therefore not be attributed to deviating lexical category assignments.

It seems as if the model had analysed *that* and *it can handle daily ...* as a circumpositioned quotation, being triggered by the occurrence of the word *said*. The per-phenomenon analysis in section 6.3.3 has already shown that the auxiliary model performs worse on circumpositioned quotations than the baseline while $\text{PIPE}_{\text{bank}}$ improves F-score for this category. Apparently the inconsistent CCGrebank gold categories introduce so much noise into the network the analysis of other types of discontinuities is negatively affected as well.

Another factor may be the fact that with increasing sentence length it may become harder for the parser to identify whether a missing argument should be filled with some successor item or with an extraposed constituent. In this case, when having reached *handle* the discontinuous constituent parser must decide whether to perform a MERGE or to SHIFT to fill the NP argument slot with some right neighbour. Therefore, it needs to know if an adjacent subsection of the remaining sequence can be reduced to an NP. Through category assignment alone this cannot be ruled out since NP appears twice as a functor in the remainder of the sentence (*its* and *the*). Therefore, the model would need to perform a shallow form of CCG parsing. This information must be construed by the multi-layer biLSTM network. I suppose that the neural model has difficulties performing such an implicit derivation in cases of long input sequences.

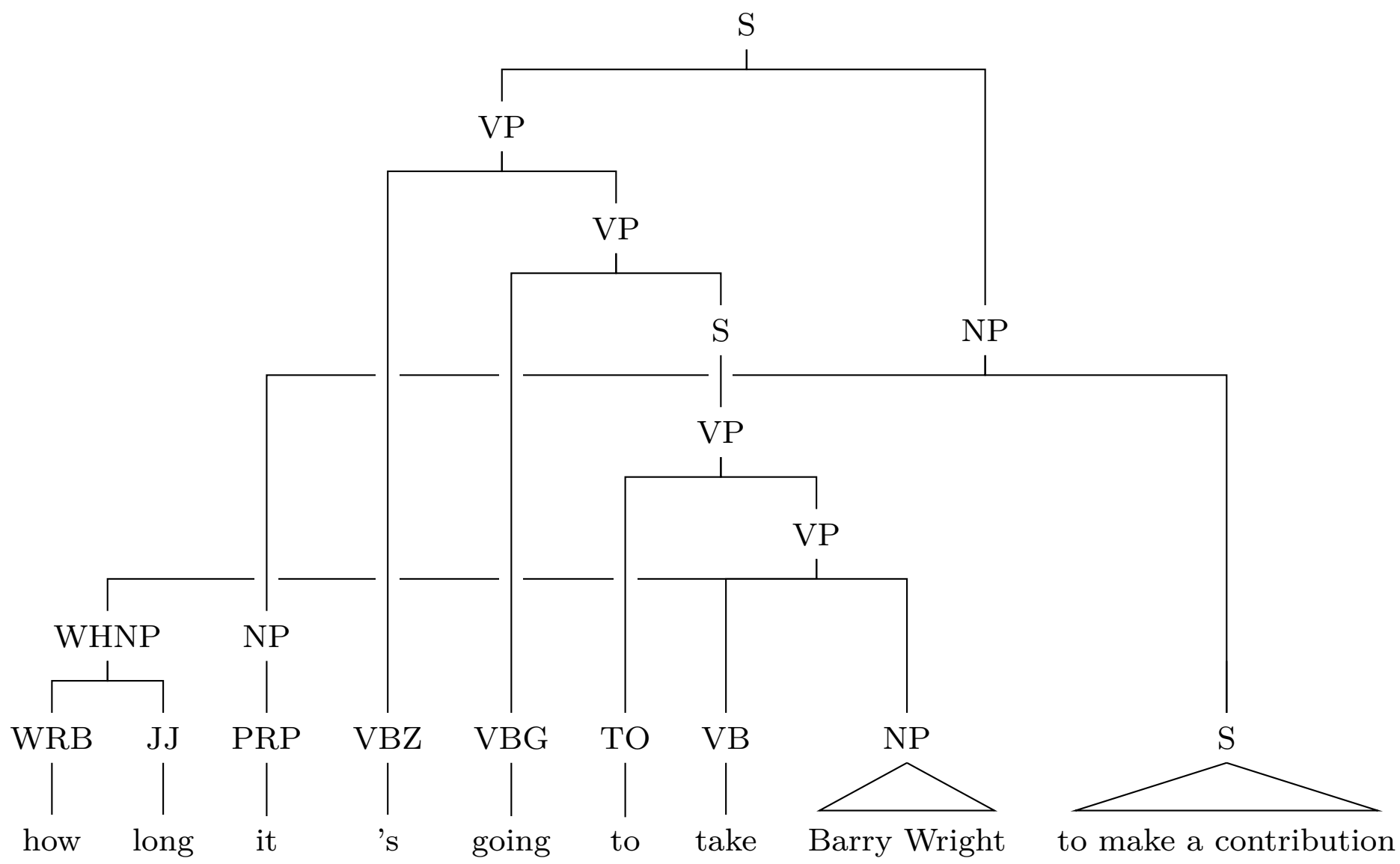

Figure 6.12: Faulty derivation of *wh*-extraction with *how* produced by the PIPE$_{\text{bank}}$ model. Instead of attaching *to make a contribution* to the VP dominating *how long*, *take* and *Barry Wright*, it predicts another discontinuity merging with *it*.

**Unexpected Decrease: Fronted quotations** Contrary to the analysis in section 5.3 which suggested a transparent correlation between the analysis of fronted quotations in the DPTB and the supertag assignments in the CCGrebank, where both the fronted quotation and the subject of the quotative verb are encoded positionally as arguments, the per-phenomenon analysis in section 6.3.3 showed that F-score for this type of discontinuity decreases in the PIPE$_{\text{bank}}$ model (-2.1) and in the CCG$^{\text{gate}}$ model. A drop in precision is the primary cause. In (6.7) I have included a sentence that features a fronted quotation which the baseline model analyses correctly (cf. figure 6.19). However, the auxiliary model produces erroneous output by including the predicate phrase *After the game* into the discontinuous VP as part of the quotation, as can be seen in figure 6.20.

(6.7) After the game ( “ Bluefield lost , 9-8 , stranding three runners in ... the ninth , ” he noted ) , trouble began .

The sentence in (6.7) is an unusual example since the quotative clause is embedded into parentheses. Assigning a CCG analysis to them apparently was problematic which is why the sentence was left out from the CCGbank and its rebanked successor. For this reason it is also not present in the CCG training dataset used for CCG$^{\text{gate}}$ so that it, and possibly similar sentences, were underrepresented during training. Therefore, the model might have not learned to treat clauses encapsulated by parentheses as single units. At the same time, most fronted quotations start at the beginning of the sentence. Thus, the network is misled by the usual analysis of a PP like *After the game* as a modifier of the following S category and attaches it projectively to the quotation.

On a similar note, I suspect that the reintegration of quotation marks in the CCGrebank is a factor for the better score of the auxiliary model compared to PIPE$_{\text{bank}}$. When providing the `depccg` standard model with quotations surrounded by quotation marks, it simply analyses them as sentential modifiers like `S/S` which may conflict with traditional sentential modifiers that behave differently while the CCGrebank assigns special categories `LQU` and `RQU`. Training the model to predict these labels may have induced knowledge about the restrictions for functional application across quotation mark boundaries.

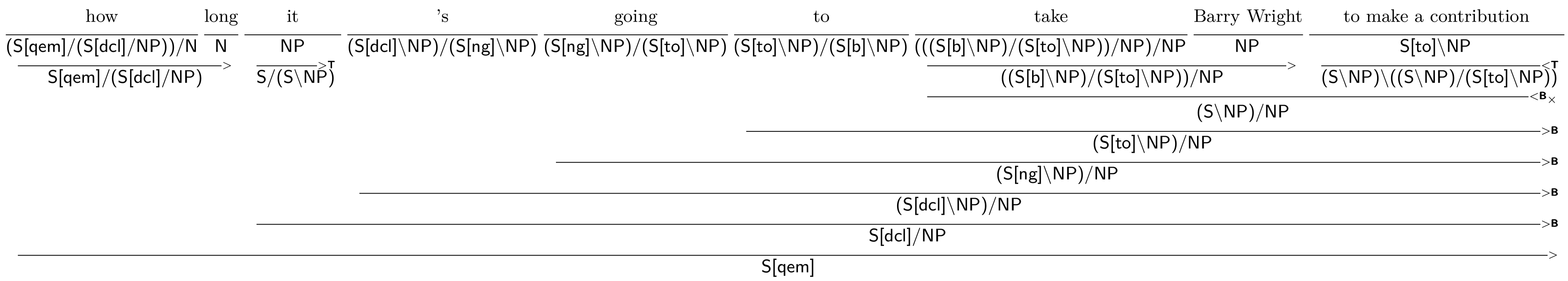

Figure 6.13: Derivation of *wh*-extraction with *how* in the CCGrebank.

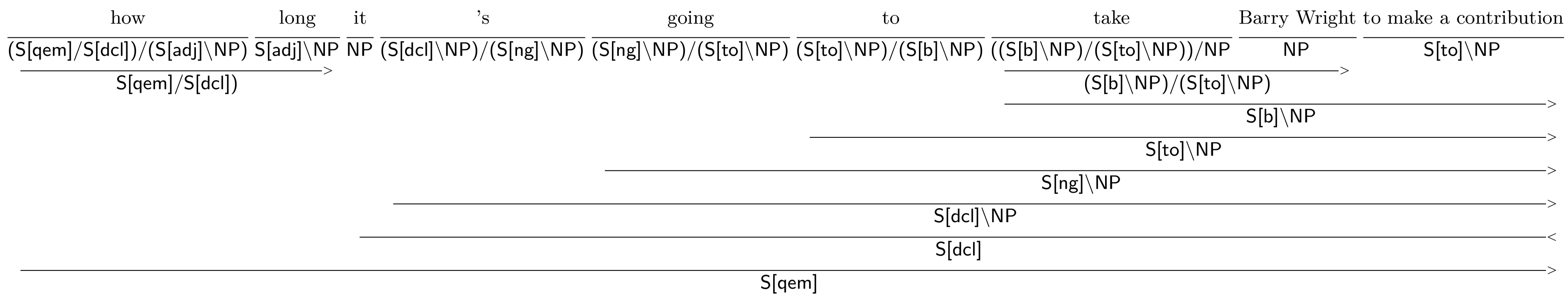

Figure 6.14: Derivation of *wh*-extraction with *how* generated by `depccg`.

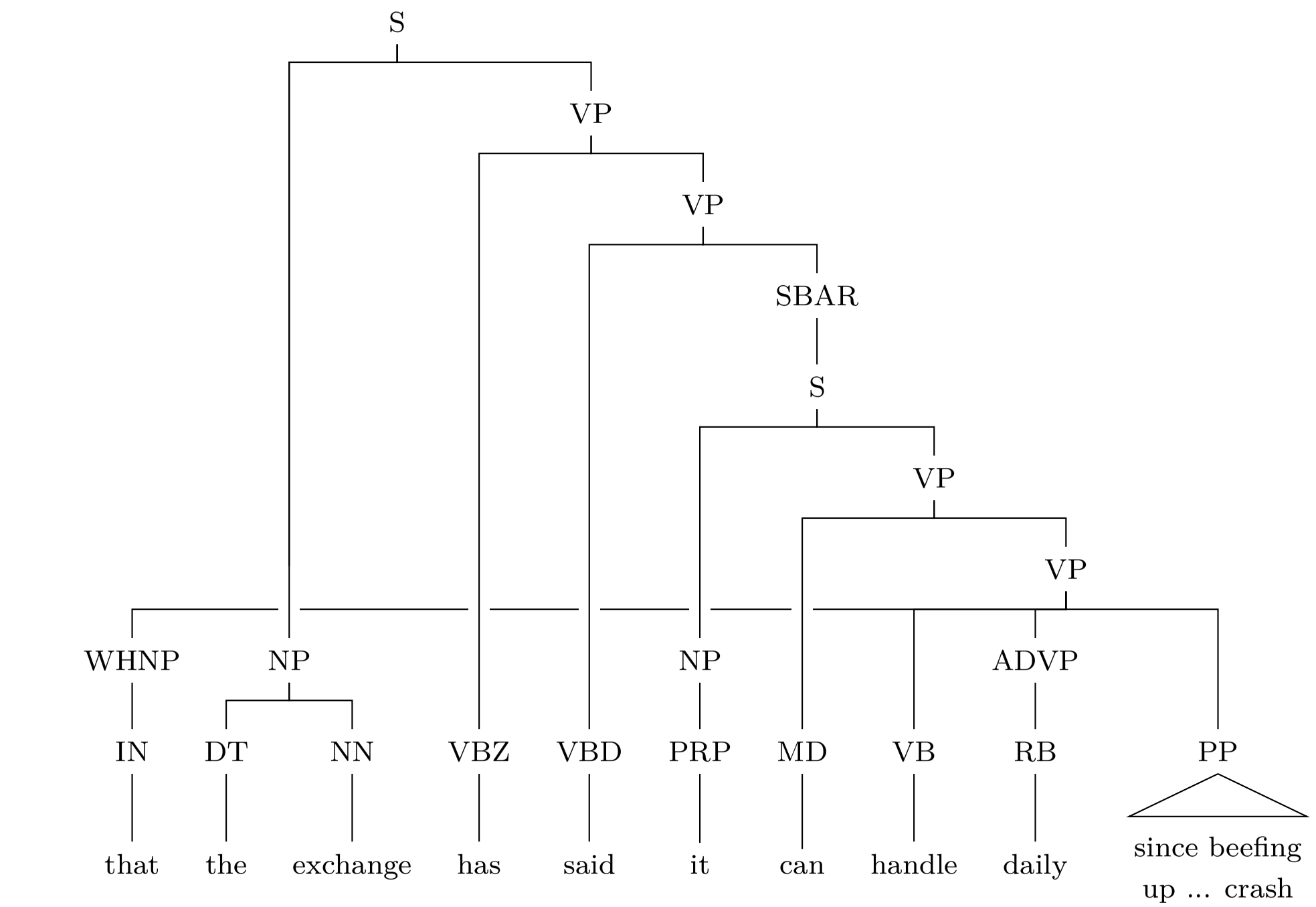

Figure 6.15: Gold derviation *wh*-extraction using *that* in the DPTB.

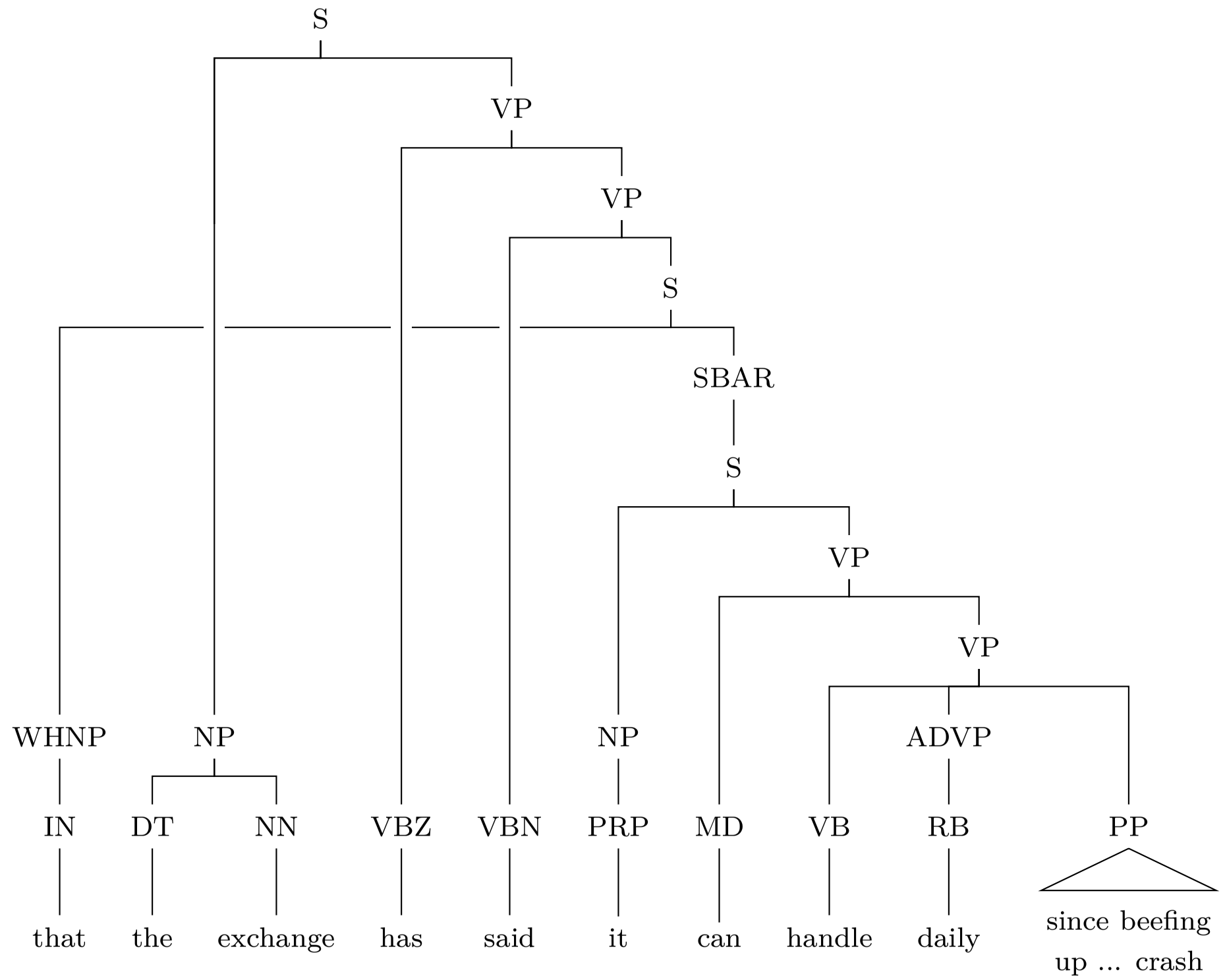

Figure 6.16: Faulty derivation of *wh*-extraction with *that* produced by the CCG$^{\text{gate}}$ model. Instead of attaching *that* to the VP dominating *handle daile since...* it predicts a circumpositioned quotation

| that | the exchange | has | said | it | can | handle | daily | since | beefing | up | its computers | after | ... crash |
|---|---|---|---|---|---|---|---|---|---|---|---|---|---|
| (N\N)/(S[dcl]/NP) | NP | (S[dcl]\NP)/(S[pt]\NP) | (S[pt]\NP)/S[dcl] | NP | (S[dcl]\NP)/(S[b]\NP) | (S[b]\NP)/NP | (S\NP)\(S\NP) | ((S\NP)\(S\NP))/(S[ng]\NP) | ((S[ng]\NP)/NP)/PR | PR | NP | ((S\NP)\(S\NP))/NP | NP |

Figure 6.17: CCG supertags from the CCGrebank for *wh*-extraction of the word *that*.

| that | the exchange | has | said | it | can | handle | daily | since | beefing | up | its ... | after | ... crash |
|---|---|---|---|---|---|---|---|---|---|---|---|---|---|
| (NP\NP)/(S[dcl]/NP) | NP[nb] | (S[dcl]\NP)/(S[pt]\NP) | (S[pt]\NP)/S[dcl] | NP | (S[dcl]\NP)/(S[b]\NP) | (S[b]\NP)/NP | (S\NP)\(S\NP) | ((S\NP)\(S\NP))/(S[ng]\NP) | (S[ng]\NP)/NP | (S\NP)\(S\NP) | NP | ((S\NP)\(S\NP))/NP | NP |

Figure 6.18: CCG supertags assigned by `depccg` for *wh*-extraction of the word *that*.

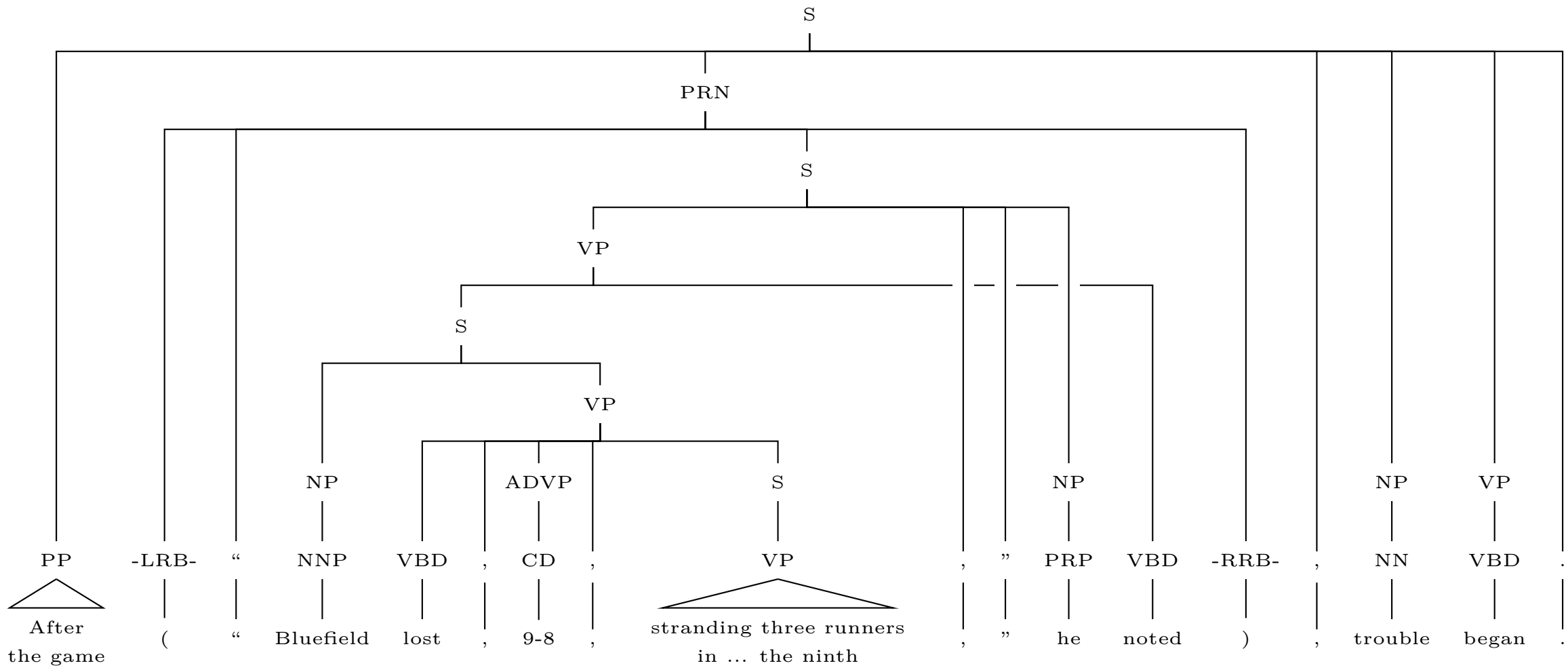

Figure 6.19: Gold derivation of a fronted quotation encapsulated in parentheses in the DPTB.

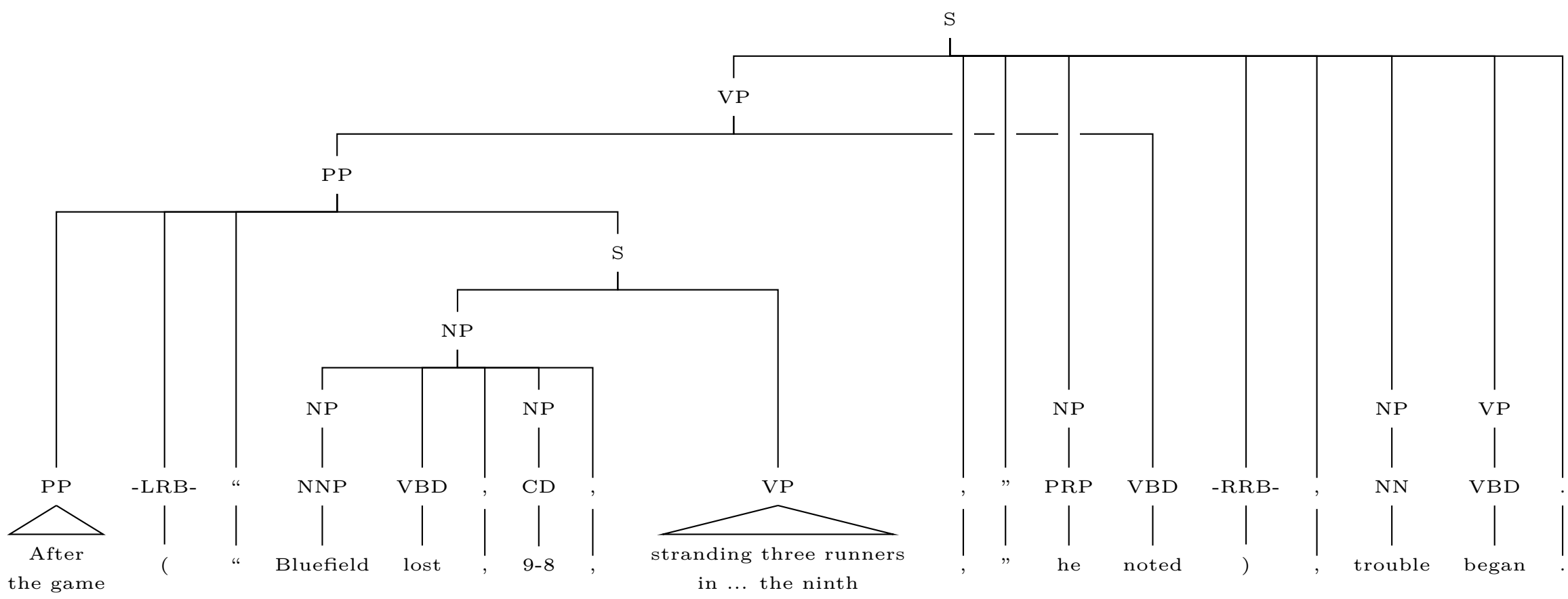

Figure 6.20: Faulty derivation of a fronted quotation encapsulated in parentheses output by the $\text{CCG}^{\text{gate}}$ model.

**Unexpected Increase: Discontinuous Dependencies** The per-phenomenon analysis in section 6.3.3 has shown consistent improvements in precision and recall for discontinuous dependencies for both $\text{PIPE}_{\text{bank}}$ and $\text{CCG}^{\text{gate}}$. F-score for the former increases by +3.5 absolute points compared to the baseline while the score for the latter rises by +4.2 points. This result is surprising since the manual analysis of this type of discontinuity in section 5.3 suggested that the CCG formalism does not feature a transparent and consistent way to encode such relationships. In the examples that were examined, the right part of the discontinuous constituent was assigned a functional category that suggests analysing it as a modifier of the middle constituent surrounded by the discontinuity (cf. figure 5.21).

In his supplementary material Coavoux (2021) includes information about the structure of extraposed dependents in the DPTB development dataset in form of the label(s) of the node(s) that are surrounded by the long-range dependency, called *gap* here, and the node label(s) of the appended subsection of the discontinuous constituent, abbreviated to *right*. I use these assignments to compute separate scores which shed light on the parsers' increased competence regarding discontinuous dependencies.

| Gap | Count | PIPE$_{bank}$ Exact match | PIPE$_{bank}$ Partial match | PIPE$_{bank}$ Precision | PIPE$_{bank}$ Recall | PIPE$_{bank}$ $F_1$ | CCG$^{gate}$ Exact match | CCG$^{gate}$ Partial match | CCG$^{gate}$ Precision | CCG$^{gate}$ Recall | CCG$^{gate}$ $F_1$ |
|---|---|---|---|---|---|---|---|---|---|---|---|
| PP | 19 | 42.1 (+10.5) | 42.1 (+10.5) | 88.9 (+22.2) | 36.4 (+9.1) | 51.6 (+12.9) | 47.4 (+15.8) | 47.4 (+15.8) | 90.9 (+24.2) | 45.5 (+18.2) | 60.6 (+21.9) |
| VP | 10 | 0.0 (-10.0) | 0.0 (-10.0) | 0.0 (-100) | 0.0 (-10.0) | 0.0 (-18.2) | 0.0 (-10.0) | 0.0 (-10.0) | 0.0 (-100) | 0.0 (-10.0) | 0.0 (-18.2) |
| NP | 5 | 60.0 (+20.0) | 60.0 (+0.0) | 100 (+0.0) | 50.0 (+0.0) | 66.7 (+0.0) | 20.0 (-20.0) | 20.0 (-40.0) | 100 (+0.0) | 16.7 (-33.3) | 28.6 (-38.1) |

Table 6.9: Labelled evaluation results for discontinuous dependency according to the node label of the gap element. Absolute improvements over the baseline are included in parentheses. Only cases with at least five occurrences are reported.

| Right | Count | PIPE$_{bank}$ Exact match | PIPE$_{bank}$ Partial match | PIPE$_{bank}$ Precision | PIPE$_{bank}$ Recall | PIPE$_{bank}$ $F_1$ | CCG$^{gate}$ Exact match | CCG$^{gate}$ Partial match | CCG$^{gate}$ Precision | CCG$^{gate}$ Recall | CCG$^{gate}$ $F_1$ |
|---|---|---|---|---|---|---|---|---|---|---|---|
| SBAR | 13 | 23.1 (+7.7) | 23.1 (+7.7) | 100 (+60.0) | 21.4 (+7.1) | 35.3 (+14.2) | 7.7 (-7.7) | 7.7 (-7.7) | 50.0 (+10.0) | 7.1 (-7.2) | 12.5 (-8.6) |
| S | 11 | 72.7 (+9.1) | 72.7 (+9.1) | 88.9 (-11.1) | 66.7 (+8.4) | 76.2 (+2.5) | 63.6 (+0.0) | 63.6 (+0.0) | 100 (+0.0) | 58.3 (+0.0) | 73.7 (+0.0) |
| PP | 7 | 0.0 (+0.0) | 14.3 (+0.0) | 50.0 (-50.0) | 9.1 (+0.0) | 15.4 (-1.3) | 14.3 (+14.3) | 28.6 (+14.3) | 100 (+0.0) | 27.3 (+18.2) | 42.9 (+26.2) |

Table 6.10: Labelled evaluation results for discontinuous dependency according to the node label of the right element. Absolute improvements over the baseline are included in parentheses. Only cases with at least five occurrences are reported.

It can be seen that the improvements in F-score for both models are caused by a solid increase in competence of the analysis of prepositional phrase gap elements (pipeline: +12.9, auxiliary: +21.9), with the improvement being strongly driven by better precision. Furthermore F-score for PP right elements increases significantly for the auxiliary model (+26.2). This confirms the previous finding from the error analysis in section 6.3.4 which showed reductions in PP attachment errors.

Both models experience a striking reduction in VP gap F-score (-18.2) resulting in a worst-possible score of 0.0. For CCG$^{gate}$ this is combined with a reduction for SBAR right elements (-8.6) which is a common combination for discontinuous NP constituents as in figure 5.20.

A manual analysis showed that the drop in F-score for cases with VP nodes in the gap indeed corresponds to misleading supertag assignments such as in figure 5.21. The appended half of the discontinuous constituent is assigned a supertag in the CCGrebank that takes as argument the gap verb with all object slots filled (`(S\NP)\(S\NP)`), i.e. it acts as a VP modifier. The parser learns to predict this supertag and is misled by the fact that standard projective VP modifiers carry the same supertag. Thus, it wrongly merges the gap element with the right constituent part, not constructing a discontinuity at all.

In other cases of discontinuous dependency, the CCG category assignments appear to be closer to the DPTB constituent structure. Take for instance (6.8a) which features a discontinuous NP category. Figure 6.21 shows part of its DPTB gold tree which is correctly retrieved by CCG$^{gate}$ except that the topmost VP and its NP daughter are merged into a single VP node.

(6.8) a. The institutions appeared confident Japanese regulators would step in to ensure orderly trading if necessary , and there was considerable speculation during the day that the Finance Ministry was working behind the scenes to do just that .

b. The need for hurried last-minute telephone negotiations among market officials will disappear once rules are in place that synchronize circuit breakers in all markets .

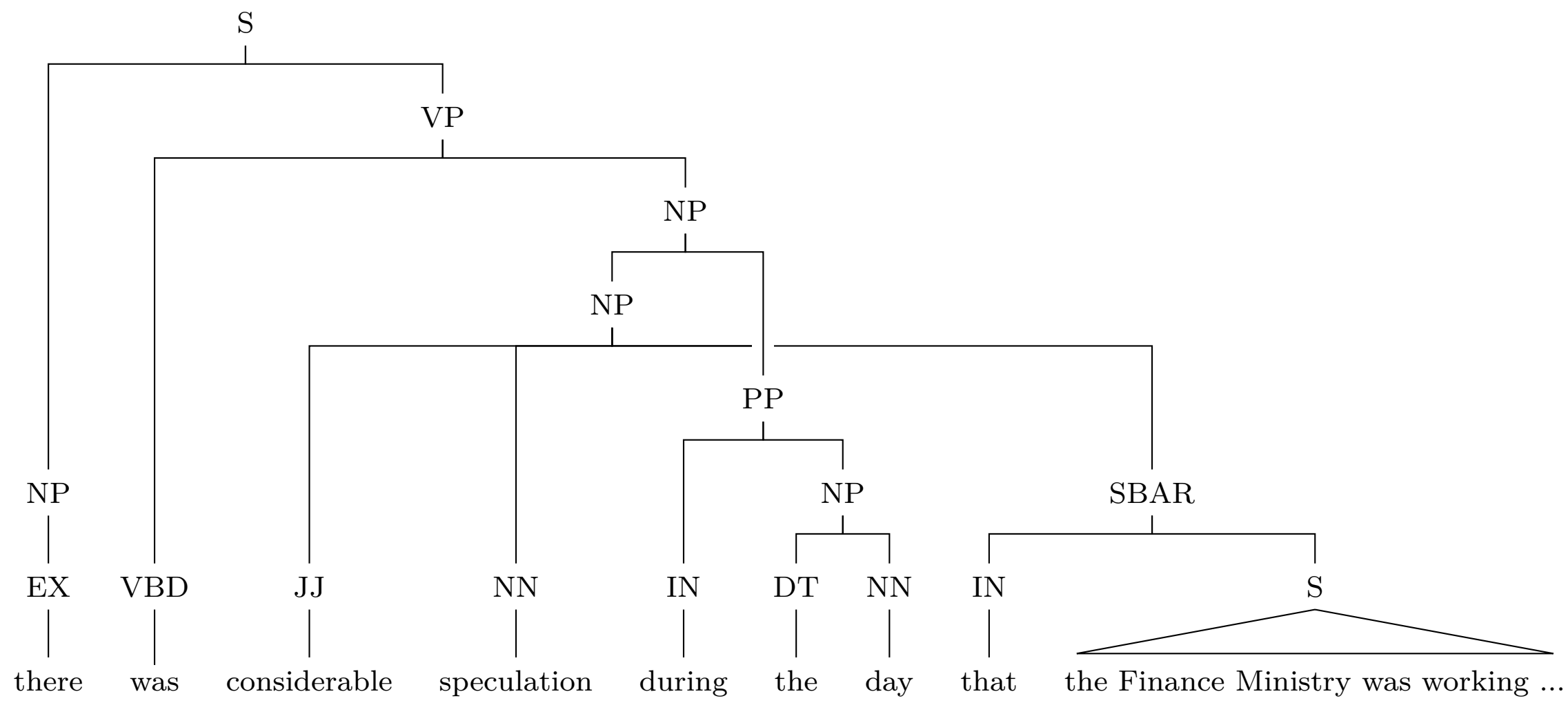

Figure 6.21: DPTB derivation including a discontinuous dependency with a PP gap element and an SBAR right element.

Figure 6.22 reproduces the corresponding CCGrebank category assignments. The appended part which can be reduced to S[em] is encoded as a core argument of *speculation* and not as a modifier of the gap. Thus, the relationship between the two parts of the discontinuous constituent has a direct correspondent in the supertag assignments. The parser is able to profit from the fact that the relative clause is encoded as a core argument of *speculation*. While *during the day* lies in between the two elements and has a category that suggests merging with *speculation* (function (N/S)\(N/S)), the parser is able to identify that this category encodes an adjunct and should be merged after attaching the core argument *that the Finance Ministry ...*. I assume that the distinction between arguments and adjuncts in CCG is a deciding factor for the model's ability to profit from supertag assignments.

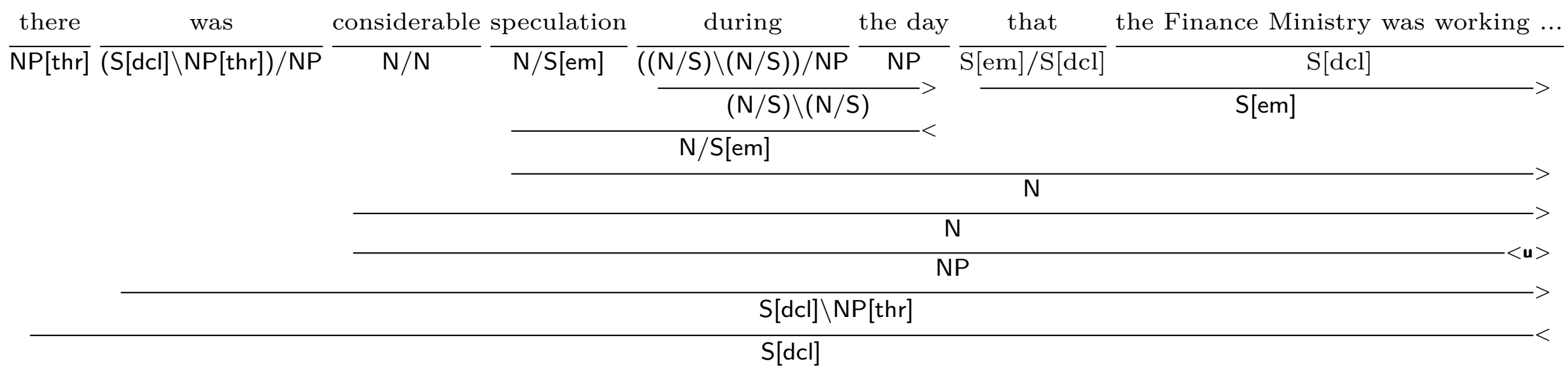

Figure 6.22: CCGrebank supertag assignment for a sentence that includes discontinuous dependency with a PP gap element and an SBAR right element.

For VP gap elements the problem lies in that the verb serves as the head of the whole clause in CCG and bridging it while retaining the relationship between the discontinuous constituent's substructures would necessitate encoding the appended S[em] constituent as an argument of *speculation*. This could be done by changing the assignments from those in figure 5.21 to figure 6.23 where N of *evidence* becomes NP/S[em] and is combined via crossed backward composition with *surfaced*, retaining the S[em] argument in the result. Alternatively, the argument and return type of *surfaced* could be changed to reflect the new S[em] argument of *evidence* which would give rise to an explosion of new verbal category types (figure 6.24). Both approaches have the problem that they come with the need of marking a new syntactic argument at the left NP category. One could justify this by viewing the right discontinuous part as an argument clause of the noun, but the linguistic motivation would definitely be questionable for relative clauses like in (6.8b). All in all, cases like these will likely remain to pose a challenge in CCG-enriched models due to the properties of the CCG formalism.

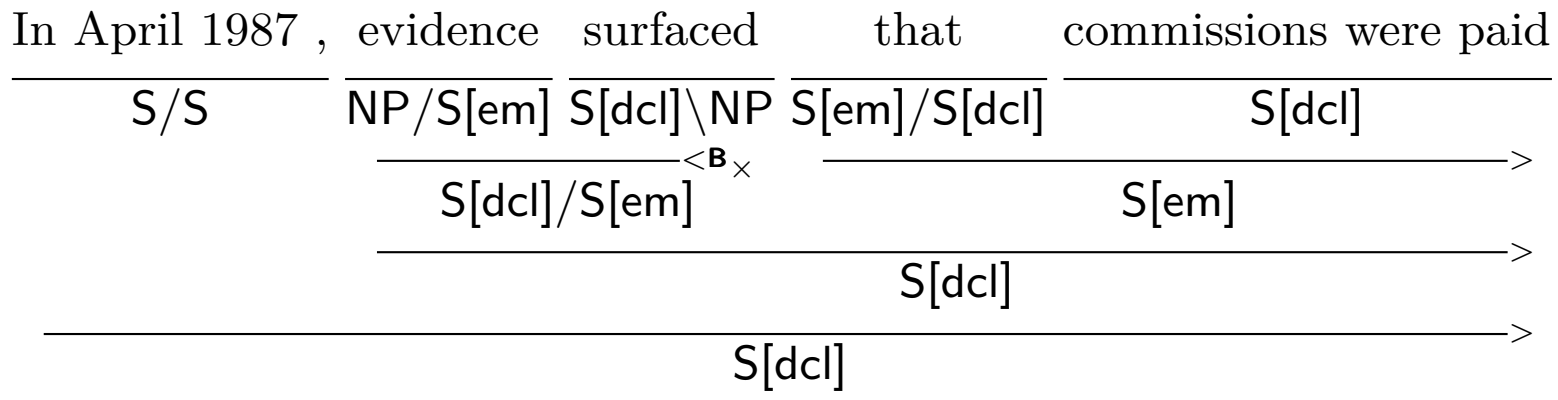

Figure 6.23: Alternative CCG supertag assignment for discontinuous dependency using crossed composition.

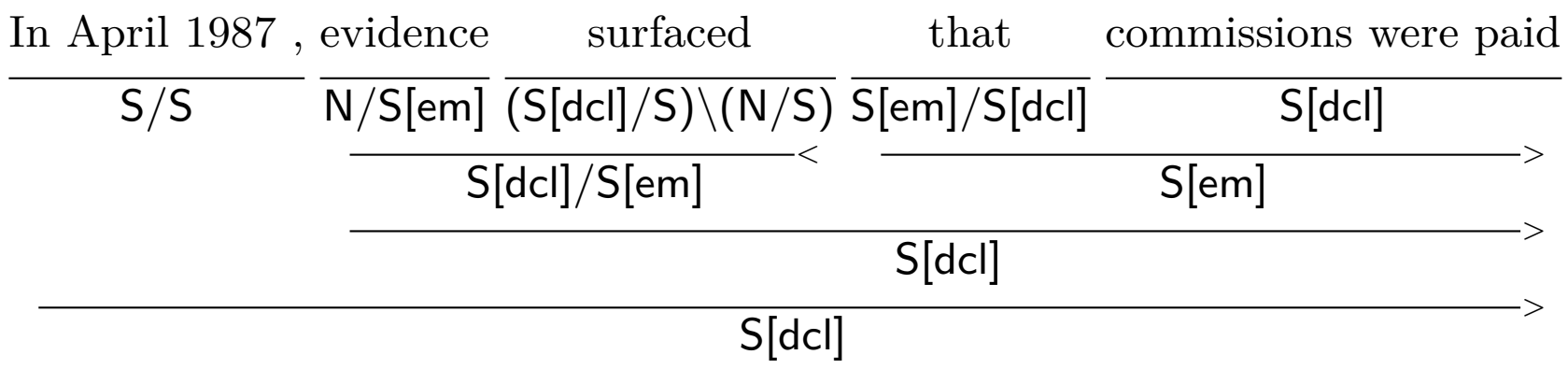

Figure 6.24: Alternative CCG supertag assignment for discontinuous dependency with a verbal category that preserves the argument of its subject.

### 6.3.6 External Comparison

Table 6.11 shows a comparison between the pipeline model $\text{PIPE}_{\text{bank}}$, the auxiliary task model scoring best on discontinuities $\text{CCG}^{\text{gate}}$ and results of previously proposed parsers on the DPTB. As can be seen, $\text{CCG}^{\text{gate}}$ provides the highest F-score for fully supervised transition-based parsers on the DPTB test split reported so far, surpassing the previous state-of-the-art model Coavoux (2021), which exhibits wider word embeddings (32 vs. 300) and character biLSTM states (100 vs. 256), by 0.1 absolute points. However, with a difference of 1.2 absolute points, the model is still far off from the overall best-scoring fully supervised model, being the grammar-less chart-based approach of Corro (2020).

Unfortunately, discontinuous F-score turns out to be significantly worse on the test dataset than on the development split. This is surprising since neither of these datasets was used in training. One can see that this effect already occurred in the original stack-free parser of Coavoux and Cohen (2019) (70.9 dev DF, 67.3 test DF) while it is absent from the GAP models from Coavoux et al. (2019) and from Coavoux (2021) which have a similar network design.

Perhaps the discontinuous phenomena in the development split have an unusual composition that the stack-free transition system is able to treat unusually well. During training, the algorithm checks every four epochs if the current network iteration scores better on the development set than the previously reported best-score and if it does, saves it. This way, a model might have been selected that does not generalise well to normal compositions of discontinuous structures.

Table 6.11 also contains a comparison of speeds where they were reported by the respective authors. Note that parsing speed is to a wide extent dependent on the type of implementation and the hardware used to run the parser. I reran the evaluation for the original model of Coavoux and Cohen (2019) to validate their reported speed of 38 sent/s which resulted in a value of 31 sent/s, much closer to the 29 sent/s for $\text{CCG}^{\text{gate}}$. I suppose that deviating from the information in their paper, they actually use two CPU threads to run the evaluation. This is also the setting found in the evaluation script which they provide as supplementary material. I ran the original model with two CPU threads which produced a speed of 36 sent/s which is close to their reported speed.

Generally, the evaluation of parsing speeds shows that while training an auxiliary task model is computationally more expensive due to the increase in training instances and model parameters (3 LSTM stacks instead of 2, gated residual connections), it only suffers minimal computational penalties after deployment when compared to its single-task counterpart (delta of 2 sent/s). Pursuing a pipeline approach speeds up training since one can outsource supertagging for the training corpus to the pre-processing pipeline but results in a model that is computationally more costly in the test setting since for every sentence to predict, two full-fledged neural networks are run which cannot be parallelised (delta of 8 sent/s).

Unfortunately, up until now, English per-phenomenon scores computed by the testsuite provided by Coavoux (2021) have only been reported for an ML-GAP-proposal by Coavoux (2021). Table 6.12 contains a comparison of the phenomenon-specific labelled F-score of the models proposed here and the ML-GAP parser from Coavoux (2021) as well as the ML-GAP+BERT variant which is enriched with BERT contextual embeddings. It can be seen that the $\text{CCG}^{\text{gate}}$ model outperforms ML-GAP in every type of discontinuity except for circumpositioned quotations. Indeed it is striking that all models from the work at hand, including the single-task baseline parser, perform only half as well on this class than ML-GAP and ML-GAP+BERT.

The improvements in F-score on discontinuous dependency attributed to the integration of BERT contextualised embeddings in the ML-GAP+BERT model remain unmatched. As Coavoux et al. (2019) note, most sentences in this class have no lexical trigger like *what* or *that* which makes recognising a discontinuity hard and requires world-knowledge. The ml-gap+bert model shows that this can be achieved to some extent by providing lexical knowledge to the model. Yet, the results of $\text{PIPE}_{\text{bank}}$ and $\text{CCG}^{\text{gate}}$ show that there is still some improvement to gain in this domain without using semi-supervised contextualised word-embeddings.

| Phenomenon | Count | baseline | $\text{PIPE}_{\text{bank}}$ | $\text{CCG}^{\text{gate}}$ | ML-GAP | ML-GAP+BERT |
|---|---|---|---|---|---|---|
| *Wh*-extraction | 91 | 84.0 | 80.8 | 86.4 | 79.5 | 89.8 |
| Fronted quotation | 71 | 94.3 | 92.2 | 93.6 | 92.3 | 95.8 |
| Discontinuous dependency | 37 | 37.9 | 41.4 | 42.1 | 33.3 | 71.1 |
| Circumpositioned quotation | 16 | 43.4 | 46.3 | 35.0 | 86.0 | 95.0 |
| *It*-extraposition | 12 | 58.8 | 66.7 | 66.7 | 55.6 | 81.8 |
| Extraction+fronted quotation | 7 | 84.2 | 84.2 | 100 | 97.3 | 100 |
| Discontinuous dependency+extraction | 5 | 62.9 | 62.1 | 68.8 | 64.5 | 88.2 |
| Extraction+extraction | 5 | 78.8 | 94.1 | 90.9 | 86.7 | 97.1 |
| Subject inversion | 5 | 75.0 | 75.0 | 75.0 | 75.0 | 100 |

Table 6.12: Comparison with previously published labelled per-phenomenon F-scores including models ML-GAP and ML-GAP+BERT from Coavoux (2021). Sentences where several discontinuous phenomena occur are treated separately. Only combinations with at least 5 occurrences in the DPTB development split are included.

| Model | Type | Complexity | Dev | | Test | | sent/s | |
|---|---|---|---|---|---|---|---|---|
| | | | $F_1$ | $DF_1$ | $F_1$ | $DF_1$ | GPU | CPU |
| **Fully supervised** | | | | | | | | |
| Evang and Kallmeyer (2011),‡, < 25 | GB | $\mathcal{O}(n^9)$ | 79† | | | | | |
| van Cranenburgh and Bod (2013), ≤ 40 | GB | $\mathcal{O}(n^9)$ | 85.2 | | 85.6 | | | |
| van Cranenburgh et al. (2016), ≤ 40 | GB | $\mathcal{O}(n^9)$ | 86.9 | | 87.0 | | <1□ | |
| Ruprecht and Mörbitz (2021) | GB | $\mathcal{O}(n^9)$ | | | 90.1 | 72.9 | $95^\gamma$ | |
| Coavoux and Cohen (2019) | TB | $\mathcal{O}(n^2)$ | 91.4 | 70.9 | 90.9 | 67.3 | | 38▼◇○ |
| Coavoux et al. (2019) | TB | $\mathcal{O}(n^2)$ | 91.2 | 72.0 | 91.0 | 71.3 | | **61**◇ |
| Coavoux (2021) | TB | $\mathcal{O}(n^2)$ | 92.0 | 75.9 | 91.4 | **74.4** | | |
| This work, $\text{PIPE}_{\text{bank}}$ | TB | $\mathcal{O}(n^2)$ | 91.5 | 70.2 | 91.4 | 66.8 | $31^\epsilon$ | 23♣♡○ |
| This work, $\text{CCG}^{\text{gate}}$ | TB | $\mathcal{O}(n^2)$ | 91.8 | 74.9 | 91.5 | 67.7 | $35^\epsilon$ | 29♣○ |
| Corro (2020) | CB | $\mathcal{O}(n^3)$ | | | **92.7** | 64.2 | **355**$^\delta$ | |
| Stanojević and Steedman (2020) | CB | $\mathcal{O}(n^6)$ | | | 90.5 | 67.1 | | |
| Corro et al. (2017) | DB | $\mathcal{O}(n^2)$ | | | 89.2 | | | ≈7.3 |
| **Semi-supervised (Pretrained embeddings)** | | | | | | | | |
| Ruprecht and Mörbitz (2021), fasttext+flair | GB | $\mathcal{O}(n^9)$ | | | 91.8 | 76.1 | $86^\gamma$ | |
| Coavoux (2021), fasttext | TB | $\mathcal{O}(n^2)$ | 92.7 | 78.1 | 92.3 | 76.5 | | |
| Corro (2020) | CB | $\mathcal{O}(n^3)$ | | | **92.9** | 64.9 | | |
| Vilares and Gómez-Rodríguez (2020), with Ling et al. (2015) | SL | $\mathcal{O}(n^2)$ | | | 89.3 | 45.2 | $572^\beta$ | **104**▲ |
| Chen and Komachi (2023), fasttext | NC | $\mathcal{O}(n^3)$ | | | 92.1 | **78.1** | **970**$^\beta$ | |
| **Semi-supervised (BERT-base)** | | | | | | | | |
| Ruprecht and Mörbitz (2021) | GB | $\mathcal{O}(n^9)$ | | | 93.3 | 80.5 | $57^\gamma$ | |
| Coavoux (2021) | TB | $\mathcal{O}(n^2)$ | 95.0 | 85.8 | **95.0** | **82.5** | | |
| Corro (2020)* | CB | $\mathcal{O}(n^3)$ | | | 94.8 | 68.9 | | |
| Vilares and Gómez-Rodríguez (2020) | SL | $\mathcal{O}(n^2)$ | | | 91.9 | 50.8 | $80^\beta$ | **2**▲ |
| Fernández-González and Gómez-Rodríguez (2021) | RCB | $\mathcal{O}(n^3)$ | | | 94.0 | 68.9 | **231**$^\alpha$ | |
| Fernández-González and Gómez-Rodríguez (2021) | RTB | $\mathcal{O}(n^2)$ | | | 94.1 | 67.2 | $152^\alpha$ | |
| **Semi-supervised (XLNet)** | | | | | | | | |
| Chen and Komachi (2023) | NC | $\mathcal{O}(n^3)$ | | | 95.0 | **83.0** | **375**$^\beta$ | |
| Fernández-González and Gómez-Rodríguez (2021) | RCB | $\mathcal{O}(n^3)$ | | | 95.1 | 74.1 | $179^\alpha$ | |
| Fernández-González and Gómez-Rodríguez (2021) | RTB | $\mathcal{O}(n^2)$ | | | **95.5** | 73.4 | $133^\alpha$ | |

Table 6.11: Global comparison of parsing results on the DPTB; rounded to the first decimal place. Complexity for grammar-based approaches assumes fan-out 3 as found in the DPTB. Types: GB: grammar-based, TB: transition-based, CB: grammar-less chart-based, SL: sequence-labelling based, DB: dependency-conversion based, NC: neural combinator, RCB: reorder and projective grammar-less chart-based, RTB: reorder and projective transition-based. *Size of BERT not reported. † Does not discount root symbols and punctuation, ‡ gold POS tags. Speeds are presented on the test set as reported by the respective authors and dependent on hardware and implementation. GPUs: $^\alpha$GeForce RTX 3090, $^\beta$GeForce GTX 1080 Ti, $^\gamma$GeForce RTX 2080, $^\epsilon$GeForce RTX 2060 SUPER 8GB, $^\delta$Tesla V100. CPUs: ▲Intel i7-7700 @ 3.60GHz, ▼Intel i7, ♠AMD Ryzen 5 3600 @ 3.60GHz. ○run on single core. ◇reported on dev set. ♡including supertagging. □measured by Ruprecht and Mörbitz (2021).

# 7. Conclusion

This work explored the introduction of lexicalised syntactic information in the form of CCG supertags into discontinuous constituent parsing. To this end, several approaches for parsing discontinuous constituent trees have been discussed: grammar-based chart parsing for linear context-free rewriting systems (LCFRS) as well as neural transition-based parsers (Swap, Gap, Merge) among which the neural stack-free approach of Coavoux and Cohen (2019) was chosen as the base for the incorporation of supertags.

After introducing combinatory categorial grammar (CCG) and the supertags it gives rise to, I manually compared correlations between lexical category assignments found in the CCGrebank corpus and the way different types of discontinuities are analysed in the discontinuous Penn Treebank (DPTB). This analysis suggested close relationships for some forms of discontinuity like *wh*-movement, fronted quotations and *it*-extrapositions while revealing potentially conflicting analyses for circumpositioned quotations, extraposed dependents and subject-verb inversion. Circumpositioned quotations showed to be especially troublesome since they do not seem to exhibit a coherent and transparent analysis in the CCGrebank. The manual analysis may be helpful for designing models in future works. It underlines the necessity to utilise architectures that can dynamically adapt to the considerably varying phenomenon-dependent task relatedness.

The main merit of this work lies in the implementation of CCG supertagging into the stack-free transition-based parser of Coavoux and Cohen (2019) following a pipeline and an auxiliary-task approach. For the latter, a variety of experimental settings was explored (gated residual connections, increased network width, deconstruction of supertags) showing that care must be exercised and adaptions must be made to benefit from shared representations for both parsing and supertagging. Both the $\text{Pipe}_{\text{bank}}$ model and the best-scoring auxiliary model $\text{CCG}^{\text{gate}}$ outperform the supertagless baseline in standard F-score. On discontinuities only, the pipeline model performs worse than the baseline while the auxiliary approach proofs to be an effective means of increasing a parser's syntactic competence regarding long-range dependencies. The auxiliary approach also outperforms its control model which confirms that the improvements can be attributed to the introduction of an auxiliary objective. It show that the complex relationship of seemingly incompatible syntactic descriptions can be effectively made use of to improve the quality of discontinuous constituent parsing. It also indicates that training a multi-task model can leverage statistical correlations between tasks to an extent a pipeline model is not able to. While the improvements are relatively modest, they suggest that research into more adapted specialised multi-task approaches for discontinuous constituent parsing that learn across representations is a worthwhile endeavour.

This result is closely connected with the utilisation of an alternative to the standard summative residual connection. Gated residual connections have been empirically shown to outperform their widely used counterpart in a hierarchical multi-task framework with stable results for biLSTM stacks up to four levels ($\text{Ctr}_4^{\text{gate}}$).

The mixed results of networks with additional CCG supertag-based tasks in the form of decomposed supertags ($\text{CCG}^{\text{gate}}_{\text{multi}}$) and head-dependency information provided by the CCGrebank ($\text{CCG}^{\text{gate}}_{\text{dep}}$) indicate that care must be exercised when enriching a model with information that goes beyond the treatment of CCG supertags as discrete categories. Head-dependency information seems to be too close to the conflicting notion of CCG derivations. While subcomponent information does not beat the best-scoring multi-task model $\text{CCG}^{\text{gate}}$, it does improve the discontinuous metrics compared to its four-layer control model. Noteworthy about this result is the slight improvement in discontinuous recall which deviates from the trend of recall drops observable in most other multi-task models. This suggests that subcomponent training could actually be of value in a multi-task approach.

Additionally, I provided a first systematic investigation of the usefulness of four different syn-tactic sequence labelling tasks as auxiliary objectives for transition-based parsing of discontinuous constituents. While CCG supertagging shows to be most beneficial for discontinuous constituents, the inclusion of chunking produces the best general F-score. Since the results may be influenced to some extent by the vastly differing numbers of distinct categories, it is difficult to draw conclusions about the individual tasks' informativeness for parsing. Nonetheless, the evaluation is a good starting point for the choice of auxiliary tasks for future projects and should be transferable to other neural parser types.

## 7.1. Open Questions

While this work has shown that introducing CCG supertagging as an auxiliary task for discontinuous constituent parsing can yield systematic improvements in parsing score, the experiments have not come close to the current state-of-the-art results which are dominated by models that use XLnet (Yang et al., 2019) and BERT (Devlin et al., 2019) contextual embeddings. Thus, the open question remains, whether CCG supertagging would provide any informative value to such models or if the syntactic competence induced by supertagging is already overtly encoded in the lexical information these embeddings provide.

Another question raised by this work deals with the viability of combining several types of unrelated auxiliary tasks. Here, I have only explored multi-task learning with one additional type of syntactic representation. The unsuccessful joint learning of CCG head-dependencies is the closest I have come to combining several representations. As explained in section 6.3, the different auxiliary tasks yield different types of improvements. For instance chunking improves general F-score while CCG supertagging affects discontinuous constituents more. Judging from this, combining several beneficial auxiliary tasks in a shared model could leverage additional synergies. However, this also raises the risk of overfitting and necessitates a computationally expensive training procedure. Furthermore, arranging several unrelated auxiliary tasks in a hierarchical architecture would pose a challenge. Overall, the potential of such an exploration is unclear.

A similar question stands for the deconstruction of supertags into subcomponent labelling tasks. Even though this approach does not perform better than the supertag-only model in this work, the unique improvement of discontinuous recall raises the question whether the kind of deconstructions or the model architecture can be improved upon to benefit from this effect.

In view of the recent successful discontinuous parsing approach using LCFRS-based supertags (Ruprecht and Mörbitz, 2021), I am looking forward to exploring ways to implement LCFRS supertagging as an auxiliary task in a neural framework effectively. Reducing the number of supertags in an informed manner as well as finding useful subtasks with smaller label set sizes might be the way to go.

Furthermore, questions regarding the neural architectures explored in this work remain. In future work, I want to explore using the gated residual connections to LSTM cell outputs proposed by Wu et al. (2016a) that inspired the simplified gated residual design in this work. Additionally, the comparison of training and development score of the $\text{CCG}^{\text{gate}}$ model in figure 6.9 as well as the results of the $\text{CCG}^{\text{gate}}_{600}$ model (cf. figure 6.10) suggested that there might be improvements to gain from better regularisation techniques. An interesting possibility is the application of variational dropout for recurrent connections as proposed by Gal and Ghahramani (2016).

Lastly, the exploration of different kinds of supertags as auxiliary tasks still lacks an important contender: lexicalised tree adjoining grammar. While the use of LTAG-spinal has not shown to be an effective means to improve discontinuous constituent parsing results, subcategorisation information as well as argument-adjunct distinctions inherent to traditional LTAG may be properties akin to the characteristics of CCG supertags that have proven helpful for the parser. I am looking forward to performing such an experiment after acquiring an LTAG-annotated corpus.

# Statutory Declaration

I affirm that I have authored the thesis titled

**Integrating Supertag Features into Neural Discontinuous Constituent Parsing**

independently and without illicit assistance from third parties. All parts of the text, figures, examples and other graphical representations that were taken from the work of others by wording or spirit are marked explicitly with regard to the sources. I have not used any sources other than those specified.

Ich versichere an Eides statt, dass die Arbeit mit dem Titel

**Integrating Supertag Features into Neural Discontinuous Constituent Parsing**

von mir selbständig und ohne unzulässige fremde Hilfe verfasst wurde. Alle Stellen der Arbeit sowie beigefügte Zeichnungen, Skizzen und andere graphische Darstellungen, die anderen Werken dem Wortlaut oder dem Sinn nach entnommen sind, habe ich unter Angabe der Quellen kenntlich gemacht. Ich habe keine anderen als die angegebenen Quellen genutzt.

Lukas Mielczarek